\documentclass[a4paper,12pt,numbered,print,index]{Classes/PhDThesisPSnPDF}
\usepackage[T1]{fontenc} 
\usepackage{blindtext}
\usepackage{nomencl}
\usepackage{natbib}
\usepackage{bibunits}
\usepackage{amsmath,amssymb,amsfonts}
\usepackage{mathtools}
\usepackage{epsfig}
\usepackage[mathscr]{euscript}
\usepackage{color}
\usepackage{booktabs} 
\usepackage{stmaryrd} 
\usepackage{bbm} 
\usepackage{blindtext, comment}
\usepackage{subcaption}
\usepackage{rotating} 
\usepackage{tikz}
\usepackage[subpreambles=false]{standalone}
\usetikzlibrary{arrows, patterns}
\usetikzlibrary{decorations.pathreplacing,angles,quotes}
\usepackage{framed, standalone}
\usepackage{pgfplots}
\usepackage{graphicx}
\usepackage{capt-of}
\usepackage{balance}
\usepackage{makecell} 
\usepackage[linesnumbered,lined,boxed,commentsnumbered,ruled,vlined]{algorithm2e}
\usepackage{setspace}
\makeatletter
\renewcommand{\fnum@figure}{Figure \thefigure}
\makeatother

\input{FrontMatter/preamble}

\title{}

\renewcommand{\submissiontext}{
\vspace{-7.5cm}
\includegraphics[scale=0.075]{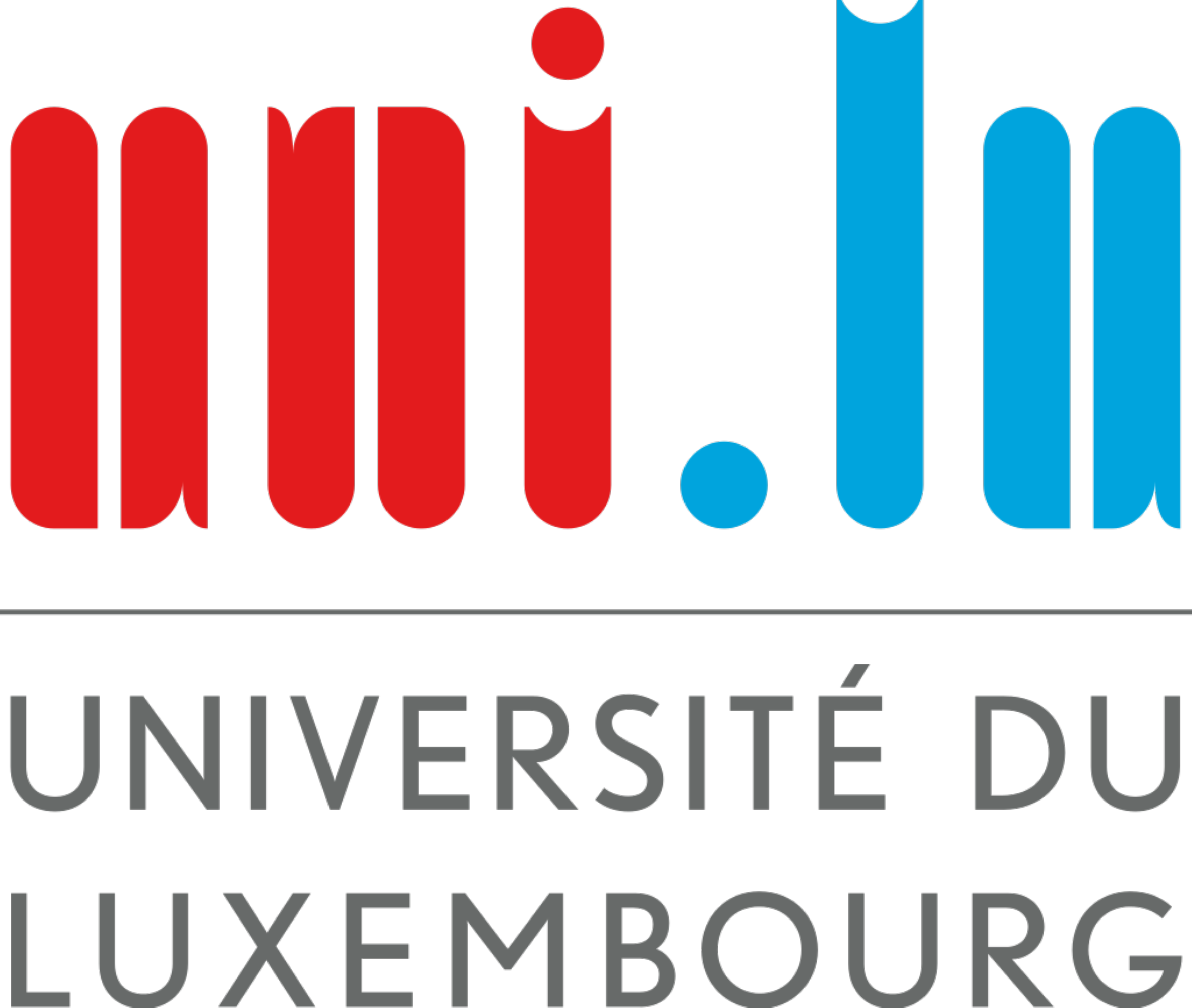}\\

\vspace{.25cm}
\small{PhD-FSTC-2019-64 \\
\vspace{-0.2cm}The Faculty of Sciences, Technology and Communication}

\vspace{.5cm}
\large{DISSERTATION} \\

\vspace{.5cm}
\small{Defence held on 22/10/2019 in Luxembourg \\
to obtain the degree of} \\

\vspace{.5cm}
\large{DOCTEUR DE L'UNIVERSIT\'E DU \\ LUXEMBOURG \\ 
\vspace{.25cm}EN INFORMATIQUE} \\

\vspace{.25cm}
\small{by}\\

\vspace{.25cm}
\Large{Jeremy Henri J CHARLIER}\\
\small{Born on 23 September 1988 in Liege (Belgium)}\\

\vspace{.75cm}
\Large{
\textsc{From Persistent Homology\\
\vspace{-0.3cm}to Reinforcement Learning\\
\vspace{-0.3cm}With Applications for Retail Banking}
}\\

\vspace{.25cm}
\flushleft{Dissertation defence committee}\\

\begin{small}
\relscale{1.25}
\vspace{.15cm}
\vspace{-.1cm} \small{ \relscale{1.2} Dr. Radu State, Dissertation Supervisor} \\
\vspace{-.2cm} \small{\textit{Professor, SnT, University of Luxembourg}}  \\

\vspace{.15cm}
\small{ \relscale{1.2} Dr. Bj\"orn Ottersten, Chairman} \\
\vspace{-.2cm} {\footnotesize{\textit{Professor, SnT, University of Luxembourg}}}  \\

\vspace{.15cm}
\small{ \relscale{1.2} Dr. Jean Hilger, Vice Chairman} \\
\vspace{-.2cm} \textit{Head of Information Technology Department, BCEE} \\

\vspace{.15cm}
\small{ \relscale{1.2} Dr. Djamila Aouada} \\
\vspace{-.2cm} \textit{Professor, SnT, University of Luxembourg} \\

\vspace{.15cm}
\small{ \relscale{1.2} Dr. Vijay Gurbani} \\
\vspace{-.2cm} \textit{Professor (Adjunct), Illinois Institute of Technology} \\
\vspace{-.6cm}\textit{Chief Data Scientist, Vail Systems, Inc.}
\end{small}
}

\ifdefineAbstract
\fi

\ifdefineChapter
\fi

\begin{document}

\frontmatter
\maketitle

\begin{dedication}
\begin{flushright}
\vspace{5cm}
To Marie-Jeanne
\end{flushright}
\end{dedication}

\begin{declaration}

I hereby declare that except where specific reference is made to the work of 
others, the contents of this dissertation are original and have not been 
submitted in whole or in part for consideration for any other degree or 
qualification in this, or any other university. This dissertation is my own 
work and contains nothing which is the outcome of work done in collaboration 
with others, except as specified in the text and Acknowledgements. This 
dissertation contains fewer than 60,000 words including appendices, 
bibliography, footnotes, tables and equations and has fewer than 150 figures.


\end{declaration}

\begin{acknowledgements}      
I would like to thank particularly Dr. Habil. Radu State and Dr. Jean Hilger for having given me the opportunity to pursue a PhD at the university of Luxembourg in collaboration with the Banque et Caisse d'\'Epargne de l'\'Etat (BCEE). 
I am grateful for the time and the dedication they put to ensure the successfullness of the research project. I would like also to thank Dr. Djamila Aouada for her precise advice and challenging questions during the PhD supervision. Additionally, I am truly grateful to Patric De Waha for taking the time to supervise the various projects for which I was involved at BCEE as well as his outstanding understanding of the difficulties of a PhD research project. Furthermore, I would like to thank Dr. Henning Schulzrinne and Dr. Gaston Ormazabal for having given me the opportunity to do a research stay within the IRT lab at Columbia university. I deeply appreciate the time they invested for the supervision of the PhD research exchange, leaving me unforgettable memories. I am genuinely appreciative to Dr. Bj\"{o}rn Ottersten for his valuable inputs during the review of this work. Moreover, I am grateful to Dr. Vijay Gurbani for his dedication and his critiques throughout the assessment of this PhD thesis. Finally, I would like to thank my fianc\'ee, Jennifer, for her patience during this long journey and my family for their continuous support. I would like to thank all my fellow colleagues of the university of Luxembourg, the Banque et Caisse d'Epargne de l'Etat and the Columbia university as well as all the others I forgot to mention who helped to contribute to this work.
\end{acknowledgements}
\begin{abstract}
The retail banking services are one of the pillars of the modern economic growth. However, the evolution of the client's habits in modern societies and the recent European regulations promoting more competition mean the retail banks will encounter serious challenges for the next few years, endangering their activities. They now face an impossible compromise: maximizing the satisfaction of their hyper-connected clients while avoiding any risk of default and being regulatory compliant. Therefore, advanced and novel research concepts are a serious game-changer to gain a competitive advantage.
\\ \vspace{-.3cm}

In this context, we investigate in this thesis different concepts bridging the gap between persistent homology, neural networks, recommender engines and reinforcement learning with the aim of improving the quality of the retail banking services. Our contribution is threefold. First, we highlight how to overcome insufficient financial data by generating artificial data using generative models and persistent homology. Then, we present how to perform accurate financial recommendations in multi-dimensions. Finally, we underline a reinforcement learning model-free approach to determine the optimal policy of money management based on the aggregated financial transactions of the clients. \\ \vspace{-.3cm}

Our experimental data sets, extracted from well-known institutions where the privacy and the confidentiality of the clients were not put at risk, support our contributions. In this work, we provide the motivations of our retail banking research project, describe the theory employed to improve the financial services quality and evaluate quantitatively and qualitatively our methodologies for each of the proposed research scenarios. 
\end{abstract}
\tableofcontents

%
%
\renewcommand{\bibname}{List of Publications}
\begin{bibunit}[unsrt]
\makeatletter
\renewcommand\@biblabel[1]{}
\makeatother
\def\bibfont{\small}
\nocite{*}
\putbib[refs1]
\end{bibunit}

\renewcommand{\bibname}{References}
%
%
%
%
%
\nomlabelwidth=30mm 

\nomenclature[1]{AE}{Auto-Encoder}
\nomenclature[1]{WAE}{Wasserstein Auto-Encoder}
\nomenclature[1]{VAE}{Variational Auto-Encoder}
\nomenclature[1]{GAN}{Generative Adversarial Network}
\nomenclature[1]{WGAN}{Wasserstein Generative Adversarial Network}
\nomenclature[1]{GP-WGAN}{Gradient Penalty Wasserstein Generative Adversarial Network}
\nomenclature[1]{TD}{Tensor Decomposition}
\nomenclature[1]{ALS}{Alternating Least Square}
\nomenclature[1]{SGD}{Stochastic Gradient Descent}
\nomenclature[1]{NAG}{Nesterov Accelerated Descent}
\nomenclature[1]{RMSProp}{Root Mean Square Propagation}
\nomenclature[1]{NCG}{Non-linear Conjugate Gradient}
\nomenclature[1]{BFGS}{Broyden–Fletcher–Goldfarb–Shanno}
\nomenclature[1]{ApHeN}{Approximate Hessian Newton}
\nomenclature[1]{MLP}{Multi-Layer Perceptron}
\nomenclature[1]{CNN}{Convolutional Neural Network}
\nomenclature[1]{LSTM}{Long Short Term Memory}
\nomenclature[1]{DT}{Decision Tree}
\nomenclature[1]{BSM}{Black-Scholes-Merton}
\nomenclature[1]{RL}{Reinforcement Learning}
\nomenclature[1]{MQLV}{Modified Q-Learning for Vasicek}
\nomenclature[1]{CP}{Candecomp$/$Parafac}
\nomenclature[1]{PHom}{Persistent Homology}
\nomenclature[1]{AI}{Artificial Intelligence}
\nomenclature[1]{ML}{Machine Learning}
\nomenclature[1]{SVM}{Support Vector Machine}
\nomenclature[1]{ReLU}{Rectified Linear Unit}
\nomenclature[1]{NN}{Neural Networks}
\nomenclature[1]{CG}{Conjugate Gradient}
\nomenclature[1]{VecHGrad}{Vector Hessian Gradient}
\nomenclature[1]{RMSE}{Root Mean Square Error}


\renewcommand*\listfigurename{List of Figures}
\listoffigures
\renewcommand*\listtablename{List of Tables}
\listoftables
\printnomenclature


\mainmatter

\chapter{Introduction}

Eleven years after the biggest financial crisis of the twenty-first century, the financial industry is facing a continuously increasing pressure from the European regulators. Either new or revised European regulations are taking effect impacting all services of the banking industry. For instance, the investment banks have to comply with the revised Markets in Financial Instruments Directive  \cite{mifidII}, MIFID 2, which tries to promote more transparency, more safety and more competition in financial markets. Furthermore, to increase the safety of the financial sector, the International Financial Reporting Standard 9, IFRS9, as adopted by the European Union \cite{ifrsIX} from the International Accounting Standards Board, IASB, and the third Basel accord \cite{baselIII}, Basel III, are now mandatory requirements. The IFRS9 accounting standard forces the banks to make provisions for future expected losses. The Basel III framework is an international framework that promotes the evaluation of risk-weighted assets triggering the amount of equity the banks should have to ensure a reliable capital strength. Directly impacting the retail banking, the revised Payment Service Directive \cite{psdII}, PSD2, aims at promoting more competition in the financial sector of retail banking services in the European Union. Under this directive, the retail banks lost the exclusive privilege of the distribution of payments solution such as credit cards or the confidential knowledge of the clients’ financial situation. Financial companies can now challenge traditional retail banks by proposing more attractive financial packages, regarding interest rate loans or insurance policies, to gain new customers. Additionally, any bank can now consult using an application program interface the financial information of any clients from any other banks without limitation. Therefore, a bank having an aggressive marketing strategy can consult the information of the clients of the other banks before prototyping financial packages targeting the clients of the competitors. The banks will be able to further compete between each other. 
Straightforwardly, this directive will have a major impact on the business opportunities and the benefits of retail banks. 
\\

\begin{figure}[b!]
  \begin{center}
  {\includegraphics[scale=.35]{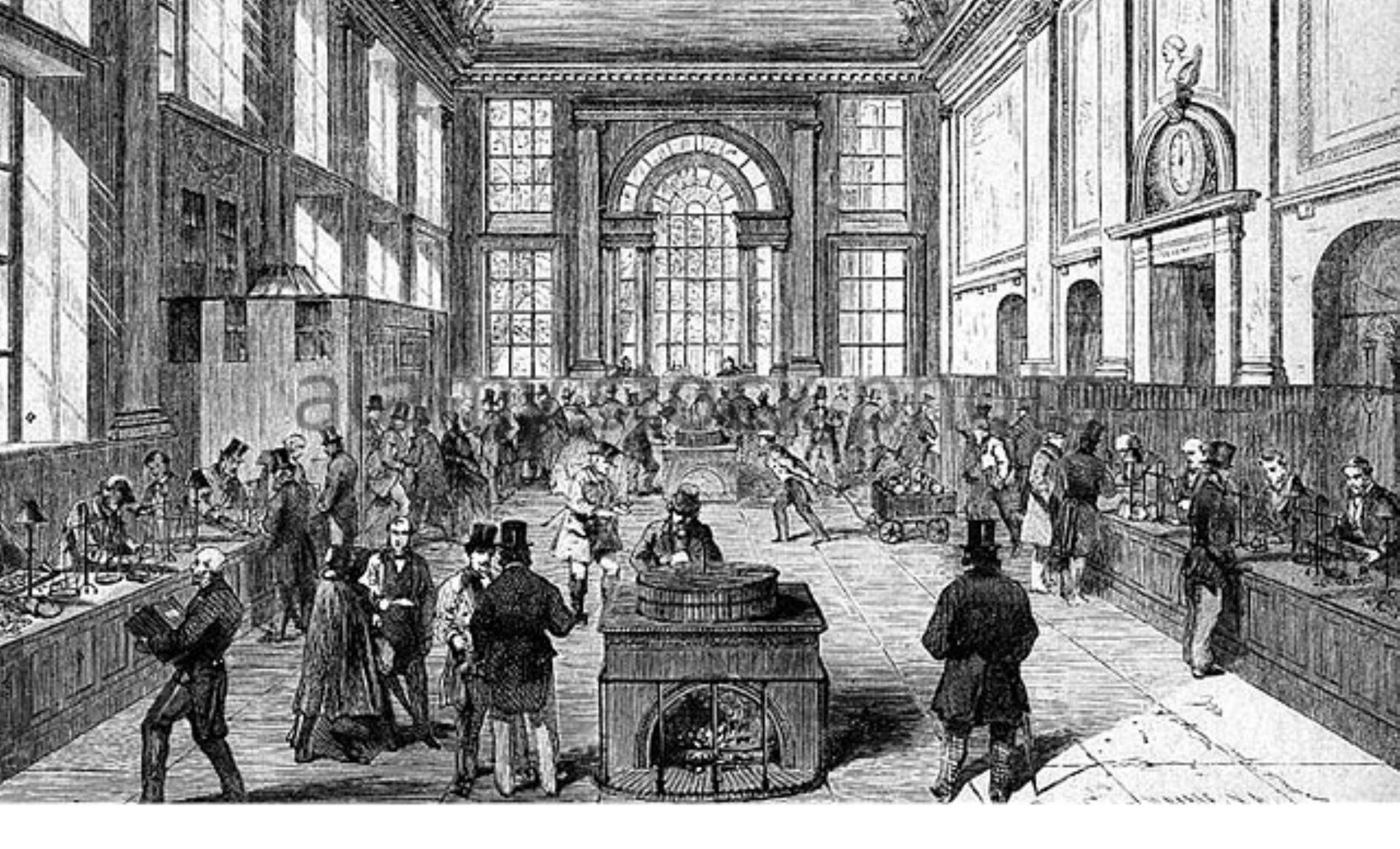}} \\ \vspace{.15cm}
  {\includegraphics[scale=2.34]{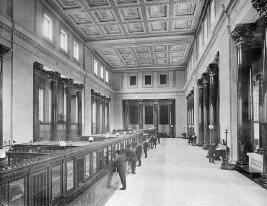}}
  {\includegraphics[scale=.225]{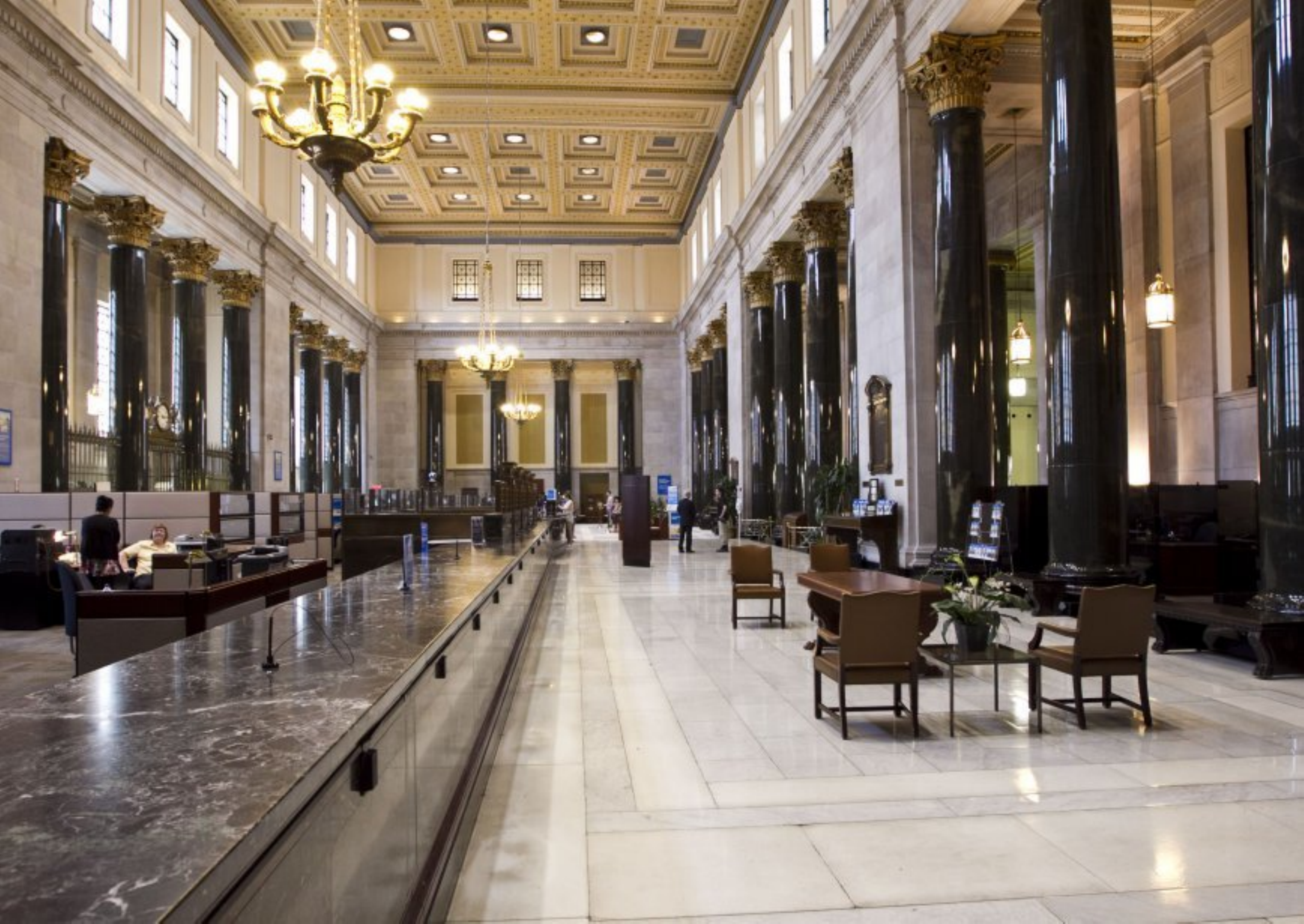}}
    \caption[Evolution of the retail banking agencies]{The retail banking offices and services have remained almost identical for the past 150 years. From left to right: the bank of England in the 19th century, the bank of Montreal at the beginning of the 20th century and the bank of Montreal nowadays.} \label{fig::bank1} 
  \end{center}
\end{figure}

Meanwhile, the society is going under a significant evolution with a new era of digitalization, contrasting the pace of evolution of the retail banking industry pictured in Figure \ref{fig::bank1}. For the past 150 years, the retail banking remained almost identical. 
The survey \cite{banklux} highlights that ninety-four percent of the people between eighteen and seventy-five years old have access to a computer while eighty-eight percent have access to a smartphone in thirty-one countries, either developed countries or countries in development. Furthermore, in the US in 2018, Americans spent an average time of more than eleven hours per day in front of a multimedia screen, either a smartphone, a tablet, a computer or a television \cite{nielsen}. It represents a significant rise of thirty minutes per day in comparison to 2017. These eleven hours are spent to watch, read, listen or write multimedia content. These drastic evolution additionally are highly dependent of the generation. For instance, millennials, people born between 1980 and 2000, are avid consumers of mobile technologies. Thirty percent of their multimedia time is spent on a smartphone. This internet generation, which will shape the future of our society, have opinions and spending habits that vary significantly from older generations \cite{millenials}. 
Strongly impacted by the 2008 financial crisis, the millenials underline that the retail banking is at its highest risk of disruption, forcing a very conservative sector to reinvent itself. As reported by the Millennial Disruption Index (MDI) in \cite{mdi}, one in three millennials are effectively opened to switch to another bank in the next ninety days. Besides, the four leading worldwide banks, JP Morgan Chase, Bank of America, Citigroup and Wells Fargo, are among the ten least loved brands among millennials. Seventy-three percent would be more excited by a new offering of financial services by Google, Amazon, Apple, Paypal or Square than their national bank, illustrating their sympathy for technology companies. A shocking number of thirty-three percent of millennials even think they do not need a bank at all. These statistics are only the emerged part of the iceberg for the retail banking services. When taking into consideration all adults being more than eighteen years old, the interactions between the clients and the banks have changed significantly for the past few years \cite{pwcbankingsurvey}. Between 2012 and 2018, visiting a retail banking agency went down from few times per month to few times per year while the mobile banking, relying on smartphones and tablets, skyrocketed from few times per year to few times a month. We recall that mobile banking denotes the use of a mobile device, such as a tablet or a smartphone, to connect online to the financial accounts while the online banking uses a computer. The instantaneous access to banking services offered by the digital channels, using online banking and mobile banking, is now more privileged by the clients than the traditional visit to the retail banking agency. This trend is particularly observable in China, especially when considering the mobile payments and the transactions market. For instance, Ant Financial, an affiliate company of the Chinese group Alibaba, and Tencent, a Chinese internet-based company, performed more than 22 trillion US dollars worth of mobile payments in 2018, outpacing the entire world's debit and credit card payments volume \cite{king2018bank}. Relying on the latest digital innovations, WeBank, China's first digital bank and affiliated to Tencent, uses a real-time blockchain core allowing unique offers and solutions for transaction payments and marketing campaigns \cite{king2018bank}. WeBank has attracted more than 100 million clients since 2014 without the requirement of paper-based applications \cite{king2018bank}. The retail banking changes happening currently in China illustrates strongly the future changes of retail banking activities in the next years to come for the rest of the world. Consequently, we can easily grasp the challenges facing the retail banks due to the evolution of the habits of the society and the further digitalization of services. 
\\

In this regulatory and digitalization context, retail banks have to enrich and propose new financial services as well as to reinvent their way of thinking. They can now rely on the new source of information coming from the digital media and digital devices. Furthermore, they can use the latest techniques inherited from computer science, such as machine learning and artificial intelligence, for their survival. More precisely, the retail banks can now benefit and use the full potential of large data sets combining exclusive information of the clients, financial information and economic predictions to create value and gain a competitive advantage. These large data sets are the fuel of the machine learning and artificial intelligence algorithms. To understand better the research problem, we have to explain in more details these generic terms and the techniques related. The expression artificial intelligence (AI) is widely used to describe advanced regressions, machine learning techniques, neural networks, deep learning and reinforcement learning. In an artificial intelligence configuration, a large data set is used with an algorithm that is capable to find rules, patterns within this data set. Although the frontier between each notion is so vague that every computer scientist has his own opinion, we can demystify and shed light among the different techniques included in this novel field. 

First, the term machine learning is very often used to describe clustering algorithms such as K-means \cite{lloyd1982least} or K-nearest-neighbor \cite{altman1992introduction}, linear algebra factorization such as the Singular Value Decomposition (SVD) \cite{golub1965calculating} or the Tensor Decomposition (TD) \cite{kolda2009tensor} used in recommender engines, decision trees \cite{belson1959matching} or Support Vector Machine (SVM) \cite{cortes1995support}. What is characteristic of these techniques is that the model fine tunes its parameters based on the data set provided as input instead of relying on hard coded and predefined rules. In most of the applications, the machine learning algorithms are defined as supervised or semi-supervised learning. In supervised learning, all of the features of the data sets are known and each example is associated with a label. In semi-supervised learning, only a fraction of the features is labeled.

\begin{figure}[t]
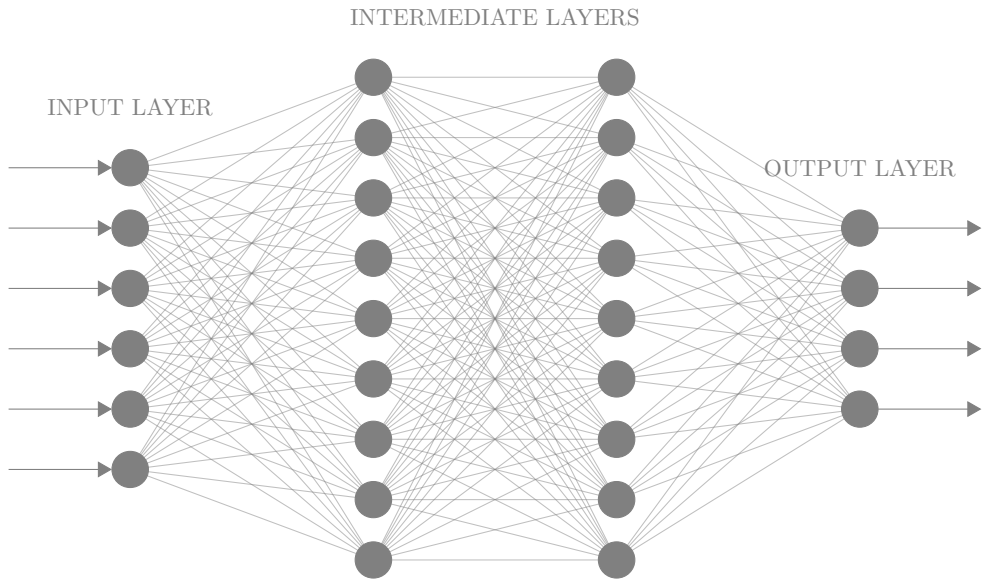

  \centering
  \includestandalone{chapter1/tikz/feedforwardnet}
  \caption[Deep learning representation of layers of neurons]{In deep learning, a large number of neurons are stacked through different layers to process the data.} 
  \label{fig::feedforward}
\end{figure}

Then, a second term widely used related to artificial intelligence is deep learning. Deep learning relies on neural networks \cite{goodfellow2016deep}. A neural network, or artificial neuron, is an elementary unit receiving a weighted input. A non linear function, or activation function, such as the sigmoid function or the Rectified Linear Unit \cite{nair2010rectified}, is used to transform the input signal to an output signal. Neural networks are highly effective to solve complex tasks because of the way they can be architected, as illustrated in Figure \ref{fig::feedforward}. Neural networks can be stacked through different layers where the first layer of neurons receives the input data and the last layer of neurons outputs the processed information contained within the data. The intermediate layer, or the hidden layer, made of at least one layer of neurons, takes care of the signal processing. The collection of neurons and layers allows to increase the problem solving capabilities and often lead to a better accuracy in numerical experiments. Furthermore, the deep learning term denotes a use of a significant number of neurons and layers. Contrary to the technique aforementioned, most neural networks and deep learning applications are classified as unsupervised learning. In an unsupervised learning framework, the algorithms try to infer properties of the data set without knowing any labels. 

\begin{figure}[t]
  \begin{center} 
  {\includegraphics[scale=.61]{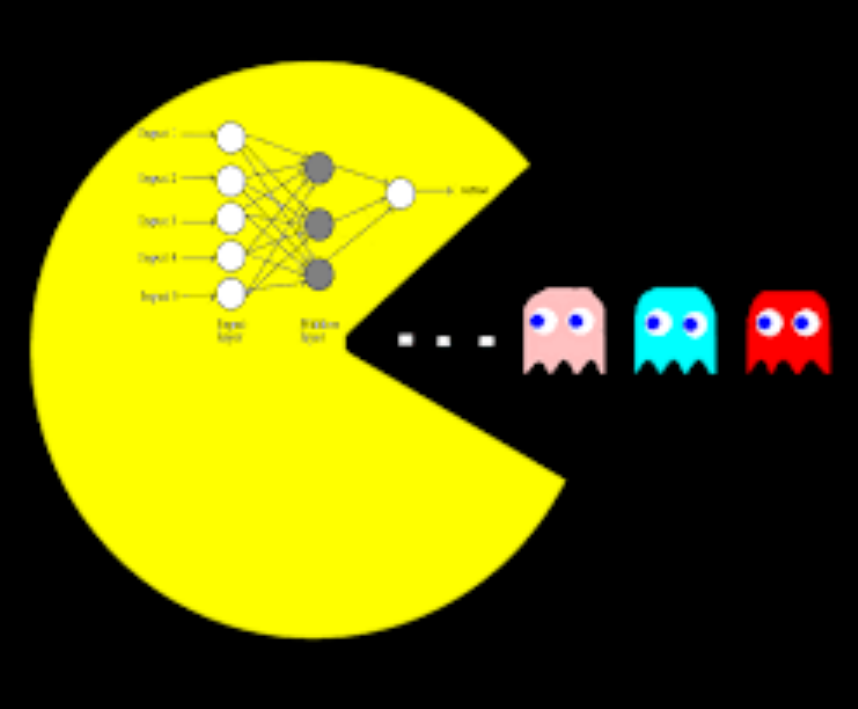}}
  {\includegraphics[scale=.225]{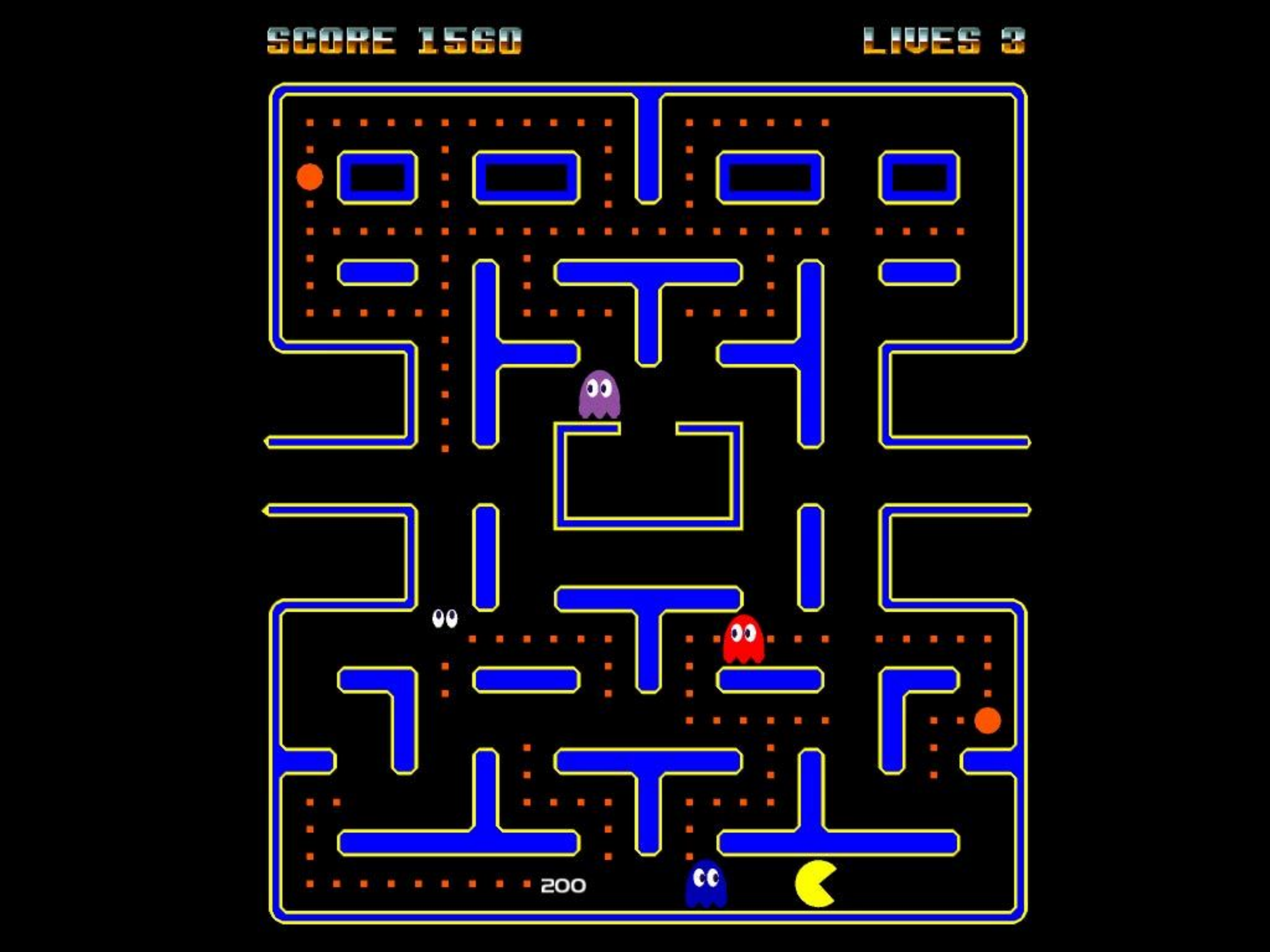}}
    \caption{Reinforcement learning applied to the Pacman Atari game.} 
  \label{fig::pacman} 
  \end{center} 
\end{figure}

Another technique that is gaining in popularity among computer scientists is reinforcement learning. In reinforcement learning, the objective is to learn an optimal policy that maximizes a reward based on a value function learned across a set of actions, or decisions. Each of the actions lead to a particular state and, consequently, to a particular reward. For instance, in a game emulator scenario, the optimal policy is learned through a replay experience under which the algorithm learns to play the optimal set of actions to maximize the overall score, as illustrated in Figure \ref{fig::pacman} with the Pacman Atari game. 
Relying on the Pacman game example, the goal is to eat all the bullets while avoiding the little monsters. With a reinforcement learning approach, the algorithm learns by itself how to play based on the maximization of the reward score, reaching human level performance. Consequently, the retail banks have now at their disposal a new field of science that they can use to face their upcoming challenges. 
\\

Throughout this chapter, we have established various challenges facing the banking industry, and especially, the retail banking industry. Furthermore, we shed some light on artificial intelligence and machine learning with explanation of basic concepts in order to demystify these concepts. 
The objective of this work is to propose innovative and novel solutions to overcome the technical advances that caused a shift in the demographics of the population impacting the banking industry. Although the problem is large and complex, we concentrate our effort to bring research innovation and open research to retail banking. We rely on the extensive information contained within the bank’s data sets in the context of evolving regulation and changing society. More precisely, we explore how to bring enriched clients' experience given limited resources, new habits and new regulations in a time of massive digitalization. Therefore, we chose to structure our approach around the following research problems:

\begin{itemize}
    \item Accurate and efficient extension of the population size under a strong European banking regulatory framework. Currently, the innovation of the retail banking services and offers is impacted by limited information for some real-world scenarios. For instance, the range of amount for car loan applications is very often incomplete and sparse, influencing the expected risk evaluation. Therefore, there is a need to extend the historical observation samples. By relying on adversarial models and neural networks, it is possible to artificially increase the number of historical samples while keeping the crucial patterns related to existing car loan applications. 
    
    \item Personalized recommendation through dimension reduction to better categorize clients' needs and clients' wills. The data sets related to the clients’ personal and financial information are large and sparse. Additionally, the banks are losing their relation with their clients because of the decrease of the visits in agencies. Consequently, the banks have less information about the center of interest of their clients. Therefore, the aim here is to propose financial products that fit the needs and the interest of a client's group while trying to compress the original information to an interpretable size. Typically, the families having young children might be interested to open a savings account for the children’s university studies while a young graduate might not be interested. The approach should highlight the difference of interests by categorizing the population while being humanly understandable.
    
    \item Learning through aggregated customer behavior with the determination of the optimal policy for money management with reinforcement learning. There are no institutions that can capitalize as much as the banks on the habits of the spendings of the different groups of population depending on the age, the personal situation or the revenues. Using a model-free approach based on reinforcement learning, it is possible to determine the optimal policy of money management for a category of population. It proposes to answer innovatively to traditional questions such as the limit of a loan amount or the monthly limit on the credit card, while offering complete transparency and being free of any human-bias. 
\end{itemize}

In Chapter 2, we present PHom-GeM, Persistent Homology for Generative Models \cite{charlier2019phomwae,charlier2019phomgem}, in the context of financial retail banking transactions. Different generative adversarial models, including Generative Adversarial Networks (GAN), Wasserstein Auto-Encoders (WAE) or Variational Auto-Encoders (VAE), are used with the persistent homology. The latter is a field of mathematics capable to reflect the density of the data points contained in a manifold and to evaluate the shape of the data manifold based on the number of holes. Our proposed method is used to generate synthetic transactions of high quality to increase the number of samples of existing transactions. Such approach is particularly interesting to boost the accuracy of the risk evaluation or of the recommendation predictions. Following the generation of financial transactions, we then propose, in Chapter 3, to perform recommendations of financial products depending on the clients’ categorizations. We reduce the dimension of large data sets through an accurate tensor resolution scheme, VecHGrad \cite{charlier2019vechgrad}, allowing to categorize a population through variables that cannot be observed, known as latent variables. Predictions are performed using neural networks on the latent categories of population for two scenarios. We first evaluate the next financial actions of the clients to help the banks to adopt a dynamic approach for the proposal of new products fitting the clients’ needs \cite{charlier2019sparse}. We then determine the future mobile banking authentication. It enables the bank to estimate which clients are more finance-oriented \cite{charlier2018user}. In Chapter 4, we discuss a retail banking methodology to establish an open, transparent and explainable decision making process. It targets the decisions made during financial product subscriptions such as the limit on a new credit card or the amount of money granted for a loan. This process should also be highly optimized to reduce the risk of the bank and to maximize the client’s satisfaction. Using a model-free reinforcement learning approach in the context of financial transactions \cite{charlier2019mqlv}, we are able to evaluate the optimal policy for money management. It enables to answer the typical question of how and where to fix the amount limit based on a complete transparent approach, free of any-model bias, depending exclusively and solely on the clients' aggregated information. In Chapter 5, we resume the accomplishments of this thesis and we address the main contributions.
\\
\chapter[PHom-GeM: Persistent Homology and Generative Models]{PHom-GeM: Persistent Homology and Generative Models for Retail Banking Transactions}

Generative models, including Generative Adversarial Networks (GANs) \cite{goodfellow2014generative} and Auto-Encoders (AE) \cite{goodfellow2016deep}, have become a cornerstone technique to generate synthetic samples. Generative models learn the data points distribution using unsupervised learning to generate new data points with slight variations. The GANs are among the most popular neural network models to generate adversarial data. A GAN is composed of two parts: a generator that produces synthetic data and a discriminator that discriminates between the generator's output and the true data. The AE are a very popular neural network architecture for adversarial samples generation and for features extraction through order reduction. They are composed of two parts: the encoder which maps the model distribution to a latent manifold and the decoder which maps the latent manifold to a reconstructed distribution. However, AE are known to provoke chaotically scattered data distribution in the latent manifold resulting in an incomplete reconstructed distribution. Furthermore, GANs also tend to provoke chaotically scattered reconstructed distribution during their training. Consequently, generative models can originate incomplete reconstructed distributions and incomplete generated adversarial distributions. Current distance measures fail to address this problem because they are not able to acknowledge the shape of the reconstructed data manifold, i.e. its topological features, and the scale at which the manifold should be analyzed. We propose Persistent Homology \cite{chazal2017tda} for Generative Models, PHom-GeM, a new methodology to assess and measure the data distribution of a generative model. PHom-GeM minimizes an objective minimization function between the true and the reconstructed distributions and uses persistent homology, the study of the topological features of a space at different spatial resolutions, to compare the nature of the true and the generated distributions. The potential of persistent homology for generative models is highlighted in two related experiments. First, we highlight the AE partial reconstructed distribution provoked by the chaotically scattered data distribution of the latent space. Then, PHom-GEM is applied on adversarial samples synthetically generated by AE and GAN generative models. Both experiments are conducted on a real-world data set particularly challenging for traditional distance measures and generative neural network models. PHom-GeM is the first methodology to propose a topological distance measure, the bottleneck distance, for generative models used to compare adversarial samples of high quality in the context of credit card transactions.  \\

\section{Motivation} \label{sec::intro} 
The field of generative models has evolved significantly for the past few years thanks to unsupervised learning and adversarial networks publications. A generative model is capable to learn and approximate any type of data distribution, thanks to neural networks, with the aim to generate new data points with small variations \cite{goodfellow2016deep}. These models are particularly useful to increase the size of the original data set or to overcome missing data points. In \cite{goodfellow2014generative}, Goodfelow et al. introduced one of the most significant generative neural networks called Generative Adversarial Network (GAN). It is a class of generative models that plays a competitive game between two networks in which the generator network must compete against an adversary according to a game theoretic scenario \cite{goodfellow2016deep}. The generator network produces samples from a noise distribution and its adversary, the discriminator network, tries to distinguish real samples from generated samples, respectively samples inherited from the training data and samples produced by the generator.  \\

Meanwhile, feature extraction through dimension reduction techniques were initially driven by linear algebra with second order matrix decompositions such as the Singular Value Decomposition (SVD) \cite{golub1989cf} and, ultimately, with tensor decompositions \cite{carroll1970analysis,kolda2009tensor}, a higher order analogue of the matrix decompositions. However, due to linear algebra limitations \cite{paatero1997weighted,bader2007temporal}, several architectures for dimension reductions based on neural networks have been proposed. Dense Auto-Encoders (AE) \cite{lecun1989backpropagation,hinton1994autoencoders} are one of the most well established approaches. An AE is a neural network trained to copy its input manifold to its output manifold through a hidden layer. The encoder function sends the input space to the hidden space and the decoder function brings back the hidden space to the input space. More recently, the Variational Auto-Encoders (VAE) presented by Kingma et al. in \cite{kingma2013auto} constitute a well-known approach but they might generate poor target distribution because of the KL divergence.  \\

Nonetheless, as explained in \cite{bengio2013generalized}, the points of the hidden space are chaotically scattered for most of the encoders, including the popular VAE. Even after proper training, groups of points of various sizes gather and cluster by themselves randomly in the hidden layer. Some features are therefore missing in the reconstructed distribution $G(Z),Z\in \mathcal{Z}$. Moreover, GANs are subject to a mode collapse during their training for which the competitive game between the discriminator and the generator networks is unfair, leading to convergence issue. Therefore, their training have been known to be very complex and, consequently, limiting their usage especially on large real world data sets. By applying some of the Optimal Transport (OT) concepts gathered in \cite{villani2003topics} and noticeably, the Wasserstein distance, Arjovsky et al. introduced the Wasserstein GAN (WGAN) in \cite{arjovsky2017wasserstein} to avoid the mode collapse of GANs and hazardous reconstructed distribution. Gulrajani et al. further optimized the concept in \cite{gulrajani2017improved} by proposing a Gradient Penalty to Wasserstein GAN (GP-WGAN) capable to generate adversarial samples of higher quality. Similarly, Tolstikhin et al. in \cite{tolstikhin2017wasserstein} applied the same OT concepts to AE and, therefore, introduced Wasserstein AE (WAE), a new type of AE generative model, that avoids the use of the KL divergence.  \\

\begin{figure}[t!]
  \begin{center}
  \vspace{-.25cm}
  \includegraphics[scale=0.7]{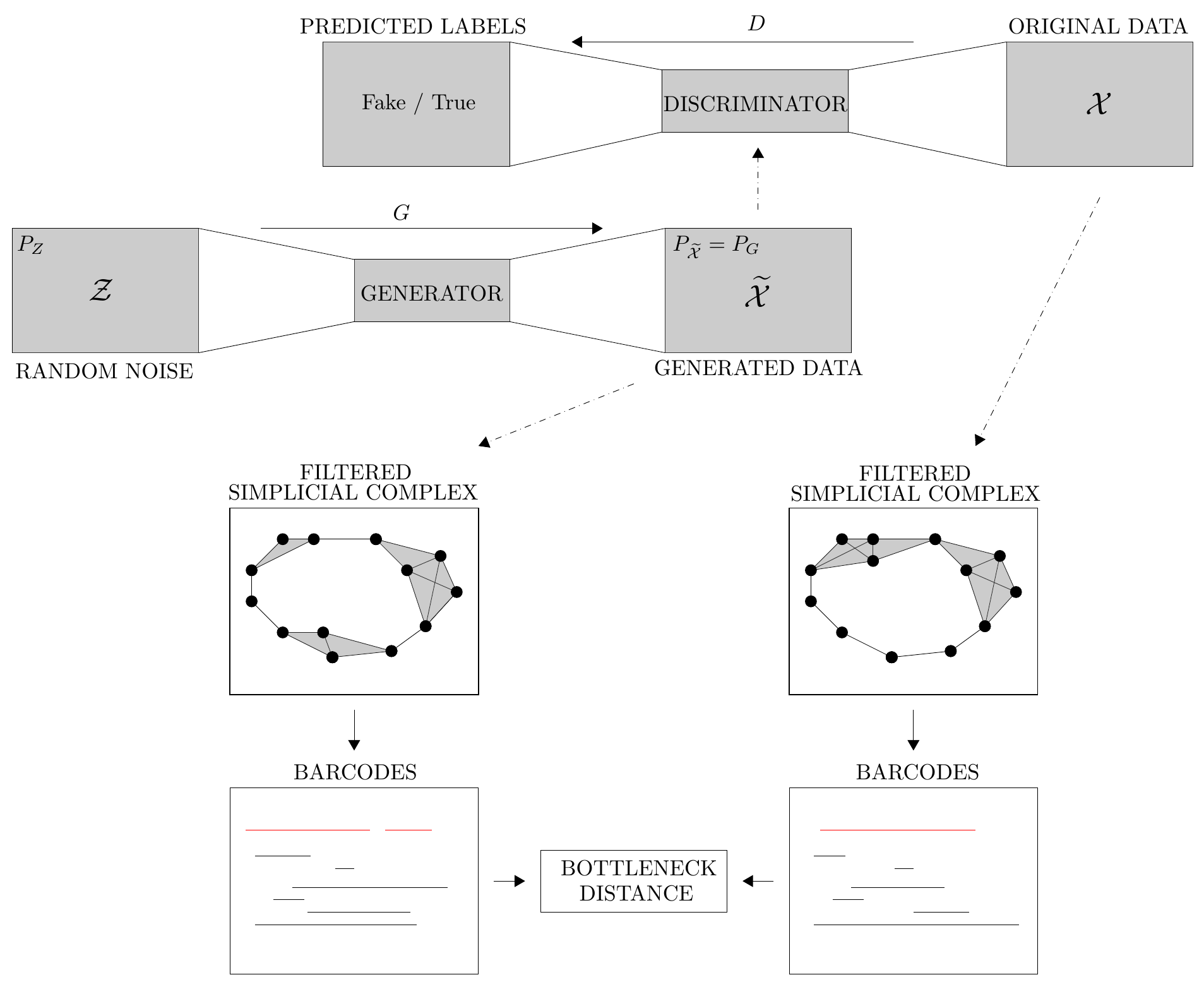}
  \captionsetup{belowskip=-0.5cm}
    \caption[PHom-GeM applied to Generative Adversarial Network]{In PHom-GeM applied to GAN, the generative model $G$ generates fake samples $\widetilde{X} \in \mathcal{\widetilde{X}}$ based on the samples $Z\in\mathcal{Z}$ from the prior random distribution $P_Z$. Then, the discriminator model $D$ tries to differentiate the fake samples $\widetilde{X}$ from the true samples $X\in\mathcal{X}$. The original manifold $\mathcal{X}$ and the generated manifold $\mathcal{\widetilde{X}}$ are transformed independently into metric space sets to obtain filtered simplicial complex. It leads to the description of topological features, summarized by the barcodes, to compare the respective topological representation of the true data distribution $P_X$ and the generative model distribution $P_G$.} \label{fig::phomgen} 
  \end{center} 
\end{figure}

\begin{figure}[t!]
	\begin{center}
	\vspace{-0.25cm}
	\includegraphics[scale=0.7]{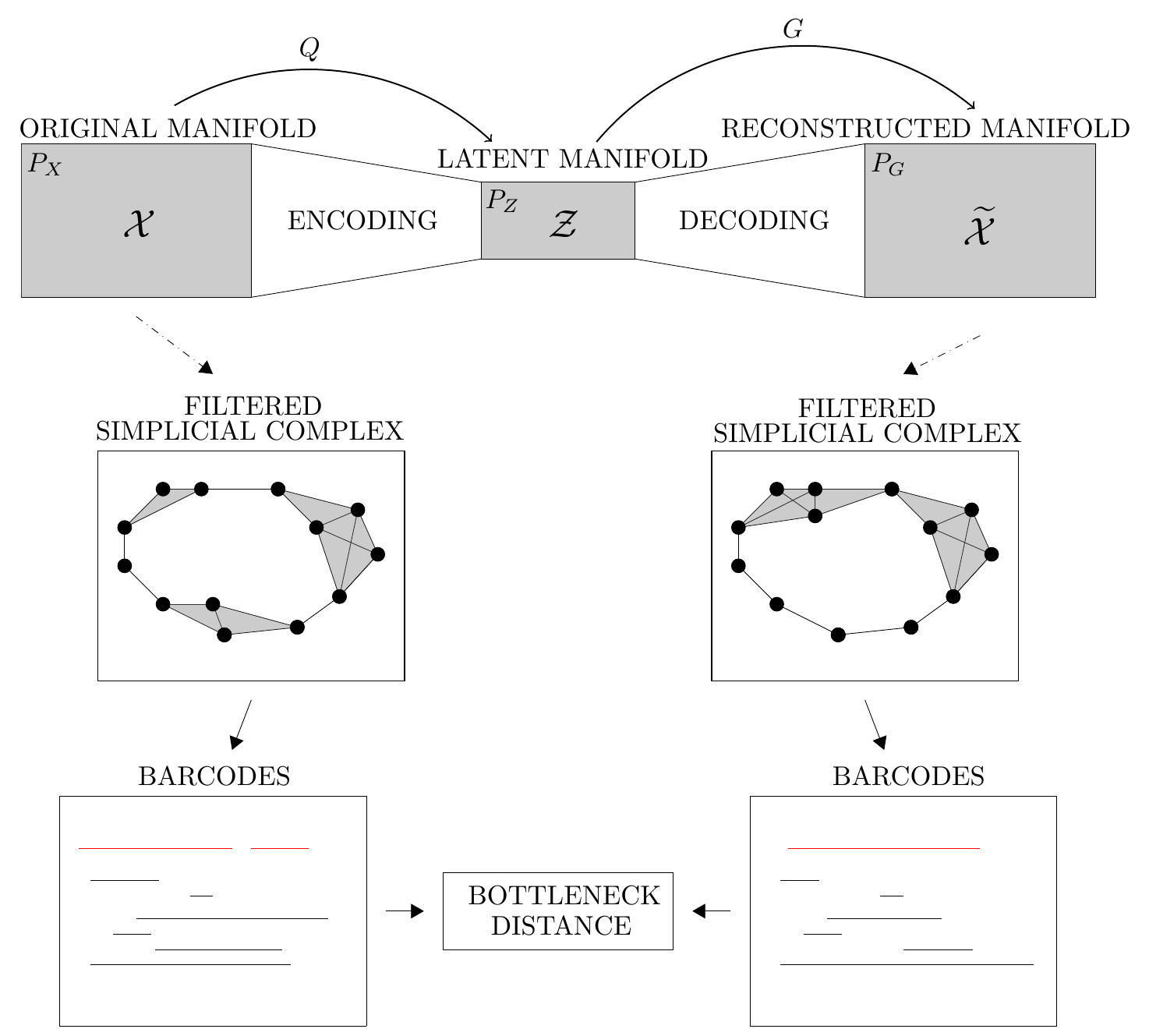}
    \caption[PHom-GeM applied to Auto-Encoders]{In PHom-GeM for AE, the generative model $G$, the decoder, generates fake samples $\widetilde{X} \in \mathcal{\widetilde{X}}$ based on the samples $Z\in\mathcal{Z}$ from a prior random distribution $P_Z$. Afterward, the original manifold $\mathcal{X}$ and the generated manifold $\mathcal{\widetilde{X}}$ are both transformed independently into metric space sets to obtain filtered simplicial complex. As for PHom-GeM applied to GAN, it leads to the description of the topological features, summarized by the barcodes, to compare the respective topological representation of the true data distribution $P_X$ and the generative model distribution $P_G$.}
    \label{fig::autoencoder}
	\end{center}
\end{figure}

Nonetheless, the description of the distribution $P_G$ of the generative models, which involves the description of the generated scattered data points \cite{bengio2013generalized} based on the distribution $P_X$ of the original manifold $\mathcal{X}$, is very difficult using traditional distance measures, such as the Euclidean distance or the Fr\'echet Inception Distance \cite{tolstikhin2017wasserstein}. We highlight the distribution and the manifold notations in Figure \ref{fig::phomgen} for GAN and in Figure \ref{fig::autoencoder} for AE. Effectively, traditional distance measures are not able to acknowledge the shapes of the data manifolds and the scale at which the manifolds should be analyzed. However, persistent homology \cite{Edelsbrunner2002tps,zomorodian2005computing} is specifically designed to highlight the topological features of the data \cite{chazal2017tda}. Therefore, building upon the persistent homology, the Wasserstein distance \cite{bousquet2017optimal} and the generative models \cite{tolstikhin2017wasserstein}, our main contribution is to propose qualitative and quantitative ways to evaluate the scattered generated distributions and the performance of the generative models. \\ 

In this work, we describe the persistent homology features of the generated model $G$ while minimizing the OT function $W_c (P_X, P_G)$ for a squared cost $c(x,y)=||x-y ||_2^2$ where $P_X$ is the model distribution of the data contained in the manifold $\mathcal{X}$, and $P_G$ the distribution of the generative model capable of generating adversarial samples. Our contributions are summarized below: \\

\begin{itemize} 
  \item A persistent homology procedure for generative models, including GP-WGAN, WGAN, WAE and VAE, which we call PHom-GeM to highlight the topological properties of the generated distributions of the data for different spatial resolutions. The objective is a persistent homology description of the generated data distribution $P_G$ following the generative model $G$.
  
  \item The use of PHom-GeM to underline the scattered latent distribution provoked by VAE in comparison to WAE on a real-word data set. We highlight the VAE's hidden layer scattered distribution using the persistent homology, confirming the original statement of VAE's scattered distribution introduced in \cite{goodfellow2016deep}.
  
  \item The application of PHom-GeM for the description of the topological properties of the encoded and decoded distribution of the data for different spatial resolutions. The objective is twofold: a persistent homology description of the encoded latent space $Q_z := \mathbb{E}_{P_X}[Q(Z|X)]$, and a persistent homology description of the decoded latent space following the generative model $P_G(X|Z)$. The latter allows to assess the information loss of the encoding-decoding process. 
  
  \item A distance measure for persistence diagrams, the bottleneck distance, applied to generative models to compare quantitatively the true and the target distributions on any data set. We measure the shortest distance for which there exists a perfect matching between the points of the two persistence diagrams. A persistence diagram is a stable summary representation of the topological features of simplicial complex, a collection of vertices, associated to the data set.
  
  \item Finally, we propose the first application of algebraic topology and generative models on a public data set containing credit card transactions, particularly challenging for this type of models and traditional distance measures while being of high interest for the retail banking industry. 
\end{itemize} 

The chapter is structured as follows. We discuss the related work in Section \ref{sec::relwork}. In Section \ref{sec::propmethod}, we review the four main generative model formulations, WGAN, GP-WGAN, WAE and VAE, derived by Arjovsky et al. in \cite{arjovsky2017wasserstein}, Gulrajani et al. in \cite{gulrajani2017improved}, Tolstikhin et al. in \cite{tolstikhin2017wasserstein} and Kingma et al. in \cite{kingma2013auto}, respectively. By using the persistent homology, we are able to compare the topological properties of the original distribution $P_X$ and the generated distribution $P_G$. We additionally compare the latent distribution $P_Z$ of the AE hidden space. We highlight experimental results in Section \ref{sec::exp}. We finally conclude in Section \ref{sec::ccl} by addressing promising directions for future work.  \\

\section{Related Work} \label{sec::relwork} 

\textbf{Literature on Persistent Homology and Topology} 
A major trend in modern data analysis is to consider that data has a shape, and more precisely, a topological structure. Topological Data Analysis (TDA) is a set of tools relying on computational algebraic topology to obtain precise information about the data structure. Two of the most important techniques are the persistent homology and the mapper algorithm \cite{chazal2017tda}.  \\ 

Data sets are usually represented as points cloud of Euclidean spaces. The shape of a data set hence depends of the scale at which it is observed. Instead of trying to find an optimal scale, the persistent homology, a method of TDA, studies the changes of the topological features (number of connected components, number of holes, ...) of a data set depending of the scale. The foundations of the persistent homology have been established in the early 2000s in \cite{Edelsbrunner2002tps} and \cite{Zomorodian2005comp}. They provide a computable algebraic invariant of filtered topological spaces (nested sequences of topological spaces which encode how the scale changes) called persistence module. This module can be decomposed into a family of intervals called persistence diagram or barcodes. This family records how the topology of the space is changing when going through the filtration \cite{ghrist2008barcodes}. The space of barcodes is measurable through the bottleneck distance. The space of persistence module is moreover endowed with a metric and under a mild assumption these two spaces are isometric \cite{Chazal2017struct}. Additionally, the \textit{Mapper} algorithm first appeared in \cite{singh2007topological}. It is a visualization algorithm which aims to produce a low dimensional representation of high-dimensional data sets in form of a graph, and therefore, capturing the topological features of the points cloud.  \\ 

Meanwhile, efficient and fast algorithms have emerged to compute the persistent homology \cite{Edelsbrunner2002tps},\cite{Zomorodian2005comp} as well as to construct filtered topological spaces using, for instance, the Vietoris-Rips complex \cite{zomorodian2010fast}. It has consequently found numerous successful applications. For instance, Nicolau et al. in \cite{nicolau2011topology} detected subgroups of cancers with unique mutational profile. In \cite{lum2013extracting}, it has been shown that computational topology could be used in medicine, social sciences or sport analysis. Lately, Bendich et al. improved statistical analyses of brain arteries of a population \cite{bendich2016persistent} while Xia et al. were capable of extracting molecular topological fingerprints for proteins in biological science \cite{xia2014persistent}.  \\ 

\textbf{Literature on Optimal Transport and Generative Models} 
The field of unsupervised learning and generative models has evolved significantly for the past few years \cite{goodfellow2016deep}. One of the first most popular generative models is auto-encoders. Nonetheless, as outlined by Bengio et al. in \cite{bengio2013generalized}, the points in the encoded hidden manifold $\mathcal{Z}$ for the majority of the encoders are chaotically scattered. Some features are missing in the reconstructed distribution $G(Z) \in \mathcal{\widetilde{X}}, Z \in \mathcal{Z}$. Thus, sampling data points for the reconstruction with traditional AE is difficult. The added constraint of Variational Auto-Encoder (VAE) in \cite{kingma2013auto} by the mean of a KL divergence, composed of a reconstruction cost and a regularization term, 
allows to decrease the impact of the chaotic scattered distribution $P_Z$ of $\mathcal{Z}$.  \\ 

Concurrently to the emergence of AE, Goodfellow et al. introduced Generative Adversarial Network (GAN) in \cite{goodfellow2014generative} in which two models are simultaneously trained. A generative model, the generator, captures the data distribution and a discriminative model, the discriminator, estimates the probability that a sample comes from the training data rather than the generator \cite{goodfellow2016deep}. However, in this game theoretic scenario, the training is complex because of the competition between the generator and the discriminator, frequently leading to a mode collapse. The concepts of optimal transport have been brought up to light following the recent work of Villani in \cite{villani2003topics}. As a solution, optimal transport \cite{villani2003topics} was therefore applied to GAN in \cite{arjovsky2017wasserstein} with the Wasserstein distance introducing consequently the Wasserstein GAN (WGAN). By adding a Gradient Penalty (GP) to the Wasserstein distance, Gulrajani et al. in \cite{gulrajani2017improved} proposed GP-WGAN, a new training for GANs capable of avoiding efficiently the mode collapse. As described in \cite{chizat2015unbalanced,liero2018optimal} in the context of unbalanced optimal transport, Tolstikhin et al. also applied these concepts to AE in \cite{tolstikhin2017wasserstein}. They proposed to add one extra divergence to the objective minimization function in the context of generative modeling leading to Wasserstein Auto-Encoders (WAE).  \\ 

Meanwhile, serious efforts have been proposed to increase the efficiency and the accuracy of the weights optimization of neural networks including Stochastic Gradient Descent (SGD) in \cite{robbins1985stochastic} applied to optimal transport for large-scale experiments \cite{genevay2016stochastic} and to machine learning \cite{bottou2010large}. Other gradient descent updates, such as Adagrad \cite{duchi2011adaptive} or Adam  \cite{kingma2014adam}, have been introduced as alternatives to the SGD. It has led ultimately to the popular RMSProp \cite{hinton2012rmsprop}, widely used for complex training of AE and GANs. \\ 

In this chapter, using the persistent homology and the bottleneck distance, we propose qualitative and quantitative ways to evaluate the performance (i) of the generated distribution of GP-WGAN, WGAN, WAE and VAE generative models and (ii) of the encoding decoding process of WAE and VAE. We build upon the work of unsupervised learning and unbalanced optimal transport with the persistent homology. We show that the barcodes, inherited from the persistence diagrams, are capable of representing the adversarial manifold $\mathcal{\widetilde{X}}$ generated by the generative models while the bottleneck distance allows us to compare quantitatively the topological features between the samples $G(Z) \in \mathcal{\widetilde{X}}$ of the generated data distribution $P_G$ and the samples $X \in \mathcal{X}$ of the true data distribution $P_X$.  \\

\section{Proposed Method} \label{sec::propmethod} 
Our method computes the persistent homology of both the true manifold $X\in\mathcal{X}$ and the generated manifold $\widetilde{X} \in \mathcal{\widetilde{X}}$ following the generative model $G$ based on the minimization of the optimal transport cost $W_c(P_X, P_G)$. In the resulting topological problem, the points of the manifolds are transformed to a metric space set for which a Vietoris-Rips simplicial complex filtration is applied (see definition 2). PHom-GeM achieves simultaneously two main goals: it computes the birth-death of the pairing generators of the iterated inclusions while measuring the bottleneck distance between the persistence diagrams of the manifolds of the generative models. 

\subsection{Preliminaries and Notations} 
We follow the notations 
introduced in \cite{gulrajani2017improved} and \cite{tolstikhin2017wasserstein}. Sets and manifolds are denoted by calligraphic letters such as $\mathcal{X}$, random variables by uppercase letters $X$, and their values by lowercase letters $x$. Probability distributions are denoted by uppercase letters $P(X)$ and their corresponding densities by lowercase letters $p(x)$. 
  
\subsection{Optimal Transport and Dual Formulation} 
Following the description of the optimal transport problem \cite{villani2003topics}, the study of the optimal allocation of resources, and relying on the Kantorovich-Rubinstein duality, the Wasserstein distance is computed as

\begin{equation} 
  W_c(P_X, P_G) = \sup_{f \in \mathcal{F}_L} \mathbb{E}_{X\sim P_X}[f(X)] - \mathbb{E}_{Y\sim P_G}[f(Y)] \quad ,
\end{equation} 

where $(\mathcal{X}, d)$ is a metric space, $\mathcal{P}(X\sim P_X, Y\sim P_G)$ is a set of all joint distributions $(X, Y)$ with marginals $P_X$ and $P_G$ respectively and $\mathcal{F}_L$ is the class of all bounded 1-Lipschitz functions on $(\mathcal{X}, d)$. \\ 

\subsection{Wasserstein GAN (WGAN)} 
The objective of the Wasserstein minimization function applied to GAN is twofold: (i) achieve appropriate calibration of the generator and the discriminator networks thanks to the transporting mass cost function while (ii) avoiding the mode collapse of standard GANs \cite{goodfellow2016deep}. By using the notations and the concept introduced in \cite{arjovsky2017wasserstein}, the WGAN objective loss function is defined by 

\begin{equation} \label{eq::WGAN} 
W_c(\mathbb{P}_r, \mathbb{P}_\theta) = \underset{w \in \mathcal{W}}{\max} \: \mathbb{E}_{x \sim \mathbb{P}_r} [f_w(x)] - \mathbb{E}_{x \sim p(z)} [f_w (g_\theta (z))] \quad ,
\end{equation} 

where $\mathbb{P}_r$ denotes the real data distribution, $\mathbb{P}_\theta$ the generative distribution of the function  $g : \mathcal{Z} \times \mathbb{R}^d \rightarrow \mathcal{X}$, denoted $g_\theta (Z)$, with $Z$ a random variable (e.g Gaussian) over another space $\mathcal{Z}$ , $\left\lbrace f_w \right\rbrace_{w \in \mathcal{W}}$ the parameterized family of functions, all K-Lipschitz for some K, and $w \in \mathcal{W}$ are the weights $w$ contained in the compact space $\mathcal{W}$.  \\

\vspace{-.35cm}
\subsection{Gradient Penalty Wasserstein GAN (GP-WGAN)}
The Gradient Penalty Wasserstein GAN (GP-WGAN) \cite{gulrajani2017improved} proposes to solve partial issues that have been left opened in \cite{arjovsky2017wasserstein} by adding a gradient penalty to the loss function of WGAN. The advantages of using a gradient penalty are twofolds: (i) reducing the WGAN's training error to improve the quality of the generated adversarial samples by backpropagating gradient information while (ii) avoiding clipping the weights during the WGAN's training. Therefore, the GP-WGAN objective loss function with gradient penalty is expressed such that  

\begin{equation} \label{eq::DWGAN} 
\begin{aligned} 
L = &\underset{\widetilde{X}\sim P_G}{\mathbb{E}}[f(\widetilde{X})] - \underset{X\sim P_X}{\mathbb{E}}[f(X)] + \lambda \underset{\widehat{X}\sim P_{\widehat{X}}}{\mathbb{E}} [(||\nabla_{\widehat{X}} f(\widehat{X})||_2 - 1)^2]  \quad ,
\end{aligned} 
\end{equation} 

where $f$ is the set of 1-Lipschitz functions on $(\mathcal{X}, d)$, $P_X$ the original data distribution, $P_G$ the generative model distribution implicitly defined by $\widetilde{X}=G(Z), Z\sim p(Z)$. The input $Z$ to the generator is sampled from a noise distribution such as a uniform distribution. $P_{\widehat{X}}$ defines the uniform sampling along straight lines between pairs of points sampled from the data distribution $P_X$ and the generative distribution $P_G$. A penalty on the gradient norm is enforced for random samples $\widehat{X}\sim P_{\widehat{X}}$. For further details, we refer to \cite{arjovsky2017wasserstein} and \cite{gulrajani2017improved}. \\ 

\vspace{-.35cm}
\subsection{Wasserstein Auto-Encoders} 
As described in \cite{tolstikhin2017wasserstein}, the WAE objective function is expressed such that  

\vspace{-.5cm}
\begin{equation} \label{eq::DWAE} 
  \begin{split} 
    D_{\text{WAE}}(P_X, P_G) := & \underset{Q(Z|X)\in\mathcal{Q}}{\inf} \mathbb{E}_{P_X} \mathbb{E}_{Q(Z|X)} [c(X, G(Z))] + \lambda \mathcal{D}_Z(Q_Z, P_Z) \quad ,
  \end{split} 
\end{equation} 

where $c(X, G(Z)): \mathcal{X}\times \mathcal{X}\rightarrow\mathcal{R}_+$ is any measurable cost function. In our experiments, we use a square cost function $c(x,y)=||x-y||_2^2$ for data points $x,y \in \mathcal{X}$. $G(Z)$ denotes the sending of $Z$ to $X$ for a given map $G:\mathcal{Z}\rightarrow \mathcal{X}$. $Q$, and $G$, are any nonparametric set of probabilistic encoders, and decoders respectively. \\ 

We use the Maximum Mean Discrepancy (MMD) for the penalty $\mathcal{D}_Z(Q_Z, P_Z)$ for a positive-definite reproducing kernel $k:\mathcal{Z}\times\mathcal{Z}\rightarrow\mathcal{R}$ 

\begin{equation} 
\setlength{\abovedisplayskip}{5pt}
\begin{split} 
  \mathcal{D}_Z(P_Z, Q_Z) & := \: \text{MMD}_k(P_Z, Q_Z) \\
  \mathcal{D}_Z(P_Z, Q_Z) & = \: || \int_\mathcal{Z}k(z, .) dP_Z(z) - \int_\mathcal{Z}k(z, .) dQ_Z(z) ||_{\mathcal{H}_k} \quad ,
\end{split} 
\end{equation} 

where $\mathcal{H}_k$ is the reproducing kernel Hilbert space of real-valued functions mapping on $\mathcal{Z}$. For details on the MMD implementation, we refer to \cite{tolstikhin2017wasserstein}.  \\ 

\subsection{Variational Auto-Encoders}
Variational Auto-Encoder (VAE) \cite{kingma2013auto} were designed to combine approximate inference and probabilistic models for efficient learning of latent variables within a data set. In the VAE framework, the input $x$ is mapped probabilistically to the latent variable $z$ by the encoder $q_\phi$ and the latent variable $z$ to $x$ by the decoder $p_\theta$ such as $x \overset{q_\phi (x|z)}{\longrightarrow} z \overset{p_\theta (z|x)}{\longrightarrow} x$. The parameters $\phi$ and $\theta$ are used to parameterize $q(x|z)$ and $p(z|x)$ respectively. The model has to learn the generative process $p(x|z)$ for the optimization problem $\underset{\theta}{\max} \int_z p(z) p_\theta (x|z)$. Because of the assumption that the probability distribution are all Gaussian, the VAE loss function is defined such that 

\begin{equation} \label{eq::vaeloss}
\begin{split}
 \mathcal{L}(\theta, \phi, \lbrace x^i \rbrace_{i=1}^M ) = & \: \dfrac{N}{M} \sum_{i=1}^M \mathcal{L}(\theta, \phi, x^{(i)})  \\
 \mathcal{L}(\theta, \phi, \lbrace x^i \rbrace_{i=1}^M ) = & \: \dfrac{N}{M} \sum_{i=1}^M \big\{
 \underbrace{\dfrac{1}{L} \sum_{l=1}^L \log p_\theta (x^i, z^{(l)})}_\text{reconstruction loss} \\
 & \quad\quad\quad\quad + 
 \underbrace{\dfrac{1}{2} \sum_{d=1}^D \left( 1 + \log_{\phi, d}(x^i) - \mu_{\phi, d}^2 (x^i) - \sigma_{\phi, d}^2 (x^i) \right) }_\text{KL divergence} \big\}
\end{split} \quad ,
\end{equation}

where the vector $z$ is determined by $z=g_\phi(\epsilon, x)= \mu_\phi (x) + \epsilon \odot \sigma_\phi(x)$ such that $z$ will have the distribution $q_\phi(z|x) \sim \mathcal{N}(z, \mu_\phi(x), \sigma_\phi(x))$ with $\odot$ the elementwise multiplication. The terms $\mu_\phi$ and $\sigma_\phi$ denote the mapping from $x$ to the mean and standard deviation vectors, respectively. Finally, $N$ is the number of data points contained in the data set and $M$ is the number of data points contained in the random sample. 
\\

\subsection{Persistence Diagram and Vietoris-Rips Complex} 
We explain the construction of the persistence module associated to a sample of a fixed distribution on a space. First, two manifold distributions are sampled from the generative models' training. Then, we construct the persistence modules associated to each sample of the points manifolds. We refer to the first subsection of section \ref{sec::relwork} for pointed reference on the persistent homology.  \\ 

We first associate to our points manifold $\mathcal{C} \subset \mathbb{R}^n$, considered as a finite metric space, a sequence of simplicial complexes. For that aim, we use Vietoris-Rips complex. \\ 

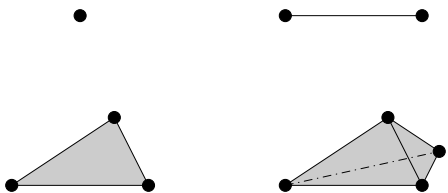
\begin{figure}[b] 
  \centering 
  \definecolor{dtsfsf}{rgb}{0.8274509803921568,0.1843137254901961,0.1843137254901961}
\definecolor{rvwvcq}{rgb}{0.08235294117647059,0.396078431372549,0.7529411764705882}

\begin{tikzpicture}[scale=0.9, transform shape, line cap=round,line join=round,>=triangle 45,x=1cm,y=1cm]

\draw [fill=black] (0,2.5) circle (2.5pt);

\draw [fill=black] (3,2.5) circle (2.5pt);
\draw [fill=black] (5,2.5) circle (2.5pt);
\draw [line width=0.25pt] (3,2.5) -- (5,2.5);

\fill [gray, opacity=.4] (-1,0) -- (1,0) -- (0.5,1) -- (-1,0);
\draw [fill=black] (-1,0) circle (2.5pt);
\draw [fill=black] (1,0) circle (2.5pt);
\draw [fill=black] (0.5,1) circle (2.5pt);
\draw [line width=0.25pt] (-1,0) -- (1,0);
\draw [line width=0.25pt] (1,0) -- (0.5,1);
\draw [line width=0.25pt] (-1,0) -- (0.5,1);

\fill [gray, opacity=.4] (3,0) -- (5,0) -- (4.5,1) -- (3,0);
\fill [gray, opacity=.4] (5,0) -- (4.5,1) -- (5.25,0.5) -- (5,0);
\draw [fill=black] (3,0) circle (2.5pt);
\draw [fill=black] (5,0) circle (2.5pt);
\draw [fill=black] (4.5,1) circle (2.5pt);
\draw [fill=black] (5.25,0.5) circle (2.5pt);
\draw [line width=0.25pt] (3,0) -- (5,0);
\draw [line width=0.25pt] (5,0) -- (4.5,1);
\draw [line width=0.25pt] (3,0) -- (4.5,1);
\draw [line width=0.25pt] (4.5,1) -- (5.25,0.5);
\draw [line width=0.25pt] (5,0) -- (5.25,0.5);
\draw [line width=0.25pt, dashdotted] (3,0) -- (5.25,0.5);

\end{tikzpicture} 
  \caption[Description of simplicial complex]{A simplical complex is a collection of numerous ``simplex" or simplices, where a 0-simplex is a point, a 1-simplex is a line segment, a 2-simplex is a triangle and a 3-simplex is a tetrahedron.} 
  \label{fig::simplex} 
\end{figure}

\textbf{Definition 1} Let $V=\lbrace 1, \cdots, |V|\rbrace$ be a set of vertices. A simplex $\sigma$ is a subset of vertices $\sigma \subseteq V$. A simplicial complex K on V is a collection of simplices $\lbrace \sigma \rbrace \:,\: \sigma \subseteq V$, such that $\tau \subseteq \sigma \in K \Rightarrow \tau \in K$. The dimension $n = |\sigma| - 1 $ of $\sigma$ is its number of elements minus 1. Simplicial complexes examples are represented in Figure \ref{fig::simplex}.  \\

\textbf{Definition 2} Let $(X,d)$ be a metric space. The Vietoris-Rips complex $\text{VR}(X, \epsilon)$ at scale $\epsilon$ associated to $X$ is the abstract simplicial complex whose vertex set is $X$, and where $\left\lbrace x_0, x_1,...,x_k\right\rbrace$ is a $k$-simplex if and only if $d(x_i, x_j ) \leq \epsilon$ for all $0\leq i, j\leq k$.  \\ 

We obtain an increasing sequence of Vietoris-Rips complex by considering the $\text{VR}(\mathcal{C}, \epsilon)$ for an increasing sequence $(\epsilon_i)_{1 \leq i \leq N}$ of value of the scale parameter $\epsilon$ 

\begin{equation}\label{diag:seqrips} 
\mathcal{K}_1 \xhookrightarrow{\iota} \mathcal{K}_2 \xhookrightarrow{\iota} \mathcal{K}_3 \xhookrightarrow{\iota} ... \xhookrightarrow{\iota} \mathcal{K}_{N-1} \xhookrightarrow{\iota} \mathcal{K}_N. 
\end{equation} 

Applying the \textit{k-th} singular homology functor $H_k(-,F)$ with coefficient in the field $F$ \cite{hatcher2002algebraic} to the equation \eqref{diag:seqrips}, we obtain a sequence of $F$-vector spaces, called the \textit{k-th} persistence module of $(\mathcal{K}_i)_{1 \leq i \leq N}$ 

\begin{multline} 
H_k(\mathcal{K}_1,F) \xrightarrow{t_1} H_k(\mathcal{K}_2,F) \xrightarrow{t_2} \cdots \xrightarrow{t_{N-2}} H_k(\mathcal{K}_{N-1},F) \xrightarrow{t_{N-1}} H_k(\mathcal{K}_N,F).
\end{multline} 

\textbf{Definition 3} $\forall ~i<j$, the \textit{(i,j)}-persistent $k$-homology group with coefficient in $F$ of $\mathcal{K}=(\mathcal{K}_i)_{1 \leq i\leq N}$ denoted $HP_k^{i \rightarrow j}(\mathcal{K},F)$ is defined to be the image of the homomorphism $t_{j-1} \circ \ldots \circ t_i : H_k(\mathcal{K}_i,F) \rightarrow H_k(\mathcal{K}_j,F)$.  \\ 

Using the interval decomposition theorem \cite{oudot2015Persistance}, we extract a finite family of intervals of $\mathbb{R}_+$ called persistence diagram. Each interval can be considered as a point in the set $D = \left\lbrace (x, y) \in \mathbb{R}_+^2 | x \leq y \right\rbrace$. Hence, we obtain a finite subset of the set $D$. This space of finite subsets is endowed with a matching distance called the bottleneck distance and defined as follow 

\begin{equation*} 
d_b(A,B)=\inf_{\phi : A^\prime \to B^\prime} \sup_{ x \in A^\prime}{\lVert x-\phi(x) \rVert}  \quad ,
\end{equation*} 

where $A^\prime= A \cup \Delta$, $B^\prime=B \cup \Delta$, $\Delta= \lbrace (x,y) \in \mathbb{R}^2_+ \vert x=y \rbrace$ and the $\inf$ is over all the bijections from $A^\prime$ to $B^\prime$.

\subsection{Practical Overview of Persistence Diagrams, Barcodes and Homology Groups}
Before describing our contribution PHom-GeM, we give a more practical overview of the concepts related to the filtration parameter, the barcodes, the persistence diagrams and the homology groups for the ease of understanding. Consequently, we emphasize the explanation of the key concepts without any formal formulation.  \\

\subsubsection{Filtration Parameter, Barcodes and Persistence Diagrams}
\begin{figure}[t!]
  \centering
  \begin{tikzpicture}[scale=0.5, transform shape, line cap=round,line join=round,>=triangle 45,x=1cm,y=1cm]
\draw [fill=black] (0,0) circle (2.5pt);
\draw [fill=black] (2.05,1.7) circle (2.5pt);
\draw [fill=black] (4,3) circle (2.5pt);
\draw [fill=black] (7,1.5) circle (2.5pt);
\draw [fill=black] (6,4.5) circle (2.5pt);
\draw [fill=black] (8,3.5) circle (2.5pt);
\draw [fill=black] (7,-1) circle (2.5pt);
\draw [fill=black] (4.5,-2) circle (2.5pt);
\draw [fill=black] (2.25,-1.5) circle (2.5pt);

\fill [gray, opacity=0.4] (0,0) circle (35pt);
\fill [gray, opacity=0.4] (2.05,1.7) circle (35pt);
\fill [gray, opacity=0.4] (4,3) circle (35pt);
\fill [gray, opacity=0.4] (7,1.5) circle (35pt);
\fill [gray, opacity=0.4] (6,4.5) circle (35pt);
\fill [gray, opacity=0.4] (8,3.5) circle (35pt);
\fill [gray, opacity=0.4] (7,-1) circle (35pt);
\fill [gray, opacity=0.4] (4.5,-2) circle (35pt);
\fill [gray, opacity=0.4] (2.25,-1.5) circle (35pt);

\draw [fill=black] (13.5,0) circle (2.5pt);
\draw [fill=black] (15.55,1.7) circle (2.5pt);
\draw [fill=black] (17.5,3) circle (2.5pt);
\draw [fill=black] (20.5,1.5) circle (2.5pt);
\draw [fill=black] (19.5,4.5) circle (2.5pt);
\draw [fill=black] (21.5,3.5) circle (2.5pt);
\draw [fill=black] (20.5,-1) circle (2.5pt);
\draw [fill=black] (18,-2) circle (2.5pt);
\draw [fill=black] (15.75,-1.5) circle (2.5pt);

\draw [line width = 0.5] (15.55,1.7) -- (17.5,3);
\draw [line width = 0.5] (19.5,4.5) -- (21.5,3.5);
\draw [line width = 0.5] (21.5,3.5) -- (20.5,1.5);
\draw [line width = 0.5] (18,-2) -- (15.75,-1.5);

\draw [gray, opacity=0.9, line width = 0.5] (0,1.25) -- (-4,1.25);
\draw [gray, opacity=0.9, line width = 0.5] (0,-1.25) -- (-4,-1.25);
\node (none) at (-4,0) {\LARGE $\varepsilon$};
\end{tikzpicture}  \\
  \begin{tikzpicture}[scale=0.5, transform shape, line cap=round,line join=round,>=triangle 45,x=1cm,y=1cm]
\draw [fill=black] (0,0) circle (2.5pt);
\draw [fill=black] (2.05,1.7) circle (2.5pt);
\draw [fill=black] (4,3) circle (2.5pt);
\draw [fill=black] (7,1.5) circle (2.5pt);
\draw [fill=black] (6,4.5) circle (2.5pt);
\draw [fill=black] (8,3.5) circle (2.5pt);
\draw [fill=black] (7,-1) circle (2.5pt);
\draw [fill=black] (4.5,-2) circle (2.5pt);
\draw [fill=black] (2.25,-1.5) circle (2.5pt);

\fill [gray, opacity=0.4] (0,0) circle (40pt);
\fill [gray, opacity=0.4] (2.05,1.7) circle (40pt);
\fill [gray, opacity=0.4] (4,3) circle (40pt);
\fill [gray, opacity=0.4] (7,1.5) circle (40pt);
\fill [gray, opacity=0.4] (6,4.5) circle (40pt);
\fill [gray, opacity=0.4] (8,3.5) circle (40pt);
\fill [gray, opacity=0.4] (7,-1) circle (40pt);
\fill [gray, opacity=0.4] (4.5,-2) circle (40pt);
\fill [gray, opacity=0.4] (2.25,-1.5) circle (40pt);

\draw [fill=black] (13.5,0) circle (2.5pt);
\draw [fill=black] (15.55,1.7) circle (2.5pt);
\draw [fill=black] (17.5,3) circle (2.5pt);
\draw [fill=black] (20.5,1.5) circle (2.5pt);
\draw [fill=black] (19.5,4.5) circle (2.5pt);
\draw [fill=black] (21.5,3.5) circle (2.5pt);
\draw [fill=black] (20.5,-1) circle (2.5pt);
\draw [fill=black] (18,-2) circle (2.5pt);
\draw [fill=black] (15.75,-1.5) circle (2.5pt);

\draw [line width = 0.5] (13.5,0) -- (15.55,1.7);
\draw [line width = 0.5] (15.55,1.7) -- (17.5,3);
\draw [line width = 0.5] (17.5,3) -- (19.5,4.5);
\draw [line width = 0.5] (19.5,4.5) -- (21.5,3.5);
\draw [line width = 0.5] (21.5,3.5) -- (20.5,1.5);
\draw [line width = 0.5] (20.5,1.5) -- (20.5,-1);
\draw [line width = 0.5] (20.5,-1) -- (18,-2);
\draw [line width = 0.5] (18,-2) -- (15.75,-1.5);
\draw [line width = 0.5] (15.75,-1.5) -- (13.5,0);

\draw [gray, opacity=0.9, line width = 0.5] (0,1.40) -- (-4,1.4);
\draw [gray, opacity=0.9, line width = 0.5] (0,-1.40) -- (-4,-1.4);
\node (none) at (-4,0) {\LARGE $\varepsilon$};
\end{tikzpicture}  \\
  \begin{tikzpicture}[scale=0.5, transform shape, line cap=round,line join=round,>=triangle 45,x=1cm,y=1cm]
\draw [fill=black] (0,0) circle (2.5pt);
\draw [fill=black] (2.05,1.7) circle (2.5pt);
\draw [fill=black] (4,3) circle (2.5pt);
\draw [fill=black] (7,1.5) circle (2.5pt);
\draw [fill=black] (6,4.5) circle (2.5pt);
\draw [fill=black] (8,3.5) circle (2.5pt);
\draw [fill=black] (7,-1) circle (2.5pt);
\draw [fill=black] (4.5,-2) circle (2.5pt);
\draw [fill=black] (2.25,-1.5) circle (2.5pt);

\fill [gray, opacity=0.4] (0,0) circle (55pt);
\fill [gray, opacity=0.4] (2.05,1.7) circle (55pt);
\fill [gray, opacity=0.4] (4,3) circle (55pt);
\fill [gray, opacity=0.4] (7,1.5) circle (55pt);
\fill [gray, opacity=0.4] (6,4.5) circle (55pt);
\fill [gray, opacity=0.4] (8,3.5) circle (55pt);
\fill [gray, opacity=0.4] (7,-1) circle (55pt);
\fill [gray, opacity=0.4] (4.5,-2) circle (55pt);
\fill [gray, opacity=0.4] (2.25,-1.5) circle (55pt);

\draw [fill=black] (13.5,0) circle (2.5pt);
\draw [fill=black] (15.55,1.7) circle (2.5pt);
\draw [fill=black] (17.5,3) circle (2.5pt);
\draw [fill=black] (20.5,1.5) circle (2.5pt);
\draw [fill=black] (19.5,4.5) circle (2.5pt);
\draw [fill=black] (21.5,3.5) circle (2.5pt);
\draw [fill=black] (20.5,-1) circle (2.5pt);
\draw [fill=black] (18,-2) circle (2.5pt);
\draw [fill=black] (15.75,-1.5) circle (2.5pt);

\draw [line width = 0.5] (13.5,0) -- (15.55,1.7);
\draw [line width = 0.5] (15.55,1.7) -- (17.5,3);
\draw [line width = 0.5] (17.5,3) -- (19.5,4.5);
\draw [line width = 0.5] (19.5,4.5) -- (21.5,3.5);
\draw [line width = 0.5] (21.5,3.5) -- (20.5,1.5);
\draw [line width = 0.5] (20.5,1.5) -- (20.5,-1);
\draw [line width = 0.5] (20.5,-1) -- (18,-2);
\draw [line width = 0.5] (18,-2) -- (15.75,-1.5);
\draw [line width = 0.5] (15.75,-1.5) -- (13.5,0);

\draw [line width = 0.5] (15.55,1.7) -- (15.75,-1.5);
\draw [line width = 0.5] (17.5,3) -- (20.5,1.5);
\draw [line width = 0.5] (19.5,4.5) -- (20.5,1.5);

\fill [gray, opacity=0.4] (13.5,0) -- (15.55,1.7) -- (15.75,-1.5);
\fill [gray, opacity=0.4] (17.5,3) -- (19.5,4.5) -- (20.5,1.5);
\fill [gray, opacity=0.4] (21.5,3.5) -- (19.5,4.5) -- (20.5,1.5);

\draw [gray, opacity=0.9, line width = 0.5] (0,1.95) -- (-4,1.95);
\draw [gray, opacity=0.9, line width = 0.5] (0,-1.95) -- (-4,-1.95);
\node (none) at (-4,0) {\LARGE $\varepsilon$};
\end{tikzpicture}
  \caption[Filtration parameter and simplex construction]{The filtration parameter $\varepsilon$ grows around each data point continuously leading to the creation of geometrical disks. When two disks intersect, a line is drawn between the two corresponding data points resulting in a connected component defined as a 1-simplex. Triangles, defined as 2-simplex, are generated as the filtration parameter $\varepsilon$ keeps growing.}
  \label{fig::scale_para_1}
\end{figure}
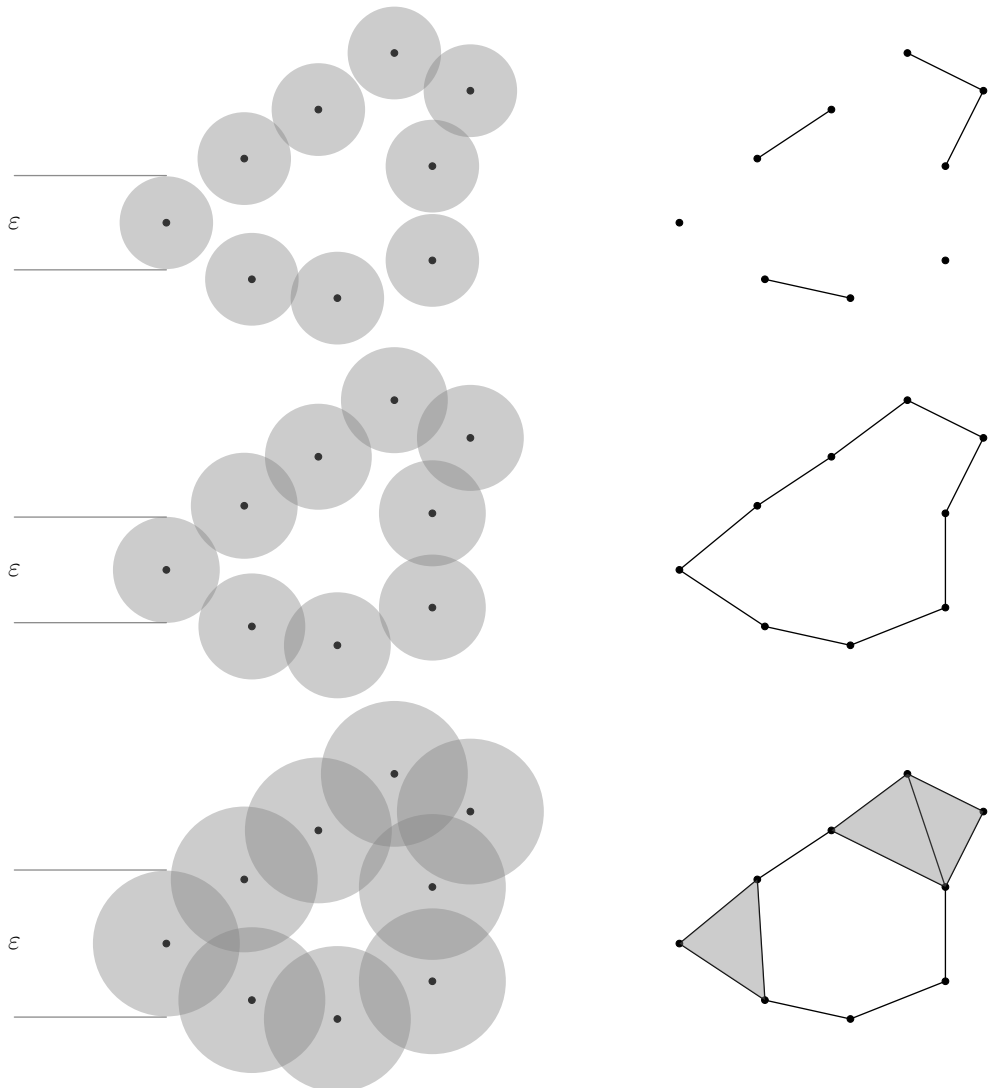

The first key notion for the construction of barcodes and persistence diagrams, used to highlight the persistent homology features of the data points cloud, is the filtration parameter denoted by $\varepsilon$. First, for every data point, the size of the points is continuously and artificially increased. Therefore, the size of the points grows as they become geometrical disks. We illustrate the concept in Figure \ref{fig::scale_para_1}. Through the continuous process, the filtration parameter $\varepsilon$ grows. When two disks intersect, a line is drawn between the two corresponding original data points, creating a connected component defined as a 1-simplex. As the filtration parameter keeps growing, more connected components are created as well as triangles defined as 2-simplex.  \\

\begin{figure}[p]
  \centering
  \includegraphics[scale=.6]{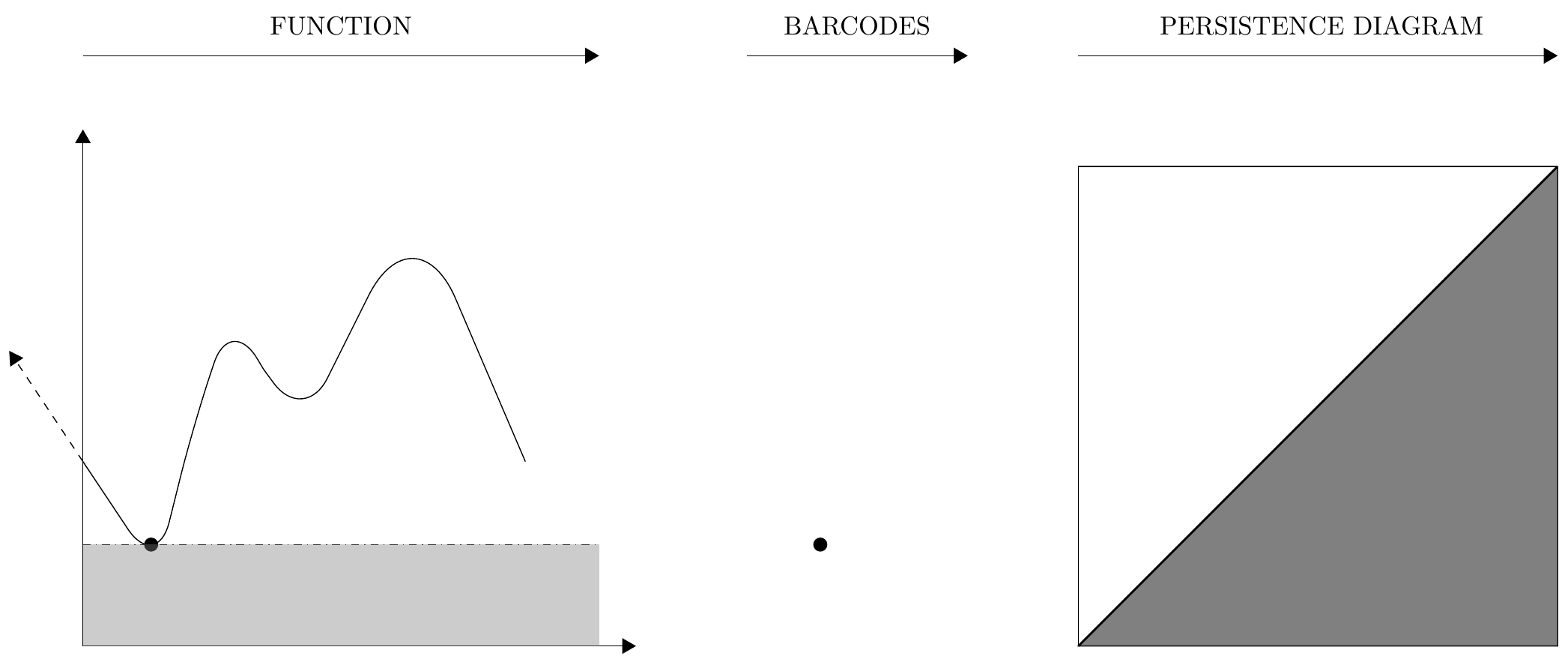}  \\ \vspace{0.25cm}
  \includegraphics[scale=.6]{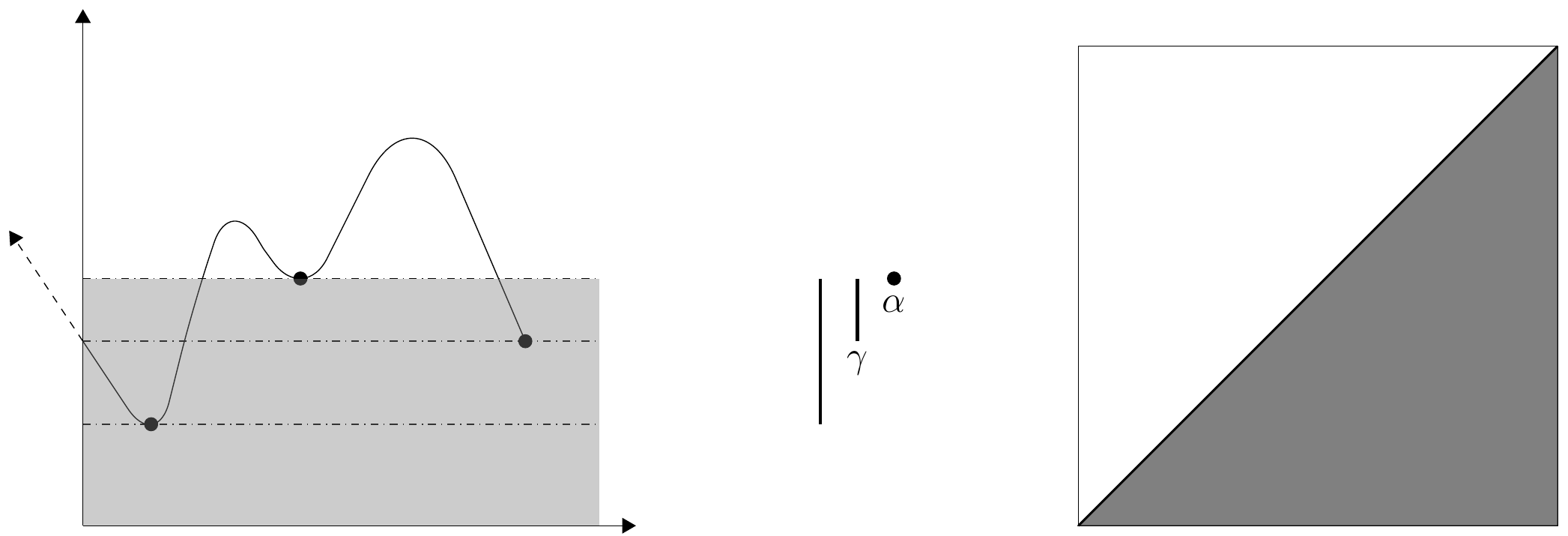}  \\ \vspace{0.25cm}
  \includegraphics[scale=.6]{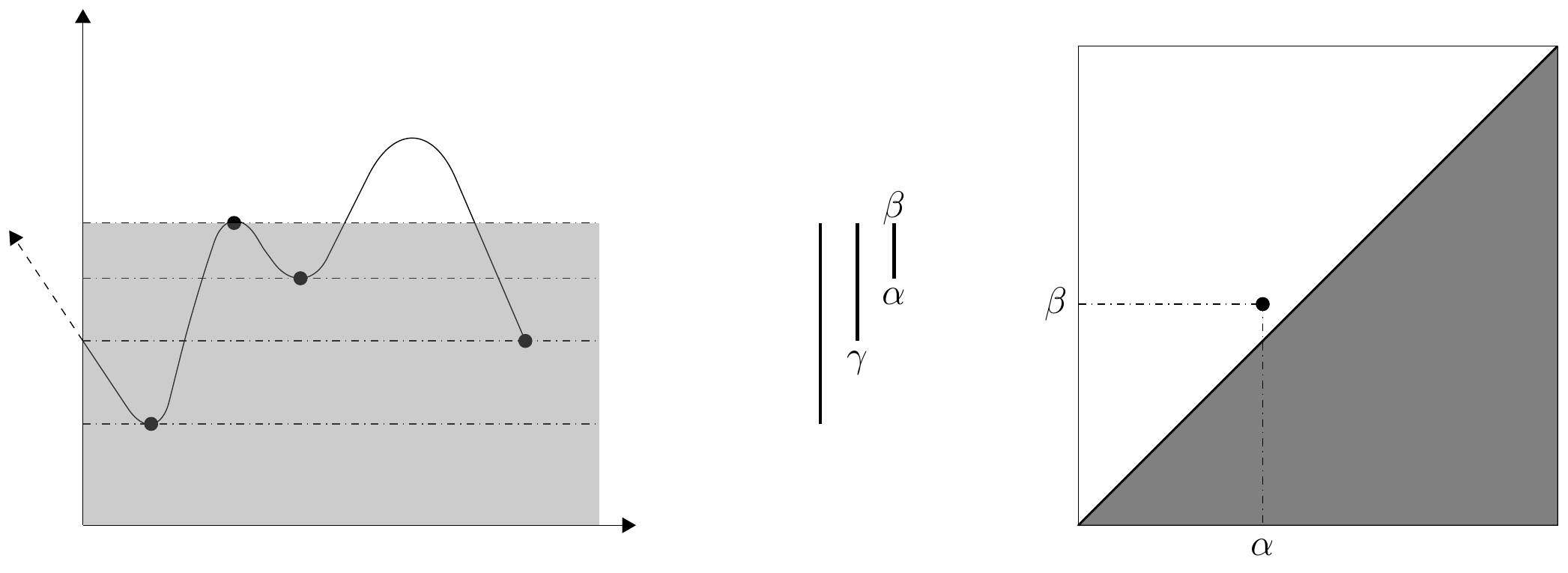}  \\ \vspace{0.25cm}
  \includegraphics[scale=.6]{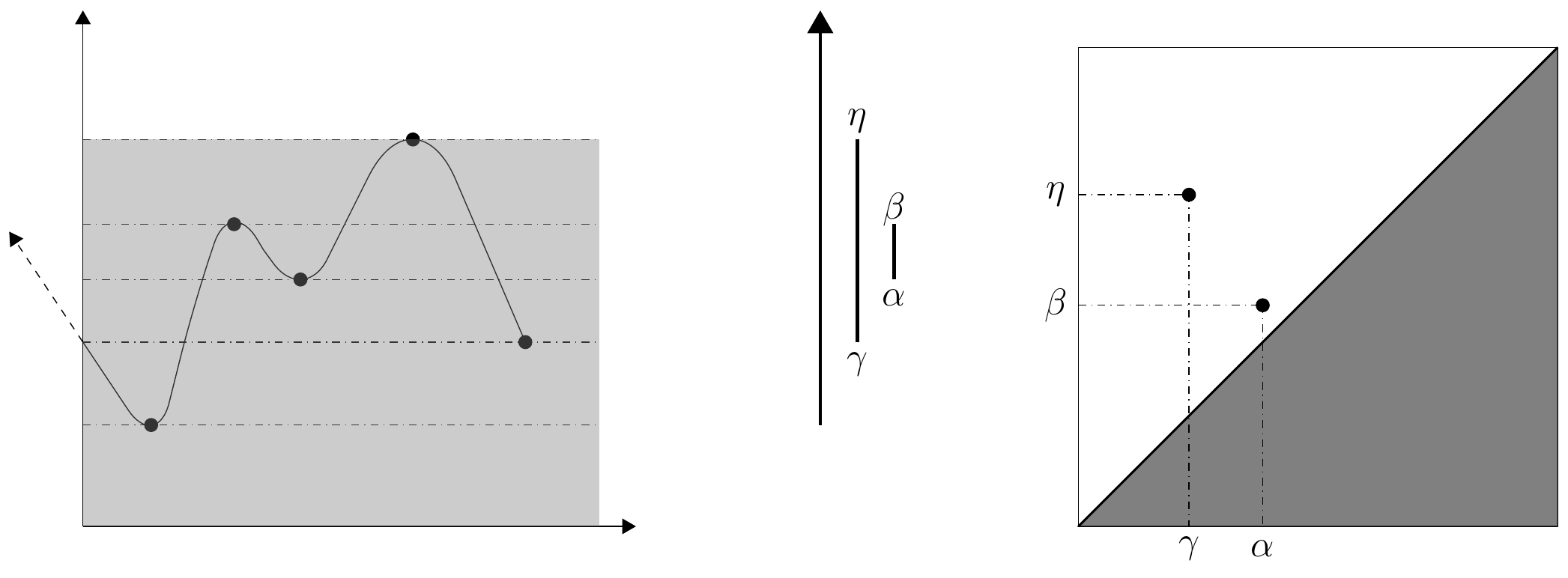}
  \caption[Barcodes and persistence diagram construction]{As the filtration evolves, the local minima of the function provoke the creation of a barcode while the local maxima induce the death of the barcode. Every birth-death cycle of each barcode can be represented in the persistence diagram, allowing the description of further persistent homology features.}
  \label{fig::bar_persdiag}
\end{figure}

By relying on this filtration parameter $\varepsilon$, we can construct barcodes and persistence diagrams. We recall that a barcode diagram is a stable summary representation of a persistence diagram. We highlight the construction of the barcodes and the persistence diagram in Figure \ref{fig::bar_persdiag}. Given a two-dimensional function, we capture the points for which there is a local extrema. The filtration starts from the bottom to reach the top. Every time a minima is observed, a barcode is created. Then, as the filtration evolves, local maxima are observed provoking the ``death'' of the barcode. Every ``birth-death'' episode can be then reported in the persistence diagram offering a slightly different visualization than the barcode diagram. Additionally, the persistence diagram can be later used to further describe the topological properties of the original data points cloud using quantitative measures, such as the bottleneck distance aforementioned.  \\

\vspace{-.5cm}
\subsubsection{Betti Numbers and Homology Groups}
\begin{figure}[b]
  \centering
  \vspace{1.25cm}
  \begin{tikzpicture}[scale=0.7, transform shape, line cap=round,line join=round,>=triangle 45,x=1cm,y=1cm]

\draw [fill=black] (-7.5,0) circle (2.5pt);
\draw [fill=black] (-4,0) circle (2.5pt);
\draw [line width=0.25pt] (-7.5,0) -- (-4,0);

\draw[line width=0.5pt] (-0.5,0) circle (2cm);

\shade[ball color = gray!40, opacity = 0.4] (5,0) circle (2cm);
\draw (5,0) circle (2cm);
\draw (3,0) arc (180:360:2 and 0.6);
\draw[dashed] (7,0) arc (0:180:2 and 0.6);
\draw (5,2) arc (90:270:0.8 and 2);
\draw[dashed] (5,-2) arc (270:450:0.8 and 2);

\end{tikzpicture}
  \caption[Visualization of homology groups]{Visualization of the first three homology groups. The line, or the connected components, belongs to the first homology group $H_0$. The second homology group $H_1$ represents the circles or the loops. Finally, the homology group $H_2$ describes the voids, the cavities or the spheres.}
  \label{fig::homology_group}
\end{figure}
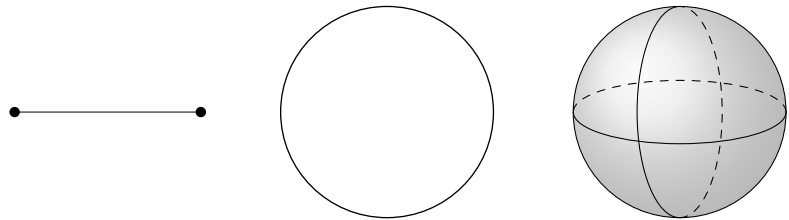

The aim of the persistent homology is to describe the shape of the data points cloud by relying on features such as connected components, loops or cavities, independent of any distance measurement. To this end, the set of features contained in the data manifold are categorized into different homological dimensions, or homology groups. We illustrate the different homology groups in Figure \ref{fig::homology_group} for the first three homology groups. The first homology group, denoted by $H_0$, is used to characterize the connected components. The second homology group, $H_1$, defines the loops or the circles. Finally, the third homology group, $H_2$, designates the voids, the cavities or the spheres. We invite the reader to \cite{hatcher2002algebraic} for more details.  \\

To complete the description of the topological features of the data points cloud, the homology groups are completed with the Betti numbers. The Betti numbers are used to measure the holes of the data point cloud for each of the homology groups \cite{zomorodian2005computing,edelsbrunner2000topological}. They more precisely reflect the topological properties of a shape as being the number of $i-$dimensional holes in a simplical complex \cite{gholizadeh2018topological}. We recall the illustration of the simplicial complex in Figure \ref{fig::simplex}. Relying on Figure \ref{fig::homology_group}, we can describe the objects with the first three Betti numbers. The connected components, highlighting the homology group $H_0$, are described with the Betti numbers $\beta_0 = 1, \beta_1 = 0, \beta_2 = 0$. Then, the circle representing the homology group $H_1$ has the following Betti numbers $\beta_0 = 1, \beta_1 = 1, \beta_2 = 0$. Finally, the sphere illustrating the homology group $H_2$ has the Betti numbers $\beta_0 = 1, \beta_1 = 0, \beta_2 = 1$.  \\

\subsubsection{Combining the Filtration Parameter, Homology Groups and Barcodes}
\begin{figure}[b!]
	\begin{center}
	\vspace{0.5cm}
	\includegraphics[scale=0.58]{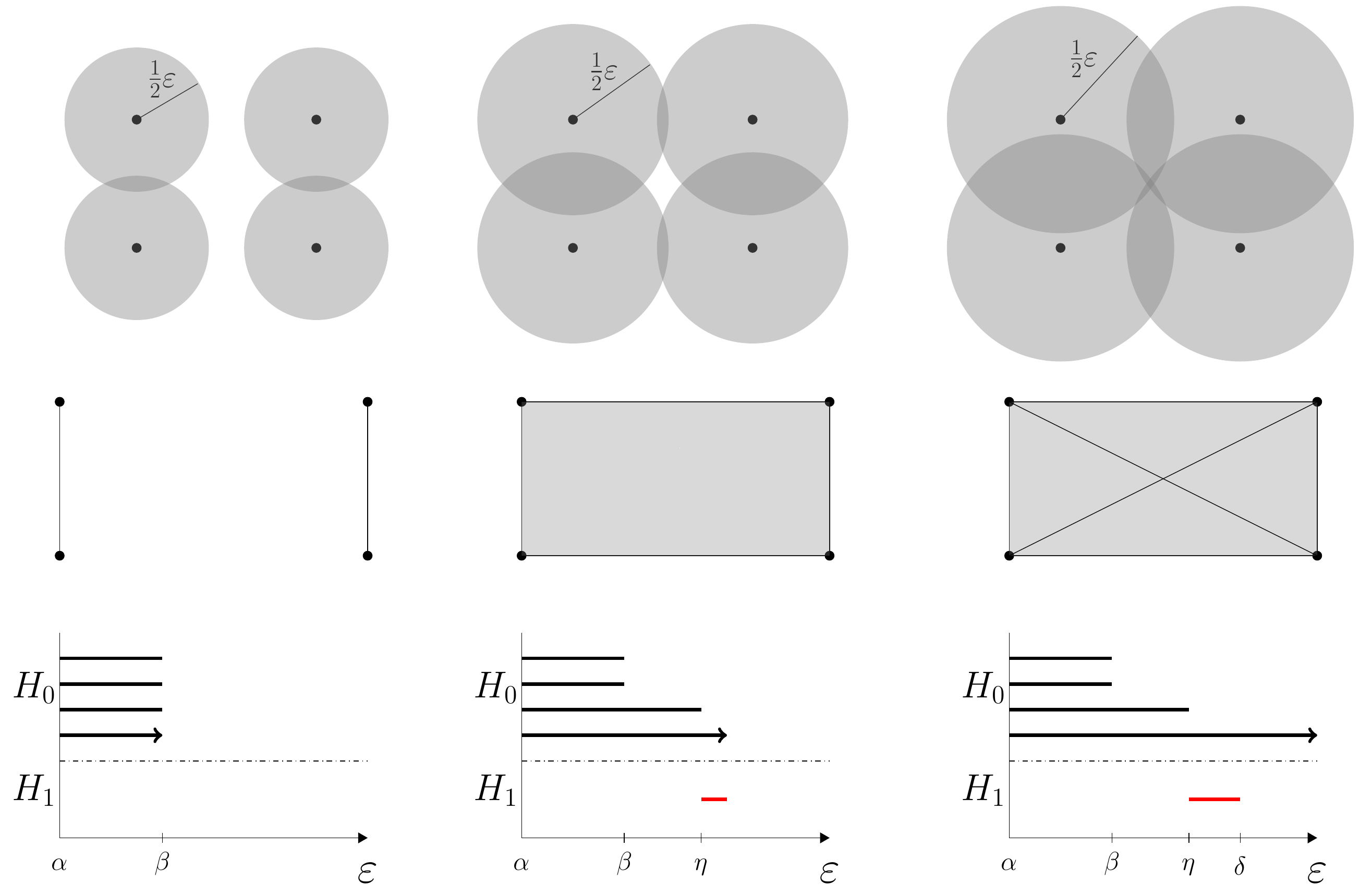}
    \caption[Combining filtration parameter, homology groups and barcodes]{Representation of the filtration parameter $\varepsilon$ with the homology groups $H_0$ and $H_1$ for data points inherited from a rectangle. As the filtration parameter $\varepsilon$ progresses, the persistent homology features are highlighted with respect to each homology groups. The homology group $H_0$ captures the connected components and the homology group $H_1$ the loops. The birth-death episodes of the persistent homology features are summarized by the barcodes.}
    \label{fig::barcode_cons}
	\end{center}
\end{figure}

\begin{figure}[t]
	\begin{center}
	\includegraphics[scale=0.8]{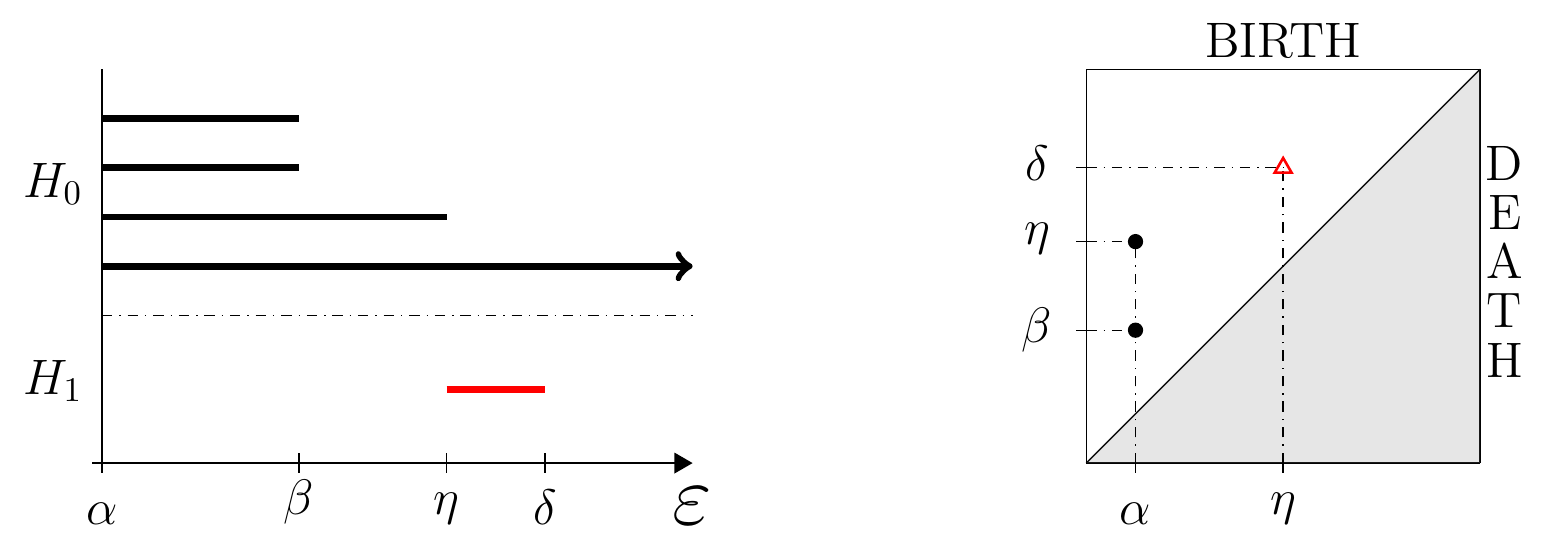}
	\caption[Summary representation of barcodes and persistence diagram]{The filtration procedure leads to the construction of the barcodes of the homological groups $H_0$ and $H_1$, a stable summary representation of the persistence diagrams. The groups $H_0$ and $H_1$ are represented by black dots and red triangles, respectively, in the persistence diagram.}
    \label{fig::persdiag_cons}
	\end{center}
\end{figure}

Following the practical description of the key concepts of the persistent homology, we emphasize how these concepts, and noticeably the filtration parameter, the simplex, the homology groups and the barcodes, relate to each other in Figures \ref{fig::barcode_cons} and \ref{fig::persdiag_cons}. Given four original data points representing a rectangle, the continuous growth of the filtration parameter $\varepsilon$ leads to the computation of various 1-simplex. Vertices, edges and faces are effectively continuously generated \cite{monod2017tropical}. The first two homology groups, $H_0$ and $H_1$, are represented. These groups, as aforementioned, describe the connected components and the loops, respectively. The different snapshots of Figure \ref{fig::barcode_cons} capture the relation between the barcodes, the homology groups and the filtration parameter with respect to the original data points and the growth of the filtration parameter $\varepsilon$. At the end of the filtration procedure, a persistence diagram is drawn to recapitulate the birth-death events observed with the barcodes, as shown in Figure \ref{fig::persdiag_cons}. The persistence diagram can be later used to describe the topological properties of the original data points cloud using quantitative measures, such as the bottleneck distance.

\subsection{Proposed Method: PHom-GeM, Persistent Homology for Generative Models}
\begin{figure}[p]
	\begin{center}
	\includegraphics[scale=.9]{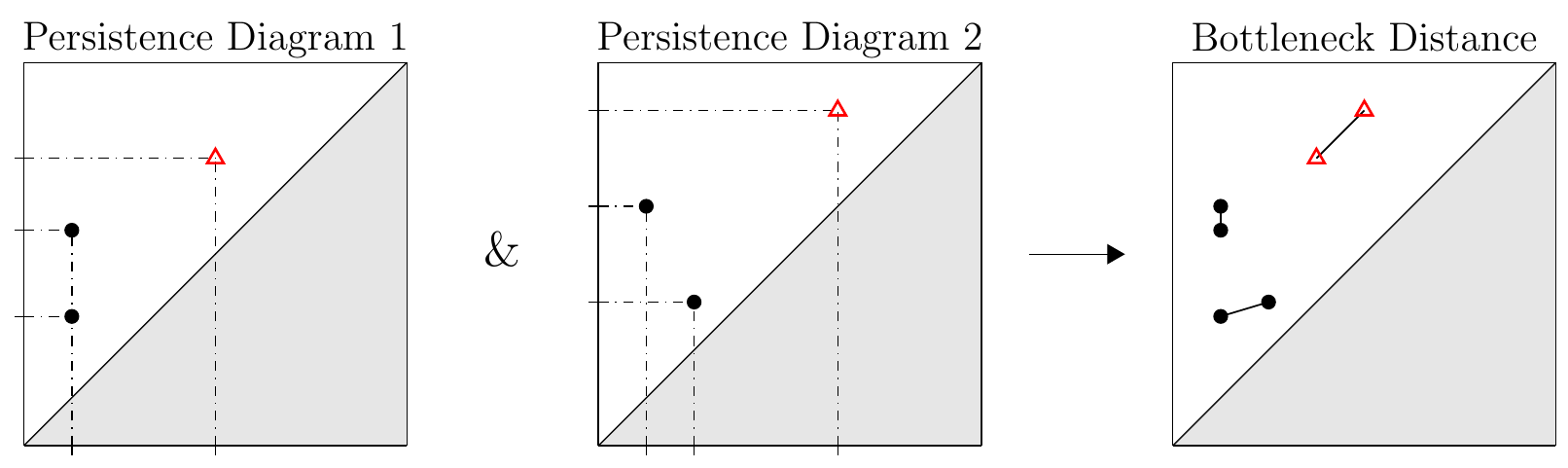}
	\caption[Persistence diagrams and the bottleneck distance]{The bottleneck distance is a measure of similarity between two persistence diagrams. The points of the first and the second persistence diagrams are gathered on one persistence diagram. The distance between each of the points is measured. The bottleneck distance is computed such that it is the shortest distance for which any couple of matched points are at distance at most b.}
    \label{fig::persdiag_bottleneck}
	\end{center}
\end{figure}

Bridging the gap between the persistent homology and the generative models, including GP-WGAN, WGAN, WAE and VAE, PHom-GeM uses a two-steps procedure. First, the minimization problem is solved for the generator $G$ and the discriminator $D$ when considering GP-WGAN and WGAN. The gradient penalty $\lambda$ in equation (\ref{eq::DWGAN}) is fixed equal to 10 for GP-WGAN and to 0 for WGAN. For auto-encoders, the minimization problem is solved for the encoder $Q$ and the decoder $G$. We use RMSProp optimizer \cite{hinton2012rmsprop} for the optimization procedure. Then, the samples of the original and generated distributions, $P_X$ and $P_G$, are mapped to the persistent homology for the description of their respective manifolds. The points contained in the manifold $\mathcal{X}$ inherited from $P_X$ and the points contained in the manifold $\mathcal{\tilde{X}}$ generated with $P_G$ are randomly selected into respective batches. Two samples, $Y_1$ from $\mathcal{X}$ following $P_X$ and $Y_2$ from $\mathcal{\tilde{X}}$ following $P_G$, are selected to differentiate the topological features of the original manifold $\mathcal{X}$ and the generated manifold $\mathcal{\tilde{X}}$. The samples $Y_1$ and $Y_2$ are contained in the spaces $\mathcal{Y}_1$ and $\mathcal{Y}_2$, respectively. The spaces $\mathcal{Y}_1$ and $\mathcal{Y}_2$ are then transformed into metric space sets $\mathcal{\widehat{Y}}_1$ and $\mathcal{\widehat{Y}}_2$ for computational purposes. We subsequently filter the metric space sets $\mathcal{\widehat{Y}}_1$ and $\mathcal{\widehat{Y}}_2$ using the Vietoris-Rips simplicial complex filtration. Given a line segment of length $\epsilon$, vertices between data points are created for data points respectively separated from a smaller distance than $\epsilon$. It leads to the construction of a collection of simplices resulting in Vietoris-Rips simplicial complex VR$(\mathcal{C}, \epsilon)$ filtration. In our case, we decide to use the Vietoris-Rips simplicial complex as it offers the best compromise between the filtration accuracy and the memory requirement \cite{chazal2017tda}. Subsequently, the persistence diagrams, $\text{dgm}_{Y_1}$ and $\text{dgm}_{Y_2}$, are constructed. We recall a persistence diagram is a stable summary representation of the topological features of simplicial complex. The persistence diagrams allow the computation of the bottleneck distance $d_b(\text{dgm}_{Y_1}, \text{dgm}_{Y_2})$ for quantitative measurements of the similarities, as illustrated in Figure \ref{fig::persdiag_bottleneck}. Finally, the barcodes represent in a simple way the birth-death of the pairing generators of the iterated inclusions detected by the persistence diagrams. We summarized the persistent homology procedure for generative models Algorithm \ref{algo::TopWGAN}. \\

\SetAlFnt{\footnotesize} 
\SetAlCapFnt{\footnotesize} 
\SetAlCapNameFnt{\footnotesize} 
\SetKwFor{Case}{case}{}{}

\begin{algorithm}[p] 
\setstretch{1.25} 
\DontPrintSemicolon 
\KwData{training and validation sets, hyperparameter $\lambda$} 
\KwResult{persistent homology description of generative manifolds} 

\Begin{  
/*\textit{\small Step 1: Generative Models Resolution}*/ 

Select samples $\left\lbrace x_1, ..., x_n\right\rbrace$ from training set 
 
Select samples $\left\lbrace z_1, ..., z_n\right\rbrace$ from validation set 
  
With RMSProp gradient descent update ($\text{lr}=0.001, \rho=0.9, \epsilon=10^{-6}$), optimize until convergence $Q$ and $G$  \\ 

\hspace{0.25cm}\lCase{GP-WGAN and WGAN:}{using equation \ref{eq::DWGAN}} 

\hspace{0.25cm}\lCase{WAE:}{using equation \ref{eq::DWAE}} 

\hspace{0.25cm}\lCase{VAE:}{using equation \ref{eq::vaeloss}}

\vspace{0.25cm} 

/*\textit{\small Step 2: Persistence Diagram and Bottleneck Distance on manifolds of generative models}*/ 

Random selection of samples $Y_1 \in \mathcal{Y}_1, Y_2 \in \mathcal{Y}_2$ from $P_X$ and $P_G$ 
 
Transform $\mathcal{Y}_1$ and $\mathcal{Y}_2$ spaces into a metric space set   

Filter the metric space set with a Vietoris-Rips simplicial complex $\text{VR}(\mathcal{C}, \epsilon)$ 

Compute the persistence diagrams $\text{dgm}_{Y_1}$ and $\text{dgm}_{Y_2}$ 

Evaluate the bottleneck distance $d_b(\text{dgm}_{Y_1}, \text{dgm}_{Y_2})$ 

Build the barcodes with respect to $Y_1$ and $Y_2$ 

} 

\KwRet{} 

\caption{Persistent Homology for Generative Models\label{algo::TopWGAN}} 
\end{algorithm}

\section{Experiments} \label{sec::exp}
We assess on a highly challenging data set for generative models whether PHom-GeM can simultaneously achieve (i) accurate persistent homology distance measurement with the bottleneck distance, (ii) detection of the AE latent manifold scattered distribution $P_Z$, (iii) accurate topological reconstruction of the data points, and (iv) precise persistent homology description of the generated data points.  \\

\textbf{Data Availability and Data Description} 
We train PHom-GeM on one real-world open data set: the credit card transactions data set from the Kaggle database\footnote{The data set is available at https://www.kaggle.com/mlg-ulb/creditcardfraud.} containing 284\,807 transactions including 492 frauds. This data set is particularly interesting because it reflects the scattered points distribution of the reconstructed manifold that are obtained during generative models' training, impacting afterward the generated adversarial samples. This data set furthermore is challenging because of the strong imbalance between normal and fraudulent transactions while being of high interest for the banking industry. To preserve the transactions confidentiality, each transaction is composed of 28 components obtained with PCA without any description and two additional features \textit{Time} and \textit{Amount} that remained unchanged. Each transaction is labeled as fraudulent or normal in a feature called \textit{Class} which takes a value of 1 in case of a fraudulent transaction and 0 otherwise.  \\ 

\textbf{Experimental Setup and Code Availability}
In our experiments, we use the Euclidean latent space $\mathcal{Z} = \mathcal{R}^2$ and the square cost function $c$ previously defined as $c(x,y)=||x-y||_2^2$ for the data points $x \in \mathcal{X}, \widetilde{x} \in \mathcal{\widetilde{X}}$. The dimensions of the true data set is $\mathcal{R}^{29}$. We kept the 28 components obtained with PCA and the amount resulting in a space of dimension 29. For the error minimization process, we used RMSProp gradient descent \cite{hinton2012rmsprop} with the parameters $\text{lr}=0.001, \rho=0.9, \epsilon=10^{-6}$ and a batch size of 64. Different values of $\lambda$ for the gradient penalty have been tested. We empirically obtained the lowest error reconstruction with $\lambda=10$ for both GP-WGAN and WAE, as summarized in Table \ref{tab::gpCalibration}. The coefficients of the persistent homology are evaluated within the field $\mathbb{Z}/2 \mathbb{Z}$. We only consider homology groups $H_0$ and $H_1$ who represent the connected components and the loops, respectively. Higher dimensional homology groups did not noticeably improve the results quality while leading to longer computational time. The simulations were performed on a computer with 16GB of RAM, Intel i7 CPU and a Tesla K80 GPU accelerator. To ensure the reproducibility of the experiments, the code is available on GitHub \cite{charlierphomgem}. 
\\ 

\begin{table}[t] 
 \centering 
 \caption[RMSE based on the gradient penalty between original and training samples]{Root Mean Square Error (RMSE) after PHom-GeM gradient penalty calibration on GP-WGAN and WAE between the predicted samples and the original samples of the model (smaller is better).
 } \label{tab::gpCalibration} 
 \begin{tabular}{ccc} 
  \toprule 
  Gradient Penalty & RMSE of GP-WGAN & RMSE of WAE\\ 
  \midrule 
  0.1 & 0.159 & 1.403 \\ 
  0.5 & 0.162 & 0.846 \\ 
  1.0 & 0.158 & 0.597 \\ 
  3.0 & 0.158 & 0.532 \\ 
  5.0 & 0.155 & 0.385 \\ 
  \textbf{10.0} & \textbf{0.153} & \textbf{0.376} \\ 
  15.0 & 0.155 & 0.381 \\ 
  20.0 & 0.158 & 0.385 \\ 
  \bottomrule 
 \end{tabular} 
\end{table} 

\textbf{Results and Discussions about PHom-GeM}
Because of the strong impact of the AE scattered distribution $P_Z$ on the quality of the encoding-decoding process, we first assess for AE, and noticeably for VAE and WAE, if PHom-GeM, Persistent Homology for Generative Models, can measure the scattered distribution $P_Z$ using the bottleneck distance. We then observe the consequence of the encoding process on the AE reconstructed distribution. Finally, in a third experiment, we reach the core of our contribution that is the use of the persistent homology on complex generated adversarial samples. We test PHom-GeM on the four different generative models: GP-WGAN, WGAN, WAE and VAE. For the second and third experiments, we compare the performance of PHom-GeM on two specificities: first, qualitative visualization of the persistence diagrams and barcodes and, secondly, quantitative estimation of the persistent homology closeness using the bottleneck distance between the generated manifolds $\mathcal{\widetilde{X}}$ of the generative models and the original manifold $\mathcal{X}$.  \\

\subsection{Detection of the Scattered Distribution of the Samples $Z$ of the Latent Manifold $\mathcal{Z}$ for AE} \label{subsec:latentsamples}
\begin{figure*}[p!]
\centering
 \begin{subfigure}{.49\textwidth}
  \centering
  Sample 1 \\ \vspace{.25cm}
  \begin{turn}{90} 
   \hspace{.25cm} Scattered Distribution
  \end{turn}
  \frame{\includegraphics[scale=.5]{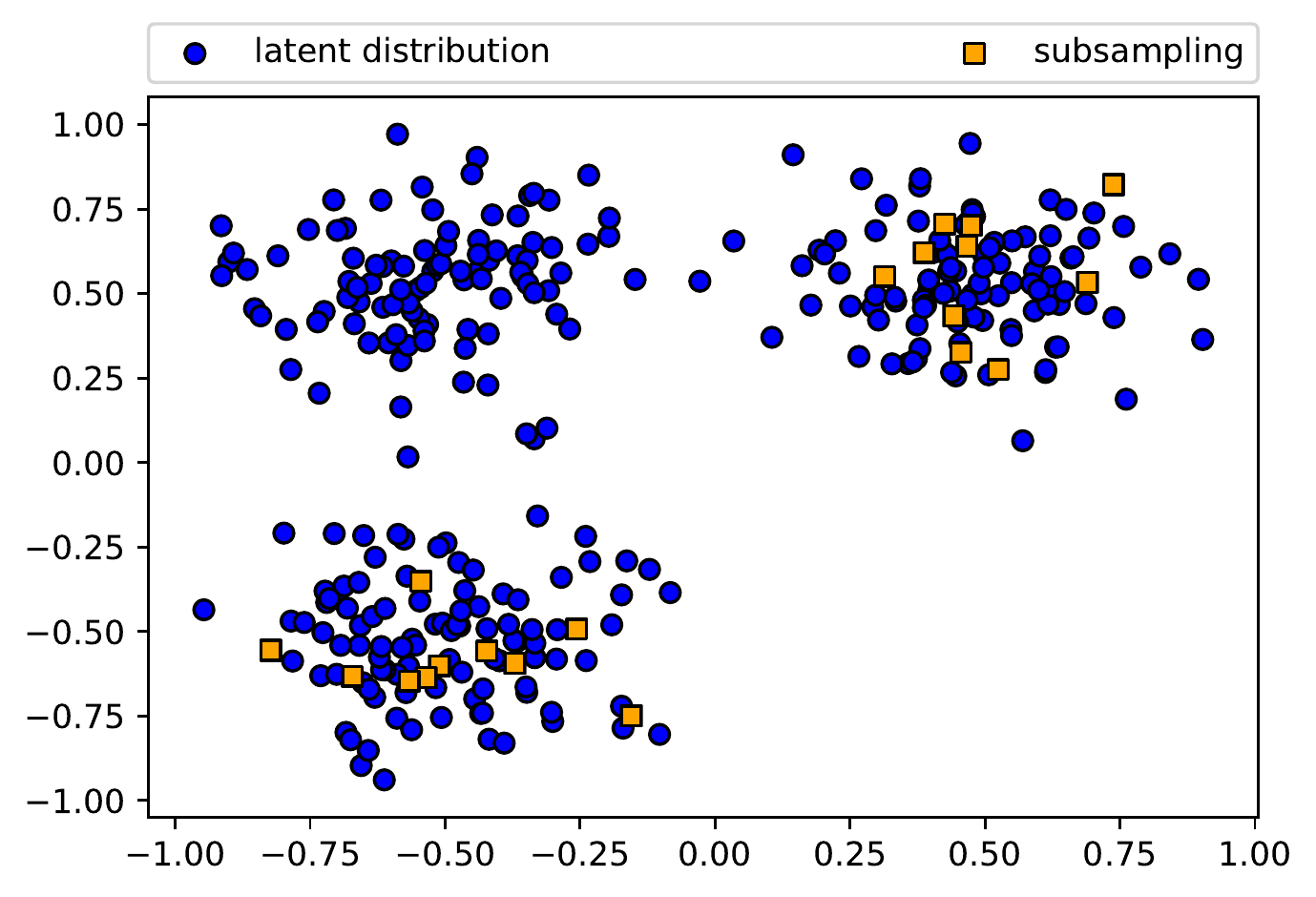}}
 \end{subfigure}\hfill
 \begin{subfigure}{.49\textwidth}
  \centering
  Sample 2 \\ \vspace{.25cm}
  \frame{\includegraphics[scale=.5]{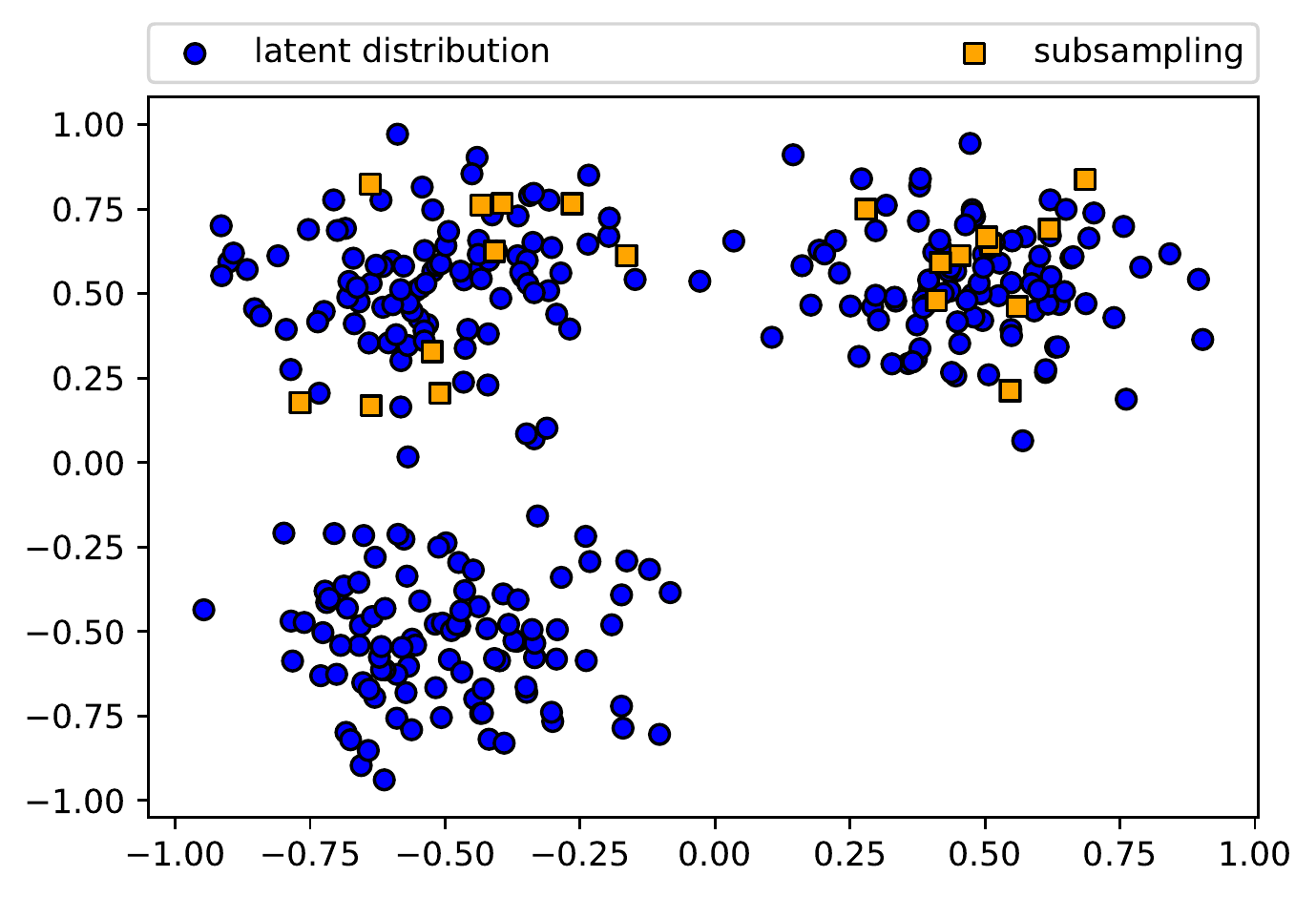}}
 \end{subfigure}\hfill
 
 \begin{subfigure}{.49\textwidth}
  \centering
  \vspace{.5cm}
  \begin{turn}{90} 
   \hspace{0.3cm} Uniform Distribution
  \end{turn}
  \frame{\includegraphics[scale=.5]{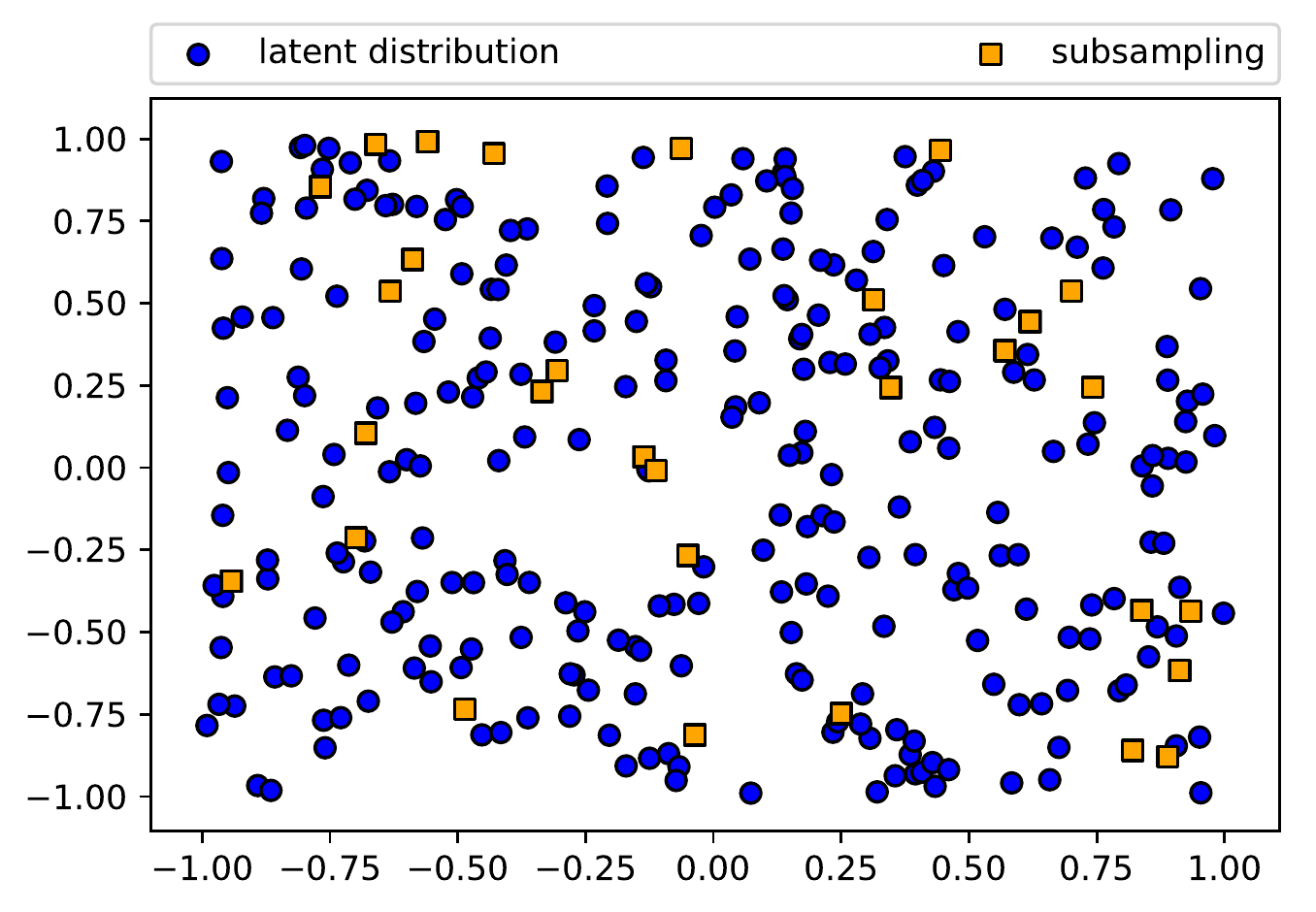}}
 \end{subfigure}\hfill
 \begin{subfigure}{.49\textwidth}
  \centering
  \vspace{.5cm}
  \frame{\includegraphics[scale=.5]{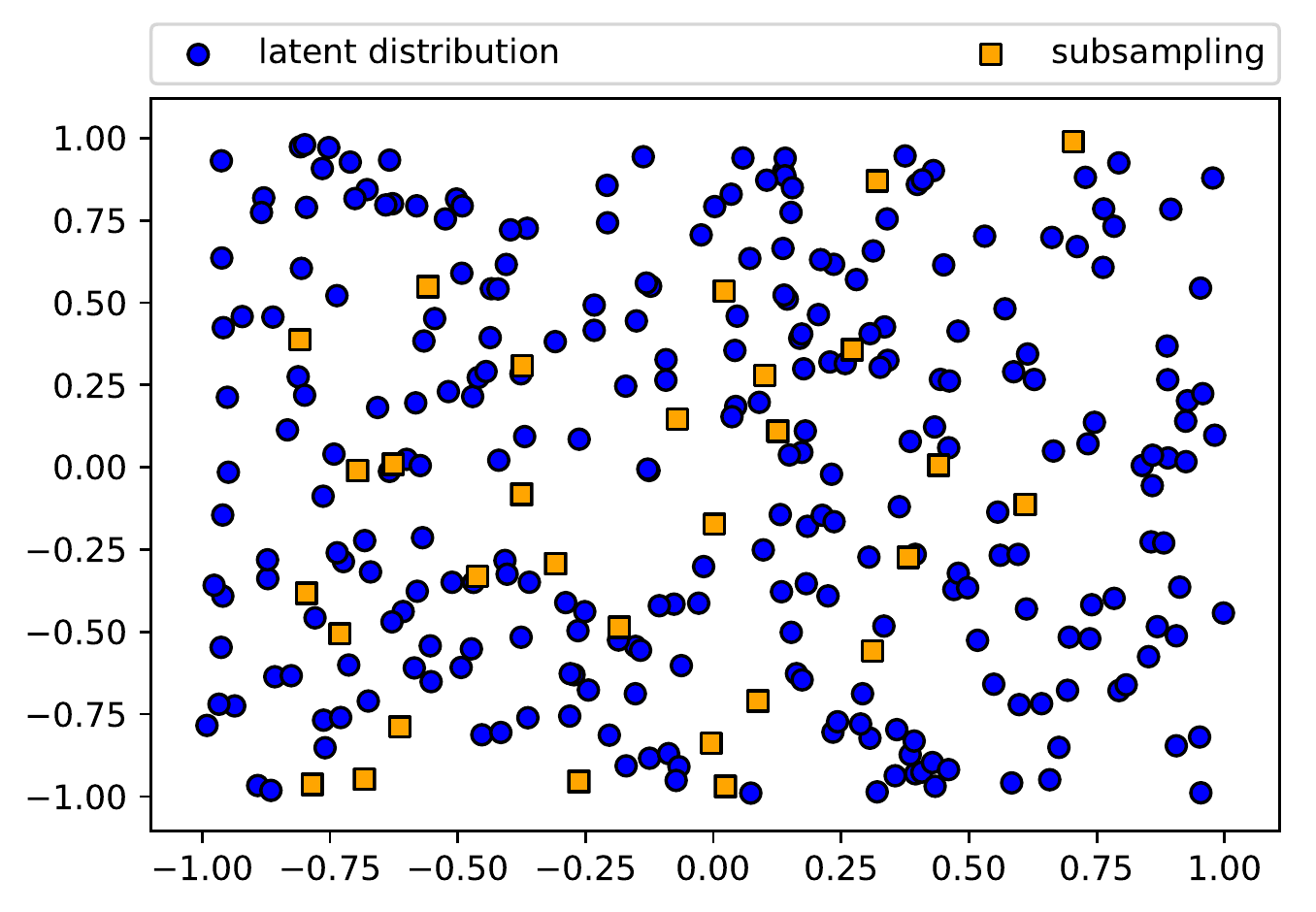}}
 \end{subfigure}\hfill
 
  \caption[Scattered and uniform distributions of the latent space]{By using the bootstrapping technique of \cite{friedman2001elements}, we can randomly select samples of the latent manifold $\mathcal{Z}$ and then, measure and aggregate the bottleneck distance between the samples. In case of a scattered distribution, there is a strong variation between the randomly selected samples of the persistent homological properties, such as the distance between the points or the number of holes between the connected components. Therefore, a scattered distribution leads to a bigger bottleneck distance than a uniform distribution.} 
  \label{fig::scatteredplot}
\end{figure*}

PHom-GeM is capable to measure the persistent homological features, for intance the number of connected components or the number of holes, between two samples of a distribution. Hereinafter, we use the persistent homological distance, the bottleneck distance, to measure and compare how much the WAE's and VAE's distributions $P_Z$ of the latent manifold $\mathcal{Z}$ is scattered from a persistent homology perspective. The strength of the bottleneck distance is to measure quantitatively the topological changes in the data, either the true or the reconstructed data, while being insensitive to the scale of the analysis. Using the bootstrapping technique of \cite{friedman2001elements}, we successively randomly select data samples $Z_i$ of the manifold $\mathcal{Z}$. The total number of selected samples $Z_i$ is at least 85\% of the total number of points contained in the manifold $\mathcal{Z}$ for a reliable statistical representation. Assuming the data is not scattered, the bottleneck distance between the samples $Z_i$ is small. On the opposite, if the data is chaotically scattered in $\mathcal{Z}$, then the topological features between the samples $Z_i$ are significantly different, and consequently, the bottleneck distance is large. We illustrate the idea in Figure \ref{fig::scatteredplot}.  \\

\begin{table}[p!]
 \centering
 \caption[PHom-GeM bottleneck distance between WAE and VAE latent samples]{Bottleneck distance (smaller is better) for PHom-GeM applied to WAE and PHom-GeM applied to VAE between the samples $Z_i$ of the latent manifold $\mathcal{Z}$ following $P_Z$ to detect the scattered distribution. The WAE better preserves the topological features during the encoding than the VAE resulting in a manifold $\mathcal{Z}$ less chaotically scattered, highlighted by the smaller bottleneck distance.}
 \label{tab::res_scarc}
 \begin{tabular}{cc|c}
  \toprule
  PHom-GeM on WAE \quad & \quad PHom-GeM on VAE \quad & \quad Difference (\%)  \\
  \midrule
  \textbf{0.0984} & 0.1372 &  28.28 \\
  \bottomrule
 \end{tabular}
\end{table}

In Table \ref{tab::res_scarc}, the bottleneck distance is significantly lower for WAE than for VAE. The level of scattered chaos observed for WAE is lower than for VAE, a direct consequence of the use of optimal transport in WAE. It also means the distribution $P_Z$ of the latent manifold $\mathcal{Z}$ is better topologically preserved for WAE than for VAE. Thus, we can expect that the reconstructed distribution $P_G(X|Z)$ of $\mathcal{X}$ is less altered for WAE than for VAE.  \\

\subsection{Reconstructed distribution $P_Z$ of the AE}
Following the results on the persistent homological features of the latent manifold $\mathcal{Z}$ of the WAE and the VAE, we assess in this section the impact of the quality of the distribution $P_Z$ on the reconstructed distribution. Effectively, before generating adversarial samples, we want to evaluate how much the WAE and the VAE are capable to reproduce at their output the data injected as input. \\

\begin{figure*}[p]
\centering
 \begin{subfigure}{1.0\textwidth}
  \centering
  \begin{turn}{90} 
   \hspace{0.8cm} Original Sample
  \end{turn}
  \frame{\includegraphics[scale=0.325]{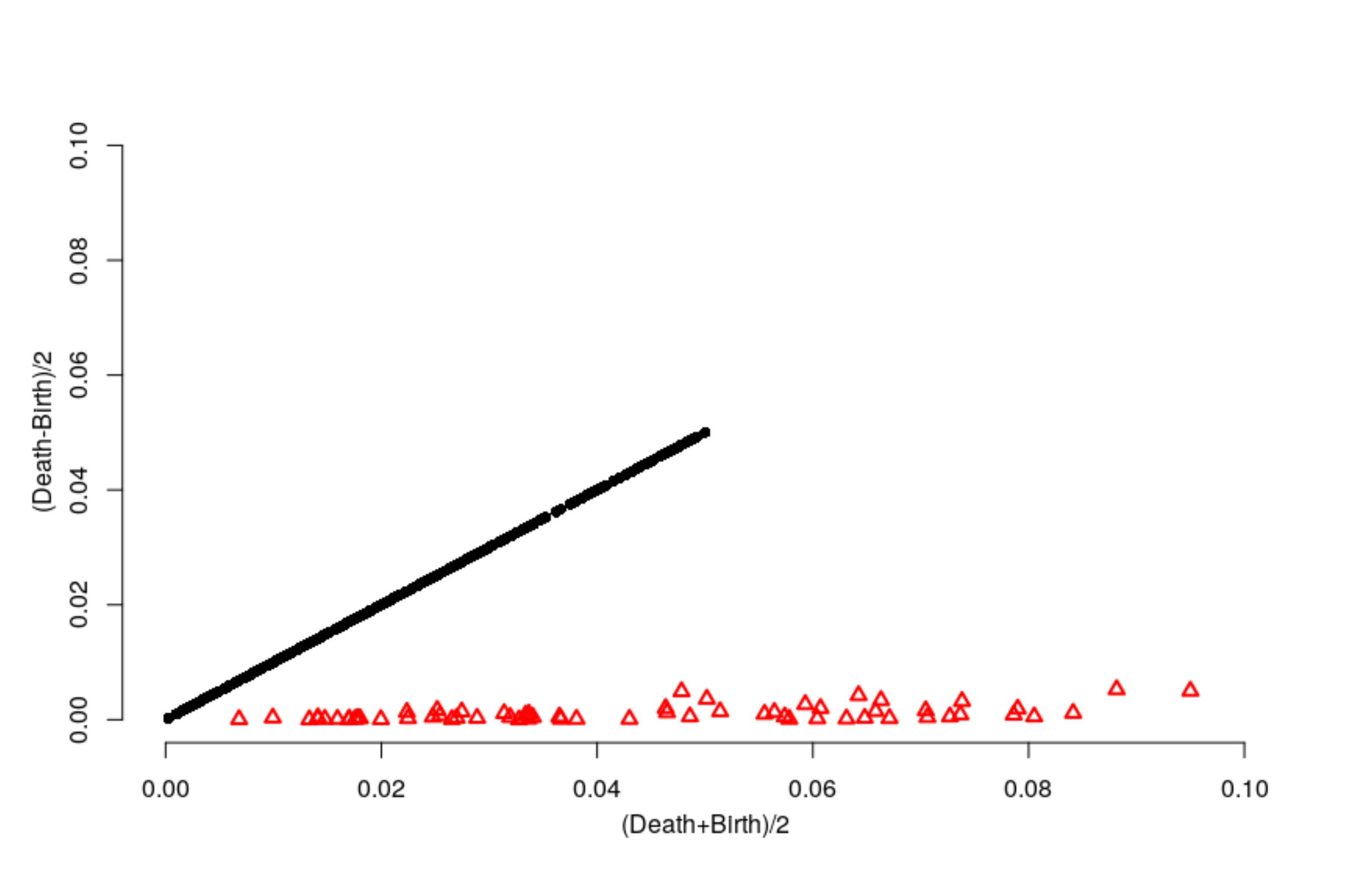}}
 \end{subfigure}\hfill
 
 \begin{subfigure}{1.0\textwidth}
  \centering
  \vspace{.5cm}
  \begin{turn}{90} 
   \hspace{1.75cm} WAE
  \end{turn}
  \frame{\includegraphics[scale=0.325]{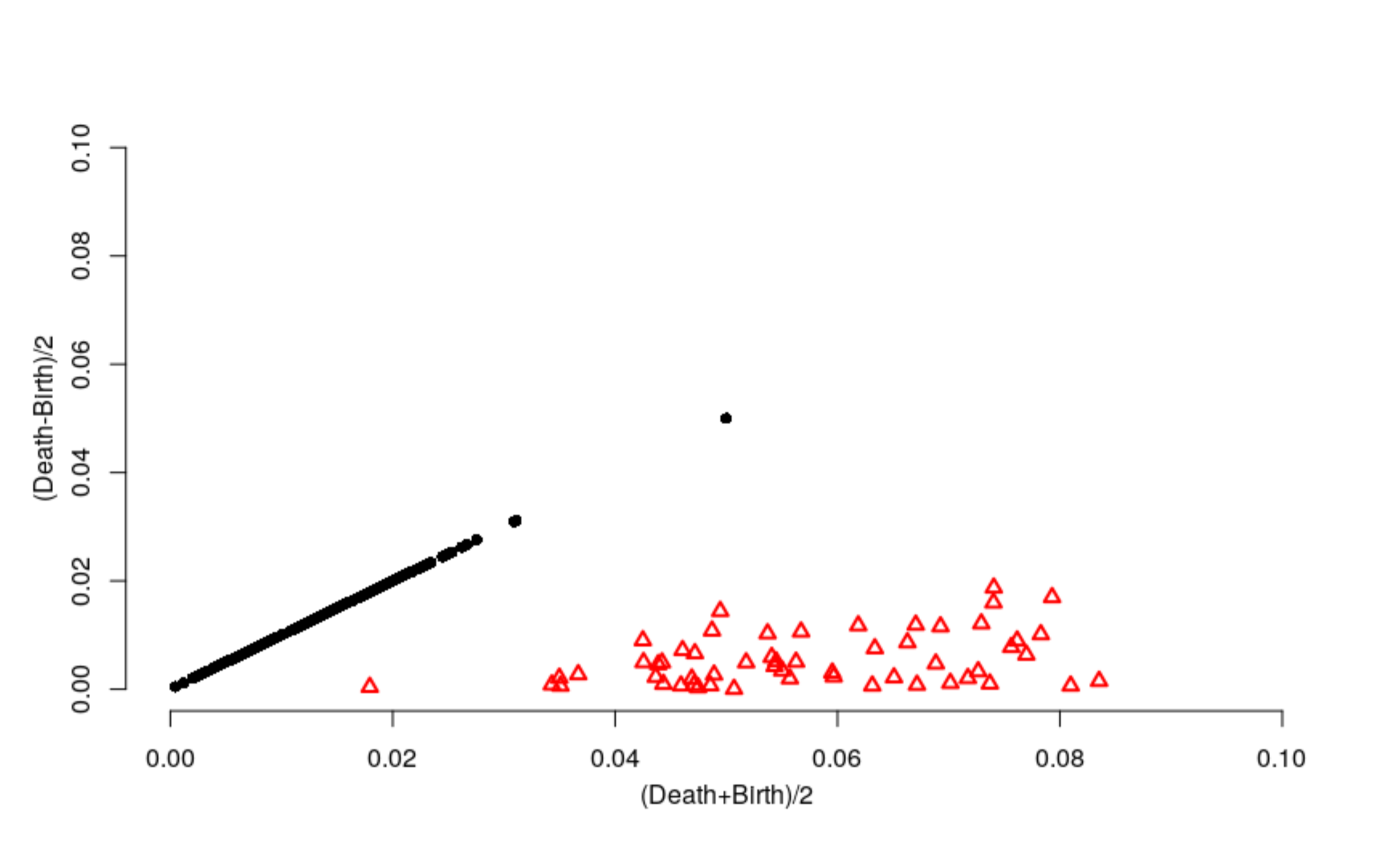}}
 \end{subfigure}\hfill
 
 \begin{subfigure}{1.0\textwidth}
  \centering
  \vspace{.5cm}
  \begin{turn}{90} 
   \hspace{1.75cm} VAE
  \end{turn}
  \frame{\includegraphics[scale=0.325]{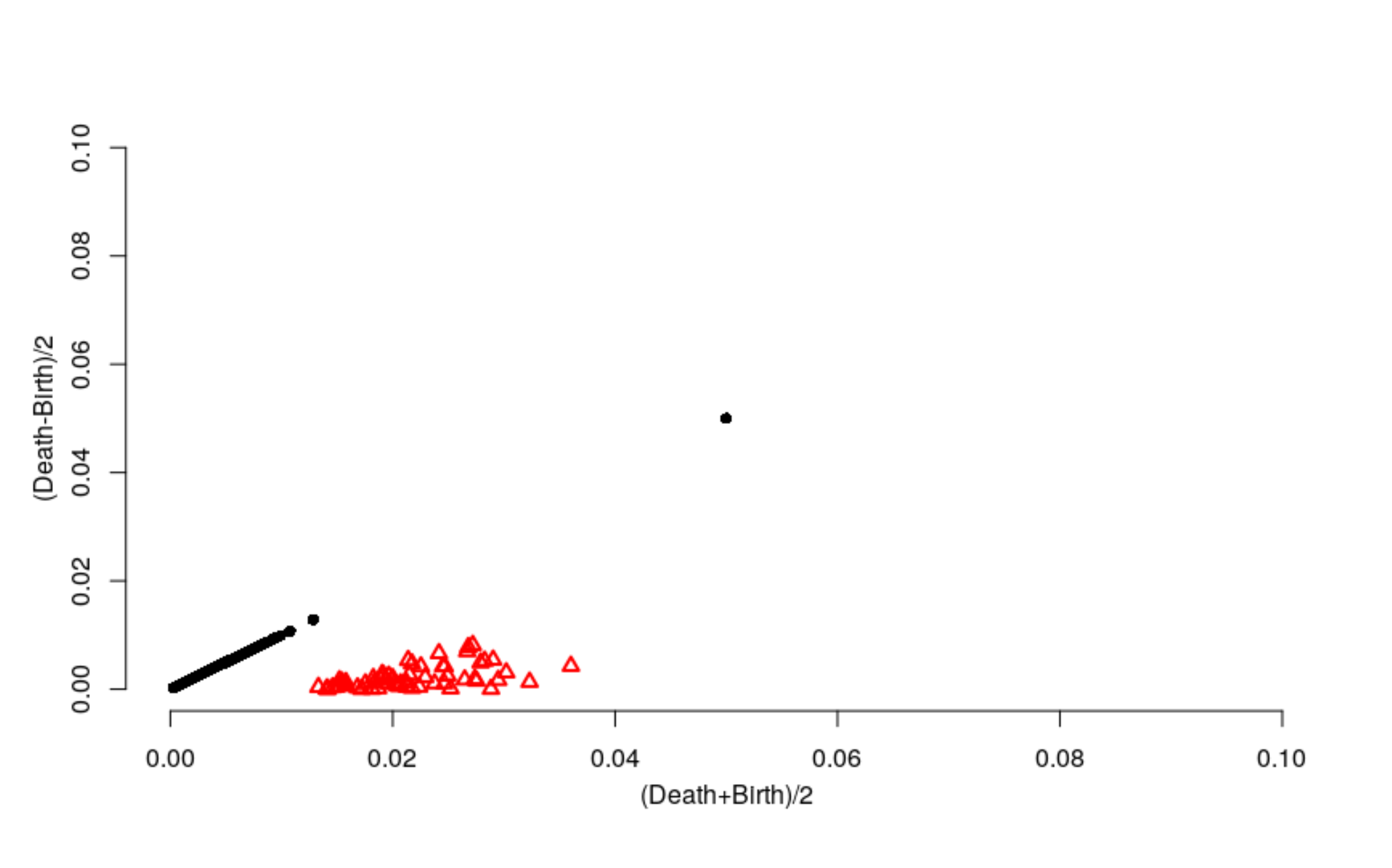}}
 \end{subfigure}\hfill
 
\caption[Rotated persistence diagrams of AE reconstructed distribution]{PHom-GeM applied to WAE and PHom-GeM applied to VAE's rotated persistence diagrams in comparison to the persistence diagram of the original sample showing the birth-death of the pairing generators of the iterated inclusions. Black points represent the 0-dimensional homology groups $H_0$ that is the connected components of the complex. Red triangles represent the 1-dimensional homology group $H_1$ that is the 1-dimensional features, the cycles.} 
\label{fig::persistence_plot}
\end{figure*}

\begin{figure*}[p]
\centering
 \begin{subfigure}{1.0\textwidth}
  \centering
  \begin{turn}{90} 
   \hspace{1.35cm} Original Sample
  \end{turn}
  \frame{\includegraphics[scale=0.28]{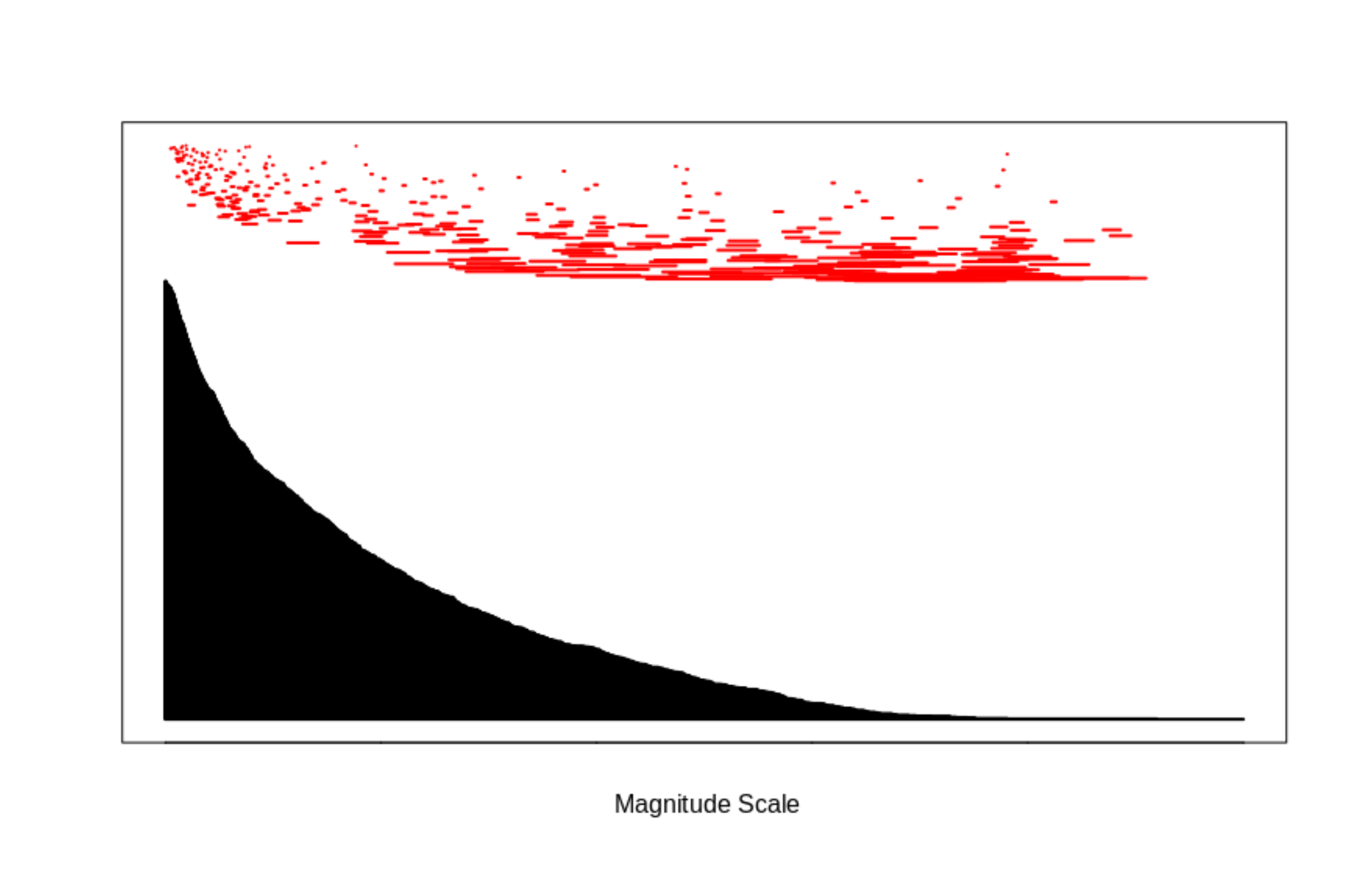}}
 \end{subfigure}\hfill
 
 \begin{subfigure}{1.0\textwidth}
  \centering
  \vspace{.5cm}
  \begin{turn}{90} 
   \hspace{2.1cm} WAE
  \end{turn}
  \frame{\includegraphics[scale=0.28]{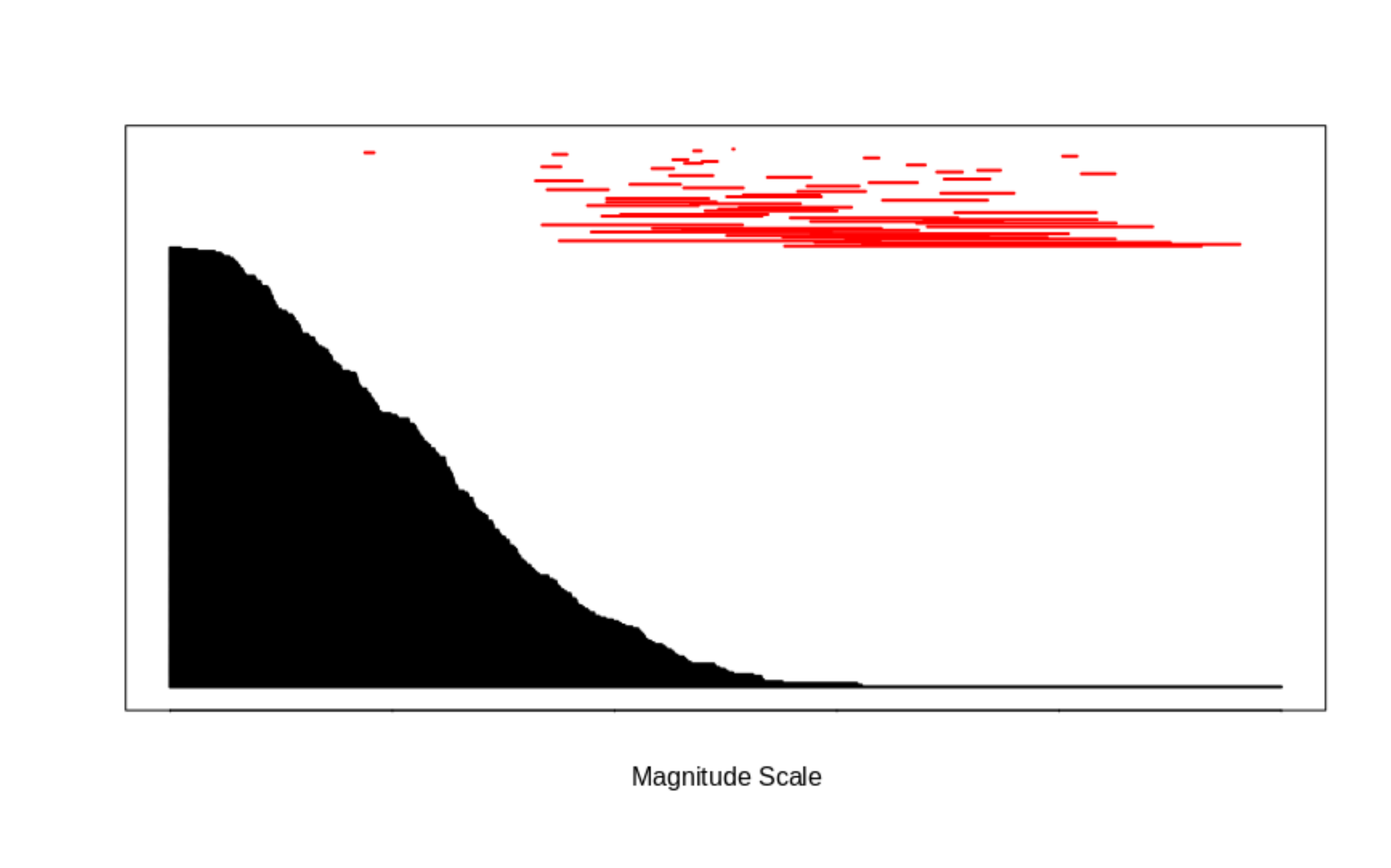}}
 \end{subfigure}\hfill
 
 \begin{subfigure}{1.0\textwidth}
  \centering
  \vspace{.5cm}
  \begin{turn}{90} 
   \hspace{2.1cm} VAE
  \end{turn}
  \frame{\includegraphics[scale=0.28]{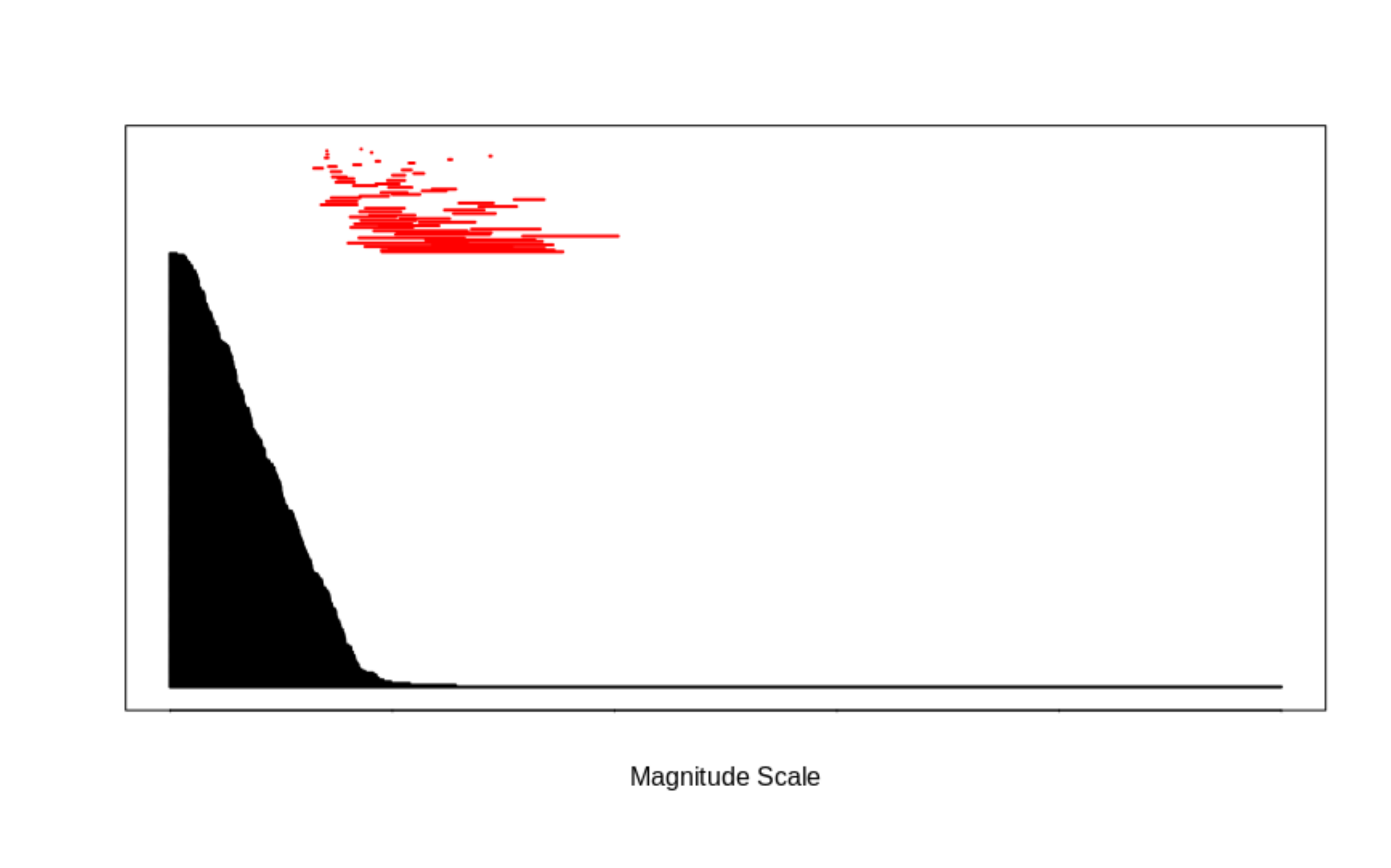}}
 \end{subfigure}\hfill
 
\caption[Barcode diagrams of AE reconstructed distribution]{PHom-GeM applied to WAE and PHom-GeM applied to VAE's barcode diagrams in comparison to the barcode diagram of the original sample based on the persistence diagrams of Figure \ref{fig::persistence_plot}. The barcodes diagrams are a simple way to represent the persistence diagrams. We refer to \cite{chazal2017tda} for further details on their generation.} 
\label{fig::latentplot}
\end{figure*}

\begin{figure*}[t!]
\centering
 \begin{subfigure}{.49\textwidth}
  \centering
  Original Samples \\ \vspace{.5cm}
  \begin{turn}{90} 
   \hspace{1.4cm} WAE
  \end{turn}
  \frame{\includegraphics[scale=0.3]{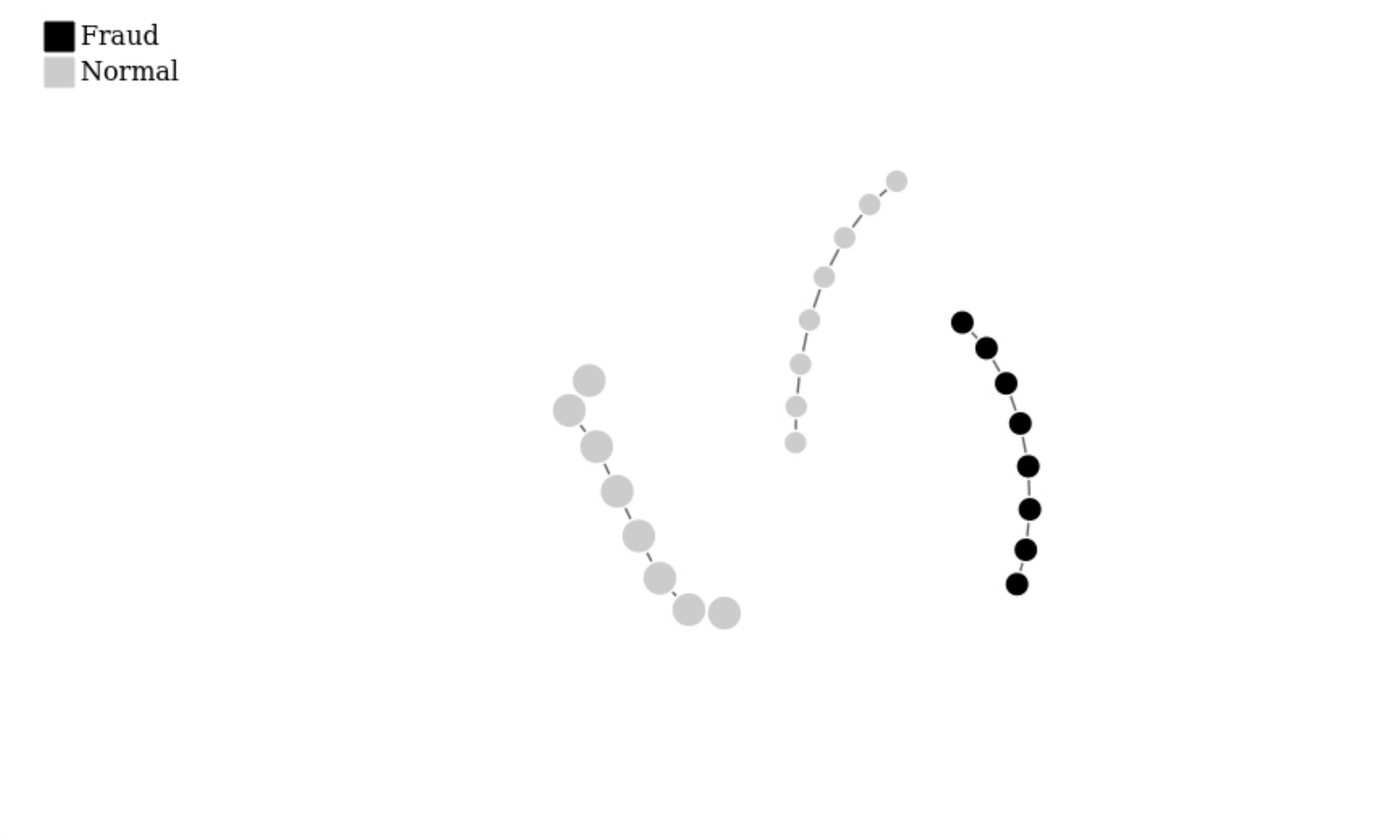}}
 \end{subfigure}\hfill
 \begin{subfigure}{.49\textwidth}
  \centering
  Reconstructed Samples \\ \vspace{.5cm}
  \frame{\includegraphics[scale=0.3]{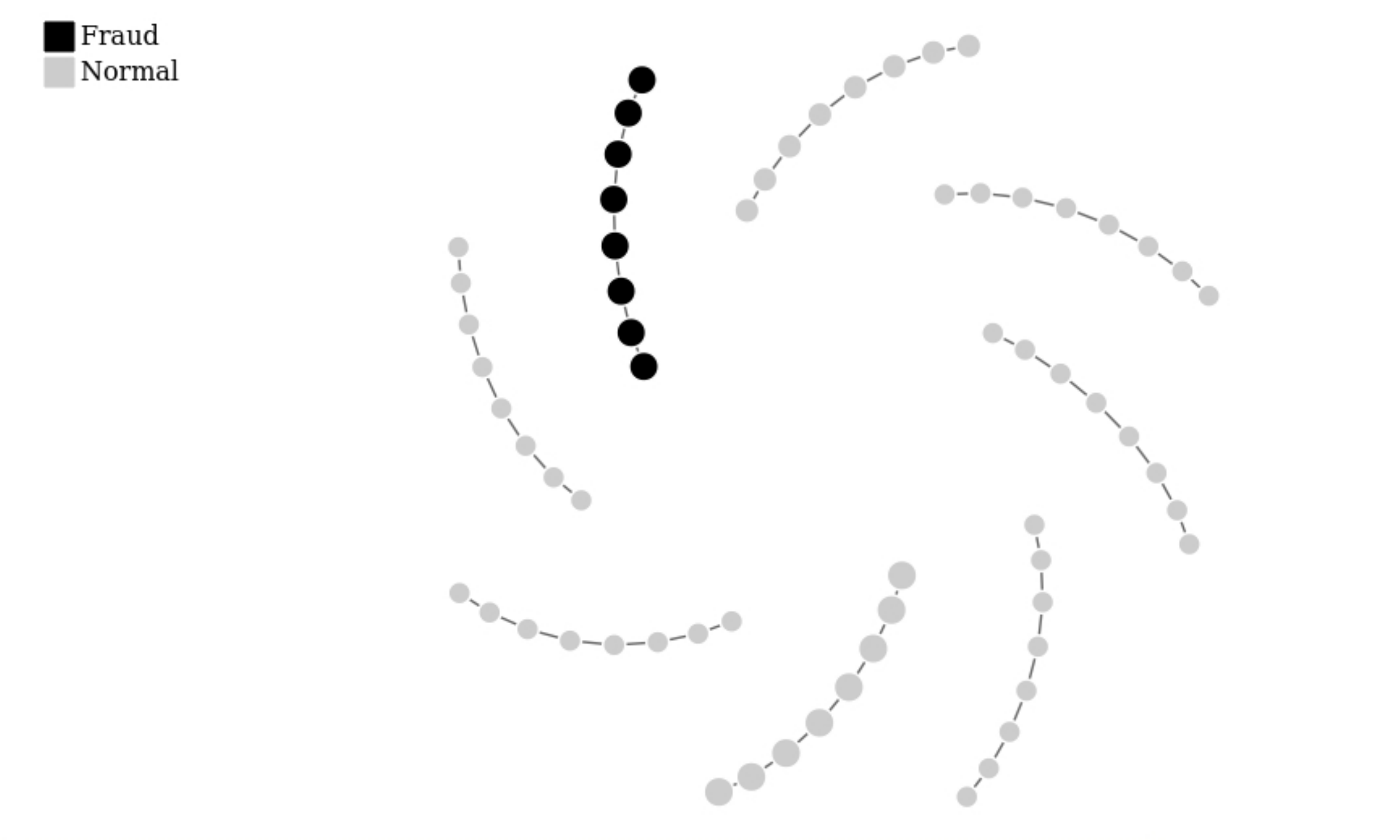}}
 \end{subfigure}\hfill
 
 \begin{subfigure}{.49\textwidth}
  \centering
  \vspace{.5cm}
  \begin{turn}{90} 
   \hspace{1.4cm} VAE
  \end{turn}
  \frame{\includegraphics[scale=0.3]{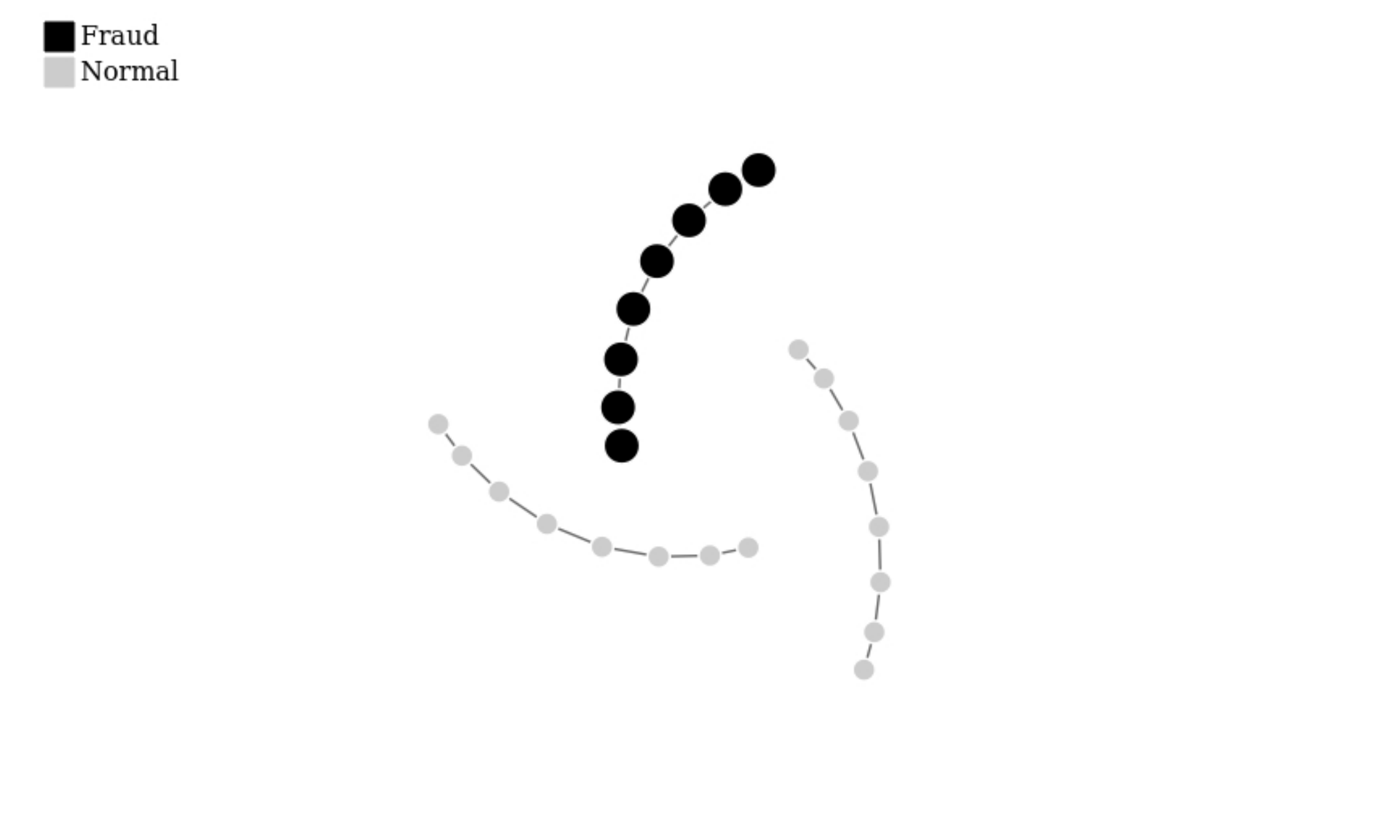}}
 \end{subfigure}\hfill
 \begin{subfigure}{.49\textwidth}
  \centering
  \vspace{.5cm}
  \frame{\includegraphics[scale=0.3]{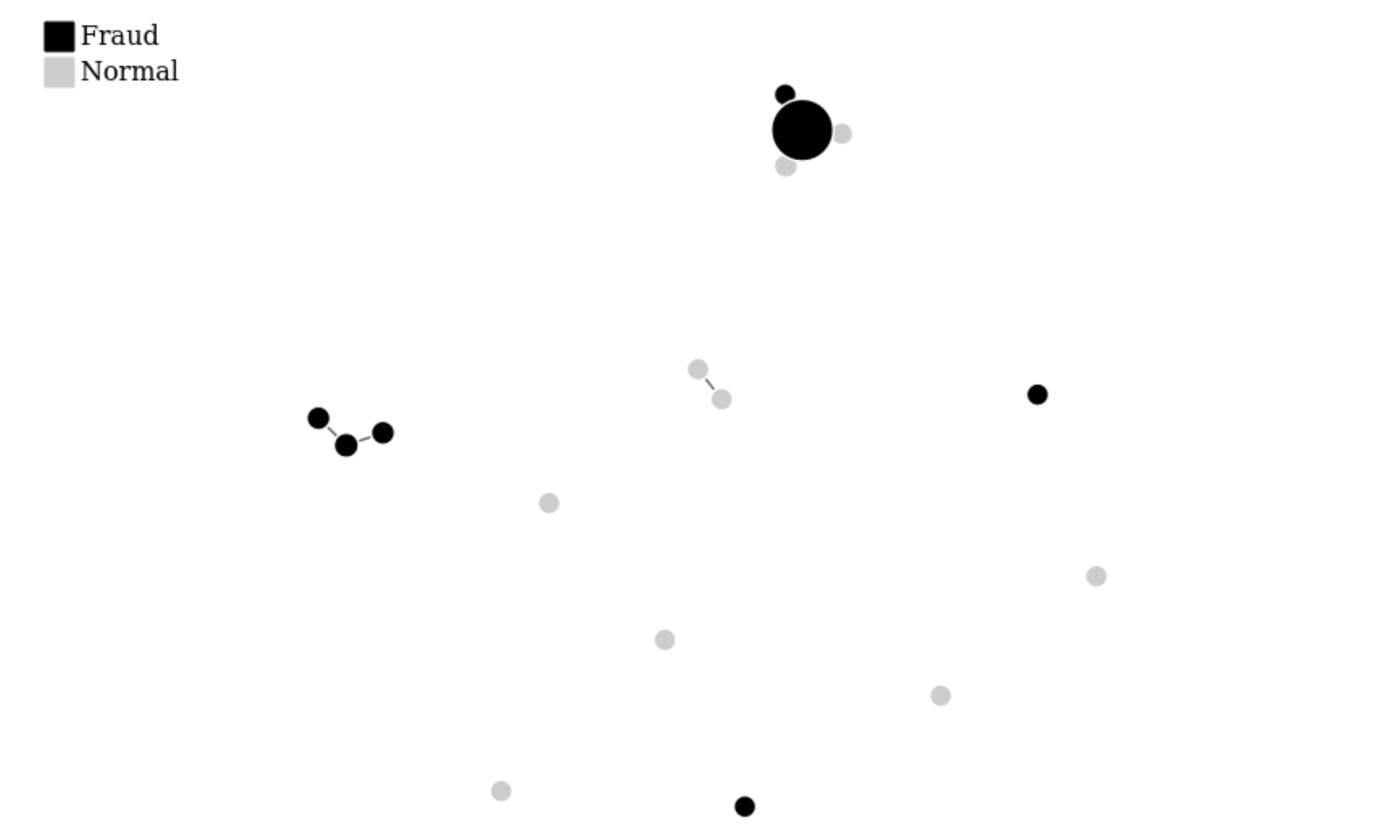}}
 \end{subfigure}\hfill
 
\caption[Topological view of the AE reconstructed distribution]{Topological representation 
of the reconstructed data distribution following $P_G(X|Z)$ for PHom-GeM applied to WAE and VAE. The encoding-decoding process of the WAE better preserves the topological features of the manifold $\mathcal{X}$.} \label{fig::reconstructedplot}
\end{figure*}

We represent the persistence diagrams and the barcode diagrams between the original and the reconstructed distributions, respectively $P_X$ and $P_G(X|Z)$, of the manifold $\mathcal{X}$ in Figures \ref{fig::persistence_plot} and \ref{fig::latentplot}. We can notice that the original and the reconstructed distributions are more widely distributed for WAE than for VAE. Additionally, the persistence diagram and the barcode diagram of WAE are qualitatively closer to those associated with the original data manifold $\mathcal{X}$. It means the topological features are better preserved for WAE than for VAE. Such result was expected based on the observation of the scattered distribution in Subsection \ref{subsec:latentsamples}. Consequently, it highlights a better encoding-decoding process thanks to the use of an optimal transport cost function in the case of WAE. Furthermore, in Figure \ref{fig::reconstructedplot}, a topological representation of the original and the reconstructed distributions is highlighted. We observe that the iterated inclusion chains are more similar for WAE than for VAE. For VAE, the inclusions of the reconstructed distribution are randomly scattered through the manifold without connected vertices. \\

In order to quantitatively assess the quality of the encoding-decoding process, we use the bottleneck distance between the persistent diagram of $\mathcal{X}$ and the persistent diagram of $G(Z)$ of the reconstructed data points. We recall the strength of the bottleneck distance is to measure quantitatively the topological changes in the data, either the true or the reconstructed data, while being insensitive to the scale of the analysis. Traditional distance measures fail to acknowledge this as they do not rely on the persistent homology and, therefore, can only reflect a measurement of the nearness relations of the data points without considering the overall shape of the data distribution. In Table \ref{tab::res_ae}, we notice the smallest bottleneck distance, and therefore, the best result, is obtained with WAE. It means WAE is capable to better preserve the topological features of the original data distribution than VAE including the nearness measurements and the overall shape.  \\

\begin{table}[b]
 \centering
 \caption[PHom-GeM bottleneck distance between WAE and VAE reconstructed samples]{Bottleneck distance (smaller is better) for PHom-GeM applied to WAE and VAE between the samples $X$ of the original manifold $\mathcal{X}$ and the reconstructed manifold $G(Z|X)$ for $Z\in\mathcal{Z}$. Because of OT, the WAE achieves better performance.}
 \label{tab::res_ae}
 \vspace{0.25cm}
 \begin{tabular}{cc|c}
  \toprule
  PHom-GeM on WAE \quad & \quad PHom-GeM on VAE \quad & \quad Difference (\%) \\
  \midrule
  \textbf{0.0788} & 0.0878 & 10.25  \\
  \bottomrule
 \end{tabular}
\end{table}

\subsection{Persistent Homology of Generated Adversarial Samples According to a Generative Distribution $P_G$}
\begin{figure*}[p]
\centering
 \begin{subfigure}{1.0\textwidth}
  \centering
  \begin{turn}{90} 
   \hspace{1.6cm} Original Sample
  \end{turn}
  \frame{\includegraphics[scale=0.4]{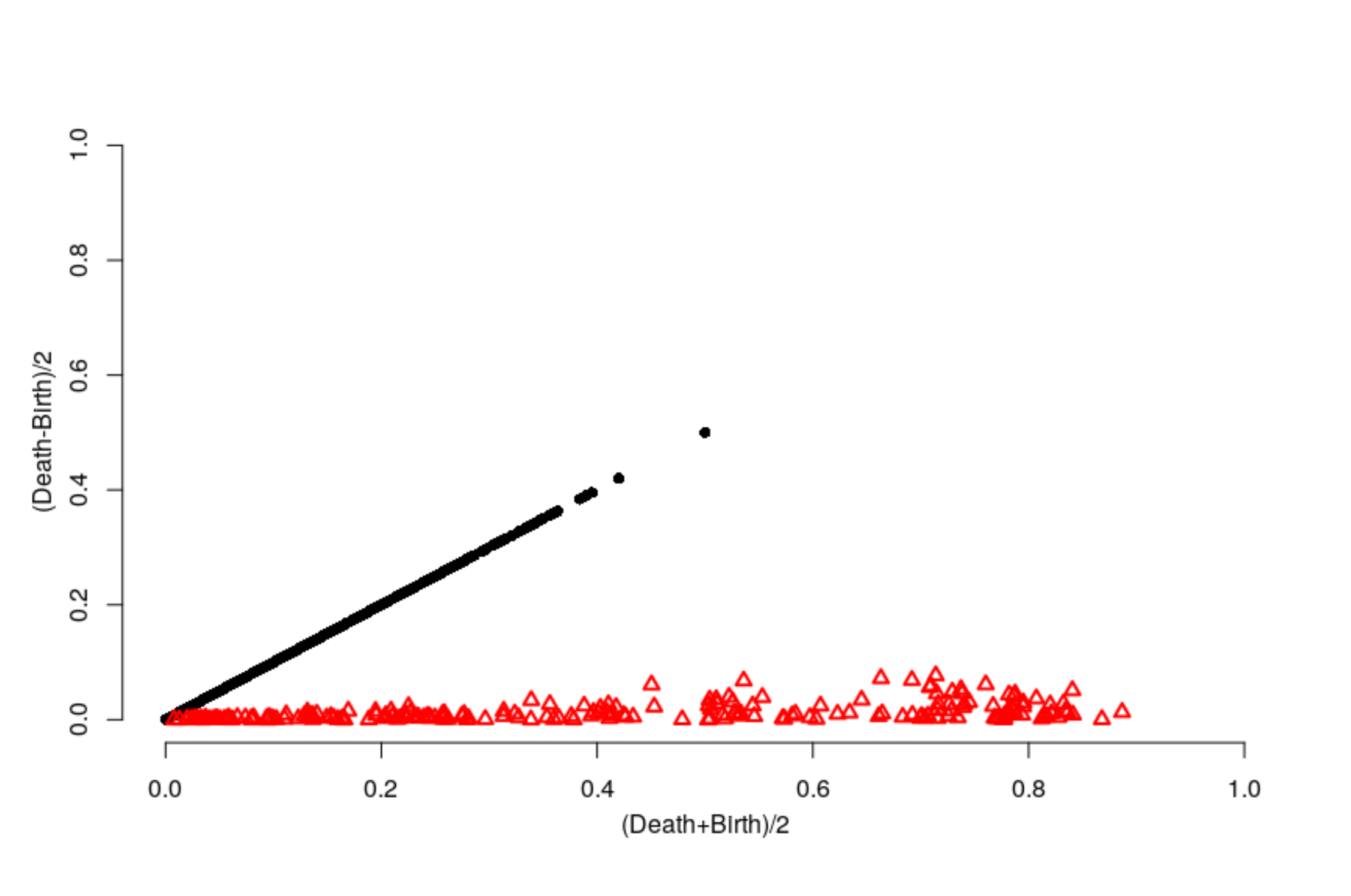}} 
 \end{subfigure}\hfill
 
 \begin{subfigure}{0.49\textwidth}
  \centering
  \vspace{0.5cm}
  \begin{turn}{90} 
   \hspace{1.15cm} GP-WGAN
  \end{turn}
  \frame{\includegraphics[scale=0.3]{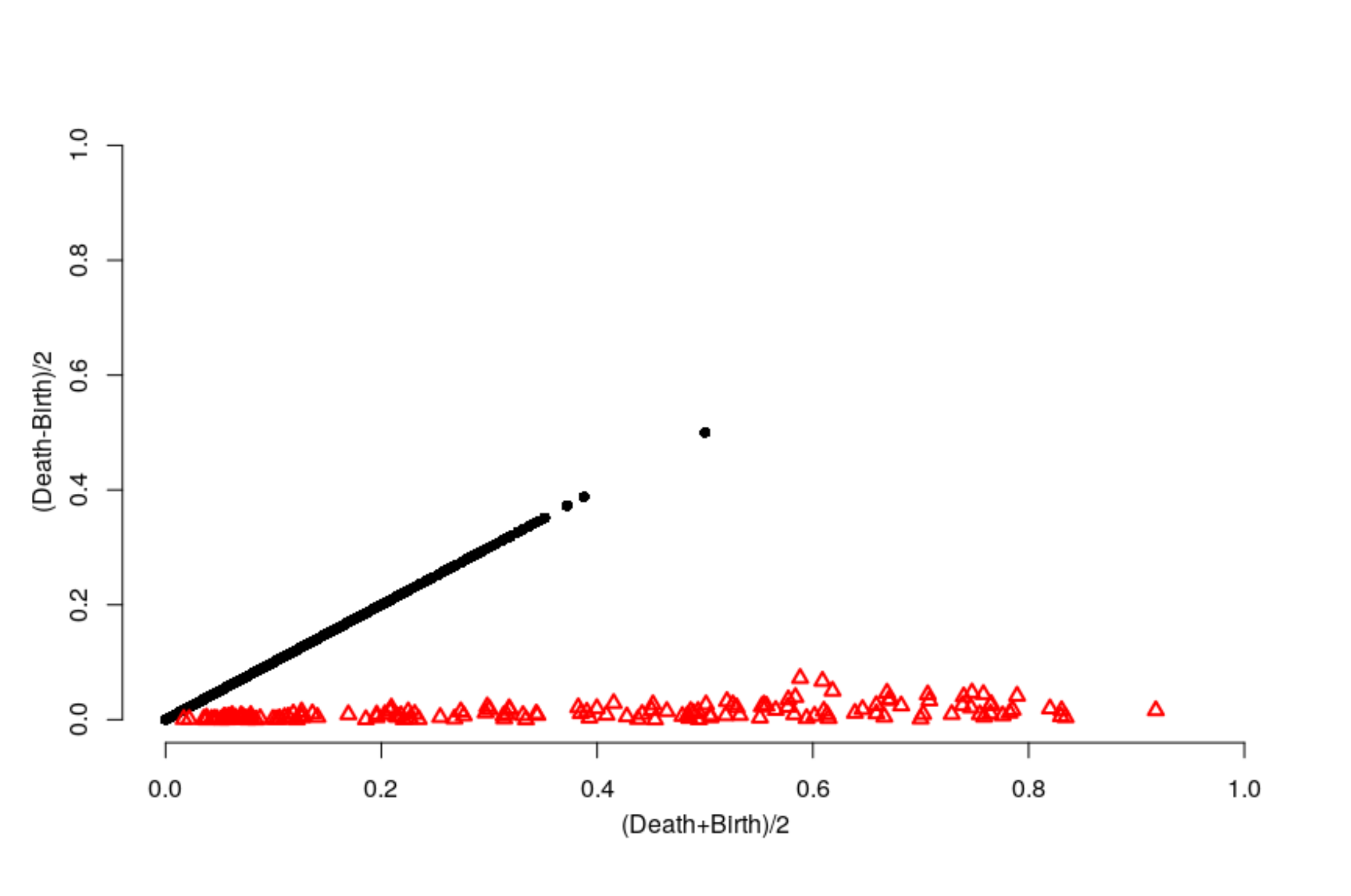}}
 \end{subfigure}\hfill
 \begin{subfigure}{0.49\textwidth}
  \centering
  \vspace{0.5cm}
  \begin{turn}{90} 
   \hspace{1.5cm} WGAN
  \end{turn}
  \frame{\includegraphics[scale=0.3]{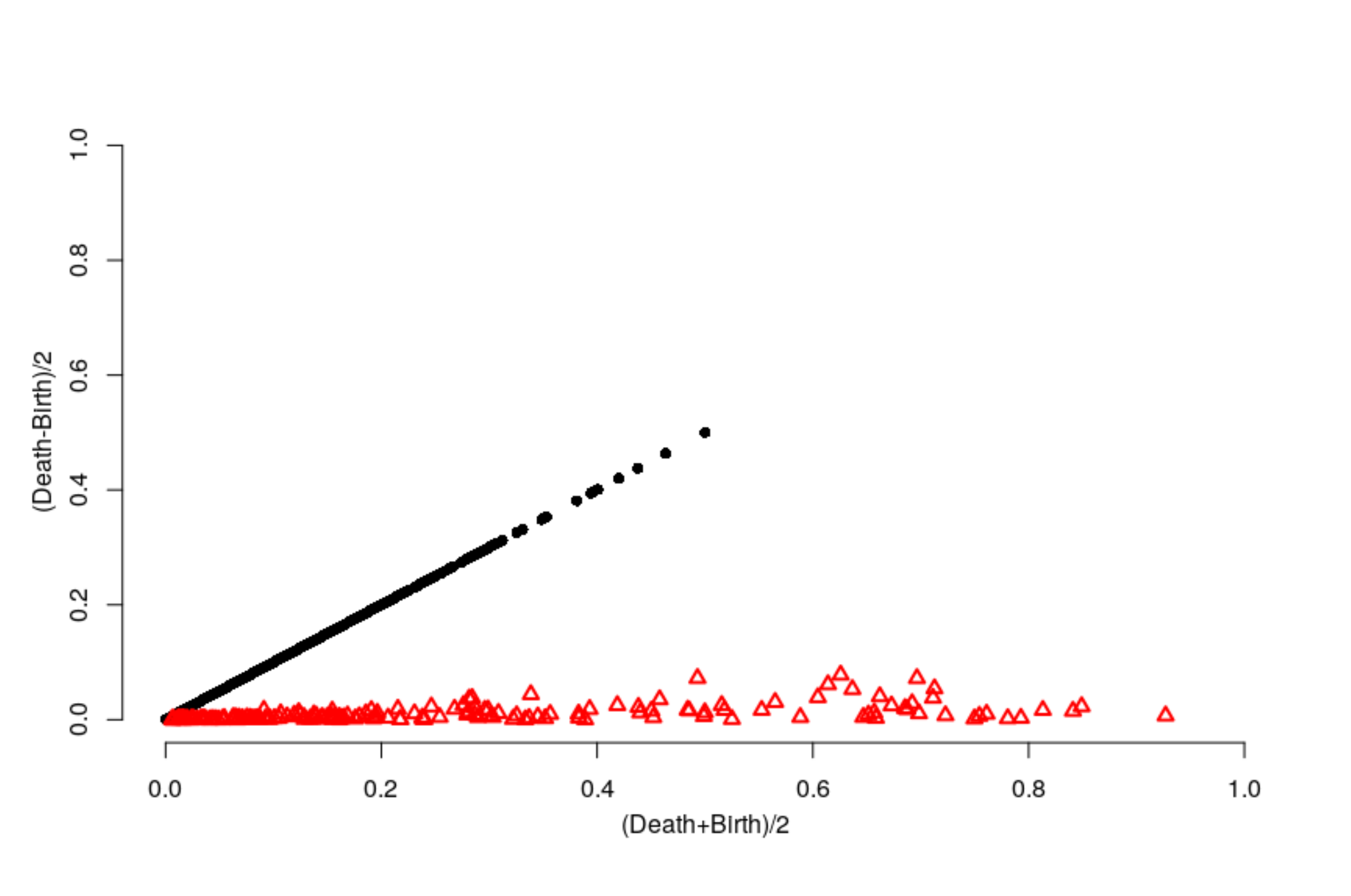}}
 \end{subfigure}\hfill
 
 \begin{subfigure}{0.49\textwidth}
  \centering
  \vspace{0.5cm}
  \begin{turn}{90} 
   \hspace{1.7cm} WAE
  \end{turn}
  \frame{\includegraphics[scale=0.3]{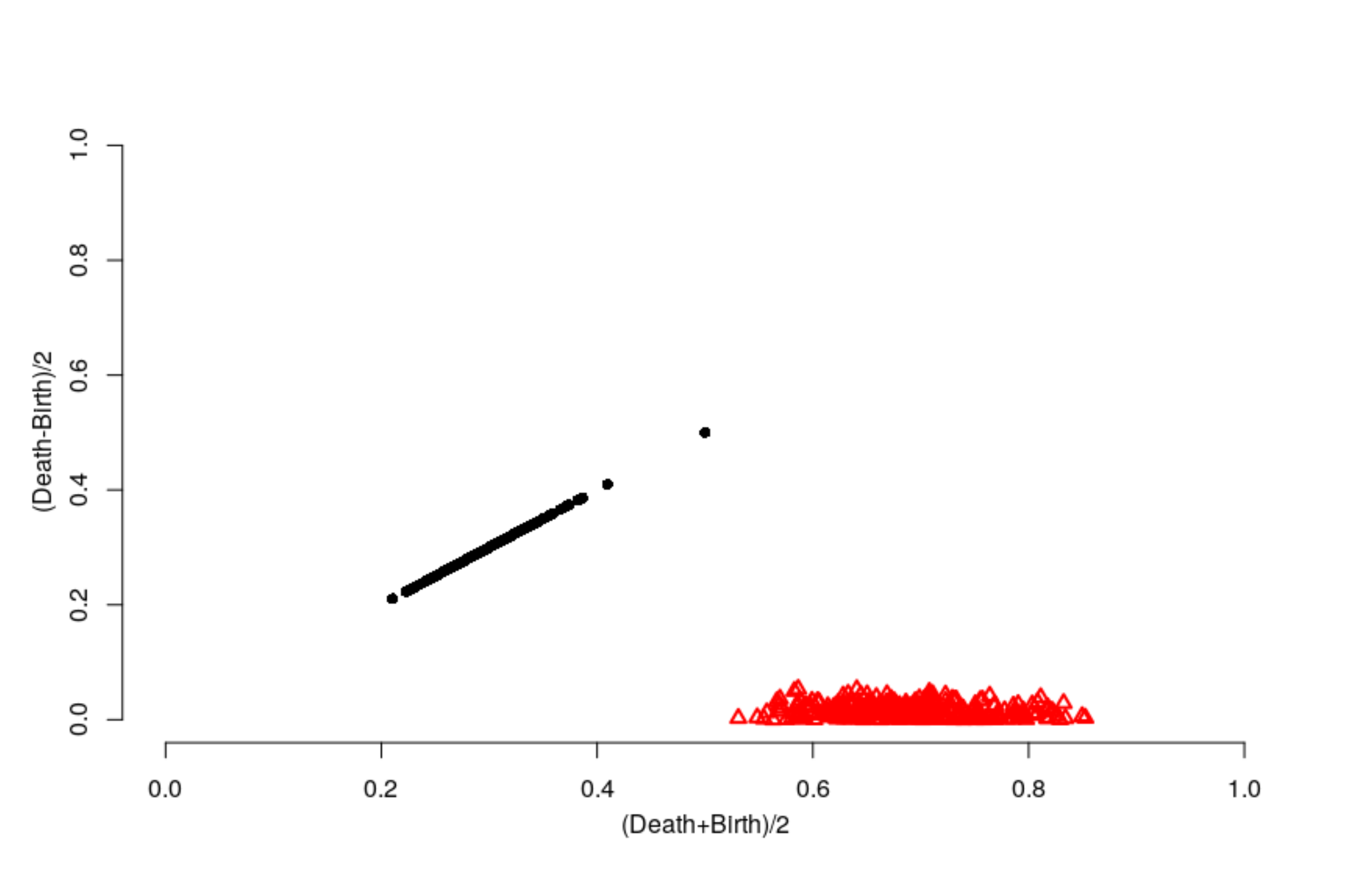}}
 \end{subfigure}\hfill
 \begin{subfigure}{0.49\textwidth}
  \centering
  \vspace{0.5cm}
  \begin{turn}{90} 
   \hspace{1.75cm} VAE
  \end{turn}
  \frame{\includegraphics[scale=0.3]{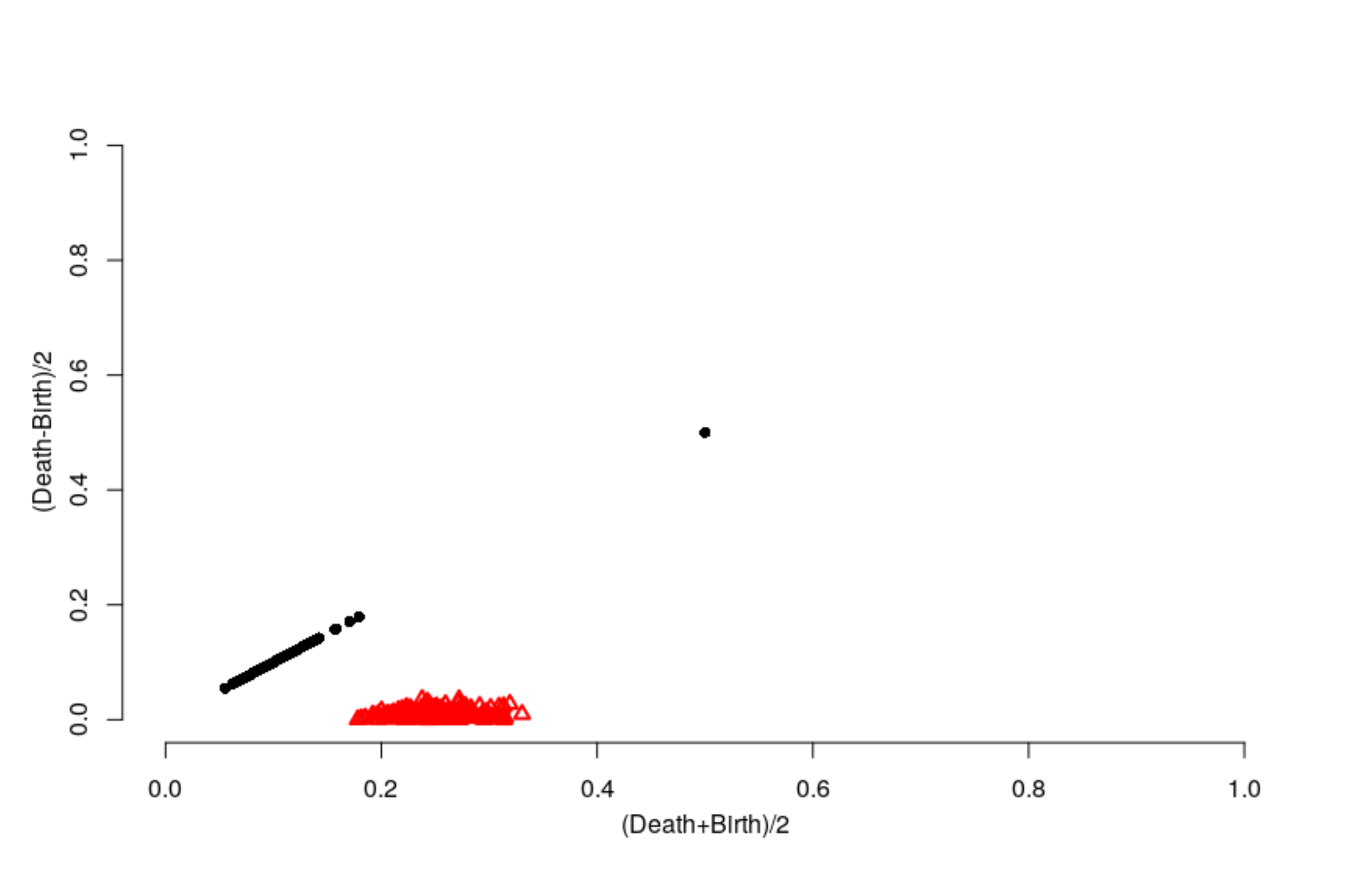}}
 \end{subfigure}\hfill
 
\caption[Rotated persistence diagrams of generated adversarial samples]{On top, the rotated persistence diagram of the original sample is represented. It illustrates the birth-death of the pairing generators of the iterated inclusions. In the persistence diagram, the black points represent the connected components of the complex and the red triangles the cycles. The GP-WGAN, WGAN, WAE and VAE's persistence diagrams allow to assess qualitatively, with the original sample persistence diagram, the persistent homology similarities between the generated and the original distribution, $P_G$ and $P_X$ respectively. 
} \label{fig::results_persdiag} 
\end{figure*}

\begin{figure*}[p]
\centering
 \begin{subfigure}{1.0\textwidth}
  \centering
  \begin{turn}{90} 
   \hspace{1.5cm} Original Sample
  \end{turn}
  \frame{\includegraphics[scale=0.3]{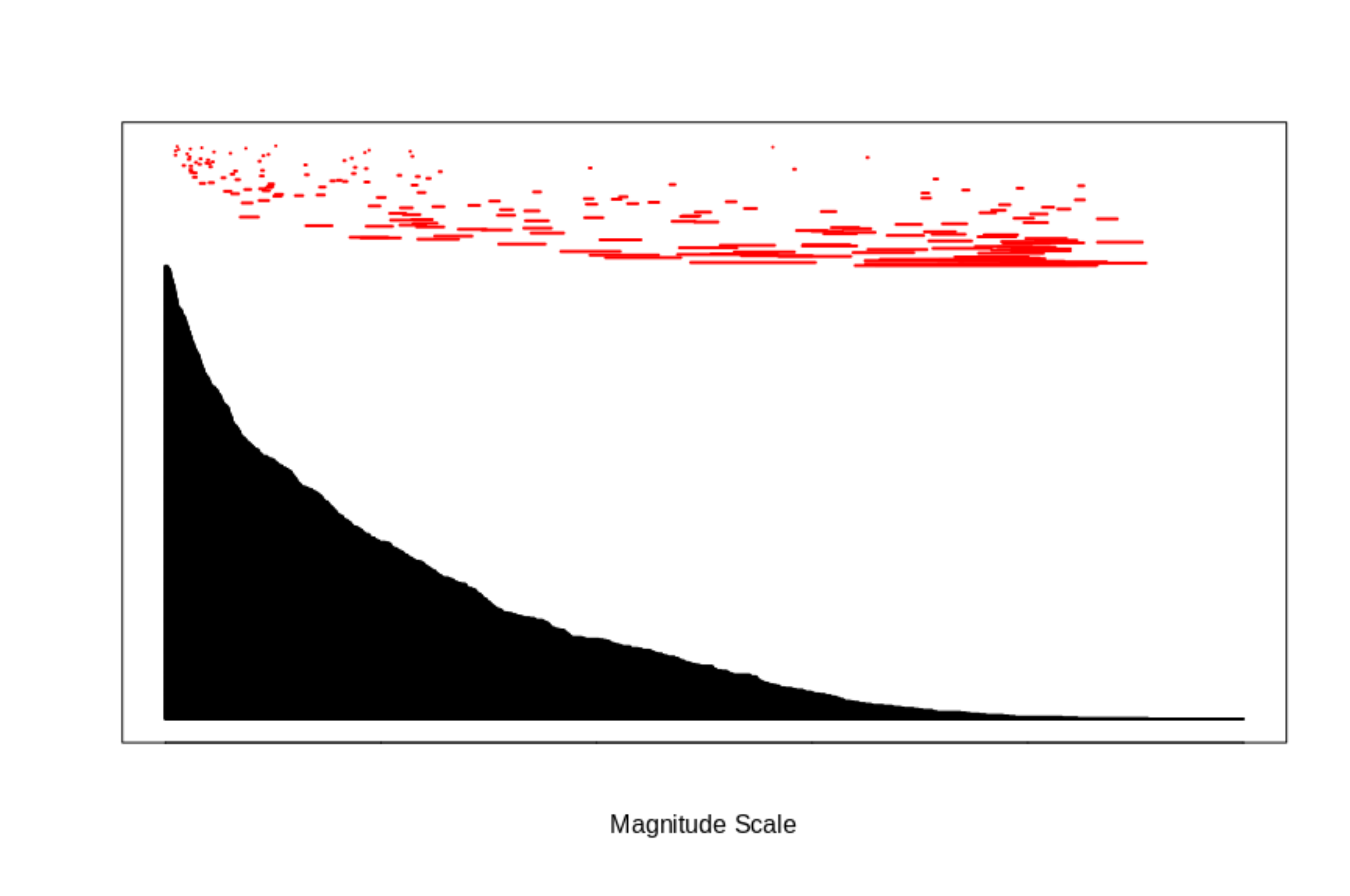}} 
 \end{subfigure}\hfill
 
 \begin{subfigure}{0.49\textwidth}
  \centering
  \vspace{.5cm}
  \begin{turn}{90} 
   \hspace{1.15cm} GP-WGAN
  \end{turn}
  \frame{\includegraphics[scale=0.225]{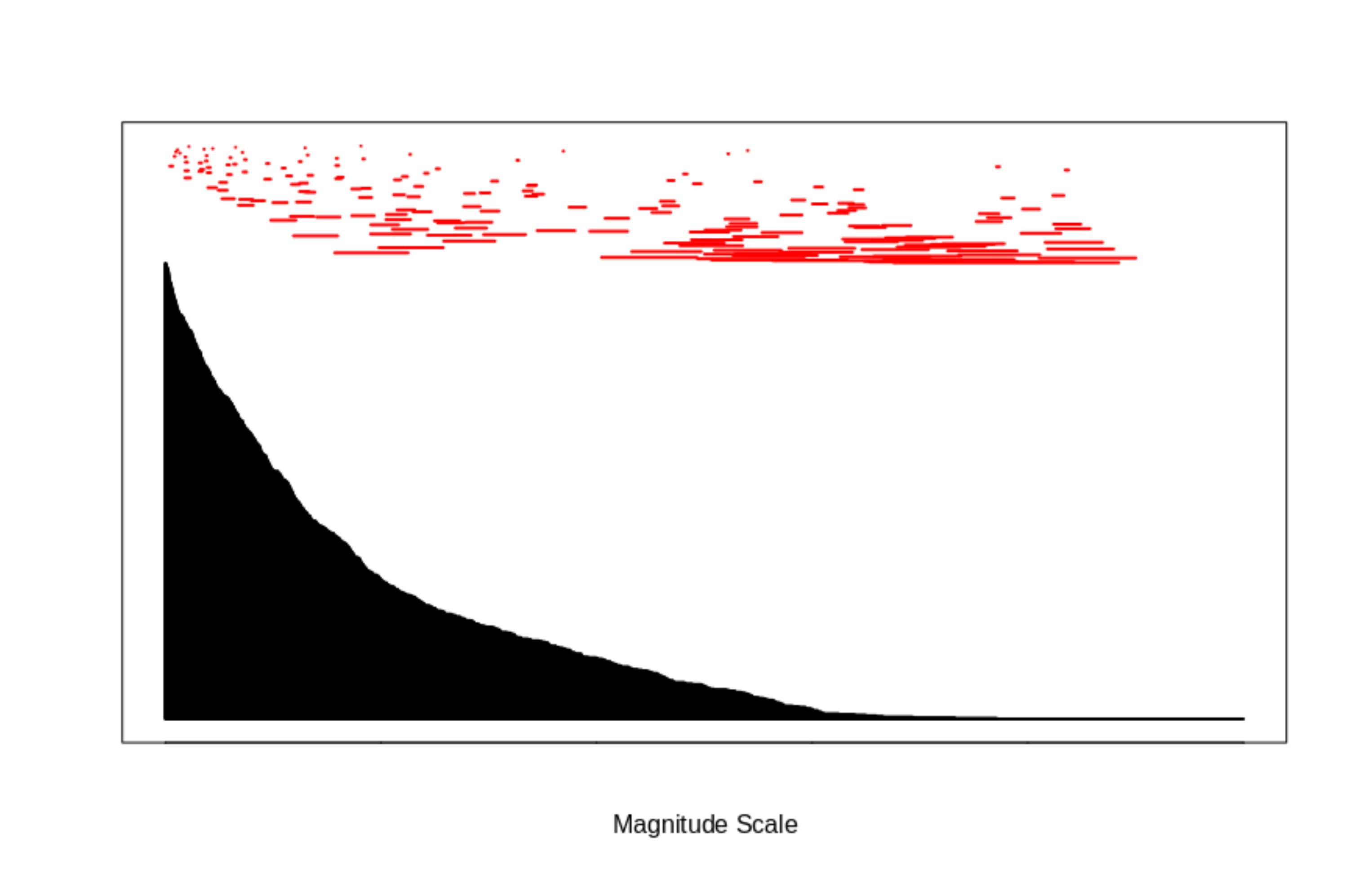}}
 \end{subfigure}\hfill
 \begin{subfigure}{0.49\textwidth}
  \centering
  \vspace{.5cm}
  \begin{turn}{90} 
   \hspace{1.45cm} WGAN
  \end{turn}
  \frame{\includegraphics[scale=0.225]{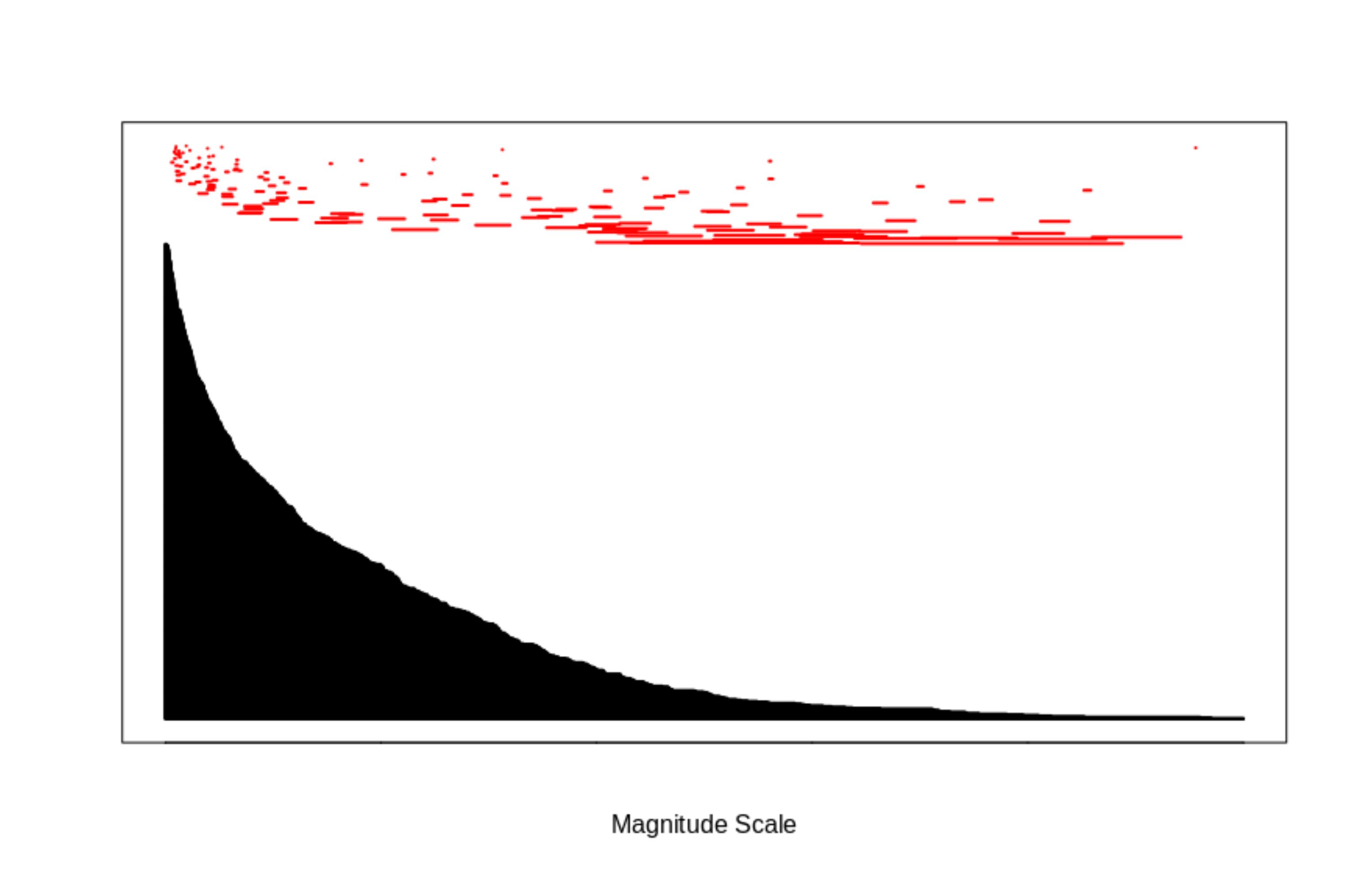}}
 \end{subfigure}\hfill
 
 \begin{subfigure}{0.49\textwidth}
  \centering
  \vspace{.5cm}
  \begin{turn}{90} 
   \hspace{1.75cm} WAE
  \end{turn}
  \frame{\includegraphics[scale=0.225]{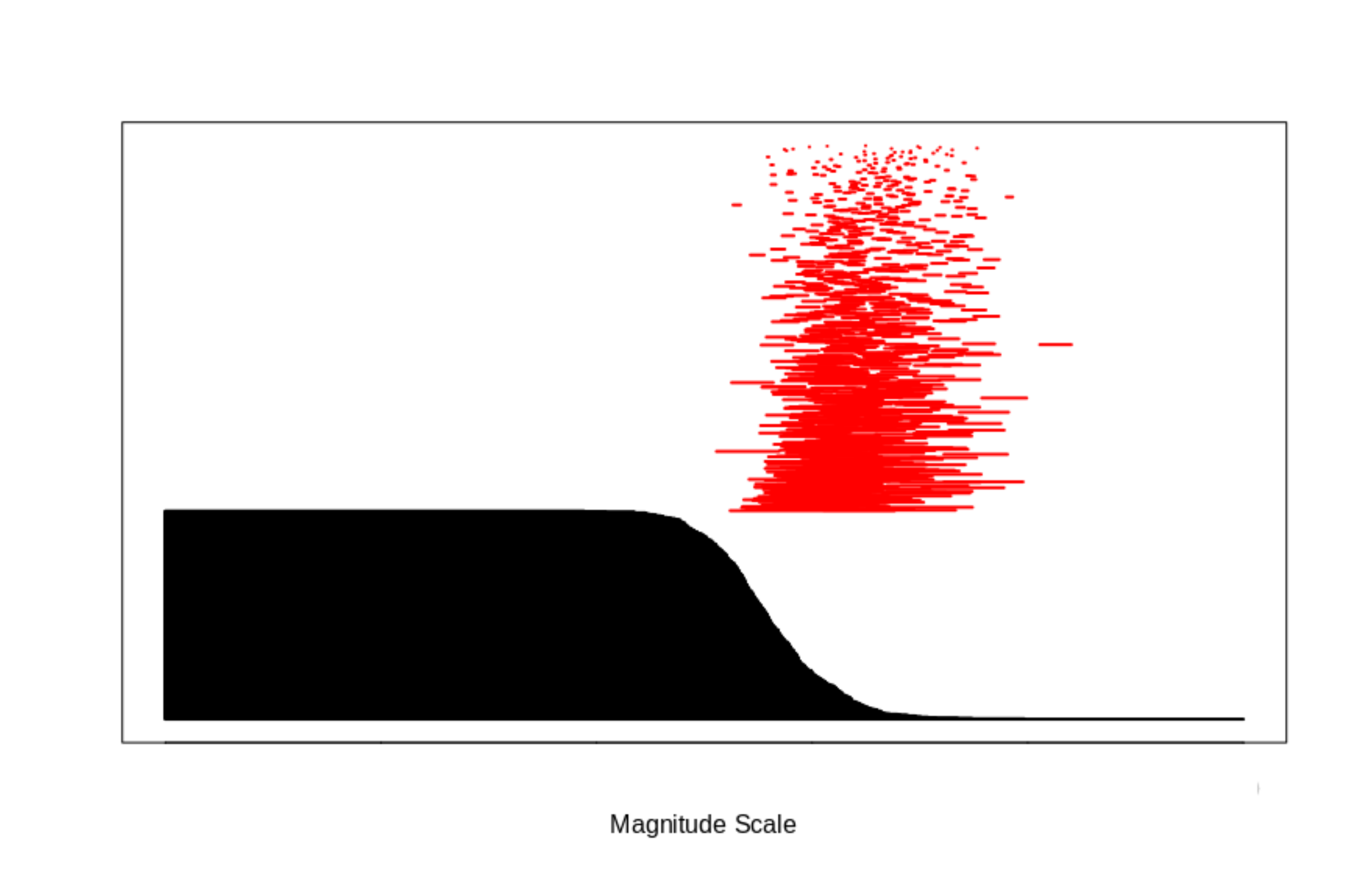}}
 \end{subfigure}\hfill
 \begin{subfigure}{0.49\textwidth}
  \centering
  \vspace{.5cm}
  \begin{turn}{90} 
   \hspace{1.75cm} VAE
  \end{turn}
  \frame{\includegraphics[scale=0.225]{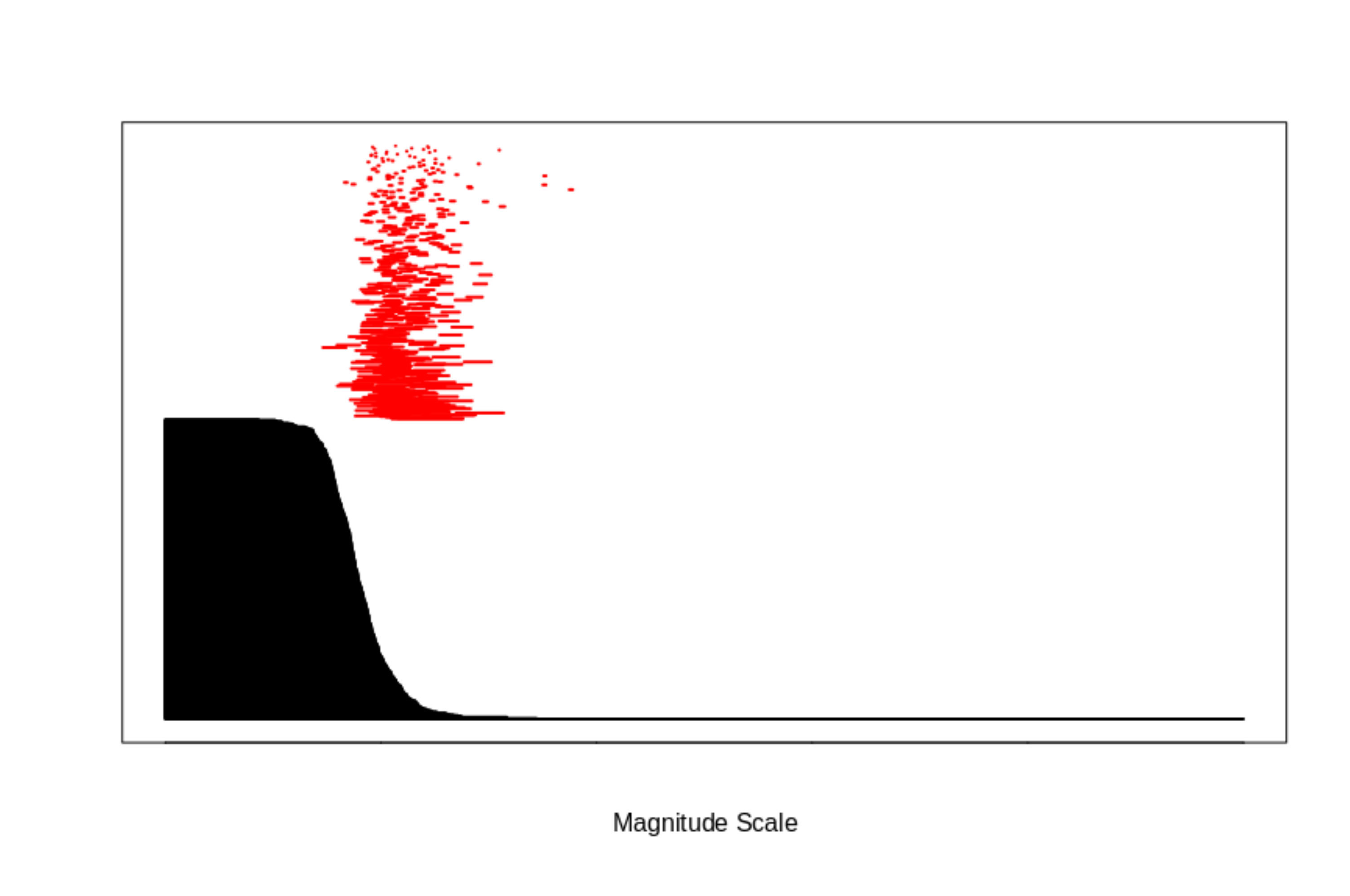}}
 \end{subfigure}\hfill
 
\caption[Barcode diagrams of generated adversarial samples]{On top, the barcode diagrams of the original sample is represented. It illustrates the birth-death of the pairing generators of the iterated inclusions. In the barcode diagram, the black lines represent the connected components of the complex and the red lines the cycles. The GP-WGAN, WGAN, WAE and VAE's barcode diagrams allow to assess qualitatively, with the original sample barcode diagram, the persistent homology similarities between the generated and the original distribution, $P_G$ and $P_X$ respectively. 
} \label{fig::results_barcodes} 
\end{figure*}

In this third experiment, we reach the core of our PHom-GeM's contribution. We evaluate on the four generative models, GP-WGAN, WGAN, WAE and VAE, with the persistent homology both the qualitative and quantitative topological properties of the generated adversarial samples with respect to the generative model distribution $P_G$. The adversarial samples are compared to the original samples $X$ of the manifold $\mathcal{X}$. On the top of Figures \ref{fig::results_persdiag} and \ref{fig::results_barcodes}, the rotated persistence and the barcode diagrams of the original sample $\mathcal{X}$ are highlighted. In the persistence diagram, as previously mentioned, black points represent the 0-dimensional homology groups $H_0$, the connected components of the complex. The red triangles represent the 1-dimensional homology group $H_1$, the 1-dimensional features known as cycles or loops. The barcode diagram is a simple way of representing the information contained in the persistence diagram. The generated distribution $P_G$ of GP-WGAN is the closest to the distribution $P_X$ followed by WGAN, WAE and VAE. The spectrum of the barcodes of GP-WGAN, effectively, is very similar to the original sample's spectrum as well as denser on the right. On the opposite, the WAE and VAE's distributions $P_G$ are not able to reproduce all of the features contained in the original distribution, underlined by the narrower and incomplete barcode spectrum.  \\ 

In order to quantitatively assess the quality of the generated distributions, we use the bottleneck distance between the persistence diagram of $\mathcal{X}$ and the persistence diagram of $G(Z)$ of the generated data points.  We recall the strength of the bottleneck distance is to measure quantitatively the topological changes in the data, either the true or the generated data, while being insensitive to the scale of the analysis. Traditional distance measures fail to acknowledge this as they do not rely on the persistent homology and, therefore, can only reflect a measurement of the nearness relations of the data points without considering the overall shape of the data distribution. Using the bootstrapping technique of \cite{friedman2001elements}, we successively randomly select data samples contained in the manifolds  $\mathcal{X}$ and $\mathcal{\widetilde{X}}$ to evaluate their bottleneck distance. The total number of selected samples is at least 85\% of the total number of points contained in the manifolds $\mathcal{X}$ and $\mathcal{\widetilde{X}}$ to ensure a reliable statistical representation. In Table \ref{tab::res_gem}, we highlight the mean value of the bottleneck distance for a 95\% confidence interval. We also underline the lower and the upper bounds of the 95\% confidence interval for each generative model. Confirming the visual observations, we notice the smallest bottleneck distance, and therefore, the best result, is obtained with GP-WGAN, followed by WGAN, WAE and VAE. It means, in our configuration, GP-WGAN is capable to generate data distribution sharing the most topological features with the original data distribution, including the nearness measurements and the overall shape. It confirms topologically on a real-world data set the claims addressed in \cite{gulrajani2017improved} of superior performance of GP-WGAN against WGAN. Furthermore, the performance of the AE cannot match the generative performance achieved by the GANs. However, the WAE, that relies on optimal transport theory, achieves better generative distribution in comparison to the popular VAE.

\begin{table}[t] 
\centering 
\caption[PHom-GeM bottleneck distance between original and generated adversarial samples]{Bottleneck distance (smaller is better) with 95\% of confidence interval between the samples $X$ of the original manifold $\mathcal{X}$ and the generated samples $\widetilde{X}$ of the manifold $\mathcal{\widetilde{X}}$. Because of the Wasserstein distance and gradient penalty, GP-WGAN achieves better performance.} \label{tab::res_gem} 
\scalebox{1}{ 
\begin{tabular}{cccc} 
  \toprule 
  Gen. Model & Mean Value & Lower Bound & Upper Bound \\ 
  \midrule
  GP-WGAN & \textbf{0.0711} & \textbf{0.0683} & \textbf{0.0738}  \\ 
  WGAN & 0.0744 & 0.0716 & 0.0772  \\ 
  WAE & 0.0821 & 0.0791 & 0.0852  \\ 
  VAE & 0.0857 & 0.0833 & 0.0881  \\ 
  \bottomrule 
\end{tabular} 
} 
\vspace{1cm}
\end{table}

\section{Conclusion} \label{sec::ccl} 
Building upon optimal transport and unsupervised learning, we introduced PHom-GeM, Persistent Homology for Generative Models, a new characterization of the generative manifolds that uses the topology and the persistent homology to highlight the manifold features and the scattered generated distributions. We discuss the relations of GP-WGAN, WGAN, WAE and VAE in the context of unsupervised learning. Furthermore, by relying on the persistent homology, the bottleneck distance has been introduced to estimate quantitatively the alteration of the topological features occurring during the encoding-decoding process and the topological features similarities between the original distribution, the AE latent manifold distributions and the generated distributions of the generative models. It is a specificity that current traditional distance measures fail to acknowledge. We used a challenging imbalanced real-world open data set containing credit card transactions, capable of illustrating the scattered generated data distributions of the generative models, particularly suitable for the banking industry. We conducted experiments showing the performance of PHom-GeM on the four generative models GP-WGAN, WGAN, WAE and VAE. We highlighted the scattered distributions of the WAE and VAE's latent manifold, an AE limitation that will be very likely addressed successively in future research. 
We furthermore showed the superior persistent homology performance of GP-WGAN in comparison to the other generative models as well as the superior performance of WAE over VAE in the context of the generation of adversarial samples.  \\

Future work will include further exploration of the topological features such as the influence of the simplicial complex. We will additionally address the use of an optimal transport persistent homological distance measure, such as the Wasserstein distance, to increase the accuracy of the topological measurements. Finally, we leave for future research how to back-propagate the persistent homology information to the objective loss function of the generative models to improve their training and the generation of adversarial samples.

\chapter{Accurate Tensor Resolution for Financial Recommendations}

The new financial European regulations, such as the revised Payment Service Directive (PSD2) \cite{psdII}, are changing the retail banking services by welcoming new joiners to promote competition. Noticeably, retail banks have lost the exclusive privilege of proposing and managing financial solutions, as for instance, the personal finance management on mobile banking applications or the credit cards distribution for the transaction payment solutions. 
Consequently, they are now looking to optimize their resources to propose a higher quality of financial services using recommender engines. The recommendations are moving from a client-request approach to a bank-proposing approach where the banks dynamically offers new financial opportunities to their clients. By being able to estimate the financial awareness of their clients or to predict their clients' financial transactions in a context-aware environment, the banks can dynamically offer the most appropriate financial solutions, targeting their clients' future needs. 
In this context, we focus on the tensor decomposition \cite{kolda2009tensor}, a collection of factorization techniques for multidimensional arrays, to build predictive financial recommender engines. Tensor decomposition are among the most general and powerful tools to perform multi-dimensional personalized recommendations. However, due to their linear algebra complexity, tensor decomposition requires accurate convex optimization schemes.  \\
Therefore, in this chapter, we describe our novel resolution algorithm, VecHGrad for Vector Hessian Gradient, for accurate and efficient stochastic resolution over all existing tensor decomposition, specifically designed for complex decomposition. We highlight the VecHGrad's performance in comparison to state of the art machine learning optimization algorithms and popular tensor resolution scheme. The description of VecHGrad is then followed by its direct application to the CP tensor decomposition \cite{harshman1970foundations, carroll1970analysis}, a multidimensional matrix decomposition that factorizes a tensor as the sum of rank-one tensors, for the predictions of the clients' actions. By predicting the clients' actions, the retail banks can adopt an aggressive approach to propose adequate new financial products to their clients. Finally, we will address the monitoring and the prediction of user-device authentication in mobile banking to estimate the financial awareness of the clients. Because of the imbalance between the number of users and devices, we will use the VecHGrad optimization algorithm to solve the PARATUCK2 tensor decomposition \cite{harshman1996uniqueness}, which expresses a tensor as a multiplication of matrices and diagonal tensors to represent and to model asymmetric relationships.  \\

\section{Motivation} \label{sec::intro3}
Endorsed by the  European objectives to promote the financial exchanges between the Euro members, new regulatory directives are now applicable such as the  Revised Payment Service Directive (PSD2). They promote more control, more transparency and more competition while trying to reduce the contagion risk in the event of a financial crisis such as in 2008. The financial services, such as Personal Finance Management (PFM) to monitor the expenses, are no longer the exclusive privilege of the retail banks. Effectively, it allows every person having a bank account to use a PFM from a third party provider to manage its personal finance, and thus transform the banks into simple vaults. This game changing directive moreover obligates the banks to provide access, via specific Application Program Interfaces (API), to the financial data of the clients. The banks can therefore compete between each other to attract new clients by proposing them new forms of credit or new financial solutions.  \\

Nonetheless, the retail banks now have the opportunity to use their clients’ digital information for recommender engines unlocking new insights about their clients to target financial product recommendations. In this context,  we concentrate on two main retail banking applications. First, we focus on the financial actions of the clients. By analyzing and predicting the financial actions of their clients, the banks gain insight knowledge of their clients’ interests and clients’ spendings. Therefore, they can dynamically propose products targeting the personal needs of their clients such as new credit cards or new loans. For instance, if a client likes to buy a new car every five years and he bought the last one five years ago, the bank can approach him to propose a new car loan with interesting rates. The predictions of the financial transactions, consequently, are at the heart of the bank’s marketing strategy to find or renew product subscriptions. In a second application, we then monitor the user-device authentication on mobile banking application. Through the regular authentication, the banks can create a financial profile awareness for every clients. The more frequently a client is authenticating to its mobile banking application, the more likely he will have a high interest for finance, and therefore, the more likely he will be interested by financial recommendation. This client will be first contacted to advert financial products for wealth and money optimization. The predictions of the user-device authentication consequently becomes involved in a strategy of financial recommendations.  \\

However, the clients' actions contains a large variety of information for both applications. Therefore, the clients’ transactions and the user-device authentication predictions in a financial context is multi-dimensional, sparse and complex. As a result, the proposed methodology has to extract information from a large sparse data set while being able to predict a sequence of actions. We rely on tensors, a higher order analogue of matrix decomposition, to answer these challenges. The tensors are able to scale down a large amount of data to an interpretable size using different types of decomposition, also called factorization, to model multidimensional interactions. Depending on the tensor decomposition, different latent variables can be highlighted with their respective asymmetric relationships. Fast and accurate tensor resolutions have nonetheless required specific numerical optimization methods related to preconditioning methods.  \\

The preconditioning gradient methods use a matrix, called a preconditioner, to update the gradient before it is used. Common well-known optimization preconditioning methods include the Newton's method, which employs the exact Hessian matrix, and the quasi-Newton methods, which do not require the knowledge of the exact Hessian matrix, as described in \cite{wright1999numerical}. Introduced to answer specifically some of the challenges facing Machine Learning (ML) and Deep Learning (DL), AdaGrad \cite{duchi2011adaptive} uses the co-variance matrix of the accumulated gradients as a preconditioner. Because of the dimensions of modern optimization, specialized variants have been proposed to replace the full preconditioning methods by diagonal approximation methods such as Adam in  \cite{kingma2014adam}, by a sketched version \cite{gonen2015faster,pilanci2017newton} or by other schemes such as Nesterov Accelerated Gradient (NAG) \cite{nesterov2007gradient} or SAGA \cite{defazio2014saga}. It is worth mentioning that the diagonal approximation methods are often preferred in practice because of the super-linear memory consumption of the other methods \cite{gupta2018shampoo}. \\

In this chapter, we take an alternative approach to preconditioning because of our aim of building accurate multidimensional financial recommendation engines. We describe how to exploit Newton's convergence using a diagonal approximation of the Hessian matrix with an adaptative line search. Our approach is motivated by the efficient and accurate resolution of tensor decomposition for which most of the ML and DL state-of-the-art optimizers fail. Our algorithm, called VecHGrad for Vector Hessian Gradient, returns the tensor structure of the gradient and uses a separate preconditioner vector. Our analysis targets non-trivial high-order tensor decomposition and relies on the extensions of vector analysis to the tensor world. We show the superior capabilities of VecHGrad over different tensor decomposition in regards to traditional resolution algorithms, such as the Alternating Least Square (ALS) or the Non-linear Conjugate Gradient (NCG) \cite{acar2011scalable}, and some popular ML and DL optimizers such as AdaGrad, Adam or RMSProp. The superior capabilities of VecHGrad are then used for our two main applications related to the clients’ financial actions and the user-device authentication. It ensures to reach a negligible numerical error at the end of the tensor factorization process with respect to the tensor decomposition used. The compressed data set inherited from the tensor factorization is then used as input for the neurons to perform the predictions. Our main contributions are summarized below: 

\begin{itemize}
    \item We propose a new resolution algorithm, called VecHGrad, that uses the gradient and the Hessian-vector product with an adaptive line search to achieve the goal of accurate and fast optimization for complex numerical tasks. We demonstrate VecHGrad's superior accuracy at convergence and compare it with traditional resolution algorithms and popular deep learning optimization algorithms for three of the most common tensor decomposition including CP, DEDICOM \cite{harshman1978models} and PARATUCK2 on five real world data sets.
    
    \item We then apply the unique property of CP decomposition for separate modeling of each order of the clients' transactions which are the time, the client reference, the transaction label and the amount. A compressed dense data set is inherited from the resolution of the CP tensor decomposition with VecHGrad. It is used as an optimized input for the neural network removing all sparse information while highlighting latent factors. The neurons then predict clients' financial activities to unlock financial recommendation.
    
    \item Finally, we have developed an approach with the PARATUCK2 tensor decomposition and its unique advantages of interactions modeling with imbalanced data for user-device authentication monitoring. In our application, one user can use several devices, and thus, we have considered the imbalance between the number of users and devices. The PARATUCK2 decomposition is solved with VecHGrad to ensure accurate identification of the latent components. A collection of neurons then predicts the users' authentication to estimate the future financial awareness of the clients. The banks can, therefore, better advertise their products by contacting the clients which will be the most likely to be interested.
\end{itemize}

The chapter is structured as follows. We discuss the related work in Section \ref{sec::relwork3}. In Section \ref{sec::propmethod3}, we describe how VecHGrad performs a numerical optimization scheme applied to tensors without the requirement of knowing the Hessian matrix. We then build upon VecHGrad for the predictions of the clients' actions with the CP tensor decomposition and neural networks. We subsequently present our user-device authentication application for the estimation of the clients' financial awareness. We use the PARATUCK2 tensor decomposition with VecHGrad to identify the latent groups and to input a compressed data set to a collection of neurons in charge of the predictions. In Section \ref{sec::exp3}, we highlight the experimental results of VecHGrad in comparison to ML optimizers followed by our two applications on clients' transaction predictions and user-device authentication monitoring. Finally, we conclude in Section \ref{sec::ccl3} by addressing promising directions for future work.  \\

\vspace{-.5cm}
\section{Related Work} \label{sec::relwork3}
In this section, we review first the literature related to the numerical optimization algorithms from the tensor and machine learning communities. The VecHGrad resolution algorithm is effectively used in the three experiments of this chapter, and consequently, is one of the main contribution of this chapter. We then review the recommender engines and the tensor decomposition in relation with our application on predictions of sparse clients' actions. Finally, we finish by highlighting briefly some research work related to user-device authentication because of our user-device monitoring application.  \\

\textbf{Literature on numerical optimization schemes}
As aforementioned, VecHGrad uses a diagonal approximation of the Hessian matrix and, therefore, it is related to other diagonal approximation such as the diagonal approximation of AdaGrad \cite{duchi2011adaptive}. The AdaGrad diagonal approximation is very popular in practice and frequently applied \cite{gupta2018shampoo}. However, it only uses gradient information, as opposed to VecHGrad which uses both gradient and Hessian information with an adaptive line search. Other approaches extremely popular in machine learning and deep learning include Adam \cite{kingma2014adam}, NAG \cite{nesterov2007gradient}, SAGA \cite{defazio2014saga} or RMSProp \cite{hinton2012rmsprop}. This non-exhaustive list of machine learning optimization methods is also applied to our study case since it offers a strong baseline comparison for VecHGrad.  \\

The methods specifically designed for tensor decomposition have to be mentioned since our study case is related to tensor decomposition. Various tensor decomposition, or tensor factorization, exist for different types of applications and different algorithms are used to solve tensor decomposition \cite{kolda2009tensor,acar2011all}. Each of the decomposition offers unique features for data compression and latent analysis. The most popular optimization scheme among the resolution of tensor decomposition is the Alternating Least Square (ALS). Under the ALS scheme \cite{kolda2009tensor}, one component of the tensor decomposition is fixed, typically a factor vector or a factor matrix. The fixed component is updated using the other components. All the components are therefore successively updated at each step of the iteration process until a convergence criteria is reached, for instance a fixed number of iteration. Such resolution does not involve any derivative computation. 
At least one ALS resolution scheme exists for every tensor decomposition. The ALS resolution scheme was introduced in \cite{carroll1970analysis} and \cite{harshman1970foundations} for the CP/PARAFAC decomposition, in \cite{harshman1978models} for the DEDICOM decomposition and in \cite{harshman1996uniqueness} for the PARATUCK2 decomposition. Welling and Weber relied on the results of \cite{lee1999learning} applied to matrix resolution to propose a general non-negative resolution of the CP decomposition in \cite{welling2001positive}. An updated ALS scheme was presented in \cite{bro1998multi} to solve PARATUCK2. Bader et al. proposed ASALSAN in \cite{bader2007temporal} to solve with non-negativity constraints the DEDICOM decomposition with the ALS scheme. While some update rules are not guaranteed to decrease the loss function, the scheme leads to overall convergence. Charlier et al. proposed recently in \cite{charlier2018non} a non-negative ALS scheme for the PARATUCK2 decomposition.  \\

Some approaches are furthermore specifically designed for one tensor decomposition using gradient information. Most frequently, it concerns CP/PARAFAC since it has been the most applied tensor decomposition \cite{kolda2009tensor}. An optimized version of the Non-linear Conjugate Gradient (NCG) for CP/PARAFAC, CP-OPT, is presented in \cite{acar2011scalable,acar2011all}. An extension of the Stochastic Gradient Descent (SGD) is described in \cite{maehara2016expected} to obtain, as mentioned by the authors, an expected CP tensor decomposition. Both approaches focus specifically on CP/PARAFAC, and consequently, their performance on other tensor decomposition is not assessed. The comparison to other numerical optimizers in the experiments is additionally limited, especially when considering existing popular machine learning and deep learning optimizers. In contrast, VecHGrad is detached of any particular model structure, including the choice of tensor decomposition, and it only relies on the gradient and on the Hessian diagonal approximation, crucial for fast convergence of complex numerical optimization. VecHGrad, hence, is easy to implement and to use in practice as it does not require to be optimized for a particular model structure.  \\

\textbf{Literature on Recommender Engines and Tensor Decomposition}
Recommender engines have become very popular in real-world applications. The strengths of the second-order matrix factorization recommender engines have been highlighted in various publications of the last decades \cite{brand2003fast,ghazanfar2013advantage,kumar2015role}. Recommender engines are generally divided into three main categories: content-based recommendations, collaborative recommendations and hybrid approaches. In content-based systems, the recommendations relies on the previous activities of a user. The collaborative recommendations engines leverages on the community. It recommends similar products to clients sharing similar interests. Finally, the hybrid recommendation combines both content-based and collaborative recommendations \cite{christou2016amore,sadanand2018movie}. The matrix factorization is limited to the unique modeling of the table \textit{clients}$\times$\textit{products} despite the development of effective and efficient matrix factorization algorithms \cite{ning2011slim}. It cannot be enriched with additional features. As a solution, tensor recommendation engines have therefore skyrocketed in the past few years \cite{lian2016regularized,zhao2016aggregated,song2017based}. The tensors offer the possibility to extend the recommender engines order \cite{zhao2015mobile,hu2015collaborative} as further information can be added to the algorithm such as the time or the location. \\

Noticeably, tensor resolution schemes are of great interest for the end-results, building efficient and accurate multidimensional recommender engines for a wide range of application domains. Their accuracy is underlined in \cite{ge2016taper} for personalized recommendations in the context of social media. The use of enriched features such as the geospatial localization or the users' social preferences significantly improved the quality of the recommendations. A regularized context-aware tensor factorization has been applied for location recommendation in \cite{lian2016regularized}. The algorithm relied on interaction regularization and on locations information by using discrete spatial distances. What is noteworthy is that accurate recommendations are intrinsically linked to accurate predictions. The tensor factorization was even used in \cite{almutairi2017context} to predict students' graduation at the university of Minnesota. Increasing the dimensionality of the data however comes at a cost as the sparsity of the information rises significantly. Different authors have therefore tried to propose various solutions to address the sparsity issues depending on the application. The tensor factorization in \cite{cai2017heterogeneous} is combined with a matrix factorization by a weight fusing method. The scheme does not nonetheless address the memory issue that comes with the sparsity of large data sets. The algorithm ParCube \cite{papalexakis2015p} introduces a parallelization method to speed up tensor decomposition while offering memory optimization for extremely large data sets. \\

Although tensor factorization achieve strong recommendations and accurate predictions, neural networks have recently been used to increase the accuracy of the predictions that allow incorporating additional features \cite{kim2016convolutional,wu2018dual}. The tensor factorization are combined with convolutional neural network or Long-Short-Term-Memory (LSTM) neural network. It is worth mentioning that the references to matrices are references to second order tensors. Even if the sparsity rises with the additional features, the gains in accuracy often overcome the additional computational cost. In fact, in \cite{wu2018neural}, neural networks became the heart of the recommender engine and the tensors an additional feature. Hidasi et al. in \cite{hidasi2015session} tried to improve the second-order tensor recommendation results by using only recurrent neural networks for short time period recommendations. However, approaches that substitute tensor recommendations have focused more on the latest deep learning progress \cite{yang2016deep,zuo2016tag,peng2017adaptive}. The deep learning scheme is similar to the tensor factorization scheme. It extracts abstract features, or latent factors in the case of tensors, to provide the most likely events to build the recommendation.
\\

\textbf{Literature on user-device authentication}
The user-device authentication has significantly evolved for the past few years thanks to the new technologies. A reliable user-device authentication was proposed in \cite{skinner2016cyber}, based on a graphical user friendly authentication. The use of Location Based Authentication (LBA) was studied in \cite{adeka2017africa}. The development of recent embedded systems within smart-devices leads to new authentication processes, which were considered as a pure fiction only few years ago. The usage of the embedded camera of smart-devices for authentication by face recognition was assessed in \cite{li2016face}. The face image taken by the camera of the mobile device was sent to a background server to perform the calculation which reverts then to the phone. In a similar approach, the use of iris recognition was proposed in \cite{chhabra2016low}. However, the authors showed this kind of authentication was not the preferred choice of the end user. The sensors embedded into smart-devices allow, additionally, other type of biometric authentication. The different biometric authentication that could be used with smart-devices were presented in \cite{spolaor2016biometric}, such as the pulse-response, the fingerprint or even the ear shape. Although biometric or LBA solutions might offer a higher level of security for authentication, their extension toward a large scale usage is complex. The authors in \cite{theofanos2016secure} developed the idea that public-key infrastructure based systems, such as strong passwords in combination with physical tokens, for example, a cell phone, would be more likely to be used and largely deployed. It is nonetheless worth mentioning that the most common procedure for mobile devices authentication is still a code of four or six digits \cite{bultel2018security}. \\

\section{Proposed Method} \label{sec::propmethod3}
In this section, we begin by describing briefly the standard form of the Newton's method, and then we introduce VecHGrad for the first order tensor optimization. We recall that a first order tensor is a vector. We then arrive at the objective of applying VecHGrad to non-trivial tensor decomposition. We next reintroduce the non-negative ALS scheme applied to the CP and the PARATUCK2 tensor decompositions, as it is the benchmark for tensor resolution algorithms. We also explain how to derive the non-negative ALS scheme for the DEDICOM decomposition based on the PARATUCK2 non-negative ALS scheme. We subsequently present the CP tensor decomposition combined with Neural Networks (NN) for the predictions of the financial activities of the clients. Finally, we reach our second application of user-device authentication in mobile banking. We explain the use of the PARATUCK2 tensor decomposition combined with NN for the predictions of authentication patterns.  \\

The terminology hereinafter follows the one described 
in \cite{kolda2009tensor} and commonly used. Scalars are denoted by lower case letters, \textit{a}. Vectors and matrices are described by boldface lowercase letters and boldface uppercase letters, respectively \textbf{a} and \textbf{A}. High order tensors are represented using Euler script notation, $\mathscr{X}$. The transpose matrix of $A\in \mathbb{R}^{I\times J}$ is denoted by $A^T$. The inverse of a matrix $A\in \mathbb{R}^{I\times I}$ is denoted by $A^{-1}$.
\\

\subsection{Classical Version of the Newton's Method}
Hereinafter, we shortly resume the classical version of Newton's method. We invite the reader to \cite{boyd2004convex,nesterov2013introductory,wright1999numerical} for a deeper review.

Let define the objective function $f : \mathbb{R}^d \rightarrow \mathbb{R}$ as a closed, convex and twice-differentiable function. 
For a convex set $\mathcal{C}$, we assume the constrained minimizer 

\begin{equation}
  x^* := \arg \underset{x \in \mathcal{C}}{\min}f(x)
\end{equation}

is uniquely defined. We define the eigenvalues of the Hessian denoted $\gamma = \lambda_{min}(\nabla^2 f(x^*))$ evaluated at the minimum. Additionally, we assume that the Hessian map $x \mapsto \nabla^2 f(x)$ is Lipschitz continuous with modulus $L$ at $x^*$, as defined below
\begin{equation}
  \parallel \nabla^2 f(x) - \nabla^2 f(x^*) \parallel \: \leq 
  L \parallel x - x^* \parallel_2  \quad .
\end{equation}
Under these conditions and given an initial random point $\Tilde{x}^0 \in \mathcal{C}$ such that $\parallel \Tilde{x}^0 - x^* \parallel_2 \leq \frac{\gamma}{2L}$, the Newton updates are guaranteed to converge quadratically as defined below

\begin{equation}
  \parallel \Tilde{x}^{t+1} - x^* \parallel_2 \: \leq \dfrac{2L}{\gamma} \parallel \Tilde{x}^{t} - x^* \parallel_2  \quad .
\end{equation}

This result is well-known, further details can be found in \cite{boyd2004convex}. In our experiments, we slightly modify Newton's method to be globally convergent by using strong Wolfe's line search, described in Subsection \ref{subsec:vechgradvectors}. \\

\subsection{Introduction to VecHGrad for Vectors} \label{subsec:vechgradvectors}
Under the Newton's method, the current iterate $\Tilde{\textbf{x}}^t \in \mathcal{C}$ is used to generate the next iterate $\Tilde{\textbf{x}}^{t+1}$ by performing a constrained minimization of the second order Taylor expansion

\begin{equation}
  \Tilde{\textbf{x}}^{t+1} = \arg \underset{\textbf{x}\in \mathcal{C}}{\min} 
  \left \{ \dfrac{1}{2} \left \langle \textbf{x}-\Tilde{\textbf{x}}^t, \nabla^2 f(\Tilde{\textbf{x}}^t)(\textbf{x}-\Tilde{\textbf{x}}^t)  \right \rangle 
  + \left \langle \nabla f(\Tilde{\textbf{x}}^t), \textbf{x}-\Tilde{\textbf{x}}^t \right \rangle
  \right \} \quad .
\end{equation}

We recall that $\nabla f$ and $\nabla^2 f$ denotes the gradient and the Hessian matrix, respectively, of the objective function $f : \mathbb{R}^d \rightarrow \mathbb{R}$. In the ML and DL community, the objective function is frequently called the loss function. 

\begin{equation}
  \nabla f = \underset{\textbf{x}\in\mathcal{C}}{grad} f = 
  [\dfrac{\partial f}{\partial x_1}, \dfrac{\partial f}{\partial x_2}, ..., 
  \dfrac{\partial f}{\partial x_d}]
\end{equation}

\vspace{0.25cm}
\begin{equation} \label{eq::Hessian_matrix}
\nabla^2 f = \textbf{Hes} = \begin{pmatrix}
\dfrac{\partial^2 f}{\partial x_1^2} & 
\dfrac{\partial^2 f}{\partial x_1 \partial x_2} & 
\cdots & 
\dfrac{\partial^2 f}{\partial x_1 \partial x_d} \\[1em]
\dfrac{\partial^2 f}{\partial x_2 \partial x_1} & 
\dfrac{\partial^2 f}{\partial x_2^2} & 
\cdots & 
\dfrac{\partial^2 f}{\partial x_2 \partial x_d} \\
\vdots & \vdots & \ddots & \vdots \\
\dfrac{\partial^2 f}{\partial x_d \partial x_1} &
\dfrac{\partial^2 f}{\partial x_d \partial x_2} & 
\cdots & 
\dfrac{\partial^2 f}{\partial x_d^2}
\end{pmatrix}
\end{equation}

When $\mathcal{C}\in\mathbb{R}^d$, which is the unconstrained form, the new iterate $\Tilde{\textbf{x}}^{t+1}$ is generated such that

\begin{equation}
  \Tilde{\textbf{x}}^{t+1} = \Tilde{\textbf{x}}^t - [\nabla^2f(\Tilde{\textbf{x}}^t)]^{-1} \nabla f(\Tilde{\textbf{x}}^t)  \quad .
\end{equation}

We use the strong Wolfe's line search which allows the Newton's method to be globally convergent. The line search is defined by the following three inequalities

\begin{equation}
\begin{split}
 \text{i) }& f(\Tilde{\textbf{x}}^t+\alpha^t \textbf{p}^t) \leq f(\Tilde{\textbf{x}}^t) + c_1 \alpha^t (\textbf{p}^t)^T \nabla f (\Tilde{\textbf{x}}^t)  \quad , \\ 
 \text{ii) }& -(\textbf{p}^t)^T \nabla f(\Tilde{\textbf{x}}^t + \alpha^t \textbf{p}^t) \leq -c_2 (\textbf{p}^t)^T \nabla f(\Tilde{\textbf{x}}^t)  \quad , \\ 
 \text{iii) }& \mid (\textbf{p}^t)^T \nabla f(\Tilde{\textbf{x}}^t + \alpha^t \textbf{p}^t) \mid \leq c_2 \mid (\textbf{p}^t)^T \nabla f(\Tilde{\textbf{x}}^t) \mid  \quad ,
\end{split}
\end{equation}

where $0\leq c_1 \leq c_2 \leq 1$, $\alpha^t > 0$ is the step length and $\textbf{p}^t = -[\nabla^2 f(\Tilde{\textbf{x}}^t)]^{-1}\nabla f(\Tilde{\textbf{x}}^t)$. Therefore, the iterate $\Tilde{\textbf{x}}^{t+1}$ becomes the following 

\begin{equation}
\begin{dcases}
   \Tilde{\textbf{x}}^{t+1} = & \Tilde{\textbf{x}}^t - \alpha^t [\nabla^2f(\Tilde{\textbf{x}}^t)]^{-1} \nabla f(\Tilde{\textbf{x}}^t) \\
   \Tilde{\textbf{x}}^{t+1} = & \Tilde{\textbf{x}}^t + \alpha^t \textbf{p}^t
\end{dcases} \quad .
\end{equation}

Computing the inverse of the exact Hessian matrix, $[\nabla^2f(\Tilde{\textbf{x}}^t)]^{-1}$, creates a certain number of difficulties. The inverse is therefore computed with a Conjugate Gradient (CG) loop. It has two main advantages: the calculations are considerably less expensive, especially when dealing with large dimensions \cite{wright1999numerical}, and the Hessian can be expressed by a diagonal approximation. The convergence of the CG loop is defined when a maximum number of iteration is reached or when the residual $\textbf{r}^t = \nabla^2 f(\Tilde{\textbf{x}}^t)\textbf{p}^t + \nabla f(\Tilde{\textbf{x}}^t)$ satisfies $\parallel \textbf{r}^t \parallel \leq \sigma \parallel \nabla f(\Tilde{\textbf{x}}^t) \parallel$ with $\sigma \in \mathbb{R}^+$. Within the CG loop, the exact Hessian matrix can be approximated by a diagonal approximation. In the CG loop, the Hessian matrix is effectively multiplied with a descent direction vector resulting in a vector. The only requirement within the main optimization loop is the descent vector. Consequently, only the results of the Hessian vector product are needed. Using the Taylor expansion, the Hessian vector product is equal to the equation below

\begin{equation} \label{eq::hesvectdot1}
\nabla^2 f(\Tilde{\textbf{x}}^t) \: \textbf{p}^t = \dfrac{\nabla f(\Tilde{\textbf{x}}^t+\eta  \: \textbf{p}^t)-\nabla f(\Tilde{\textbf{x}}^t)}{\eta}  \quad .
\end{equation}

The term $\eta$ is the perturbation and the term $\textbf{p}^t$ the descent direction vector, fixed equal to the negative of the gradient at initialization. The extensive computation of the inverse of the exact full Hessian matrix is bypassed using only gradient diagonal approximation. Finally, we reached the objective of describing VecHGrad for first order tensor, summarized in Algorithm \ref{algo:vector}.

\SetAlFnt{\scriptsize}
\SetAlCapFnt{\scriptsize}
\SetAlCapNameFnt{\scriptsize}

\begin{algorithm}[b!]
\setstretch{1.25}
\DontPrintSemicolon

\Repeat{$t = \text{maxiter} \:$ 
  \textsc{or} $\: f(\Tilde{\textbf{x}}^t)\leq \epsilon_1\:$ 
  \textsc{or} $\: \parallel \nabla f(\Tilde{\textbf{x}}^t) \parallel \leq \epsilon_2$}{
  Receive loss function $f:\mathbb{R}^d \rightarrow \mathbb{R}$

  Compute gradient $\nabla f(\Tilde{\textbf{x}}^t) \in \mathbb{R}^d$

  Fix $\textbf{p}_0^t = - \nabla f(\Tilde{\textbf{x}}^t)$
  
  \Repeat{$k = \textit{cg}_{\text{maxiter}} \:$ \textsc{or} $\: \parallel \textbf{r}_k \parallel \leq \sigma \parallel \nabla f(\Tilde{\textbf{x}}^t)) \parallel$}{
  Update $\textbf{p}_k^t$ with CG loop:
      $\textbf{r}_k = \nabla^2 f(\Tilde{\textbf{x}}^t)p_k^t + \nabla f(\Tilde{\textbf{x}}^t)$  \\
  }
  
  $\alpha^t \leftarrow $ Wolfe's line search
  
  Update parameters:
      $\Tilde{\textbf{x}}^{t+1} = \Tilde{\textbf{x}} + \alpha^t \textbf{p}^t_{\text{opt}}$  \\
}

\caption{VecHGrad, vector case \label{algo:vector}}
\end{algorithm}

\subsection{VecHGrad for Fast and Accurate Resolution of Tensor Decomposition}
Hereinafter, we introduce VecHGrad in its general form, which is applicable to tensors of any dimension, of any order for any decomposition. We review further definitions and operations involving tensors before presenting the algorithm and its theoretical convergence rate. 

\subsubsection{Tensor Definitions and Tensor Operations} ~\\
A slice is a two-dimensional section of a tensor, defined by fixing all but two indices. Alternatively, the \textit{k}-th frontal slice of a third-order tensor, $\mathscr{X}_{::k}$, may be denoted more compactly as $\textbf{X}_{k}$. \\

The vectorization operator flattens a tensor of $n$ entries to a column vector $\mathbb{R}^n$. The ordering of the tensor elements is not important as long as it is consistent \cite{kolda2009tensor}. For a third order tensor $\mathscr{X}\in\mathbb{R}^{I\times J \times K}$, the vectorization of $\mathscr{X}$ is equal to
\begin{equation}
  \text{vec} (\mathscr{X}) = 
  \begin{bmatrix} 
    x_{111} & x_{112} & \cdots & x_{IJK}
  \end{bmatrix}^T .
\end{equation}

The n-mode product of a tensor $\mathscr{X}\in\mathbb{R}^{I_1\times I_2 \times ... \times I_N}$ with a matrix $\textbf{U}\in\mathbb{R}^{J\times I_n}$ is denoted by $\mathscr{X}\times_n \textbf{U}$. The n-mode product can be expressed either elementwise or in terms of unfolded tensors.

\begin{equation}
  \begin{split}
    (\mathscr{X} \times_n \textbf{U})_{i_1 ...i_{n-1}ji_{n+1}...i_N }
    \:=\: \sum_{i_n=1}^{I_n} x_{i_1 i_2 ... i_N} u_{j i_n} \\
    \mathscr{Y}\: = \: \mathscr{X} \times \textbf{U} \quad \Leftrightarrow \quad
    \textbf{Y}_{(n)} \:=\: \textbf{U}\textbf{X}_{(n)}
  \end{split} \quad .
\end{equation}

The outer product between two vectors, $\textbf{u}, \textbf{v} \in\mathbb{R}^I, \mathbb{R}^J$, is denoted by the symbol $\circ$
\begin{equation}
\textbf{u} \circ \textbf{v} = 
\begin{bmatrix}
u_1v_1 & u_1v_2 & \cdots & u_1v_J\\ 
u_2v_1 & u_2v_2 & \cdots & u_2v_J\\ 
\vdots & \vdots & \vdots & \vdots \\
u_Iv_1 & u_Iv_2 & \cdots & u_Iv_J 
\end{bmatrix}
= \textbf{u}_i\textbf{v}_j  \quad .
\end{equation}

The Kronecker product between two matrices \textbf{A}$\in\mathbb{R}^{I\times J}$ and \textbf{B}$\in\mathbb{R}^{K\times L}$, denoted by \textbf{A}$\otimes$\textbf{B}, results in a matrix \textbf{C}$\in\mathbb{R}^{IK\times KL}$ such that
\begin{equation} \label{eq::kron}
\textbf{C}=\textbf{A}\otimes\textbf{B}=
\begin{bmatrix}
 a_{11}\textbf{B}& a_{12}\textbf{B} & \cdots & a_{1J}\textbf{B}\\ 
 a_{21}\textbf{B}& a_{22}\textbf{B} & \cdots & a_{2J}\textbf{B}\\
 \vdots & \vdots & \ddots & \vdots \\ 
 a_{I1}\textbf{B}& a_{I2}\textbf{B} & \cdots & a_{IJ}\textbf{B}
\end{bmatrix} \quad .
\end{equation}

The Khatri-Rao product between two matrices \textbf{A}$\in\mathbb{R}^{I\times K}$ and \textbf{B}$\in\mathbb{R}^{J\times K}$, denoted by \textbf{A}$\odot$\textbf{B}, results in a matrix \textbf{C} of size $\mathbb{R}^{IJ\times K}$. It is the column-wise Kronecker product
\begin{equation} \label{eq::kr}
\textbf{C}=\textbf{A}\odot\textbf{B}=
[\textbf{a}_1\otimes \textbf{b}_1 \quad \textbf{a}_2\otimes \textbf{b}_2 \quad \cdots \quad \textbf{a}_K\otimes \textbf{b}_K] \quad .
\end{equation}

The square root of the sum of all tensor entries squared of the tensor $\mathscr{X}$ defines its norm such that 
\begin{equation} \label{eq::norm}
\parallel \mathscr{X}\parallel \:=\: \sqrt{\sum_{j=1}^{I_1}\sum_{j=2}^{I_2} ... \sum_{j=n}^{I_n}x_{j_1, j_2, ..., j_n}^2} \quad .
\end{equation}

The rank-\textit{R} of a tensor $\mathscr{X}\in\mathbb{R}^{I_1\times I_2\times ...\times I_N}$ is the number of linear components that could fit $\mathscr{X}$ exactly such that 
\begin{equation} \label{eq::rank}
\mathscr{X}=\sum_{r=1}^R \textbf{a}_r^{(1)} \circ \textbf{a}_r^{(2)} \circ ... \circ \textbf{a}_r^{(N)} \quad .
\end{equation}

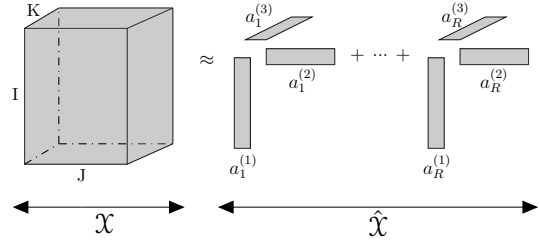
\begin{figure}[t]
  \centering
  \begin{tikzpicture}[scale=0.6, transform shape]
\draw (-7.25,-0.55) rectangle (-5,2.45);
\fill [gray, opacity=.4] (-7.25,-0.55) rectangle (-5,2.45);
\draw (-5,2.45) -- (-4,2.9);
\draw (-7.25,2.45) -- (-6.5,2.9);
\draw (-5,-0.55) -- (-4,-0.1);
\draw (-6.5,2.9) -- (-4,2.9);
\draw (-4,-0.1) -- (-4,2.9);
\draw [dashdotted] (-6.5,-0.1) -- (-4,-0.1);
\draw [dashdotted] (-6.5,-0.1) -- (-6.5,2.9);
\draw [dashdotted] (-7.25,-0.55) -- (-6.5,-0.1);

\fill [gray, opacity=.4] (-4,-0.1) -- (-4,2.9) -- (-5,2.45) -- (-5,-0.55)  -- cycle;
\fill [gray, opacity=.4] (-5,2.45) -- (-4,2.9) -- (-6.5,2.9) -- (-7.25,2.45)  -- cycle;

\node (none) at (-3.25,1.75) {$\approx$};

\coordinate (A) at (-2.65,-0.2) {} {} {} {} {} {};
\coordinate (B) at (-2.3,1.8) {} {} {} {} {} {};
\coordinate (C) at (-1.95,1.65) {} {} {} {} {} {};
\coordinate (D) at (-0.45,2) {} {} {} {} {} {};
\coordinate (e1) at (-2.4,2.2) {} {} {} {} {} {};
\coordinate (e2) at (-1.95,2.2) {} {} {} {} {} {};
\coordinate (e3) at (-0.95,2.7) {} {} {} {} {} {};
\coordinate (e4) at (-1.4,2.7) {} {} {} {} {} {};
\draw (A) rectangle (B);
\fill [gray, opacity=.4] (A) rectangle (B);
\draw (C) rectangle (D);
\fill [gray, opacity=.4] (C) rectangle (D);
\draw (e1) -- (e2) -- (e3) -- (e4) -- cycle;
\fill [gray, opacity=.4] (e1) -- (e2) -- (e3) -- (e4) -- cycle;

\node (none) at (0.05,1.85) {$+$};
\node (none) at (0.55,1.85) {$...$};
\node (none) at (1.05,1.85) {$+$};

\coordinate (A) at (1.6,-0.2) {} {} {} {} {} {} {} {};
\coordinate (B) at (1.95,1.8) {} {} {} {} {} {} {} {};
\coordinate (C) at (2.3,1.65) {} {} {} {} {} {} {} {};
\coordinate (D) at (3.8,2) {} {} {} {} {} {} {} {};
\coordinate (e1) at (1.85,2.2) {} {} {} {} {} {} {} {};
\coordinate (e2) at (2.3,2.2) {} {} {} {} {} {} {} {};
\coordinate (e3) at (3.3,2.7) {} {} {} {} {} {} {} {};
\coordinate (e4) at (2.85,2.7) {} {} {} {} {} {} {} {};
\draw (A) rectangle (B);
\fill [gray, opacity=.4] (A) rectangle (B);
\draw (C) rectangle (D);
\fill [gray, opacity=.4] (C) rectangle (D);
\draw (e1) -- (e2) -- (e3) -- (e4) -- cycle;
\fill [gray, opacity=.4] (e1) -- (e2) -- (e3) -- (e4) -- cycle;

\draw [line width=0.25pt, -triangle 45] (-6,-1.5) -- (-3.75,-1.5);
\draw [line width=0.25pt, -triangle 45] (-6,-1.5) -- (-7.5,-1.5);

\draw [line width=0.25pt, -triangle 45] (3,-1.5) -- (-3,-1.5);
\draw [line width=0.25pt, -triangle 45] (3,-1.5) -- (4.0,-1.5);

\node (none) at (-5.5,-1.85) {\Large $\mathscr{X}$};
\node (none) at (0.5,-1.85) {\Large $\mathscr{\hat{X}}$};
\node (none) at (-7.45,1) {\small{I}};
\node (none) at (-6,-0.8) {\small{J}};
\node (none) at (-7.05,2.85) {\small{K}};
\node (none) at (-2.45,-0.6) {$a_1^{(1)}$};
\node (none) at (-1.15,1.3) {$a_1^{(2)}$};
\node (none) at (-2.1,2.75) {$a_1^{(3)}$};
\node (none) at (1.8,-0.6) {$a_R^{(1)}$};
\node (none) at (3.05,1.3) {$a_R^{(2)}$};
\node (none) at (2.2,2.75) {$a_R^{(3)}$};

\end{tikzpicture}
  \caption{Third order CP/PARAFAC tensor decomposition}
  \label{fig::cptd}
\end{figure}

The CP/PARAFAC decomposition, shown in Figure \ref{fig::cptd}, was introduced in \cite{harshman1970foundations, carroll1970analysis}. The tensor $\mathscr{X}\in\mathbb{R}^{I\times I\times K}$ is defined as a sum of rank-one tensors. The number of rank-one tensors is determined by the rank, denoted by $R$, of the tensor $\mathscr{X}$. The CP decomposition is expressed as 

\begin{equation} \label{eq::cpequation}
  \mathscr{X} = \sum_{r=1}^{R} \textbf{a}_r^{(1)} \circ \textbf{a}_r^{(2)} \circ \textbf{a}_r^{(3)} \circ... \circ \textbf{a}_r^{(N)} \quad ,
\end{equation}

where $\textbf{a}_r^{(1)}, \textbf{a}_r^{(2)}, \textbf{a}_r^{(3)}, ..., \textbf{a}_r^{(N)}$ are factor vectors of size $\mathbb{R}^{I_1}, \mathbb{R}^{I_2}, \mathbb{R}^{I_3}, ..., \mathbb{R}^{I_N}$. Each factor vector $\textbf{a}_r^{(n)}$ with $n\in \left\lbrace 1, 2, ..., N \right\rbrace$ and $r \in \left\lbrace 1, ..., R \right\rbrace$ refers to one order and one rank of the tensor $\mathscr{X}$. \\

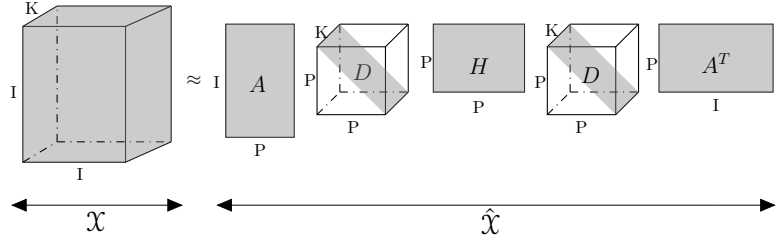
\begin{figure}[t]
  \centering
  \begin{tikzpicture}[scale=0.6, transform shape]

\draw (-7.25,-0.55) rectangle (-5,2.45);
\fill [gray, opacity=.4] (-7.25,-0.55) rectangle (-5,2.45);
\draw (-5,2.45) -- (-4,2.9);
\draw (-7.25,2.45) -- (-6.5,2.9);
\draw (-5,-0.55) -- (-4,-0.1);
\draw (-6.5,2.9) -- (-4,2.9);
\draw (-4,-0.1) -- (-4,2.9);
\draw [dashdotted] (-6.5,-0.1) -- (-4,-0.1);
\draw [dashdotted] (-6.5,-0.1) -- (-6.5,2.9);
\draw [dashdotted] (-7.25,-0.55) -- (-6.5,-0.1);

\fill [gray, opacity=.4] (-4,-0.1) -- (-4,2.9) -- (-5,2.45) -- (-5,-0.55)  -- cycle;
\fill [gray, opacity=.4] (-5,2.45) -- (-4,2.9) -- (-6.5,2.9) -- (-7.25,2.45)  -- cycle;

\node (none) at (-3.5,1.25) {$\approx$};

\draw  (-2.8,2.5) rectangle (-1.3,0);
\fill [gray, opacity=.4]  (-2.8,2.5) rectangle (-1.3,0);
\draw (-2.1,1.5) node[below]{\large{$A$}};

\draw (4.25,0.5) rectangle (5.75,2) node (v1) {};
\draw (5.75,2) -- (6.25,2.5);
\draw (4.25,2) -- (4.75,2.5);
\draw (5.75,0.5) -- (6.25,1);
\draw (4.75,2.5) -- (6.25,2.5);
\draw (6.25,1) -- (6.25,2.5);
\draw [dashdotted] (4.75,1) -- (6.25,1);
\draw [dashdotted] (4.75,1) -- (4.75,2.5);
\draw [dashdotted] (4.25,0.5) -- (4.75,1);
\fill [gray, opacity=.4] (5.75,0.5) -- (6.25,1) -- (4.75,2.5) -- (4.25,2)  -- cycle;
\draw (0.2,1.75) node[below]{\large{$D$}};

\draw (1.75,2.5) rectangle (3.75,1);
\fill [gray, opacity=.4] (1.75,2.5) rectangle (3.75,1);
\draw (2.75,1.9) node[below]{\large{$H$}};

\draw (-0.8,0.5) rectangle (0.7,2) node (v1) {};
\draw (0.7,2) -- (1.2,2.5);
\draw (-0.8,2) -- (-0.3,2.5);
\draw (0.7,0.5) -- (1.2,1);
\draw (-0.3,2.5) -- (1.2,2.5);
\draw (1.2,1) -- (1.2,2.5);
\draw [dashdotted] (-0.3,1) -- (1.2,1);
\draw [dashdotted] (-0.3,1) -- (-0.3,2.5);
\draw [dashdotted] (-0.8,0.5) -- (-0.3,1);
\fill [gray, opacity=.4] (0.7,0.5) -- (1.2,1) -- (-0.3,2.5) -- (-0.8,2)  -- cycle;
\draw (5.2,1.7) node[below]{\large{$D$}};

\draw  (6.7,2.5) rectangle (9.2,1);
\fill [gray, opacity=.4] (6.7,2.5) rectangle (9.2,1);
\draw (7.95,2) node[below]{\large{$A^T$}};

\draw [line width=0.25pt, -triangle 45] (-6.5,-1.5) -- (-7.5,-1.5);
\draw [line width=0.25pt, -triangle 45] (-6.5,-1.5) -- (-3.75,-1.5);

\draw [line width=0.25pt, -triangle 45] (8,-1.5) -- (-3,-1.5);
\draw [line width=0.25pt, -triangle 45] (8,-1.5) -- (9.25,-1.5);

\node (none) at (-5.65,-1.85) {\Large $\mathscr{X}$};
\node (none) at (3,-1.85) {\Large $\mathscr{\hat{X}}$};

\node (none) at (-7.45,1) {\small{I}};
\node (none) at (-6,-0.8) {\small{I}};
\node (none) at (-7.05,2.85) {\small{K}};

\draw (-3,1.5) node[below]{\small{I}};
\draw (-2.05,0) node[below]{\small{P}};
\draw (0,0.5) node[below]{\small{P}};
\draw (-0.95,1.55) node[below]{\small{P}};
\draw (-0.7,2.6) node[below]{\small{K}};
\draw (4.35,2.65) node[below]{\small{K}};
\draw (1.6,1.95) node[below]{\small{P}};
\draw (2.75,0.95) node[below]{\small{P}};
\draw (4.1,1.5) node[below]{\small{P}};
\draw (5,0.5) node[below]{\small{P}};
\draw (6.55,1.9) node[below]{\small{P}};
\draw (7.95,1) node[below]{\small{I}};

\end{tikzpicture}
  \caption{Third order DEDICOM tensor decomposition}
  \label{fig::dedicomtd}
\end{figure}

The DEDICOM decomposition \cite{harshman1978models}, illustrated in Figure \ref{fig::dedicomtd}, describes the asymmetric relationships between $I$ objects of the tensor $\mathscr{X}\in \mathbb{R}^{I\times I \times K}$.  The decomposition groups the $I$ objects into $R$ latent components (or groups) and describe their pattern of interactions by computing $\textbf{A}\in\mathbb{R}^{I\times R}$, $\textbf{H}\in\mathbb{R}^{R\times R}$ and $\mathscr{D}\in\mathbb{R}^{R\times R \times K}$ such that
\begin{equation}
  \textbf{X}_k = \textbf{A}\textbf{D}_k \textbf{H} \textbf{D}_k \textbf{A}^T
  \quad \textrm{with} \quad k=\left\lbrace 1, ..., K \right\rbrace \quad .
\end{equation}

The matrix $\textbf{A}$ indicates the participation of object $i = 1, ..., I$ in the group $r = 1, ..., R$, the matrix $\textbf{H}$ the interactions between the different components $r$ and the tensor $\mathscr{D}$ represents the participation of the $R$ latent component according to the third order $K$. \\

\begin{figure}[t]
  \centering
  \begin{tikzpicture}[scale=0.6, transform shape]

\draw (-7.25,-0.55) rectangle (-5,2.45);
\fill [gray, opacity=.4] (-7.25,-0.55) rectangle (-5,2.45);
\draw (-5,2.45) -- (-4,2.9);
\draw (-7.25,2.45) -- (-6.5,2.9);
\draw (-5,-0.55) -- (-4,-0.1);
\draw (-6.5,2.9) -- (-4,2.9);
\draw (-4,-0.1) -- (-4,2.9);
\draw [dashdotted] (-6.5,-0.1) -- (-4,-0.1);
\draw [dashdotted] (-6.5,-0.1) -- (-6.5,2.9);
\draw [dashdotted] (-7.25,-0.55) -- (-6.5,-0.1);

\fill [gray, opacity=.4] (-4,-0.1) -- (-4,2.9) -- (-5,2.45) -- (-5,-0.55)  -- cycle;
\fill [gray, opacity=.4] (-5,2.45) -- (-4,2.9) -- (-6.5,2.9) -- (-7.25,2.45)  -- cycle;

\node (none) at (-3.5,1.25) {$\approx$};

\draw  (-2.8,2.5) rectangle (-1.3,0);
\fill [gray, opacity=.4]  (-2.8,2.5) rectangle (-1.3,0);
\draw (-2.1,1.5) node[below]{\large{$A$}};

\draw (4.2,0) rectangle (6.2,2) node (v1) {};
\draw (6.2,2) -- (6.7,2.5);
\draw (4.2,2) -- (4.7,2.5);
\draw (6.2,0) -- (6.7,0.5);
\draw (4.7,2.5) -- (6.7,2.5);
\draw (6.7,0.5) -- (6.7,2.5);
\draw [dashdotted] (4.7,0.5) -- (6.7,0.5);
\draw [dashdotted] (4.7,0.5) -- (4.7,2.5);
\draw [dashdotted] (4.2,0) -- (4.7,0.5);
\fill [gray, opacity=.4] (6.2,0) -- (6.7,0.5) -- (4.7,2.5) -- (4.2,2)  -- cycle;
\draw (0.25,1.8) node[below]{\large{$D^A$}};

\draw (1.75,2.5) rectangle (3.75,1);
\fill [gray, opacity=.4] (1.75,2.5) rectangle (3.75,1);
\draw (2.75,1.9) node[below]{\large{$H$}};

\draw (-0.8,0.5) rectangle (0.7,2) node (v1) {};
\draw (0.7,2) -- (1.2,2.5);
\draw (-0.8,2) -- (-0.3,2.5);
\draw (0.7,0.5) -- (1.2,1);
\draw (-0.3,2.5) -- (1.2,2.5);
\draw (1.2,1) -- (1.2,2.5);
\draw [dashdotted] (-0.3,1) -- (1.2,1);
\draw [dashdotted] (-0.3,1) -- (-0.3,2.5);
\draw [dashdotted] (-0.8,0.5) -- (-0.3,1);
\fill [gray, opacity=.4] (0.7,0.5) -- (1.2,1) -- (-0.3,2.5) -- (-0.8,2)  -- cycle;
\draw (5.5,1.55) node[below]{\large{$D^B$}};

\draw  (7.25,2.5) rectangle (9.75,0.5);
\fill [gray, opacity=.4] (7.25,2.5) rectangle (9.75,0.5);
\draw (8.55,1.75) node[below]{\large{$B^T$}};

\draw [line width=0.25pt, -triangle 45] (-6.5,-1.5) -- (-7.5,-1.5);
\draw [line width=0.25pt, -triangle 45] (-6.5,-1.5) -- (-3.75,-1.5);

\draw [line width=0.25pt, -triangle 45] (8,-1.5) -- (-3,-1.5);
\draw [line width=0.25pt, -triangle 45] (8,-1.5) -- (10,-1.5);

\node (none) at (-5.65,-1.85) {\Large $\mathscr{X}$};
\node (none) at (3.5,-1.85) {\Large $\mathscr{\hat{X}}$};

\node (none) at (-7.45,1) {I};
\node (none) at (-6,-0.8) {J};
\node (none) at (-6,-0.8) {J};
\node (none) at (-7.05,2.85) {K};

\draw (-3,1.5) node[below]{\small{I}};
\draw (-2.05,0) node[below]{\small{P}};
\draw (0,0.5) node[below]{\small{P}};
\draw (-0.95,1.45) node[below]{\small{P}};
\draw (-0.7,2.6) node[below]{\small{K}};
\draw (4.3,2.65) node[below]{\small{K}};
\draw (1.6,2) node[below]{\small{P}};
\draw (2.75,0.95) node[below]{\small{Q}};
\draw (4.05,1.25) node[below]{\small{Q}};
\draw (5.2,0) node[below]{\small{Q}};
\draw (7.1,1.75) node[below]{\small{Q}};
\draw (8.5,0.5) node[below]{\small{J}};

\end{tikzpicture}
  \caption{Third order PARATUCK2 tensor decomposition}
  \label{fig::paratuck2td}
\end{figure}
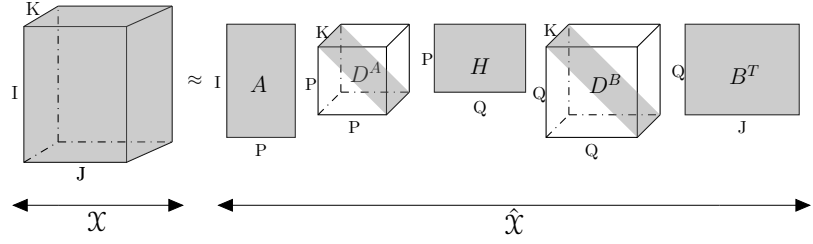

The PARATUCK2 decomposition \cite{harshman1996uniqueness}, represented in Figure \ref{fig::paratuck2td}, expresses the original tensor $\mathscr{X} \in \mathbb{R}^{I\times J\times K}$ as a product of matrices and tensors

\begin{equation} \label{eq::paratuck2}
\textbf{X}_k = \textbf{A}\textbf{D}^A_k\textbf{H}\textbf{D}^B_k\textbf{B}^T \quad \textrm{with} \quad k=\left\lbrace 1, ..., K \right\rbrace \quad ,
\end{equation}

where $\textbf{A}$, $\textbf{H}$ and $\textbf{B}$ are matrices of size $\mathbb{R}^{I\times P}$, $\mathbb{R}^{P \times Q}$ and $\mathbb{R}^{J\times Q}$. The matrices $\textbf{D}^A_k\in \mathbb{R}^{P\times P}$ and $\textbf{D}^B_k\in \mathbb{R}^{Q\times Q}\:\forall k\in\left\lbrace 1, ...,K \right\rbrace$ are the slices of the tensors $\mathscr{D}^A\in \mathbb{R}^{P\times P \times K}$ and $\mathscr{D}^B\in \mathbb{R}^{Q\times Q \times K}$. The columns of the matrices $\textbf{A}$ and $\textbf{B}$ represent the latent factors $P$ and $Q$, and therefore the rank of each object set. The matrix $\textbf{H}$ underlines the asymmetry between the $P$ latent components and the $Q$ latent components. The tensors $\mathscr{D}^A$ and $\mathscr{D}^B$ measures the evolution of the latent components regarding the third order.

\subsubsection{VecHGrad for Tensor Resolution}
Here, we describe the main application of VecHGrad that is the resolution of non-trivial tensor decomposition. \\

The loss function, also called the objective function, is denoted by $f$ and it is equal to 

\begin{equation} \label{eq::errorminimization}
  f(\Tilde{\textbf{x}}) = \min_{\mathscr{\hat{X}}} ||\mathscr{X}-\mathscr{\hat{X}}|| \quad .
\end{equation}
The tensor $\mathscr{X}$ is the original tensor and the tensor $\mathscr{\hat{X}}$ is the approximated tensor built from the decomposition. For instance, if we consider the CP/PARAFAC tensor decomposition applied on a third order tensor, the tensor $\mathscr{\hat{X}}$ is the tensor built with the factor vectors $\textbf{a}_r^{(1)}, \textbf{a}_r^{(2)}, \textbf{a}_r^{(3)}$ for $r=1,...,R$ initially randomized such that 

\begin{equation}
  \mathscr{\hat{X}} = \sum_{r=1}^{R} \textbf{a}_r^{(1)} \circ \textbf{a}_r^{(2)} \circ \textbf{a}_r^{(3)} \quad .
\end{equation}

The vector $\Tilde{\textbf{x}}$ is a flattened vector containing all the entries of the decomposed tensor $\mathscr{\hat{X}}$. If we consider the previous example of a third order tensor $\mathscr{\hat{X}}$ of rank $R$ factorized with the CP/PARAFAC tensor decomposition, we obtain the following vector $\Tilde{\textbf{x}}\in\mathbb{R}^{d=R(I + J + K)}$ such that 

\begin{equation} \label{eq::vectorize_x}
\setlength{\abovedisplayskip}{-1pt}
\begin{split}
  \Tilde{\textbf{x}} = \: & \text{vec}(\mathscr{\hat{X}}) 
  =
  \begin{bmatrix}
    \textbf{a}^{(1)}_{1}, \; \textbf{a}^{(1)}_{2}, ..., \textbf{a}^{(R)}_{I} ,
    \textbf{a}^{(2)}_{1}, \; \textbf{a}^{(2)}_{2}, ..., \textbf{a}^{(R)}_{J} ,
    \textbf{a}^{(3)}_{1}, \; \textbf{a}^{(3)}_{2}, ..., \textbf{a}^{(R)}_{K}
  \end{bmatrix}^T
\end{split} \quad .
\end{equation}

Since the objective is to propose a universal approach for any tensor decomposition, we rely on the finite difference method to compute the gradient of the loss function of any tensor decomposition. Thus, the method can be transposed to any decomposition just by changing the decomposition equation. The approximate gradient is based on a fourth order formula (\ref{eq::fourth_order_diff}) to ensure a reliable approximation \cite{schittkowski2001nlpqlp} of the exact gradient, defined as

\begin{equation} \label{eq::fourth_order_diff}
\setlength{\abovedisplayskip}{-1pt}
\begin{split}
\dfrac{\partial}{\partial x_i} f(\Tilde{\textbf{x}}^t) 
\approx \dfrac{1}{4!\eta} & \big( 
2 [f(\Tilde{\textbf{x}}^t-2\eta \textbf{e}_i) - f(\Tilde{\textbf{x}}^t + 2 \eta \textbf{e}_i)]
+ 16 [f(\Tilde{\textbf{x}}^t + \eta \textbf{e}_i) - f(\Tilde{\textbf{x}}^t - \eta \textbf{e}_i)]
\big)
\end{split}
\quad .
\end{equation}

In (\ref{eq::fourth_order_diff}), the index $i$ is the index of the variables for which the derivative is to be evaluated. The variable $\textbf{e}_i$ is the $i-$th unit vector. The term $\eta$, the perturbation, is fixed small enough to achieve the convergence of the iterative process.  \\

The Hessian diagonal approximation is evaluated as described in Subsection \ref{subsec:vechgradvectors}. During the CG optimization loop, the Hessian matrix is multiplied with a descent direction vector resulting in a vector. Therefore, only the results of the Hessian vector product is required. Using the Taylor expansion, this product is equal to the equation below

\begin{equation} \label{eq::hesvectdot2}
\nabla^2 f(\Tilde{\textbf{x}}^t) \: \textbf{p}^t = \dfrac{\nabla f(\Tilde{\textbf{x}}^t+\eta  \: \textbf{p}^t)-\nabla f(\Tilde{\textbf{x}}^t)}{\eta} \quad .
\end{equation}

The perturbation term is denoted by $\eta$ and the descent direction vector by $\textbf{p}^t$, fixed equal to the negative of the gradient at initialization. Consequently, the extensive computation of the inverse of the exact full Hessian matrix is bypassed using only gradient diagonal approximation. Finally, we reached the core  objective of describing VecHGrad for tensors, summarized in Algorithm \ref{algo:vechgrad}.

\SetAlFnt{\scriptsize}
\SetAlCapFnt{\scriptsize}
\SetAlCapNameFnt{\scriptsize}

\begin{algorithm}[b!]
\setstretch{1.25}
\DontPrintSemicolon

Receive tensor decomposition equation:
    $g: \left\{\begin{matrix*}[l]
  \mathbb{R}^d \rightarrow  \mathbb{R}^{I_1\times I_2\times ...\times I_n}  \\ 
  \Tilde{\textbf{x}}^t \mapsto \mathscr{\Tilde{X}} 
  \end{matrix*}\right.$ \\
Receive $\Tilde{\textbf{x}}^0 \:=\: \text{vec}(\mathscr{\hat{X}})$  \\
\Repeat{$t = \text{maxiter} \:$ 
  \textsc{or} $\: f(\Tilde{\textbf{x}}^t)\leq \epsilon_1\:$ 
  \textsc{or} $\: \parallel \nabla f(\Tilde{\textbf{x}}^t) \parallel \leq \epsilon_2$}{
  Receive loss function: 
  $f: \left\{\begin{matrix*}[l]
  \mathbb{R}^d \rightarrow \mathbb{R} \\ 
  \Tilde{\textbf{x}}^t \mapsto \parallel \mathscr{X} - g(\Tilde{\textbf{x}}^t) \parallel
  \end{matrix*}\right. $\\
  
  Compute gradient $\nabla f(\Tilde{\textbf{x}}^t) \in \mathbb{R}^d$

  Fix $\textbf{p}_0^t = - \nabla f(\Tilde{\textbf{x}}^t)$
  
  \Repeat{$k = \textit{cg}_{\text{maxiter}} \:$ \textsc{or} $\: \parallel \textbf{r}_k \parallel \leq \sigma \parallel \nabla f(\Tilde{\textbf{x}})^t) \parallel$}{
  Update $\textbf{p}_k^t$ with CG loop:
      $\textbf{r}_k = \nabla^2 f(\Tilde{\textbf{x}}^t)\textbf{p}_k^t + \nabla f(\Tilde{\textbf{x}}^t)$ \\
  }
  
  $\alpha^t \leftarrow $ Wolfe's line search
  
  Update parameters:
      $\Tilde{\textbf{x}}^{t+1} = \Tilde{\textbf{x}}^t + \alpha^t \textbf{p}^t_{\text{opt}}$  \\
  
}

\caption{VecHGrad, tensor case \label{algo:vechgrad}}
\end{algorithm}

\subsubsection{Theoretical Convergence Rate of VecHGrad}
VecHGrad is based on the Newton's method but it relies on a diagonal approximation of the Hessian matrix instead of the full exact Hessian matrix with an adaptive line search. The reason is that although the exact Newton's method convergence is quadratic \cite{wright1999numerical}, the computation of the exact Hessian matrix is too time consuming for ML and DL large-scale applications, including tensor application. Therefore, VecHGrad has a superlinear convergence such that 

\begin{equation} \label{eq::superlinconv}
\underset{n\rightarrow \infty}{lim} \dfrac{\left \| \textbf{B}_n - \nabla^2 f(\Tilde{\textbf{x}}_\text{opt})\textbf{p}_n \right \|}{\left \| \textbf{p}_n \right \|} = 0 \quad ,
\end{equation}

with $\Tilde{\textbf{x}}_\text{opt}$ the point of convergence, $\textbf{p}_n$ the search direction and $\textbf{B}_n$ the approximation of the Hessian matrix. Practically, the convergence rate is described according to the equation below

\begin{equation} \label{eq::convergencerate}
q \approx 
\log \dfrac{|\Tilde{\textbf{x}}^{t+1}-\Tilde{\textbf{x}}^{t}|}{|\Tilde{\textbf{x}}^{t}-\Tilde{\textbf{x}}^{t-1}|} 
\begin{bmatrix}
  \log \dfrac{|\Tilde{\textbf{x}}^{t}-\Tilde{\textbf{x}}^{t-1}|}{|\Tilde{\textbf{x}}^{t-1}-\Tilde{\textbf{x}}^{t-2}|}
\end{bmatrix}^{-1} \quad .
\end{equation}

\subsection{Alternating Least Square for Tensor Decomposition}
Because we designed VecHGrad for accurate tensor resolution, we have to compare its performance with the most popular numerical optimizer in the tensor world that is the Alternating Least Square (ALS) \cite{kolda2009tensor}. One component of the tensor decomposition, a factor vector or a factor matrix, is fixed and updated using the other components of the decomposition. Every component is successively updated until a convergence criteria is reached. In our experiments, we furthermore emphasize on the non-negative ALS scheme. It allows an easier interpretation of the tensor decomposition latent factors, especially in our case involving financial recommendations. We first present the standard ALS scheme \cite{carroll1970analysis,harshman1970foundations} followed by the non-negative ALS scheme based on \cite{lee1999learning,welling2001positive} for the CP tensor decomposition. We then adopt the same methodology for the PARATUCK2 tensor decomposition \cite{bro1998multi,charlier2018non}. Finally, we explain how to derive the non-negative ALS scheme for the DEDICOM tensor decomposition based on the scheme of the PARATUCK2 tensor decomposition.  \\

Under the ALS scheme, the following minimization equation has to be solved for both the standard and the non-negative update rules
\begin{equation} \label{eq::minimztn}
\min_{\mathscr{\hat{X}}} ||\mathscr{X}-\mathscr{\hat{X}}|| \quad ,
\end{equation}
with $\mathscr{\hat{X}}$ the approximate tensor of the decomposition and $\mathscr{X}$ the original tensor. All the factor matrices and the factor tensors are updated iteratively to solve the equation \eqref{eq::minimztn}.  \\

\subsubsection{Standard and Non-negative ALS for CP}
We recall that the CP tensor decomposition is defined by 

\begin{equation} \label{eq::cpequation2}
  \mathscr{X} = \sum_{r=1}^{R} \textbf{a}_r^{(1)} \circ \textbf{a}_r^{(2)} \circ \textbf{a}_r^{(3)} \circ... \circ \textbf{a}_r^{(N)}
\end{equation}

for a $N$-order tensor $\mathscr{X}\in\mathbb{R}^{I_1 \times I_2 \times ... \times N}$, a rank $R$, and the factor vectors $\textbf{a}_r^{(1)}, \textbf{a}_r^{(2)}, \textbf{a}_r^{(3)}, ..., \textbf{a}_r^{(N)}$ of size $\mathbb{R}^{I_1}, \mathbb{R}^{I_2}, \mathbb{R}^{I_3}, ..., \mathbb{R}^{I_N}$.  \\

\textbf{Standard ALS Method for CP}  \\
We denote the latent factor vectors by $\textbf{A} = \sum_{r=1}^R a_r^{(1)}$, $\textbf{B} = \sum_{r=1}^R a_r^{(2)}$ and $\textbf{C} = \sum_{r=1}^R a_r^{(3)}$. For a third order tensor $\mathscr{X}\in\mathbb{R}^{I\times J \times K}$, the standard update rules are the following

\begin{equation} \label{eq::cp_std_als}
\begin{cases}
A \leftarrow & \textbf{X}_{(1)} (\textbf{C} \odot \textbf{B}) ( \textbf{C}^T \textbf{C} \ast \textbf{B}^T \textbf{B} )^\dagger  \\
B \leftarrow & \textbf{X}_{(2)} (\textbf{C} \odot \textbf{A}) ( \textbf{C}^T \textbf{C} \ast \textbf{A}^T \textbf{A} )^\dagger  \\
C \leftarrow & \textbf{X}_{(3)} (\textbf{B} \odot \textbf{A}) ( \textbf{B}^T \textbf{B} \ast \textbf{A}^T \textbf{A} )^\dagger 
\end{cases} \quad .
\end{equation}

Based on the standard ALS update rules for the CP tensor decomposition, we can deduce the non-negative ALS update rules.  \\

\textbf{Non-negative ALS Method for CP}  \\
We use a tensor of size $\mathscr{X}\in\mathbb{R}^{I\times J \times K}$ with the latent factor vectors $\textbf{A} = \sum_{r=1}^R a_r^{(1)}$, $\textbf{B} = \sum_{r=1}^R a_r^{(2)}$ and $\textbf{C} = \sum_{r=1}^R a_r^{(3)}$. Based on the non-negative update rules for matrices \cite{lee1999learning}, the non-negative ALS method for CP follows the below update rules

{
\allowdisplaybreaks
\begin{align} \label{eq::cp_nn_als}
a_{ir} \leftarrow & a_{ir} \dfrac{[\textbf{X}_{(1)} ( \textbf{C} \odot \textbf{B} ) ]_{ir}}{[ \textbf{A} ( \textbf{C} \odot \textbf{B} )^T( \textbf{C} \odot \textbf{B} )]_{ir}} \quad , \\
b_{jr} \leftarrow & b_{jr} \dfrac{[\textbf{X}_{(2)} ( \textbf{C} \odot \textbf{A} ) ]_{jr}}{[ \textbf{B} ( \textbf{C} \odot \textbf{A} )^T ( \textbf{C} \odot \textbf{A} )]_{jr}} \quad , \\
c_{kr} \leftarrow & c_{kr} \dfrac{[\textbf{X}_{(3)} ( \textbf{B} \odot \textbf{A} ) ]_{kr}}{[ \textbf{C} ( \textbf{B} \odot \textbf{A} )^T ( \textbf{B} \odot \textbf{A} ) ]_{kr}} \quad .
\end{align}
}

We will use the non-negative ALS update rules in our experiments involving the CP tensor decomposition and the ALS algorithm.  \\

\subsubsection{Standard and Non-negative ALS for PARATUCK2 and DEDICOM}
We recall that the equation for the PARATUCK2 tensor decomposition is equal to 

\begin{equation} \label{eq::par_for_als}
\textbf{X}_k = \textbf{A}\textbf{D}^A_k\textbf{H}\textbf{D}^B_k\textbf{B}^T \quad \textrm{with} \quad k=\left\lbrace 1, ..., K \right\rbrace \quad ,
\end{equation}

where $\mathscr{X} \in \mathbb{R}^{I\times J\times K}$ is the original tensor, $\textbf{A}$, $\textbf{H}$ and $\textbf{B}$ are matrices of size $\mathbb{R}^{I\times P}$, $\mathbb{R}^{P \times Q}$ and $\mathbb{R}^{J\times Q}$. The matrices $\textbf{D}^A_k\in \mathbb{R}^{P\times P}$ and $\textbf{D}^B_k\in \mathbb{R}^{Q\times Q}\:\forall k\in\left\lbrace 1, ...,K \right\rbrace$ are the slices of the tensors $\mathscr{D}^A\in \mathbb{R}^{P\times P \times K}$ and $\mathscr{D}^B\in \mathbb{R}^{Q\times Q \times K}$.  \\

\textbf{Standard ALS Method for PARATUCK2}  \\
We consider one level \textit{k} of \textit{K}, the third dimension of the tensor to ease the explanation for the resolution of the PARATUCK2 tensor decomposition. Under the standard ALS method, the equation \eqref{eq::par_for_als} is rearranged to update $\textbf{A}$ such that
\begin{equation} \label{eq::upd_A}
\textbf{X}_k=\textbf{A}\textbf{F}_k 
\quad \textrm{with} \quad
\textbf{F}_k=\textbf{D}_k^A \textbf{H} \textbf{D}_k^B \textbf{B}^T \quad .
\end{equation}

The simultaneous least square solution for all \textit{k} leads to 
\begin{equation}
\textbf{A} = \textbf{X}(\textbf{F}^\dag)^T
\quad \textrm{with} \quad
\begin{cases}
\textbf{X} = & \left[ \textbf{X}_1 \: \textbf{X}_2 \cdots \textbf{X}_k \right] \\ 
\textbf{F} = & \left[ \textbf{F}_1 \: \textbf{F}_2 \cdots \textbf{F}_k \right] \\ 
\end{cases} \quad .
\end{equation}

To update $\mathscr{D}^A$, the equation \eqref{eq::par_for_als} is rearranged such that 
\begin{equation} \label{eq::upd_DA}
\textbf{X}_k=\textbf{A} \textbf{D}_k^A \textbf{F}_k^T
\quad \textrm{with} \quad
\textbf{F}_k=\textbf{B} \textbf{D}_k^B \textbf{H}^T \quad .
\end{equation}

The matrix $\textbf{D}_k^A$ is a diagonal matrix which leads to the below update
\begin{equation}
\textbf{D}_{(k,:)}^A=\left[ (\textbf{F}_k \odot \textbf{A}) \textbf{x}_k \right]^T
\quad \textrm{with} \quad 
\textbf{x}_k = \textrm{vec}(\textbf{X}_k) \quad .
\end{equation}

The notation $(k,:)$ represents the \textit{k-th} row of $\textbf{D}_{(k,:)}^A$. To update $\textbf{H}$, the equation \eqref{eq::par_for_als} is rearranged as
\begin{equation} \label{eq::upd_H}
\textbf{x}_k = (\textbf{B}\textbf{D}_k^B \otimes \textbf{A}\textbf{D}_k^A) \textbf{h}
\quad \textrm{with} \quad
\begin{cases}
\textbf{x}_k = & \textrm{vec}(\textbf{X}_k) \\ 
\textbf{h} = & \textrm{vec}(\textbf{H}) \\ 
\end{cases} \quad ,
\end{equation}

which brings the solution 

\begin{equation}
\textbf{h} = \textbf{Z}^\dag \textbf{x}
\quad \textrm{with} \quad
\textbf{Z} = \begin{pmatrix}
\textbf{B}\textbf{D}_1^B \otimes \textbf{A}\textbf{D}_1^A \\ 
\textbf{B}\textbf{D}_2^B \otimes \textbf{A}\textbf{D}_2^A \\ 
\vdots \\
\textbf{B}\textbf{D}_k^B \otimes \textbf{A}\textbf{D}_k^A
\end{pmatrix} \quad .
\end{equation}

The methodology presented for the update of $\textbf{A}$ and $\mathscr{D}^A$ is reproduced to update $\textbf{B}$ and $\mathscr{D}^B$. We can now build upon the standard ALS update rules for the PARATUCK2 tensor decomposition to define the non-negative ALS update rules.  \\

\textbf{Non-negative ALS Method for PARATUCK2}  \\
In the experiments, we use the non-negative PARATUCK2 decomposition adapted from the non-negative matrix factorization presented by Lee and Seung in \cite{lee1999learning}. The matrices, $\textbf{A}$, $\textbf{B}$ and $\textbf{H}$, and the tensors, $\mathscr{D}^A$ and $\mathscr{D}^B$, are computed according to the following multiplicative update rules

{
\allowdisplaybreaks
\begin{align} \label{eq::als_upd}
a_{ip}\leftarrow a_{ip} 
\dfrac{\left[ \textbf{X} \textbf{F}^T\right]_{ip}}
{\left[ \textbf{A}(\textbf{FF}^T) \right]_{ip}} 
\: &, \:
\textbf{F}=\mathscr{D}^A \textbf{H}\mathscr{D}^B \textbf{B}^T \quad , \\
d^a_{pp}\leftarrow d^a_{pp}
\dfrac{\left[ \textbf{Z}^T \textbf{x} \right]_{pp}}
{\left[ \mathscr{D}^A(\textbf{ZZ}^T) \right]_{pp}} 
\: &, \:
\textbf{Z}=(\textbf{B}\mathscr{D}^B \textbf{H}^T) \odot\textbf{A} \quad , \\
h_{pq}\leftarrow h_{pq}
\dfrac{\left[ \textbf{Z}^T \textbf{x} \right]_{pq}}
{\left[ \textbf{H}(\textbf{ZZ}^T) \right]_{pq}} 
\: &, \:
\textbf{Z}=\textbf{B} \mathscr{D}^B \otimes \textbf{A} \mathscr{D}^A \quad , \\
d^b_{qq}\leftarrow d^b_{qq}
\dfrac{\left[ \textbf{x} \textbf{Z} \right]_{qq}}
{\left[ \mathscr{D}^B (\textbf{Z}^T\textbf{Z})\right]_{qq}}
\: &, \:
\textbf{Z}=\textbf{B} \odot (\textbf{H}^T \mathscr{D}^A \textbf{A}^T)^T \quad , \\
b_{qj}\leftarrow b_{qj} 
\dfrac{\left[ \textbf{X}^T \textbf{F}^T \right]_{qj}}
{\left[ \textbf{B}(\textbf{F} \textbf{F}^T) \right]_{qj}} 
\: &, \:
\textbf{F}= (\textbf{A} \mathscr{D}^A \textbf{H}\mathscr{D}^B)^T \quad , \\
\end{align}
}
with 
\begin{equation}
\begin{cases}
\textbf{X} &= \left[ \textbf{X}_1 \: \textbf{X}_2 \cdots \textbf{X}_k \right] \\ 
\textbf{x} &= \textrm{vec}(\mathscr{X}) 
\end{cases} \quad .
\end{equation}

The multiplicative update rules help to an easier interpretation of the tensor decomposition since all the numbers are positive or null. We used this non-negative ALS update in our experiments involving the PARATUCK2 tensor decomposition and the ALS scheme.  \\

\textbf{Non-negative ALS method for DEDICOM}  \\
We recall that the equation for the DEDICOM tensor decomposition is equal to 

\begin{equation}
  \textbf{X}_k = \textbf{A}\textbf{D}_k \textbf{H} \textbf{D}_k \textbf{A}^T
  \quad , \quad \forall \: k = 1, ..., K \quad ,
\end{equation}

with the original tensor $\mathscr{X} \in \mathbb{R}^{I\times I\times K}$, the factor matrices $\textbf{A}\in\mathbb{R}^{I\times R}$, $\textbf{H}\in\mathbb{R}^{R\times R}$ and the factor tensor $\mathscr{D}\in\mathbb{R}^{R\times R \times K}$. 

The DEDICOM and the PARATUCK2 tensor decomposition are very similar. Therefore, we can deduce the non-negative ALS update rule for DEDICOM based on the PARATUCK2 update rules 

{
\allowdisplaybreaks
\begin{align} \label{eq::als_ded}
a_{ip}\leftarrow a_{ip} 
\dfrac{\left[ \textbf{X} \textbf{F}^T\right]_{ip}}
{\left[ \textbf{A}(\textbf{FF}^T) \right]_{ip}} 
\: &, \:
\textbf{F}=\mathscr{D} \textbf{H} \mathscr{D} \textbf{A}^T \quad , \\
d_{pp}\leftarrow d_{pp}
\dfrac{\left[ \textbf{Z}^T \textbf{x} \right]_{pp}}
{\left[ \mathscr{D}(\textbf{ZZ}^T) \right]_{pp}} 
\: &, \:
\textbf{Z}=(\textbf{A}\mathscr{D} \textbf{H}^T) \odot \textbf{A} \quad , \\
h_{pp}\leftarrow h_{pp}
\dfrac{\left[ \textbf{Z}^T \textbf{x} \right]_{pp}}
{\left[ \textbf{H}(\textbf{ZZ}^T) \right]_{pp}} 
\: &, \:
\textbf{Z}=\textbf{A} \mathscr{D} \otimes \textbf{A} \mathscr{D} \quad , \\
d_{pp}\leftarrow d_{pp}
\dfrac{\left[ \textbf{x} \textbf{Z} \right]_{pp}}
{\left[ \mathscr{D}^B  (\textbf{Z}^T\textbf{Z})\right]_{pp}}
\: &, \:
\textbf{Z}=\textbf{A} \odot (\textbf{H}^T \mathscr{D} \textbf{A}^T)^T \quad , \\
a_{pi}\leftarrow a_{pi} 
\dfrac{\left[ \textbf{X}^T \textbf{F}^T \right]_{pi}}
{\left[ \textbf{A}(\textbf{F} \textbf{F}^T) \right]_{pi}} 
\: &, \:
\textbf{F}= (\textbf{A} \mathscr{D} \textbf{H} \mathscr{D})^T \quad , \\
\end{align}
}
with 
\begin{equation}
\begin{cases}
\textbf{X} &= \left[ \textbf{X}_1 \: \textbf{X}_2 \cdots \textbf{X}_k \right] \\ 
\textbf{x} &= \textrm{vec}(\mathscr{X}) 
\end{cases} \quad .
\end{equation}

It is noteworthy that some factors update might not lead to a minimization of the objective function for a given iteration. However, the overall convergence through the successive iterations is always obtained. We used this scheme in our experiments involving the ALS method and the DEDICOM tensor decomposition.  \\

\subsection{The CP Tensor Decomposition and Third Order Financial Predictions}
\begin{figure}[t]
\begin{center}
\begin{tikzpicture}[scale=0.9, transform shape]
\draw (-7.25,-0.55) rectangle (-5,2.45);
\fill [gray, opacity=.4] (-7.25,-0.55) rectangle (-5,2.45);
\draw (-5,2.45) -- (-4,2.9);
\draw (-7.25,2.45) -- (-6.5,2.9);
\draw (-5,-0.55) -- (-4,-0.1);
\draw (-6.5,2.9) -- (-4,2.9);
\draw (-4,-0.1) -- (-4,2.9);
\draw [dashdotted] (-6.5,-0.1) -- (-4,-0.1);
\draw [dashdotted] (-6.5,-0.1) -- (-6.5,2.9);
\draw [dashdotted] (-7.25,-0.55) -- (-6.5,-0.1);

\fill [gray, opacity=.4] (-4,-0.1) -- (-4,2.9) -- (-5,2.45) -- (-5,-0.55)  -- cycle;
\fill [gray, opacity=.4] (-5,2.45) -- (-4,2.9) -- (-6.5,2.9) -- (-7.25,2.45)  -- cycle;

\coordinate (A) at (-0.8,-0.8) {} {} {} {};
\coordinate (B) at (-0.45,1.2) {} {} {} {};
\coordinate (C) at (-0.1,1.05) {} {} {} {};
\coordinate (D) at (1.4,1.4) {} {} {} {};
\coordinate (e1) at (-0.55,1.6) {} {} {} {};
\coordinate (e2) at (-0.1,1.6) {} {} {} {};
\coordinate (e3) at (0.9,2.1) {} {} {} {};
\coordinate (e4) at (0.45,2.1) {} {} {} {};
\draw (A) rectangle (B);
\fill [gray, opacity=.4] (A) rectangle (B);
\draw (C) rectangle (D);
\fill [gray, opacity=.4] (C) rectangle (D);
\draw (e1) -- (e2) -- (e3) -- (e4) -- cycle;
\fill [gray, opacity=.4] (e1) -- (e2) -- (e3) -- (e4) -- cycle;

\draw[- triangle 60] (-3.5,1.2) node (v1) {} -- (-2,1.2);

\draw[gray, opacity=1]  (3,3) circle (0.3);
\fill[gray, opacity=1]  (3,3) circle (0.3);
\draw[gray, opacity=1]  (3,1.2) circle (0.3);
\fill[gray, opacity=1]  (3,1.2) circle (0.3);
\draw[gray, opacity=1]  (3,-0.5) circle (0.3);
\fill[gray, opacity=1]  (3,-0.5) circle (0.3);

\draw[gray, opacity=1]  (4.5,2.2) circle (0.3);
\fill[gray, opacity=1]  (4.5,2.2) circle (0.3);
\draw[gray, opacity=1]  (4.5,1.2) circle (0.3);
\fill[gray, opacity=1]  (4.5,1.2) circle (0.3);
\draw[gray, opacity=1]  (4.5,0.2) circle (0.3);
\fill[gray, opacity=1]  (4.5,0.2) circle (0.3);

\draw[gray, opacity=1]  (5.95,1.2) circle (0.3);
\fill[gray, opacity=1]  (5.95,1.2) circle (0.3);

\draw[- triangle 60] (1.1,2.2) -- (2.6,2.8);
\draw[- triangle 60] (1.6,1.2) -- (2.6,1.2);
\draw[- triangle 60] (-0.2,-0.2) -- (2.6,-0.5);

\draw[gray, opacity=.5] (3,3) -- (4.5,2.2);
\draw[gray, opacity=.5] (3,3) -- (4.5,1.2);
\draw[gray, opacity=.5] (3,3) -- (4.5,0.2);

\draw[gray, opacity=.5] (3,1.2) -- (4.5,2.2);
\draw[gray, opacity=.5] (3,1.2) -- (4.5,1.2);
\draw[gray, opacity=.5] (3,1.2) -- (4.5,0.2);

\draw[gray, opacity=.5] (3,-0.5) -- (4.5,0.2);
\draw[gray, opacity=.5] (3,-0.5) -- (4.5,2);
\draw[gray, opacity=.5] (3,-0.5) -- (4.5,1.2);

\draw[gray, opacity=.5] (4.5,2.2) -- (5.95,1.2);
\draw[gray, opacity=.5] (4.5,1.2) -- (5.95,1.2);
\draw[gray, opacity=.5] (4.5,0.2) -- (5.95,1.2);

\draw[- triangle 60] (-7.25,4) -- (1.6,4);
\draw[- triangle 60] (2.6,4) -- (6.5,4);

\draw[decoration={brace, mirror,raise=5pt, amplitude=7pt},decorate] (-7,-1.0) -- (-4.5,-1.0);
\draw[decoration={brace, mirror,raise=5pt, amplitude=7pt},decorate] (-1.5,-1.0) -- (1.25,-1.0);
\draw[decoration={brace, mirror,raise=5pt, amplitude=7pt},decorate] (2.75,-1.0) -- (6.25,-1.0);

\node (none) at (-5.5,-1.85) {\large $\mathcal{X}_{true}$};
\node (none) at (-1.35,1.3)[scale=2] {\large $\sum$};
\node (none) at (0.25,-1.85) {\large $\mathcal{X}_{target}$};
\node (none) at (4.5,-1.85) {\tiny NEURAL NETWORK};

\node (none) at (0,0.2) {\small $a_i^{(1)}$};
\node (none) at (0.7,0.65) {\small $a_i^{(2)}$};
\node (none) at (-0.35,2.05) {\small $a_i^{(3)}$};

\node (none) at (-2.7,1.45) {\small $W_c$};

\node (none) at (4.5,4.5) {\tiny \textbf{NEURAL NETWORK}};
\node (none) at (4.5,4.25) {\tiny \textbf{FOR PREDICTIONS}};
\node (none) at (-2.8,4.25) {\tiny \textbf{REMOVE SPARSE INFORMATION}};
\end{tikzpicture}
\caption[Neural networks predictions with the CP tensor decomposition]{The CP tensor decomposition factorizes a third-order tensor $\mathcal{X}_{true}$ of rank $R$ into a sum of rank-one tensors, denoted by $\mathcal{X}_{target}$, removing the sparsity of the transactions. The factor vectors are denoted by $\textbf{a}^{(i)}_r$ with $r=1,...,R$ and $i=\left\lbrace1,2,3\right\rbrace$. The neural networks are used to predict the next financial transactions of the clients.}
\label{fig::cp_sparsepredictions}
\end{center}
\end{figure}
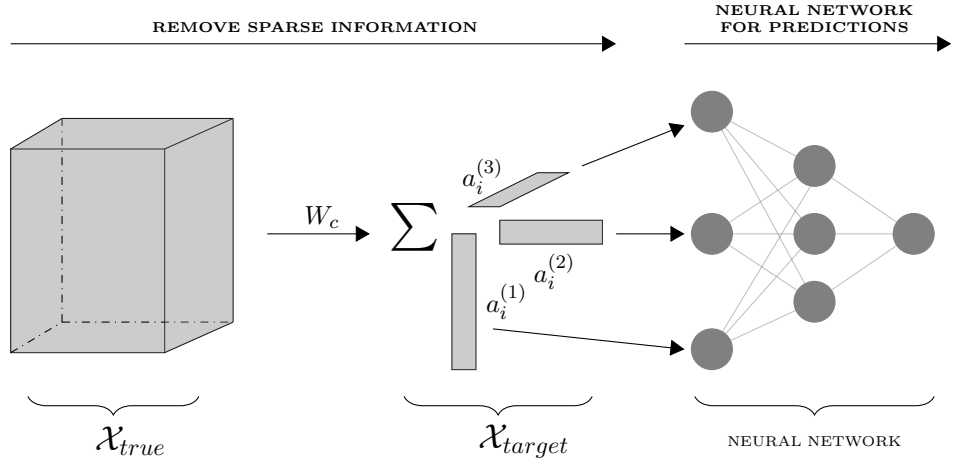

Building upon VecHGrad, the CP tensor decomposition and the neural networks, we use a two-steps procedure for the predictions of the financial transactions of the clients, as illustrated in \ref{fig::cp_sparsepredictions}. First, the clients' transactions are stored in a third order tensor, denoted by $\mathcal{X}_{true}$. The tensor $\mathcal{X}_{true}$ is then decomposed using the CP tensor factorization to remove the sparsity contained in the transactions. The VecHGrad resolution algorithm is used with the equations (\ref{eq::cpequation}) and (\ref{eq::errorminimization}) to ensure an accurate CP tensor decomposition. The sparse information contained in $\mathcal{X}_{true}$ is removed from the factor vectors $\textbf{a}^{(1)}, ..., \textbf{a}^{(N)}$ of $\mathcal{X}_{target}$. The factor vectors furthermore highlight the different latent groups of clients and transactions. In a second step, the factor vectors $\textbf{a}^{(1)}, ..., \textbf{a}^{(N)}$ of $\mathcal{X}_{target}$ are then mapped to the inputs of the neural networks to achieve third order financial predictions. The neural networks are able to predict the financial activities of the bank's clients through the training of the factor vectors data set to learn the function $g(.): \mathbb{R}^3 \rightarrow \mathbb{R}^1$.
\\

Given the shape of the factor vectors used for the predictions, it is impossible to guess at first glance which neural networks would predict the most accurately the financial transactions. We therefore use and compare different neural networks architecture and a machine learning regression for the predictions. The Decision Trees (DT) are a widely used machine learning technique \cite{lior2014data} to predict the value of a variable by learning simple decision rules from the data \cite{kim2015decision,breiman2017classification}. Their regression decision rules however have some limitations. Outpacing DT capabilities, neural networks including Multi-Layer Perceptron (MLP), Convolutional Neural Network (CNN) and Long-Short-Term-Memory (LSTM), and their applications, have therefore skyrocketed for the past few years \cite{liu2017survey}. MLP consists of at least three layers: one input layer, one output layer and one or more hidden layer \cite{goodfellow2016deep}. Each neuron of the hidden layer transforms the values of the previous layer with a non-linear activation function. Although MLP is applied in deep learning, it lacks the possibility of modeling short term and long term events. This feature is found in LSTM \cite{goodfellow2016deep}. The LSTM has a memory block connected to the input gate and the output gate. The memory block is activated through a forget gate, resetting the memory information. Moreover, CNN is worth considering for classification and computer vision. In a CNN, the neurons are capable of extracting high order features in successive layers \cite{hubel1959receptive}. Through proper classification, the CNN is able to detect and predict various tasks including activities recognition \cite{abdel2012applying, zheng2014time}. We summarized the methodology of the financial predictions with the CP tensor decomposition in Algorithm \ref{algo:cp_nn}.

%

\SetAlFnt{\scriptsize}
\SetAlCapFnt{\scriptsize}
\SetAlCapNameFnt{\scriptsize}

\begin{algorithm}[t]
\setstretch{1.25}
\DontPrintSemicolon

Receive the CP tensor decomposition equation:
    $g: \left\{\begin{matrix*}[l]
  \mathbb{R}^d \rightarrow  \mathbb{R}^{I_1\times I_2\times ...\times I_n}  \\ 
  \Tilde{\textbf{x}}^t \mapsto \mathscr{\Tilde{X}} \approx 
  \sum\limits_{r=1}^{R} \textbf{a}_r^{(1)} \circ \textbf{a}_r^{(2)} \circ \textbf{a}_r^{(3)} \circ... \circ \textbf{a}_r^{(N)}
  \end{matrix*}\right.$ \\

Receive $\Tilde{\textbf{x}}^0 \:=\: \text{vec}(\mathscr{\hat{X}})$  \\
\Repeat{convergence criteria is reached}{
  VecHGrad optimization loop for $t$ steps
  
  $\quad$ with loss function: 
  $f: \left\{\begin{matrix*}[l]
  \mathbb{R}^d \rightarrow \mathbb{R} \\ 
  \Tilde{\textbf{x}}^t \mapsto \parallel \mathscr{X} - g(\Tilde{\textbf{x}}^t) \parallel
  \end{matrix*}\right. $\\

  Update parameters:
      $\Tilde{\textbf{x}}^{t+1}$
}

\tcc{$\textbf{A} = \textbf{a}^{(1)}, \textbf{B} = \textbf{a}^{(2)}, \textbf{C} = \textbf{a}^{(3)}$}

$\textbf{A}, \textbf{B}, \textbf{C} \gets$ unflatten($\Tilde{\textbf{x}}^{opt}$)

Send $\textbf{A}, \textbf{B}, \textbf{C}$ to the input of the NN

Training of the NN to learn the function $g(.): \mathbb{R}^3\rightarrow \mathbb{R}^1$

$\textbf{y} \in \mathbb{R}^1$ $\gets$ NN prediction of financial activities

\caption{Financial predictions with the CP tensor decomposition \label{algo:cp_nn}}
\end{algorithm}

\subsection{Monitoring of User-Device Authentication with the PA\-RA\-TUCK2 Tensor Decomposition}
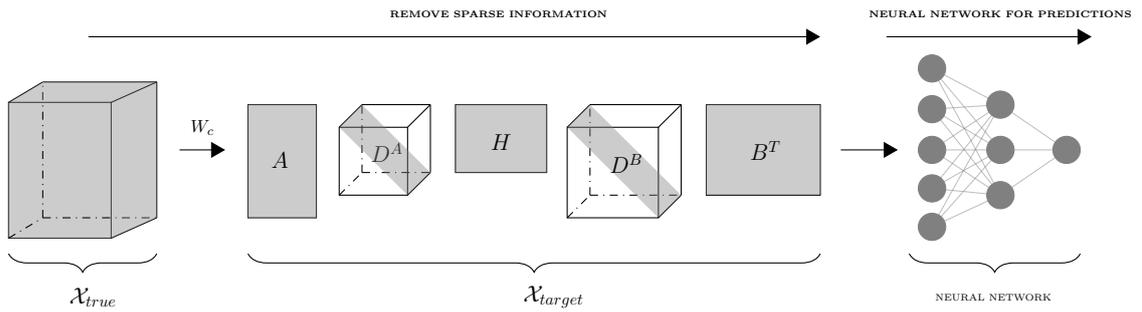
\begin{figure}[b!]
\begin{center}
\begin{tikzpicture}[scale=0.6, transform shape]

\draw[- triangle 60] (-6.5,5) node (v1) {} -- (-5.5,5);
\node (none) at (-6.0,5.5) {\small $W_c$};

\draw[- triangle 60] (-8.5,7.5) -- (7.55,7.5);
\draw[- triangle 60] (9,7.5) -- (13.5,7.5);

\draw[decoration={brace, mirror,raise=5pt, amplitude=7pt},decorate] (-10.25,3.0) -- (-7.0,3.0);
\draw[decoration={brace, mirror,raise=5pt, amplitude=7pt},decorate] (-5.0,3.0) -- (7.55,3.0);
\draw[decoration={brace, mirror,raise=5pt, amplitude=7pt},decorate] (9.5,3.0) -- (13.25,3.0);

\node (none) at (-8.40,1.75) {\large $\mathcal{X}_{true}$};
\node (none) at (1.7,1.75) {\large $\mathcal{X}_{target}$};
\node (none) at (11.35,1.75) {\tiny NEURAL NETWORK};
\node (none) at (11.5,8.0) {\tiny \textbf{NEURAL NETWORK FOR PREDICTIONS}};
\node (none) at (.5,8.0) {\tiny \textbf{REMOVE SPARSE INFORMATION}};

\draw (-10.25,3.05) rectangle (-8,6.05);
\fill [gray, opacity=.4] (-10.25,3.05) rectangle (-8,6.05);
\draw (-8,6.05) -- (-7,6.5);
\draw (-10.25,6.05) -- (-9.5,6.5);
\draw (-8,3.05) -- (-7,3.5);
\draw (-9.5,6.5) -- (-7,6.5);
\draw (-7,3.5) -- (-7,6.5);
\draw [dashdotted] (-9.5,3.5) -- (-7,3.5);
\draw [dashdotted] (-9.5,3.5) -- (-9.5,6.5);
\draw [dashdotted] (-10.25,3.05) -- (-9.5,3.5);

\fill [gray, opacity=.4] (-7,3.5) -- (-7,6.5) -- (-8,6.05) -- (-8,3.05)  -- cycle;
\fill [gray, opacity=.4] (-8,6.05) -- (-7,6.5) -- (-9.5,6.5) -- (-10.25,6.05)  -- cycle;

\draw  (-5,6) rectangle (-3.5,3.5);
\fill [gray, opacity=.4]  (-5,6) rectangle (-3.5,3.5);
\draw (-4.3,5.1) node[below]{\large{$A$}};

\draw (2,3.5) rectangle (4,5.5) node (v1) {};
\draw (4,5.5) -- (4.5,6);
\draw (2,5.5) -- (2.5,6);
\draw (4,3.5) -- (4.5,4);
\draw (2.5,6) -- (4.5,6);
\draw (4.5,4) -- (4.5,6);
\draw [dashdotted] (2.5,4) -- (4.5,4);
\draw [dashdotted] (2.5,4) -- (2.5,6);
\draw [dashdotted] (2,3.5) -- (2.5,4);
\fill [gray, opacity=.4] (4,3.5) -- (4.5,4) -- (2.5,6) -- (2,5.5)  -- cycle;
\draw (-1.95,5.3) node[below]{\large{$D^A$}};

\draw (-0.45,6) rectangle (1.55,4.5);
\fill [gray, opacity=.4] (-0.45,6) rectangle (1.55,4.5);
\draw (0.55,5.5) node[below]{\large{$H$}};

\draw (-3,4) rectangle (-1.5,5.5) node (v1) {};
\draw (-1.5,5.5) -- (-1,6);
\draw (-3,5.5) -- (-2.5,6);
\draw (-1.5,4) -- (-1,4.5);
\draw (-2.5,6) -- (-1,6);
\draw (-1,4.5) -- (-1,6);
\draw [dashdotted] (-2.5,4.5) -- (-1,4.5);
\draw [dashdotted] (-2.5,4.5) -- (-2.5,6);
\draw [dashdotted] (-3,4) -- (-2.5,4.5);
\fill [gray, opacity=.4] (-1.5,4) -- (-1,4.5) -- (-2.5,6) -- (-3,5.5)  -- cycle;
\draw (3.3,5.05) node[below]{\large{$D^B$}};

\draw  (5.05,6) rectangle (7.55,4);
\fill [gray, opacity=.4] (5.05,6) rectangle (7.55,4);
\draw (6.35,5.35) node[below]{\large{$B^T$}};

\draw[gray, opacity=1]  (10,6.8) circle (0.3);
\fill[gray, opacity=1]  (10,6.8) circle (0.3);
\draw[gray, opacity=1]  (10,5.9) circle (0.3);
\fill[gray, opacity=1]  (10,5.9) circle (0.3);
\draw[gray, opacity=1]  (10,5) circle (0.3);
\fill[gray, opacity=1]  (10,5) circle (0.3);
\draw[gray, opacity=1]  (10,4.15) circle (0.3);
\fill[gray, opacity=1]  (10,4.15) circle (0.3);
\draw[gray, opacity=1]  (10,3.3) circle (0.3);
\fill[gray, opacity=1]  (10,3.3) circle (0.3);

\draw[gray, opacity=1]  (11.5,6) circle (0.3);
\fill[gray, opacity=1]  (11.5,6) circle (0.3);
\draw[gray, opacity=1]  (11.5,5) circle (0.3);
\fill[gray, opacity=1]  (11.5,5) circle (0.3);
\draw[gray, opacity=1]  (11.5,4) circle (0.3);
\fill[gray, opacity=1]  (11.5,4) circle (0.3);

\draw[gray, opacity=1]  (12.95,5) circle (0.3);
\fill[gray, opacity=1]  (12.95,5) circle (0.3);

\draw[- triangle 60] (8,5) -- (9.25,5);

\draw[gray, opacity=.5] (10,6.8) -- (11.5,6);
\draw[gray, opacity=.5] (10,6.8) -- (11.5,5);
\draw[gray, opacity=.5] (10,6.8) -- (11.5,4);

\draw[gray, opacity=.5] (10,5.9) -- (11.5,6);
\draw[gray, opacity=.5] (10,5.9) -- (11.5,5);
\draw[gray, opacity=.5] (10,5.9) -- (11.5,4);

\draw[gray, opacity=.5] (10,5) -- (11.5,6);
\draw[gray, opacity=.5] (10,5) -- (11.5,5);
\draw[gray, opacity=.5] (10,5) -- (11.5,4);

\draw[gray, opacity=.5] (10,4.15) -- (11.5,6);
\draw[gray, opacity=.5] (10,4.15) -- (11.5,5);
\draw[gray, opacity=.5] (10,4.15) -- (11.5,4);

\draw[gray, opacity=.5] (10,3.3) -- (11.5,4);
\draw[gray, opacity=.5] (10,3.3) -- (11.5,6);
\draw[gray, opacity=.5] (10,3.3) -- (11.5,5);

\draw[gray, opacity=.5] (11.5,6) -- (12.95,5);
\draw[gray, opacity=.5] (11.5,5) -- (12.95,5);
\draw[gray, opacity=.5] (11.5,4) -- (12.95,5);

\end{tikzpicture}
\caption[Neural networks predictions with the PARATUCK2 tensor decomposition]{The PARATUCK2 tensor decomposition factorizes a third-order tensor $\mathcal{X}_{true}$ into a multiplication of latent matrices and tensors, denoted by $\mathcal{X}_{target}$. It removes the sparsity of the user-device authentication while preserving the data set imbalance. The factor matrices and factor tensors are denoted by $\textbf{A}, \textbf{H}, \textbf{B}$ and $\mathcal{D}^\textbf{A}, \mathcal{D}^\textbf{B}$. The neural networks are used to predict the next user-device authentication.}
\label{fig::par_authen}
\end{center}
\end{figure}

By adapting the methodology of the third order financial activities predictions for the user-device authentication monitoring with the PARATUCK2 tensor decomposition, we use a two-steps procedure. The procedure is capable to reduce the dimensions of the initial data set by keeping only the useful information while being able to perform accurate predictions. The proposed approach is illustrated in Figure \ref{fig::par_authen}. 
\\

The user-device authentication are first stored in a third order tensor, denoted by $\mathcal{X}_{true}$. The tensor $\mathcal{X}_{true}$ is then decomposed into three factor matrices $\textbf{A}, \textbf{B}$ and $\textbf{H}$ and two factor tensors $\mathcal{D}^\textbf{A}$ and $\mathcal{D}^\textbf{B}$ using the PARATUCK2 tensor decomposition. It allows to reduce the dimensions of the data set to an interpretable size while keeping all the useful authentication patterns contained in the original tensor $\mathcal{X}_{true}$. The reason to choose the PARATUCK2 tensor decomposition additionally is that a client, a user of the mobile application, can authenticate several times per day using different devices. Our tensor decomposition has therefore to be capable of modeling the asymmetric relationships between the users and the devices. This is the case for the PARATUCK2 tensor decomposition, contrarily to the CP tensor decomposition. We rely on the VecHGrad resolution algorithm using the equations (\ref{eq::paratuck2}) and (\ref{eq::errorminimization}) to ensure an accurate PARATUCK2 tensor decomposition. Each of the factor matrices and of the factor tensors highlight respectively the user authentication and the device authentication. We provide more practical details in the experiments section.
\\

In a second step, the factor matrices and the factor tensors are mapped to the inputs of the neural networks to achieve the user authentication predictions. We recall we aim at building a financial awareness score based on the client's habits on the mobile banking application. We consequently concentrate our effort on the user authentication that has been isolated from the device authentication thanks to the PARATUCK2 tensor decomposition. The factor matrix $\textbf{A}$ and tensor $\mathcal{D}^\textbf{A}$ are used to train the neural networks to learn the function $g(.): \mathbb{R}^3 \rightarrow \mathbb{R}^1$ in order to perform predictions of the next user authentications. Similarly to the financial actions predictions of the clients, we cannot guess which neural networks will predict best the next user authentication. We therefore train independently and compare the three neural networks aforementioned including a MLP \cite{goodfellow2016deep}, a CNN\cite{hubel1959receptive} and a LSTM \cite{goodfellow2016deep} as well as a DT \cite{lior2014data}. We summarized our approach for the user-device authentication monitoring with the PARATUCK2 tensor decomposition in Algorithm \ref{algo:par_nn}. 
\\

\SetAlFnt{\scriptsize}
\SetAlCapFnt{\scriptsize}
\SetAlCapNameFnt{\scriptsize}

\begin{algorithm}[t!]
\setstretch{1.25}
\DontPrintSemicolon

Receive the PARATUCK2 tensor decomposition equation:
    $g: \left\{\begin{matrix*}[l]
  \mathbb{R}^d \rightarrow  \mathbb{R}^{I_1\times I_2\times ...\times I_n}  \\ 
  \Tilde{\textbf{x}}^t \mapsto \Tilde{\textbf{X}} _k \approx 
  \textbf{A}\textbf{D}^A_k\textbf{H}\textbf{D}^B_k\textbf{B}^T \quad \textrm{with} \quad k=\left\lbrace 1, ..., K \right\rbrace
  \end{matrix*}\right.$ \\

Receive $\Tilde{\textbf{x}}^0 \:=\: \text{vec}(\mathscr{\hat{X}})$  \\
\Repeat{convergence criteria is reached}{
  VecHGrad optimization loop for $t$ steps $=0,1,2,...,T$ to determine $\Tilde{\textbf{x}}^{opt}$
  
  $\quad$ with loss function: 
  $f: \left\{\begin{matrix*}[l]
  \mathbb{R}^d \rightarrow \mathbb{R} \\ 
  \Tilde{\textbf{x}}^t \mapsto \parallel \mathscr{X} - g(\Tilde{\textbf{x}}^t) \parallel
  \end{matrix*}\right. $\\

  At each step $t$, update vector $\Tilde{\textbf{x}}^{t+1}$
}

$\textbf{A}, \mathcal{D}^\textbf{A}, \textbf{H}, \mathcal{D}^\textbf{B}, \textbf{B} \gets$ unflatten($\Tilde{\textbf{x}}^{opt}$)

Send $\textbf{A}$ and $\mathcal{D}^\textbf{A}$ to the input of the NN

Training of the NN to learn the function $g(.): \mathbb{R}^3\rightarrow \mathbb{R}^1$

$\textbf{y} \in \mathbb{R}^1$ $\gets$ NN prediction of financial activities

\caption{User-device authentication monitoring with the PARATUCK2 tensor decomposition \label{algo:par_nn}}
\end{algorithm}

\section{Experiments} \label{sec::exp3}
In this section, we present our three experiments in the context of accurate resolution of tensor decomposition for financial recommendation. We first present the results of the VecHGrad optimization algorithm. We then apply the superior accuracy of the VecHGrad optimization algorithm to the task of predictions of sparse financial transactions of the bank's clients. In the last experiment, we use VecHGrad to perform efficient and precise monitoring of user-device authentication for mobile banking.
\\

\subsection{VecHGrad for Accurate Tensor Resolution} \label{subsection::vechgrad}
Hereinafter, we investigate the convergence behavior of VecHGrad in comparison to other popular numerical resolution methods inherited from both the tensor and the machine learning communities. We compare VecHGrad with ten different algorithms applied to the three main tensor decomposition with increasing linear algebra complexity, CP/PARAFAC, DEDICOM and PARATUCK2: 
\begin{itemize} \setlength{\itemsep}{0pt} \setlength{\parskip}{0pt}
\item ALS, Alternating Least Squares \cite{bro1998multi,charlier2018non};
\item SGD, Gradient Descent \cite{wright1999numerical};
\item NAG, Nesterov Accelerated Gradient \cite{nesterov2007gradient};
\item Adam \cite{kingma2014adam};
\item RMSProp \cite{hinton2012rmsprop};
\item SAGA \cite{defazio2014saga};
\item AdaGrad \cite{duchi2011adaptive};
\item CP-OPT and the Non-linear Conjugate Gradient (NCG) \cite{acar2011scalable,acar2011all};
\item L-BFGS \cite{liu1989limited} inherited from BFGS \cite{broyden1970convergence,fletcher1970new,goldfarb1970family,shanno1970conditioning}.
\end{itemize}

\textbf{Data Availability and Code Availability}
We highlight VecHGrad using popular data sets including CIFAR10, CIFAR100, MNIST, LFW and COCO. All the data sets are available online. Each data set has different intrinsic characteristics such as the size or the sparsity. A quick overview of the data set features is presented in Table \ref{tab::dataset}. We chose to use different data sets as the performance of the different optimizers might vary slightly depending on the data. The overall conclusion of the experiments therefore is independent of one particular data set. 
The implementation and the code of the experiments are available on GitHub\footnote{The code is available at https://github.com/dagrate/vechgrad.}.  \\

\begin{table}[t]
  \caption[Description of the data sets used]{Description of the data sets used (K: thousands).}
  \label{tab::dataset}
  \centering
  \scalebox{0.9}{
    \setlength{\tabcolsep}{10pt}
    \begin{tabular}{cccc}
	\toprule
    Data Set & Labels & Size & Batch Size\\
    \midrule
    CIFAR-10 & image $\times$ pixels $\times$ pixels & 50K $\times$ 32 $\times$ 32 & 64 \\
    CIFAR-100 & image $\times$ pixels $\times$ pixels & 50K $\times$ 32 $\times$ 32 & 64 \\
    MNIST & image $\times$ pixels $\times$ pixels & 60K $\times$ 28 $\times$ 28 & 64 \\
    COCO & image $\times$ pixels $\times$ pixels & 123K $\times$ 64 $\times$ 64 & 32 \\
    LFW & image $\times$ pixels $\times$ pixels & 13K $\times$ 64 $\times$ 64 & 32 \\
    \bottomrule
  \end{tabular}
  }
\end{table}

\textbf{Experimental Setup}
In our experiments, we use the standard parameters for the popular ML and DL gradient optimization methods. We use $\eta = 10^{-4}$ for SGD, $\gamma = 0.9$ and $\eta = 10^{-4}$ for NAG, $\beta_1 = 0.9, \beta_2 = 0.999, \epsilon = 10^{-8}$ and $\eta = 0.001$ for Adam, $\gamma = 0.9, \eta = 0.001$ and $\epsilon = 10^{-8}$ for RMSProp, $\eta = 10^{-4}$ for SAGA, $\eta = 0.01$ and $\epsilon = 10^{-8}$ for AdaGrad. We use the Hestenes-Stiefel update \cite{hestenes1952methods} for the NCG resolution. The convergence criteria is moreover reached when $f^{i+1} - f^i \leq 0.001$ or when the maximum number of iteration is reached. We use 100,000 iterations for the gradient-free methods, 10,000 iterations for the gradient methods and 1,000 iterations for the Hessian-based methods. We additionally fixed the number of iterations to 20 for the VecHGrad's inner CG loop, used to determine the descent direction. We invite the reader to review the code available on GitHub\footnote{The code is available at https://github.com/dagrate/vechgrad.} for further knowledge about the parameters used. The simulations are conducted on a server with 50 Intel Xeon E5-4650 CPU cores and 50GB of RAM. All the resolution schemes have been implemented in Julia and are compatible for the ArrayFire GPU accelerator library.  \\

\textbf{Results and Discussions}
Two different experiments are performed to highlight the strengths and the weaknesses of the optimizers aforementioned. The evaluation of the optimizers' performance is first based on a visual sample and, then, we provide a detailed overview of the mean loss function at convergence and the mean calculation time over all batches for the 
data sets CIFAR10, CIFAR100, MNIST, LFW and COCO.
\\
%

\begin{figure}[t!]
\centering
\includegraphics[scale=.25]{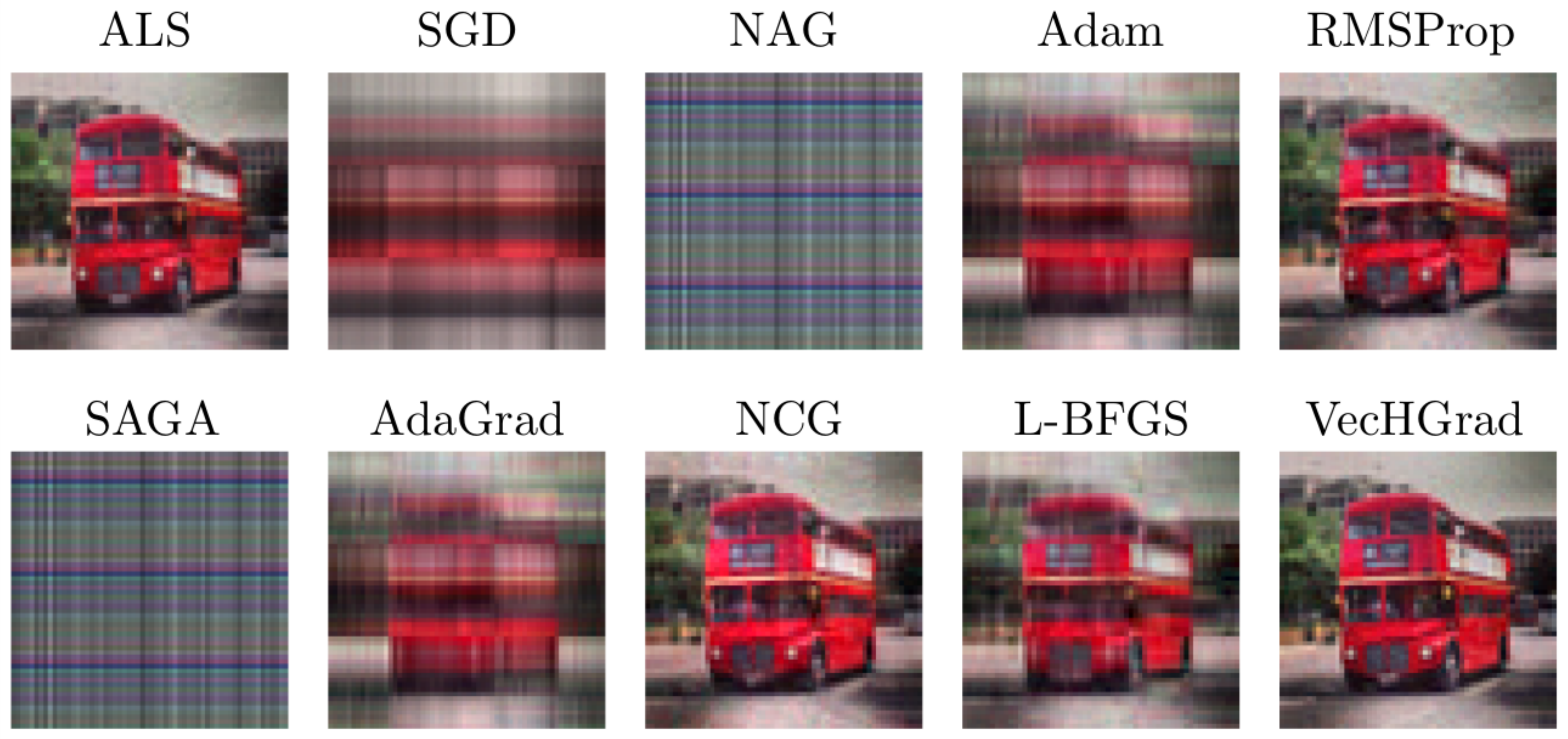}
\caption[Visual output at convergence of the different optimizers]{Visual simulation of the accuracy at convergence of the different optimizers for the PARATUCK2 decomposition. The accuracy at convergence is highlighted by how blurry the image is (the less blurry, the better). The popular gradient optimizers AdaGrad, NAG and SAGA failed to converge to a solution close to the original image, contrarily to VecHGrad or RMSProp.} 
\label{fig::visualplot}
\end{figure}

In our first experiment, we highlight visually the strengths of each of the optimization algorithms aforementioned. Figure \ref{fig::visualplot} depicts the resulting error of the loss function of each of the methods at convergence for the PARATUCK2 tensor decomposition. We voluntarily chose latent components for which the numerical optimization would be difficult since we are interested to highlight the differences of convergence for complex optimization, and not to reproduce a good image quality. The error of the loss function, or how accurate a method is, is reflected by the blurriness of the picture. The less the image is blurry, the lower the loss function error at convergence. As it can be noticed, some numerical methods, including ALS, RMSProp or VecHGrad, offer the best observable image quality at convergence, given our choice of parameters. However, other popular schemes, including NAG and SAGA, fail to converge to a solution resulting in a noisy image, far from being close to the original image.  \\

In a second experiment, we compare in Tables \ref{tab::accuracyparatuck2} and \ref{tab::timeparatuck2} the loss function errors and the calculation times of the numerical optimization methods on the five ML data sets, CIFAR-10, CIFAR-100, MNIST, COCO and LFW, for the three tensor decomposition CP, DEDICOM and PARATUCK2. Both the loss function errors and the calculation times are computed based on the mean of the loss function errors and the mean of the calculation times over all batches. As it can be observed, the numerical schemes of the NAG, SAGA and AdaGrad algorithms fail to minimize the error of the loss function accurately. We have furthermore to mention that the ALS scheme offers a good compromise between the resulting errors and the required calculation times, explaining its major success across tensor decomposition applications. The gradient descent optimizers, Adam and RMSProp, and the Hessian based optimizers, VecHGrad and L-BFGS, are capable to minimize the most accurately the loss function. The NCG method achieves satisfying errors for the CP and the DEDICOM decomposition but its performance decreases significantly when trying to solve the complex PARATUCK2 decomposition. Surprisingly, the calculation times of the Adam and RMSProp gradient descents are greater than the calculation times of VecHGrad. VecHGrad is capable to outperform the gradient descent schemes on both accuracy and speed thanks to the use of the vector Hessian approximation, inherited from gradient information, and its adaptive strong Wolfe's line search. We can therefore conclude that VecHGrad is capable to solve accurately and efficiently complex numerical optimizations, including complex tensor decomposition, whereas, surprisingly, some popular machine learning gradient descents, such as SAGA, NAG or AdaGrad, fail or show average performance.  \\

\begin{table}[t!]
\centering
\caption[Mean of the loss function errors of the different optimizers]{Mean of the loss function errors at convergence over all batches. The lower, the better (best result in bold).}
\label{tab::accuracyparatuck2}
\scalebox{0.9}{
\setlength{\tabcolsep}{6.25pt}
\begin{tabular}{ccccccccccc}
\toprule
  Decomposition & Optimizer & CIFAR-10 & CIFAR-100 & MNIST & COCO & LFW \\
\midrule
  CP & ALS & 318.667 & 428.402 & 897.766 & 485.138 & 4792.605 \\
  CP & SGD & 2112.904 & 2825.710 & 2995.528 & 3407.415 & 7599.458 \\
  CP & NAG & 4338.492 & 5511.272 & 4916.003 & 8187.315 & 18316.589 \\
  CP & Adam & 1578.225 & 2451.217 & 1631.367 & 2223.211 & 6644.167 \\
  CP & RMSProp & 127.961 & 128.137 & 200.002 & 86.792 & 4205.520 \\
  CP & SAGA & 4332.879 & 5501.528 & 4342.708 & 6327.580 & 13242.181 \\
  CP & AdaGrad & 3142.583 & 4072.551 & 2944.768 & 4921.861 & 10652.488 \\
  CP & NCG & 41.990 & 37.086 & 23.320 & 76.478 & 4130.942 \\
  CP & L-BFGS & 195.298 & 525.279 & 184.906 & 596.160 & 4893.815 \\
  CP & VecHGrad & \textbf{$\leq$ 0.100} & \textbf{$\leq$ 0.100} & \textbf{$\leq$ 0.100} & \textbf{$\leq$ 0.100} & \textbf{$\leq$ 0.100} \\[.15cm]

DEDICOM & ALS & 1350.991 & 	1763.718 & 	1830.830 & 	1894.742 & 	3193.685  \\
DEDICOM & SGD & 435.780 & 	456.051 & 	567.503 & 	406.760 & 	511.093   \\
DEDICOM & NAG & 4349.151 & 	5722.073 & 	4415.687 & 	6325.638 & 	9860.454  \\
DEDICOM & Adam & 579.723 & 	673.316 & 	575.341 & 	743.977 & 	541.515   \\
DEDICOM & RMSProp & 63.795 & 	236.974 & 	96.240 & 	177.419 & 	33.224    \\
DEDICOM & SAGA & 4285.512 & 	5577.981 & 	4214.771 & 	5797.562 & 	8128.724  \\ 
DEDICOM & AdaGrad & 1962.966 & 	2544.436 & 	1452.278 & 	2851.649 & 	3033.965  \\ 
DEDICOM & NCG & 550.554 & 	321.332 & 	171.181 & 	583.430 & 	711.549   \\
DEDICOM & L-BFGS & 423.802 & 	561.689 & 	339.284 & 	435.188 & 	511.620   \\
DEDICOM & VecHGrad & \textbf{$\leq$ 0.100} & \textbf{$\leq$ 0.100} & \textbf{$\leq$ 0.100} & \textbf{$\leq$ 0.100} & \textbf{$\leq$ 0.100} \\[.15cm]

PARATUCK2 & ALS & 408.724 & 	480.312 & 	1028.250 & 	714.623 & 	658.284   \\ 
PARATUCK2 & SGD & 639.556 & 	631.870 & 	1306.869 & 	648.962 & 	495.188   \\
PARATUCK2 & NAG & 4699.058 & 	6046.024 & 	5168.824 & 	8205.223 & 	14546.438 \\ 
PARATUCK2 & Adam & 512.725 & 	680.653 & 	591.156 & 	594.687 & 	615.731   \\ 
PARATUCK2 & RMSProp & 133.416 & 	145.766 & 	164.709 & 	134.047 & 	174.769   \\ 
PARATUCK2 & SAGA & 4665.435 & 	5923.178 & 	4934.328 & 	6350.172 & 	8847.886  \\ 
PARATUCK2 & AdaGrad & 1775.433 & 	2310.402 & 	1715.316 & 	2752.348 & 	2986.919  \\ 
PARATUCK2 & NCG & 772.634 & 	1013.032 & 	270.288 & 	335.532 & 	15181.961 \\ 
PARATUCK2 & L-BFGS & 409.666 & 	522.158 & 	464.259 & 	467.139 & 	416.761   \\ 
PARATUCK2 & VecHGrad & \textbf{$\leq$ 0.100} & \textbf{$\leq$ 0.100} & \textbf{$\leq$ 0.100} & \textbf{$\leq$ 0.100} & \textbf{$\leq$ 0.100} \\
\bottomrule
\end{tabular}
}
\end{table}

\begin{table}[t!]
\centering
\caption[Mean calculation times of the different optimizers to reach convergence]{Mean calculation times (sec.) to reach convergence over all batches. The convergence is reached when the evolution of the loss function errors between the iterations is smaller than 0.05. The results have to be analyzed with Table \ref{tab::accuracyparatuck2} as a small calculation time does not ensure a small loss function error.
}
\label{tab::timeparatuck2}
\scalebox{0.9}{
\setlength{\tabcolsep}{6.25pt}
\begin{tabular}{ccccccccccc}
\toprule
  Decomposition & Optimizer & CIFAR-10 & CIFAR-100 & MNIST & COCO & LFW \\
\midrule
CP & ALS & 5.289 & 	4.584 & 	2.710 & 	5.850 & 	4.085	 \\ 
CP & SGD & 1060.455 & 	1019.432 & 	0.193 & 	2335.060 & 	6657.985 \\
CP & NAG & 280.432 & 	256.196 & 	0.400 & 	1860.660 & 	1.317    \\ 
CP & Adam & 2587.467 & 	2771.068 & 	2062.562 & 	6667.673 & 	6397.708 \\ 
CP & RMSProp & 2013.424 & 	2620.088 & 	2082.481 & 	5588.660 & 	4975.279 \\ 
CP & SAGA & 1141.374 & 	1160.775 & 	0.191 & 	3504.593 & 	3692.471 \\ 
CP & AdaGrad & 1768.562 & 	2324.147 & 	959.408 & 	3729.306 & 	6269.536 \\ 
CP & NCG & 315.132 & 	165.983 & 	4.910 & 	778.279 & 	716.355  \\ 
CP & L-BFGS & 2389.839 & 	2762.555 & 	2326.405 & 	5936.053 & 	5494.634 \\ 
CP & VecHGrad & 200.417 & 	583.117 & 	644.445 & 	1128.358 & 	1866.799 \\[.15cm]

DEDICOM & ALS & 21.280 & 	70.820 & 	14.469 & 	55.783 & 	158.946   \\
DEDICOM & SGD & 1826.214 & 	1751.355 & 	1758.625 & 	1775.100 & 	1145.594  \\
DEDICOM & NAG & 30.847 & 	25.820 & 	240.587 & 	43.003 & 	49.518    \\
DEDICOM & Adam & 2105.825 & 	2128.626 & 	1791.295 & 	2056.036 & 	1992.987  \\
DEDICOM & RMSProp & 1233.237 & 	1129.172 & 	993.429 & 	1140.844 & 	1027.007  \\
DEDICOM & SAGA & 27.859 & 	30.970 & 	64.440 & 	28.319 & 	32.154    \\
DEDICOM & AdaGrad & 196.208 & 	266.057 & 	1856.267 & 	2020.417 & 	2027.370  \\
DEDICOM & NCG & 2524.762 & 	644.067 & 	236.868 & 	1665.704 & 	4219.446  \\
DEDICOM & L-BFGS & 1568.677 & 	1519.808 & 	1209.971 & 	1857.267 & 	1364.027  \\
DEDICOM & VecHGrad & 592.688 & 	918.439 & 	412.623 & 	607.254 & 	854.839   \\[.15cm]
  
PARATUCK2 & ALS & 225.952 & 	209.978 & 	230.392 & 	589.437 & 	625.668   \\
PARATUCK2 & SGD & 1953.609 & 	2625.722 & 	2067.727 & 	3002.172 & 	2745.380  \\ 
PARATUCK2 & NAG & 48.468 & 	48.724 & 	285.679 & 	76.811 & 	72.068    \\ 
PARATUCK2 & Adam & 2628.211 & 	2657.387 & 	2081.996 & 	2719.519 & 	2709.638  \\ 
PARATUCK2 & RMSProp & 1407.752 & 	1156.370 & 	1092.156 & 	1352.057 & 	1042.899  \\ 
PARATUCK2 & SAGA & 74.248 & 	70.952 & 	120.861 & 	71.398 & 	86.682    \\ 
PARATUCK2 & AdaGrad & 2595.478 & 	2626.939 & 	2073.777 & 	292.564 & 	2795.260  \\ 
PARATUCK2 & NCG & 150.196 & 	1390.013 & 	928.071 & 	1586.523 & 	82.701    \\ 
PARATUCK2 & L-BFGS & 2780.658 & 	2656.062 & 	2188.253 & 	3522.249 & 	2822.661  \\
PARATUCK2 & VecHGrad & 885.246 & 	1149.594 & 	1241.425 & 	1075.570 & 	1222.827  \\ 

\bottomrule
\end{tabular}
}
\end{table}

\subsection{Predicting Sparse Clients' Actions in the Banking Environment}
New regulations have opened the competition in retail banking, especially in Europe. Retail banks lost the exclusive privilege of proposing financial solutions such as, for instance, transaction payments. They are consequently looking to propose a higher quality of financial services using recommender engines. The recommendations are moving from a client-request approach to a bank-proposing approach where the banks dynamically offers new financial opportunities to their clients. We address this problem in our experiments with our approach. It allows the banks to predict the financial transactions of the clients in a context-aware environment and, therefore, to offer the most appropriate financial solutions to their clients for their future needs.  \\

\textbf{Clients' Transactions and Data Availability}
In 2016, the Santander bank released an anonymized public data set containing the financial activities of its clients\footnote{The data set is available at https://www.kaggle.com/c/santander-product-recommendation}. The file contains activities of 2.5 millions of clients classified in 22 transactions labels for a 16 months period. The first reported transaction is on 28 January 2015 and the last one on 28 April 2016. The total number of transactions is slightly more than 17 millions. The three most common transactions are the main account transfers, direct debits and online transactions. Since the transactions are classified per month, we track monthly activities to predict the activities of the following months.  \\

\textbf{Experimental Setup and Code Availability}
In our simulation, we choose the 200 clients having the most frequent financial activities during the 16 months since regular activities are more interesting for the predictions modeling. The transactions are categorized into 22 different labels such as, for instance, credit card activity, payroll activity, savings or interest payments. Consequently, all the information is gathered in the tensor $\mathcal{X}_{true}$, with a size equal to 200$\times$22$\times$16. We define the tensor rank equal to 25, considering the sparsity of the information and the number of transaction labels. For the CP decomposition with the VecHGrad resolution, we define the stopping condition as the relative change in the objective function $W_c$. 
The CP decomposition stops when $|W_{c_{n}}-W_{c_{n-1}}|(W_{c_{n-1}})^{-1}\leq 10^{-6}$ where $W_{c_{n}}$ and $W_{c_{n-1}}$ are the values of $W_c$ at the current and the previous iterations, respectively. We rely on the keras library for the neural network implementation. We use the Adam solver with the default parameters $\beta_1 = 0.5, \beta_2 =0.999$ for the neural network training. The experiments were performed on a computer with 16GB of RAM, Intel i7 CPU and a Tesla K80 GPU accelerator. 
\\

\textbf{Results and Discussions}
We first highlight the superior accuracy of the VecHGrad resolution algorithm on our financial data set. We then perform the predictions using three different type of neural networks: Multi-Layer Perceptron (MLP), Convolutional Neural Network (CNN) and Long-Short Term Memory (LSTM) network. We additionally cross-validate the performance of the neural network with a Decision Tree (DT). The performance of each neural network and the DT is based on four metrics: the Mean Absolute Error (MAE), the Jaccard distance, the cosine similarity and the Root Mean Square Error (RMSE).  \\

Based on the findings of Subsection \ref{subsection::vechgrad}, we compare the performance between VecHGrad and other popular machine learning and deep learning optimizers on our clients' transactions data set. As it can be noticed in Table \ref{tab::cpoptnetals}, the numerical errors of the objective function $W_c$ of the CP tensor decomposition are the lowest at convergence for the VecHGrad resolution algorithm, thanks to its adaptive line search and the Hessian update. The results furthermore are very similar to the results of the subsection \ref{subsection::vechgrad}. The numerical optimizers NCG, BFGS, ALS and RMSProp lead to lowest numerical errors after VecHGrad while the updates SGD, NAG and SAGA achieve poor performance. It is worth mentioning, additionally, that the NCG optimizer allows to reach the second lowest numerical error, explaining its success within the scientific community when using the CP tensor decomposition. Our findings legitimate the use of the VecHGrad algorithm for the task of predicting the next sparse clients' actions since, in our configuration, it outperforms the popular numerical optimizers.  \\

\begin{figure}[!p]
  \centering
  \includegraphics[scale=0.5]{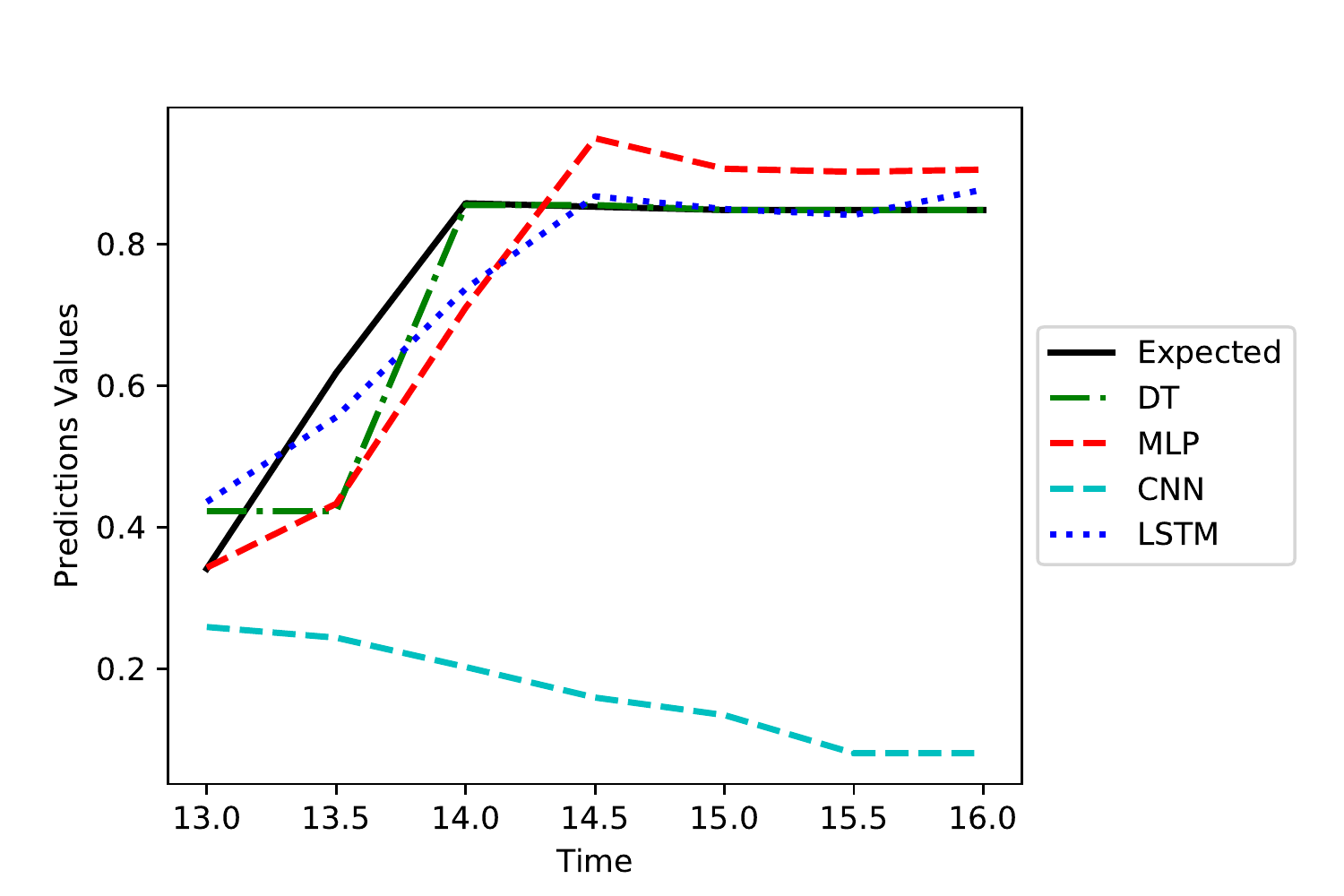}
  \caption[Personal savings predictions]{Three months predictions of the evolution of the personal savings of one latent group of clients. We can observe the prediction differences between the neural networks.}
  \label{fig::exp_plotpred}
\end{figure}

\begin{figure}[!p]
  \centering
  \includegraphics[scale=0.5]{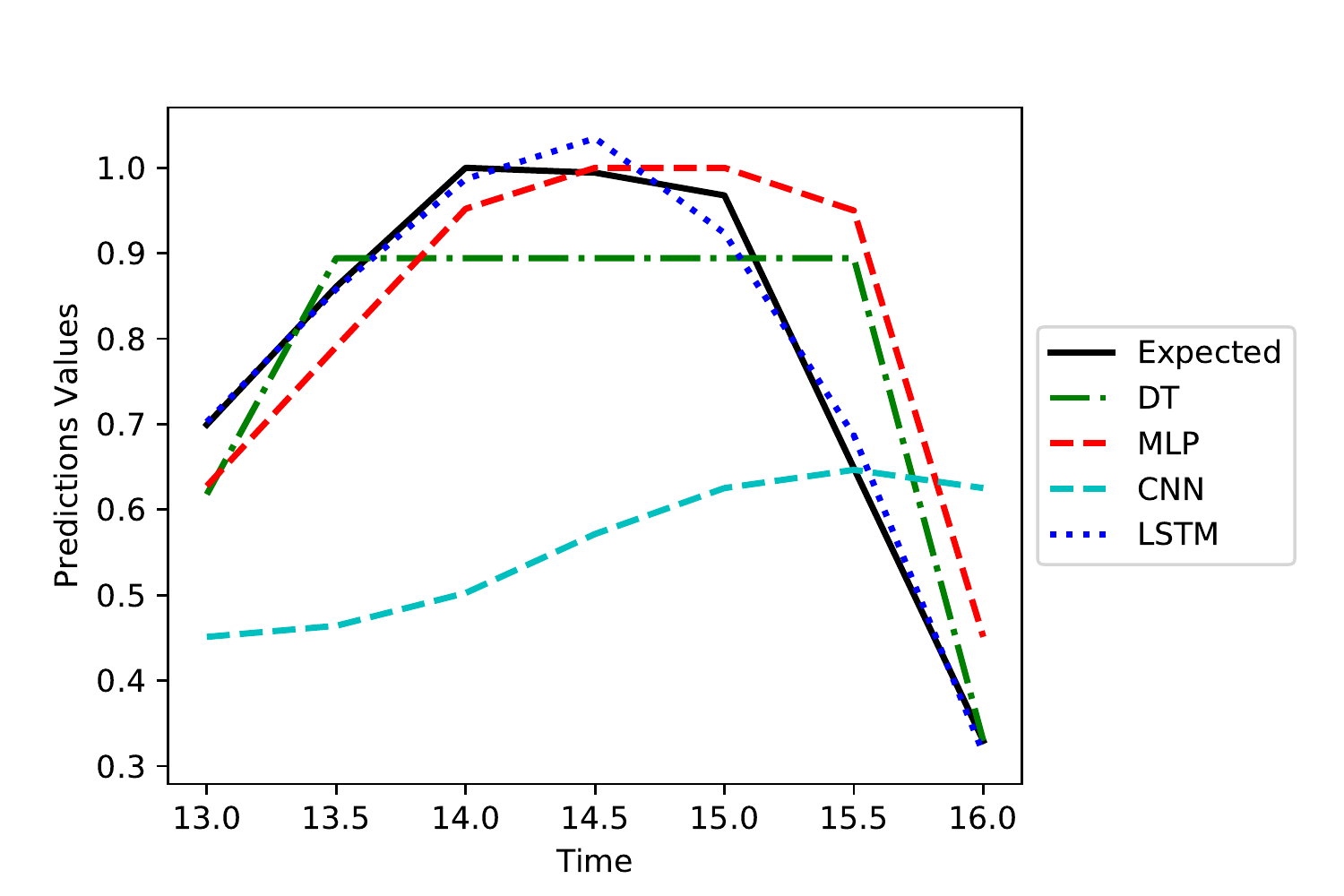}
  \caption[Credit cards spendings prediction]{Three months predictions of the evolution of the credit card spendings of one latent group of clients. The prediction differences between the methods are highlighted.}
  \label{fig::exp_plotpred2}
\end{figure}

\begin{figure}[!p]
 \centering
 \includegraphics[scale=0.5]{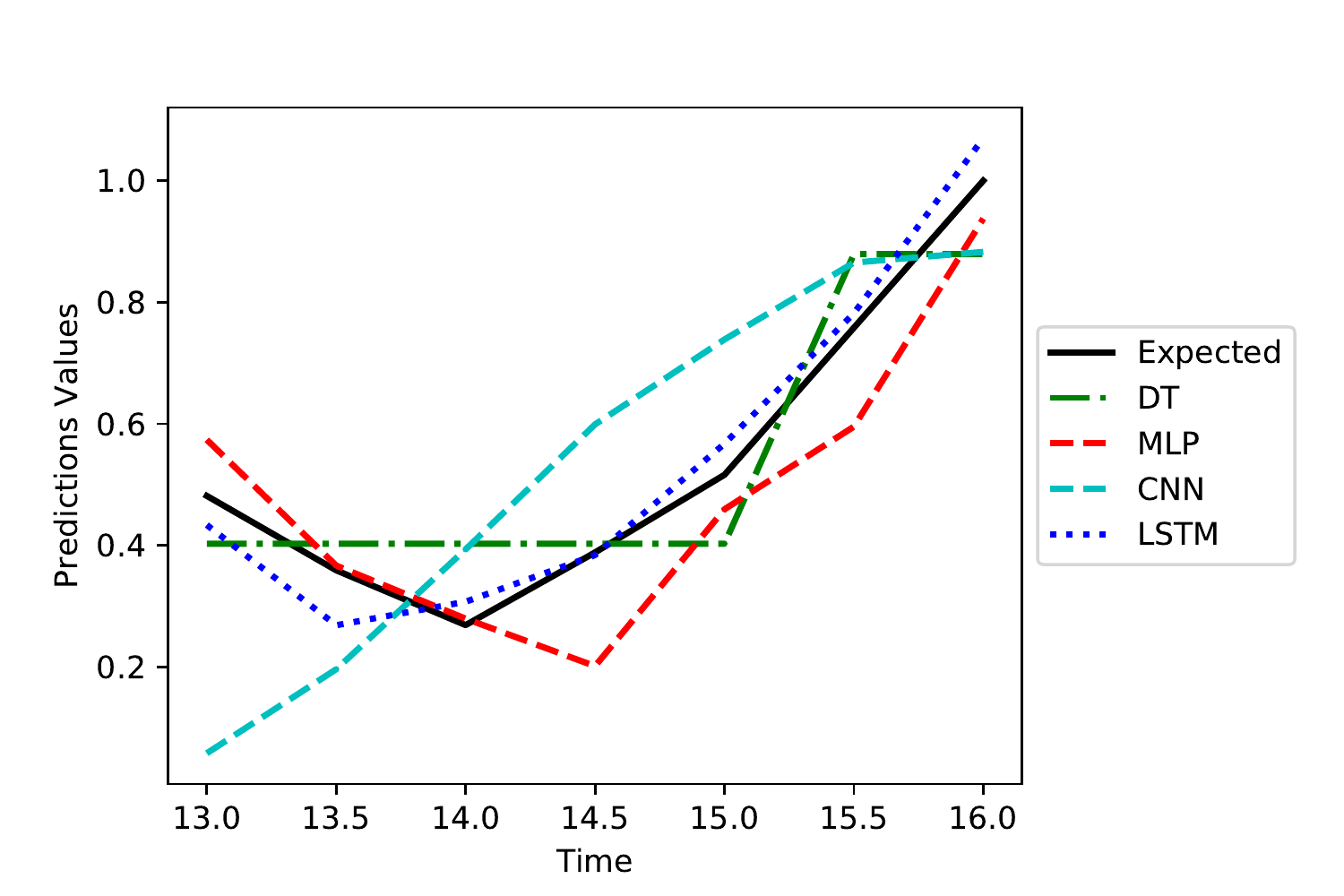}
 \caption[Debit cards spendings prediction]{Three months predictions of the evolution of the debit card spendings of one latent group of clients. The prediction differences between the methods are highlighted.}
 \label{fig::exp_plotpred3}
\end{figure}

The predictive models, MLP, CNN, LSTM and DT, have been trained on one year period from 28 January 2015 until 28 January 2016. The activities for the next three months are then predicted with a rolling time window of one month. As shown in Figure \ref{fig::exp_plotpred}, the LSTM models the most accurately the future personal savings activities followed by the MLP, the DT, and finally the CNN. The CNN fails visually to predict accurately the savings activity in comparison to the other three methods, while the LSTM seems to achieve the most accurate predictions. We highlight this preliminary conclusion in Table \ref{tab::predictionerrors1} by reporting the previously described four metrics for Figure \ref{fig::exp_plotpred}. In Figures \ref{fig::exp_plotpred2} and \ref{fig::exp_plotpred3}, we present the predictions of the credit card and the debit card spendings. The best results are similarly obtained with the LSTM, followed by the MLP, the DT and the CNN. In Table \ref{tab::predictionerrors2}, we show the aggregated metrics among all transaction predictions. In all the experiments, the LSTM network predicts the activities the most accurately having the lowest errors, followed by the MLP, the DT and the CNN. \\

\begin{table}[t]
 \caption[Residual errors of the loss function between the optimizers for the clients' actions]{Residual errors of the objective function $W_c$ between the different optimizers for the prediction of the clients' actions at convergence (the smaller, the better). All methods have similar computation time.}
 \centering
 \label{tab::cpoptnetals}
 \scalebox{1.0}{
 \begin{tabular}{cc}
 \toprule
 Optimizer & $W_c$ Error  \\
 \midrule
 CP-VecHGrad & $\leq$ \textbf{0.100}  \\
 CP-ALS & 222.280  \\
 CP-SGD & 8585.511  \\
 CP-NAG & 9930.186  \\
 CP-Adam & 5941.293  \\
 CP-RMSProp & 475.649  \\
 CP-SAGA & 10646.734  \\
 CP-Adagrad & 8055.968  \\
 CP-NCG & 16.792  \\
 CP-BFGS & 101.833  \\
 \bottomrule
 \end{tabular}
 }
\end{table}

To resume, we obtain the best results when using LSTM for the predictions. Considering that most of the clients' financial actions are cyclic and occur on a monthly basis, we find legitimate that the LSTM gives the best predictions. By using our approach, the banks therefore are able to remove the sparsity of the financial transactions of the clients wile being able to predict the actions for the future weeks, a key aspect for marketing personalized financial products.

\begin{table}[t!]
 \caption[Latent predictions errors on personal savings]{Latent predictions errors on personal savings. LSTM achieves superior performance.}
 \centering
 \label{tab::predictionerrors1}
 \begin{tabular}{ccccc}
 \toprule
 Error Measure & 	DT & 		MLP & 	 	CNN & 		LSTM  \\
 \midrule
 MAE &				\textbf{0.040} & 	0.083 & 	0.579 & 	0.047  \\
 Jaccard dist. &	0.053 & 	0.109 & 	0.777 & 	\textbf{0.061}  \\
 cosine sim. &		0.946 & 	0.928 & 	0.832 & 	\textbf{0.966}  \\
 RMSE &				0.080 & 	0.121 & 	0.626 & 	\textbf{0.063}  \\
 \bottomrule
 \end{tabular}
\end{table}

\begin{table}[t!]
 \caption[Aggregated prediction errors on all transactions]{Aggregated predictions errors on all transactions. LSTM achieves superior performance.}
 \centering
 \label{tab::predictionerrors2}
 \begin{tabular}{ccccc}
 \toprule
 Error Measure &    DT &        MLP &       CNN &       LSTM \\
 \midrule
 MAE &              0.024 & 	0.021 & 	0.270 & 	\textbf{0.010}  \\
 Jaccard dist. &    0.034 & 	0.025 & 	0.288 & 	\textbf{0.017}  \\
 cosine sim. &      0.831 & 	0.908 & 	0.883 & 	\textbf{0.953}  \\
 RMSE &             0.026 & 	0.021 & 	0.298 & 	\textbf{0.014}  \\
 \bottomrule
 \end{tabular}
\end{table}

\subsection{User-Device Authentication in Mobile Banking}
Because of the increasing competition in the banking sector brought by the PSD2 directive, the banking actors are now looking to use all the digital information available from their clients. More especially, the banks can split their clients into different groups and build different financial awareness profile to target product recommendations by monitoring and predicting the user-device authentication of their mobile banking application. We consequently highlight how we can analyze the trace of the authentication through tensor decomposition. We discuss how we take into account the imbalance between the number of users and devices. We then highlight the accurate removal of sparse information using our VecHGrad resolution algorithm followed by the results of the predictions of the authentication with our methodology.
\\

\textbf{User-Computer Authentication and Data Availability}
For the sake of the reproducibility of the experiments, we present the approach with a public data set. In 2014, the Los Alamos National Laboratory enterprise network published the anonymized user-computer authentication logs of their laboratory \cite{hagberg-2014-credential}, and available at {\color{blue} \underline{\url{https://csr.lanl.gov/data/auth/}}}. Each authentication event is composed of the authentication time (in Unix time), the computer label and the user label such as, for instance, "1,U1,C1". In total, more than 11,000 users and 22,000 computers are listed representing 13 GB of data. \\

\textbf{Construction of the user-computer authentication tensor}
We randomly select 150 users and 300 computers within the dataset representing more than 60 millions lines. The first two months of authentication events have been compressed into 50 time intervals, corresponding to 25 working days per month. A tensor $\mathscr{X}\in \mathbb{R}^{I\times J\times K}$ of size of 150$\times$300$\times$50 is built. The first dimension, denoted by $I$, represents the users, the second dimension, denoted by $J$, the computers and the last dimension, $K$, stands for the time intervals. \\

\textbf{Limitations of the CP tensor decomposition}
The CP tensor decomposition expresses the original tensor into a sum of rank one tensors. The user-computer authentication tensor is therefore decomposed as a sum of user-computer-time rank-one tensors. However, in the case of strong imbalance, CP leads to underfitting or overfitting one of the dimension \cite{acar2011scalable}. Within the dataset, we can find 2 users that connect to at least 20 different computers. Therefore, a rank equal to 2, one per user, consequently underfits the computer connections. A rank equal to 20, one per machine, overfits the number of users. In Table \ref{tab::overfitcp}, the underfitting is underlined by significant residual errors at convergence. The overfitting is detected by a good understanding of the data since the residual errors tend to be small. Hence, the PARATUCK2 decomposition is chosen to model properly each dimension of the original tensor. \\

\begin{table}[b]
  \caption[Underfitting and overfitting errors of the CP decomposition]{In CP, for imbalanced dataset, underfitting one dimension is highlighted by significant residual errors. Overfitting is difficult to measure because of the low residual errors. A good understanding of the data is required to estimate it.}
  \label{tab::overfitcp}
  \centering
    \begin{tabular}{cccc}
	\toprule
    Tensor Size & Rank & Residual Errors & $\frac{|f(x_n) - f(x_{n-1})|}{|f(x_n)|} $\\
    \midrule
    2$\times$20$\times$30 & 2 & \textbf{50.275} & $<10^{-6}$ \\
    2$\times$20$\times$30 & 20 & \textbf{1.147} & $<10^{-6}$ \\
    \bottomrule
  \end{tabular}
\end{table}

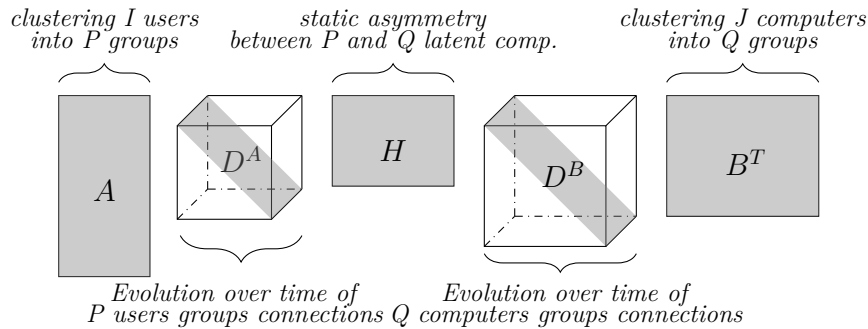
\begin{figure}[t!]
\begin{center}
\begin{tikzpicture}[scale=0.8, transform shape]
\draw  (-4.5,2) rectangle (-3,-1);
\fill [gray, opacity=.4]  (-4.5,2) rectangle (-3,-1);
\draw (-3.75,0.75) node[below]{\large{$A$}};

\draw (2.5,-0.5) rectangle (4.5,1.5) node (v1) {};
\draw (4.5,1.5) -- (5,2);
\draw (2.5,1.5) -- (3,2);
\draw (4.5,-0.5) -- (5,0);
\draw (3,2) -- (5,2);
\draw (5,0) -- (5,2);
\draw [dashdotted] (3,0) -- (5,0);
\draw [dashdotted] (3,0) -- (3,2);
\draw [dashdotted] (2.5,-0.5) -- (3,0);
\fill [gray, opacity=.4] (4.5,-0.5) -- (5,0) -- (3,2) -- (2.5,1.5)  -- cycle;
\draw (-1.5,1.3) node[below]{\large{$D^A$}};

\draw (0,2) rectangle (2,0.5);
\fill [gray, opacity=.4] (0,2) rectangle (2,0.5);
\draw (1,1.4) node[below]{\large{$H$}};

\draw (-2.55,-0.05) rectangle (-1,1.5) node (v1) {};
\draw (-1,1.5) -- (-0.5,2);
\draw (-2.55,1.5) -- (-2.05,2);
\draw (-1,-0.05) -- (-0.5,0.45);
\draw (-2.05,2) -- (-0.5,2);
\draw (-0.5,0.45) -- (-0.5,2);
\draw [dashdotted] (-2.05,0.45) -- (-0.5,0.45);
\draw [dashdotted] (-2.05,0.45) -- (-2.05,2);
\draw [dashdotted] (-2.55,-0.05) -- (-2.05,0.45);
\fill [gray, opacity=.4] (-1,-0.05) -- (-0.5,0.45) -- (-2.05,2.0) -- (-2.55,1.5)  -- cycle;
\draw (3.8,1.05) node[below]{\large{$D^B$}};

\draw  (5.5,2) rectangle (8,0);
\fill [gray, opacity=.4] (5.5,2) rectangle (8,0);
\draw (6.8,1.25) node[below]{\large{$B^T$}};

\draw[decoration={brace,raise=5pt, amplitude=7pt},decorate] (-4.5,2.0) -- node[below=-35pt] {\textit{into P groups}} (-3,2.0);
\node at (-3.75,3.25) {\textit{clustering I users}};

\draw[decoration={brace,mirror,raise=5pt, amplitude=10pt},decorate] (-2.5,-0.10) -- (-0.5,-0.10);
\node at (-1.6,-1.3) {\textit{Evolution over time of}};
\node at (-1.6,-1.65) {\textit{P users groups connections}};

\draw[decoration={brace,raise=5pt, amplitude=7pt},decorate] (0,2.0) -- node[below=-35pt] {\textit{between P and Q latent comp.}} (2,2.0);
\node at (1,3.25) {\textit{static asymmetry}};

\draw[decoration={brace,mirror,raise=5pt, amplitude=7pt},decorate] (2.5,-0.40) -- (5,-0.40);
\node at (3.85,-1.3) {\textit{Evolution over time of }};
\node at (3.85,-1.65) {\textit{Q computers groups connections}};

\draw[decoration={brace,raise=5pt, amplitude=7pt},decorate] (5.5,2.0) -- node[below=-35pt] {\textit{into Q groups}} (8,2.0);
\node at (6.75,3.25) {\textit{clustering J computers}};

\end{tikzpicture}
\caption[PARATUCK2 decomposition applied to user-computer authentication]{PARATUCK2 decomposition applied to user-computer authentication. The neural network predictions are performed on the tensor $\mathscr{D}^A$.}
\label{fig::PARATUCK2_exp}
\end{center}
\end{figure}

\textbf{PARATUCK2 Tensor Completion and Resolution}
PARATUCK2 decomposes the main tensor $\mathscr{X}\in\mathbb{R}^{I\times J \times K}$ into a product of matrices and diagonal tensors as shown in Figure \ref{fig::PARATUCK2_exp}. The matrix \textbf{A} factorizes the users into $P$ groups. We observe 15 different groups of users, and therefore, $P$ equals to 15. The diagonal tensor $\mathscr{D}^A$ reflects the temporal evolution of the connections of the $P$ users groups. The matrix $\textbf{H}$ represents the asymmetry between the $P$ users groups and the $Q$ computers groups. We notice 25 different groups of machines related to different authentication profiles, and consequently, $Q$ equals to 25. The diagonal tensor $\mathscr{D}^B$ illustrates the temporal evolution of the connections of the $Q$ computers groups. Finally, the matrix $\textbf{B}$ factorizes the computers into $Q$ latent groups of computers. \\

\textbf{Predictions for Financial Recommendation}
To achieve higher subscription rates during the advertising campaign of financial products, we explore the latent predictions for targeted recommendation based on the future user-computer authentication. The results of PARATUCK2 contain the users' temporal information and the computers' temporal information in the diagonal tensors $\mathscr{D}^A$ and $\mathscr{D}^B$, respectively. Predicting the users' authentication allows the banks to build a more complete financial awareness profile of their clients for optimized advertisement.
\\

The first step of our approach is to remove the sparsity of the information and to ensure the accurate resolution of the PARATUCK2 tensor decomposition on our user-device authentication data set. We hereinafter compare the error at convergence between VecHGrad and other machine learning and deep learning optimizers. The results are presented in Table \ref{tab::par_acc}. We recall that the PARATUCK2 tensor decomposition is one of the most complex tensor decomposition to solve, and therefore, it can lead to numerical instabilities during the resolution optimization process. Similarly to the previous experiments, the VecHGrad algorithm achieves the lowest numerical errors at convergence followed by the SGD, BFGS, ALS, and RMSProp algorithms. The numerical instability of the NCG algorithm is highlighted here. Effectively, the NCG algorithm diverges when applied on the complex PARATUCK2 tensor decomposition with our user-device authentication data set. The optimizers NAG and SAGA additionally lead to the biggest numerical errors. These findings confirm the results of the previous experiments. We consequently used the VecHGrad resolution algorithm to solve the objective minimization function $W_c$, as illustrated in Figure \ref{fig::par_authen}, since it is the algorithm delivering the lowest numerical errors at convergence.  \\

\begin{table}[t]
 \caption[Residual errors of the loss function between the optimizers for user-device authentication]{Residual errors of the objective function $W_c$ between the different optimizers for the monitoring of the user-device authentication at convergence (the smaller, the better). All the methods have similar computation time.}
 \centering
 \label{tab::par_acc}
 \scalebox{1.0}{
\setlength{\tabcolsep}{20pt}
 \begin{tabular}{cc}
 \toprule
 Optimizer & $W_c$ Error  \\
 \midrule
 PARATUCK2 - VecHGrad & $\leq$ \textbf{0.100}  \\
 PARATUCK2 - ALS & 99.834  \\
 PARATUCK2 - SGD & 8.651  \\
 PARATUCK2 - NAG & 513.860  \\
 PARATUCK2 - Adam & 419.812  \\
 PARATUCK2 - RMSProp & 146.086  \\
 PARATUCK2 - SAGA & 513.561  \\
 PARATUCK2 - Adagrad & 461.413  \\
 PARATUCK2 - NCG & diverge  \\
 PARATUCK2 - BFGS & 67.152  \\
 \bottomrule
 \end{tabular}
 }
\vspace{1cm}
\end{table}

\begin{figure}[!p]
  \centering
 \includegraphics[scale=0.6]{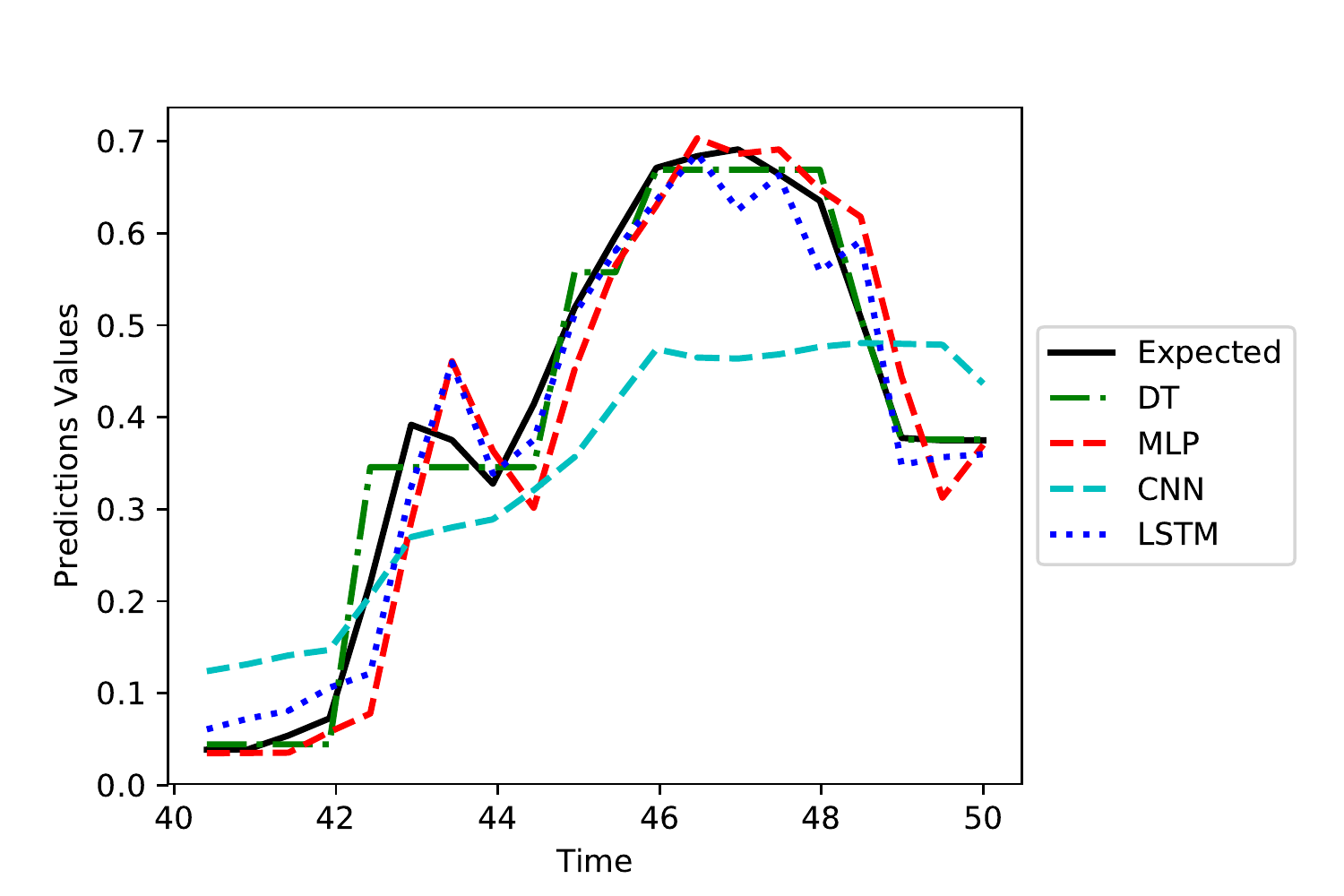}
  \caption[Prediction of the first group of latent users' authentication]{Two weeks prediction of the evolution of a latent users' authentication group according to the different models used.}
  \label{fig::preds}
\end{figure}

\begin{figure}[!p]
  \centering
 \includegraphics[scale=0.6]{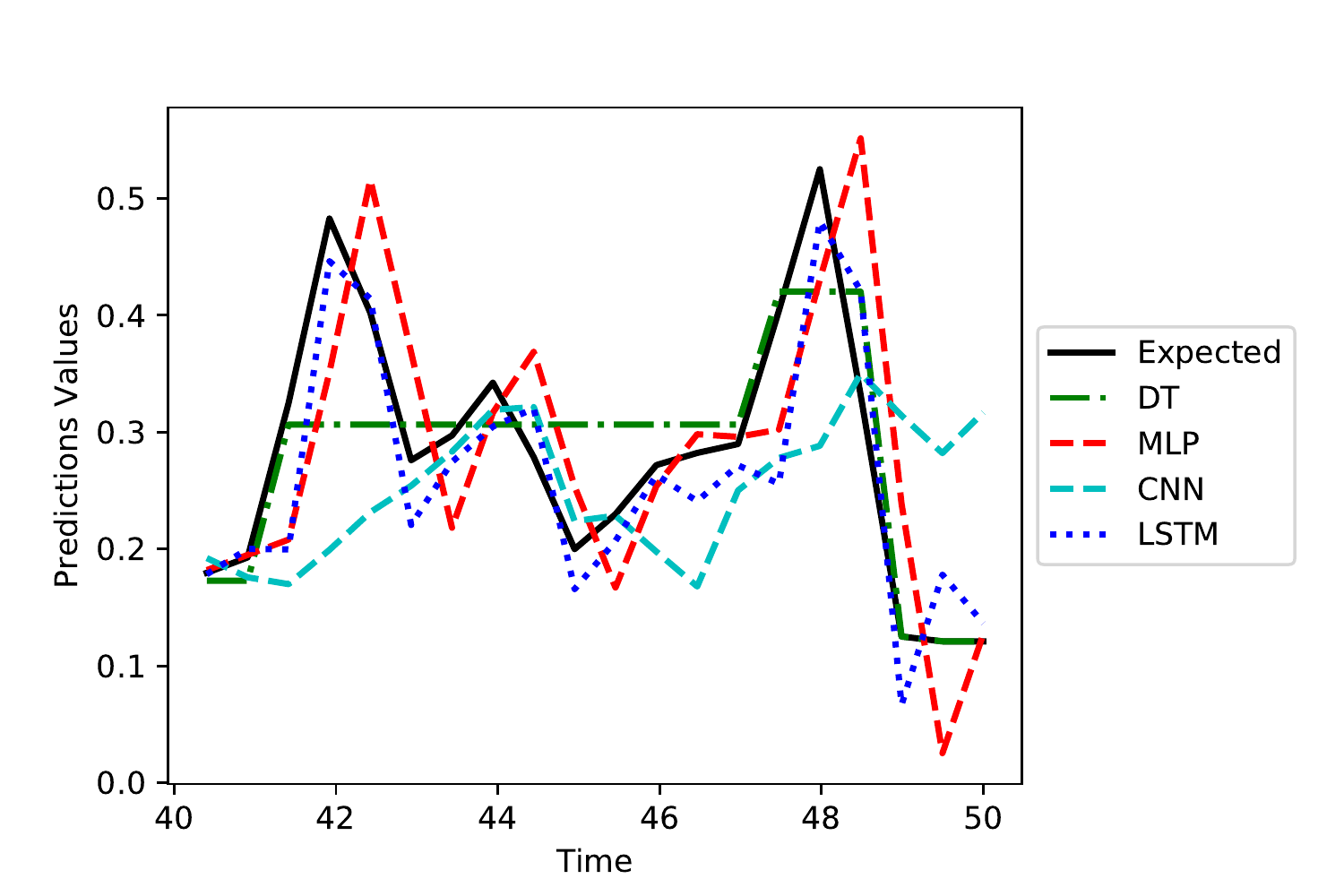}
  \caption[Prediction of the second group of latent users' authentication]{Two weeks prediction of the evolution of another latent users' authentication group according to the different models used.}
  \label{fig::preds2}
\end{figure}

In Figures \ref{fig::preds} and \ref{fig::preds2}, we highlight the results of the predictions of the users' authentication for a specific group of clients, corresponding to one specific latent factor $P$. Four different methods have been used for the predictions, DT, MLP, CNN and LSTM. All the methods have been trained on a six weeks period. The users' authentication for the next two weeks are then predicted with a rolling time window of one day. Figures \ref{fig::preds} and \ref{fig::preds2} highlight visually that the LSTM models the most accurately the future users' authentication. It is followed by the MLP, the DT, and finally the CNN. We underline this preliminary statement using six well-known error measures. The Mean Absolute Error (MAE), the Mean Directional Accuracy (MDA), the Pearson correlation, the Jaccard distance, the cosine similarity and the Root Mean Square Error (RMSE) are used to determine objectively the most accurate predictive method. 
Table \ref{tab::prederrors1} describes the aggregated error measures of the users' authentication. As previously seen, the LSTM is the closest to the true authentication since it has the lowest error values. Then, the MLP comes second, the DT third, and the CNN last.
\\

\begin{table}[!p]
\caption[Aggregated errors of the users' authentication predictions]{Aggregated latent predictions errors of the users' authentication with decision tree and neural networks}
\centering
\label{tab::prederrors1}
\begin{tabular}{ccccc}
\toprule
Error Measure & DT & MLP & CNN & LSTM \\
\midrule
MAE & 0.0965 & 0.0506 & 0.1106 & \textbf{0.0379} \\
MDA & 0.1579 & 0.7447 & 0.5263 & \textbf{0.6842} \\
Pearson corr. & 0.8537 & 0.9598 & 0.8885 & \textbf{0.9753} \\
Jaccard dist. & 0.2257 & 0.1206 & 0.2648 & \textbf{0.0911} \\
cosine sim. & 0.9587 & 0.9891 & 0.9745 & \textbf{0.9914} \\
RMSE & 0.1306 & 0.0695 & 0.3140 & \textbf{0.0477} \\
\bottomrule
\end{tabular}
\end{table}

We can conclude that LSTM combined with PARATUCK2 models the best the future users' authentication with the aim to better target the clients that might be interested by financial products during the bank's advertising campaigns. As the majority of the user's authentication are sequence-based, it is legitimate to find out LSTM gives the best results for the predictions. Effectively, each user has a recurrent pattern in the authentication process depending on its activities of the day. By using VecHGrad for PARATUCK2 and the LSTM for the predictions, the bank can therefore earn a significant competitive advantage for the personalized products recommendation, relying only on its clients' authentication on the mobile application.

\section{Conclusion} \label{sec::ccl3}
Building upon tensor decomposition, numerical optimization, preconditioning methods and neural networks, we proposed a predictive method to target personalized recommendations for the banking industry. Because of the requirement of precise and accurate resolution of tensor decomposition, we introduced VecHGrad, a Vector Hessian Gradient optimization method. VecHGrad uses partial information of the second derivative and an adaptive strong Wolfe's line search to ensure faster convergence. We conducted experiments on five real world data sets, CIFAR10, CIFAR100, MNIST, COCO and LFW, very popular in machine learning and deep learning. We highlighted that VecHGrad is capable to outperform the accuracy of the widely used gradient based resolution methods such as Adam, RMSProp or Adagrad, and the linear algebra update ALS on three different tensor decomposition, CP, DEDICOM and PARATUCK2, offering different levels of complexity. Based on the results of this experiment, we used VecHGrad resolution algorithm for the task of predicting the next financial actions of the bank’s clients in a sparse environment. We rely on a public data set proposed by the Santander bank. We proposed a predictive method in which the sparsity of the financial transactions is removed before performing the predictions on future client's transactions. The sparsity of the financial transactions is removed using the popular CP tensor decomposition, which decomposes the initial tensor containing the financial transaction into a sum of rank-one tensors. The predictions are performed using different type of neural networks including LSTM, CNN, MLP and a decision tree. Due to the recurrent activities of most of the financial transactions, we underlined the best results were found when the CP tensor decomposition was used with LSTM.  We furthermore presented our methodology for a novel application  in the context of mobile banking application. We introduced the use of the PARATUCK2 tensor decomposition and neural networks for the monitoring and the predictions of imbalanced user-device authentication. The PARATUCK2 tensor decomposition expresses a tensor as a multiplication of matrices and diagonal tensors and, therefore, it is highly suitable for imbalanced data sets. The user-device authentication on mobile banking allows to build a financial awareness profile to better target potential subscription by the clients on new products. We rely on a public data set proposed by the Los Alamos National Laboratory enterprise network for the experiments. The resolution of the PARATUCK2 tensor decomposition was performed using VecHGrad to ensure a decomposition of high accuracy. We performed users’ authentication  predictions using LSTM, CNN, MLP and a decision tree, evaluated on different distance measures. Similarly to the predictions of the next clients' transactions 
the best results were obtained with LSTM because of the recurrent and cyclic patterns of the user-device authentication.
\\

Future work will concentrate on the influence of the adaptive line search for the VecHGrad algorithm. We effectively observed that the performance of the algorithm is strongly correlated with the performance of the the adaptive line search optimization. We will simultaneously look to reduce the memory cost of the adaptive line search as it has a crucial impact for a GPU implementation as well as a limited memory resolution for a usage on very large data sets. Concerning the predictions of the financial activities for personal financial recommendation, a smaller time frame discretization, weekly or daily, will be assessed with other financial transactions. It will offer a larger choice of financial product recommendations depending on the clients' mid-term and long-term interests. Finally, the financial recommendation depending of the user-device authentication on a mobile banking application will be further extended by incorporating additional features. Noticeably, the navigation usage, the time gap between each action and the type of device used will be monitored to further improve the bank's advertising campaigns of their products to the appropriate clients. 
\chapter{MQLV: Q-learning for Optimal Policy of Money Management in Retail Banking}

Reinforcement learning \cite{sutton2018reinforcement} is one of the best approaches to train a computer game emulator capable of human level performance. In a reinforcement learning approach, an optimal value function is learned across a set of actions, or decisions, that leads to a set of states giving different rewards, with the objective to maximize the overall reward. A policy assigns to each state-action pairs an expected return. We call an optimal policy a policy for which the value function is optimal. QLBS \cite{halperin2017qlbs}, Q-Learner in the Black-Scholes(-Merton) Worlds, applies the reinforcement learning concepts, and noticeably, the popular Q-learning algorithm \cite{watkins1992q}, to the financial stochastic model of Black, Scholes and Merton \cite{black1973pricing,merton1973theory}. It is, however, specifically optimized for the geometric Brownian motion and the vanilla options. Its range of application is, therefore, limited to vanilla option pricing within the financial markets. We propose MQLV, Modified Q-Learner for the Vasicek model, a new reinforcement learning approach that determines the optimal policy of money management based on the aggregated financial transactions of the clients. It unlocks new frontiers to establish personalized credit card limits or to fulfill bank loan applications, targeting the retail banking industry. MQLV extends the simulation to mean reverting stochastic diffusion processes and it uses a digital function, a Heaviside step function expressed in its discrete form, to estimate the probability of a future event such as a payment default.  In our experiments, we first show the similarities between a set of historical financial transactions and Vasicek generated transactions and, then, we underline the potential of MQLV on generated Monte Carlo simulations. MQLV is the first Q-learning Vasicek-based methodology addressing transparent decision making processes in retail banking. 

\section{Introduction} \label{4_sec::intro}
A major goal of the reinforcement learning (RL) and Machine Learning (ML) community is to build efficient representations of the current environment to solve complex tasks. In RL, an agent relies on multiple sensory inputs and past experience to derive a set of plausible actions to solve a new situation \cite{mnih2013playing}. While the initial idea around reinforcement learning is far from new \cite{sutton1984temporal,watkins1989learning,williams1987class}, significant progress has been achieved recently by combining neural networks and Deep Learning (DL) with RL, to either increase the quality of the signals from the multiple sensory inputs or the number of linear function approximators. DL \cite{krizhevsky2012imagenet,sermanet2013pedestrian} has, for instance, allowed the development of a novel agent combining RL with a class of deep artificial neural networks \cite{mnih2013playing,mnih2015human} resulting in Deep Q-Network (DQN). The Q refers to the Q-learning algorithm introduced in \cite{watkins1992q}. It is an incremental method that successively improves its evaluations of the quality of the state-action pairs. The DQN approach achieves human level performance on Atari video games using unprocessed pixels as inputs. In \cite{van2016deep}, deep RL with double Q-Learning was proposed to challenge the DQN approach while trying to reduce the overestimation of the action values, a well-known drawback of the Q-learning and DQN methodologies. Meanwhile, the extension of the DQN approach from discrete action domain to continuous action domain, directly from the raw pixels to inputs, was successfully achieved for various simulated tasks \cite{lillicrap2015continuous}.  \\

\begin{figure}[t!]
 \centering
 \includegraphics[scale=1]{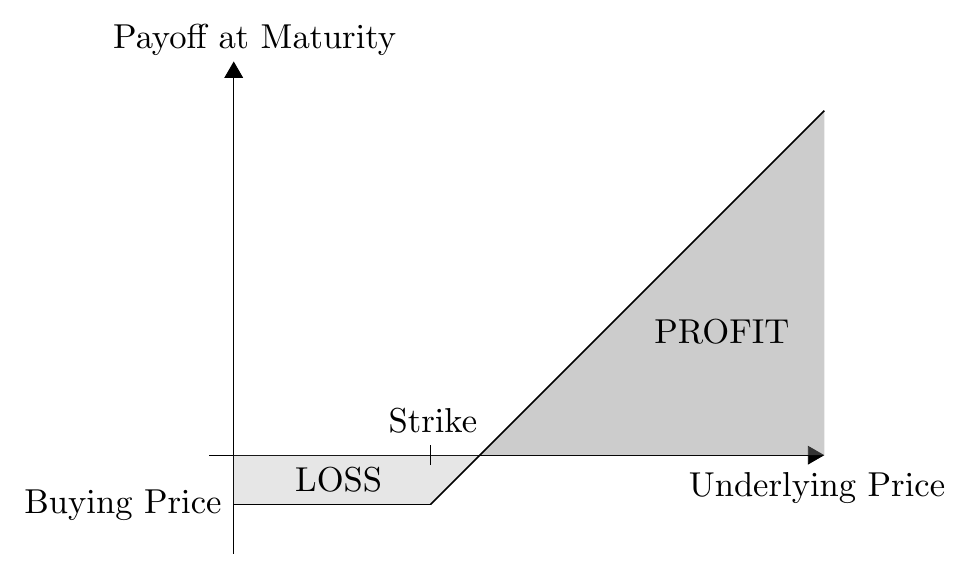}
 \caption[Payoff of a European vanilla call option at maturity]{Payoff at maturity of a vanilla call option. The option allows to buy the underlying asset, a stock, at maturity for a certain price, the strike price. A profit is generated if the underlying price is higher than the strike at maturity, otherwise the buyers makes a loss.}
 \label{fig::vanilla_payoff}
\end{figure}

Nonetheless, most of the proposed models focused on gaming theory and computer game simulation and very few to the financial world. 
In QLBS \cite{halperin2017qlbs}, a RL approach is applied to the Black, Scholes and Merton (BSM) financial framework for derivatives \cite{black1973pricing,merton1973theory}, a cornerstone of the modern quantitative finance. In the BSM model, the dynamic of a stock market is defined as following a Geometric Brownian Motion (GBM) to estimate the price of a vanilla option on a stock \cite{wilmott2013paul}. A vanilla option is an option that gives the holder the right to buy or sell the underlying asset, a stock, at maturity for a certain price, the strike price. In Figure \ref{fig::vanilla_payoff}, we describe the payoff $(S_T - K)^+$ at maturity of a vanilla call option with $S_T$ the spot price at maturity of the stock and $K$ the strike price. The holder of the vanilla option will execute his right to buy the stock at the strike price as soon as the payoff is positive and compensates the buying cost of the option. QLBS is one of the first approach to propose a complete RL framework 
for finance. As mentioned by the author, a certain number of topics are, however, not covered in the approach. For instance, it is specifically designed for vanilla options and it fails to address any other type of financial 
applications. Additionally, the initial generated paths rely on the popular GBM but there exist a significant number of other popular stochastic models depending on the market dynamics \cite{hull2003options}.  \\

In this work, we describe a RL approach tailored for personal recommendation in retail banking regarding money management to be used for loan applications or credit card limits. The method is part of a banking strategy trying to reduce the customer churn in a context of a competitive retail banking market. We rely on the Q-learning algorithm and on a mean reverting diffusion process to address this topic. It leads ultimately to a fitted Q-iteration update and a model-free and off-policy setting. The diffusion process reflects the time series observed in retail banking such as transaction payments or credit card transactions. 
We furthermore introduce a new terminal digital function, $\Pi$, defined as a Heaviside step function in its discrete form for a discrete variable $n \in \mathbb{R}$.
The digital function is at the core of our approach for retail banking since it can evaluate the future probability of an event including, for instance, the future default probability of a client based on his spendings. Our method converges to an optimal policy, and to optimal sets of actions and states, respectively the spendings and the available money. The retail banks can, consequently, determine the optimal policy of money management based on the aggregated financial transactions of the clients. The banks are able to compare the difference between the MQLV's optimal policy and the individual policy of each client. It contributes to an unbiased decision making process while offering transparency to the client. Our main contributions are summarized below:

\begin{itemize}
\item A new RL framework called MQLV, Modified Q-Learning for Vasicek, extending the initial QLBS framework \cite{halperin2017qlbs}. MQLV uses the theoretical foundation of RL learning and Q-Learning to build a financial RL framework based on a mean reverting diffusion process, the Vasicek model  \cite{vasicek1977equilibrium}, to simulate data, in order to reach ultimately a model-free and off-policy RL setting.   \\
\item The definition of a digital function to estimate the future probability of an event. 
The aim is to widen the application perspectives of MQLV by using a characteristic terminal function that is usable for a decision making process in retail banking  such as the estimation of the default probability of a client.   \\
\item The first application of Q-learning to determine the clients' optimal policy of money management in retail banking. MQLV leverages the clients aggregated financial transactions to define the optimal policy of money management, targeting the risk estimation of bank loan applications or credit cards.   \\
\item The introduction of an update function applied to MQLV to compensate the overestimation of the action values, characteristic of the Q-Learning algorithm. The objective is to minimize the overestimation of the event probabilities measured by the digital function.   \\
\end{itemize}

The chapter is structured as follows. We discuss the related work in Section \ref{sec::relatedwork}. We review QLBS and the Q-Learning formulations derived by Halperin in \cite{halperin2017qlbs} in the context of the Black, Scholes and Merton model in Section \ref{sec::background}. We describe MQLV according to the Q-Learning algorithm that leads to a model-free and off-policy setting in Section \ref{sec::algorithm}. 
We highlight experimental results in Section \ref{sec::experiments}. We conclude in Section \ref{sec::conclusion} by addressing promising directions for future work.

\section{Related Work} \label{sec::relatedwork}
The foundations of modern reinforcement learning 
\cite{sutton1984temporal,williams1987class} established the theoretical framework to learn policies for sequential decision problems by proposing a formulation of the cumulative future reward signal. The Q-learning algorithm introduced in \cite{watkins1989learning} is one of the cornerstone of reinforcement learning. 
However, the convergence of the Q-Learning algorithm was solved several years later. It was shown that the Q-Learning algorithm with non-linear function approximators \cite{tsitsiklis1997analysis} with off-policy learning \cite{baird1995residual} could provoke a divergence of the Q-network. The reinforcement learning community therefore focused on linear function approximators \cite{tsitsiklis1997analysis} to ensure convergence.   \\

The emergence of neural networks and deep learning \cite{goodfellow2016deep} contributed to address the use of reinforcement learning with neural networks. At an early stage, deep auto-encoders were used to extract feature spaces to solve reinforcement learning tasks \cite{lange2010deep}. Thanks to the release of the Atari 2600 emulator \cite{bellemare2013arcade}, a public data set was then available answering the needs of the RL community for larger simulation. The Atari emulator allowed a proper performance benchmark of the different reinforcement learning algorithms and offered the possibility to test various architectures. The Atari games were used to introduce the concept of deep reinforcement learning \cite{mnih2013playing,mnih2015human}. The authors used a convolutional neural network trained with a variant of Q-learning to successfully learn control policies directly from high dimensional sensory inputs. They reached human-level performance on many of the Atari games. Shortly after, the deep reinforcement learning was challenged by double Q-Learning within a deep reinforcement learning framework \cite{van2016deep}. The double Q-Learning algorithm was initially introduced in \cite{hasselt2010double} in a tabular setting. The double deep Q-Learning gave more accurate estimates and lead to much higher scores than the one observed in \cite{mnih2013playing,mnih2015human}. An ongoing work is consequently to further improve the results of the double deep Q-learning algorithms through different variants. The authors used a quantile regression to approximate the full quantile function for the state-action return distribution in \cite{dabney2018implicit}, leading to a large class of risk-sensitive policies. It allowed them to further improve the scores on the Atari 2600 games simulator. Similarly, a new algorithm, C51, which applies the Bellman's equation to the learning of the approximate value distribution was designed in \cite{bellemare2017distributional} and showed state-of-the-art performance.
\\

Meanwhile, a certain number of publications focused on model-free policies and actor-critic framework. Stochastic policies were trained in \cite{wawrzynski2013autonomous} with a replay buffer to avoid divergence. It was showed in \cite{silver2014deterministic} that deterministic policy gradients (DPG) exist, even in a model-free environment. The DPG approach was subsequently extended in \cite{balduzzi2015compatible} using a deviator network. A deviator network backpropagates different signals to train the network. Continuous control policies were learned using backpropagation, and therefore, introducing the Stochastic Value Gradient SVG(0) and SVG(1) in \cite{heess2015learning}. Recently, Deep Deterministic Policy Gradient (DDPG) was presented in \cite{lillicrap2015continuous} to learn competitive policies using an actor-critic model-free algorithm based on a DPG algorithm that can operate over continuous action spaces.  \\

\section{Background} \label{sec::background}
We define $A_t \in \mathcal{A}$ the action taken at time $t$ for a given state $X_t \in \mathcal{X}$ and the immediate reward by $R_{t+1}$. To avoid any confusion between the stochastic diffusion process and the different states of the environment, the ongoing state is denoted by $X_t \in \mathcal{X}$ and the stochastic diffusion process by $S_t \in \mathcal{S}$ at time $t$. The discount factor that trades off the importance of immediate and later rewards is expressed by $\gamma \in [0;1]$.  \\

We recall a policy is a mapping from states to probabilities of selecting each possible action \cite{sutton2018reinforcement}. By following the notations of \cite{halperin2017qlbs}, the policy $\pi$ such that 

\begin{equation} \label{eq::policy}
\pi : \left\lbrace 0, \ldots, T-1 \right\rbrace \times \mathcal{X}\rightarrow \mathcal{A}
\end{equation}

maps at time $t$ the current state $X_t=x_t$ into the action $a_t \in \mathcal{A}$

\begin{equation} \label{eq::policymap}
a_t = \pi(t,x_t) \quad .
\end{equation}

The value of a state $x$ under a policy $\pi$, denoted by $v_\pi(x)$ when starting in $x$ and following $\pi$ thereafter, is called the state-value function for policy $\pi$ and it is defined as 

\begin{equation} \label{eq::statevalue}
v_\pi = \mathbb{E}_\pi \left[ \sum_{k=0}^{\infty} \gamma^k R_{t+k+1} | X_t = x\right] \quad .
\end{equation}

The action-value function, $q_\pi (x,a)$ for policy $\pi$ defines the value of taking action $a$ in state $x$ under a policy $\pi$	as the expected return starting from $x$, taking the action $a$, and thereafter following policy $\pi$ such that

\begin{equation} \label{eq::actionvalue}
q_\pi(x,a) = \mathbb{E}_\pi \left[ \sum_{k=0}^{\infty} \gamma^k R_{t+k+1} | X_t = x, A_t = a \right] \quad .
\end{equation} 

The optimal policy, $\pi_t^*$, is the policy that maximizes the state-value function

\begin{equation} \label{eq::optimalpolicy}
\pi_t^* (X_t) = \arg \max_\pi V_t^\pi (X_t) \quad .
\end{equation}

The optimal state-value function, $V_t^*$, satisfies the Bellman optimality equation such that

\begin{equation} \label{eq::optimalvalue}
V_t^* (X_t) = \mathbb{E}_t^{\pi^*} \left[ R_t(X_t, u_t=\pi_t^*(X_t), X_{t+1})+\gamma V_{t+1}^*(X_{t+1}) \right] \quad .
\end{equation}

The Bellman equation for the action-value function, the Q-function, is defined as 

\begin{equation} \label{eq::bellmanQfunction}
Q_t^\pi (x,a) = \mathbb{E}_t \left[ R_t(X_t, a_t, X_{t+1}) | X_t = x, a_t = a\right] + \gamma \mathbb{E}_t^\pi \left[ V_{t+1}^\pi (X_{t+1}) | X_t = x \right] \quad .
\end{equation}

The optimal action-value function, $Q_t^*$, is obtained for the optimal policy with 

\begin{equation} \label{eq::optimalQfunction}
\pi_t^* = \arg \max_\pi Q_t^\pi (x,a) .
\end{equation}

The optimal state-value and action-value functions are connected by the following system of equations

\begin{equation} \label{eq::actstate}
\begin{cases}
 V_t^* =  \max_a Q^*(x,a) &  \\ 
 Q_t^* = \mathbb{E}_t \left[ R_t(X_t,a,X_{t+1}) \right] + \gamma \mathbb{E}_t \left[ V_{t+1}^* (X_{t+1} | X_t = x) \right] &  \\ 
\end{cases} \quad .
\end{equation} 

Therefore, we can obtain the Bellman optimality equation
\begin{equation} \label{eq::bellmanoptimal}
Q_t^*(x,a) = \mathbb{E}_t \left[ R_t(X_t,a_t,X_{t+1}) + \gamma \max_{a_{t+1} \in \mathcal{A}} Q_{t+1}^* (X_{t+1}, a_{t+1}) | X_t=x, a_t=a \right] \quad .
\end{equation} 

Using the Robbins-Monro update \cite{robbins1985stochastic}, the update rule for the optimal Q-function with on-line Q-learning on the data point $(X_t^{(n)}, a_t^{(n)}, R_t^{(n)}, X_{t+1}^{(n)})$ is expressed by the following equation with $\alpha$ a constant step-size parameter

\begin{equation} \label{eq::qlearning_update}
\begin{split}
Q_t^{*,k+1} (X_t, a_t) = & (1-\alpha^k) Q_t^{*,k}(X_t,a_t) + \\ 
& \: \alpha^k \left[ R_t(X_t, a_t, X_{t+1}) + \gamma \max_{a_{t+1} \in \mathcal{A}} Q_{t+1}^{*,k} (X_{t+1}, a_{t+1}) \right]
\end{split} \quad .
\end{equation}

\section{Algorithm} \label{sec::algorithm}
\begin{figure}[b!]
 \centering
 \includegraphics[scale=0.55]{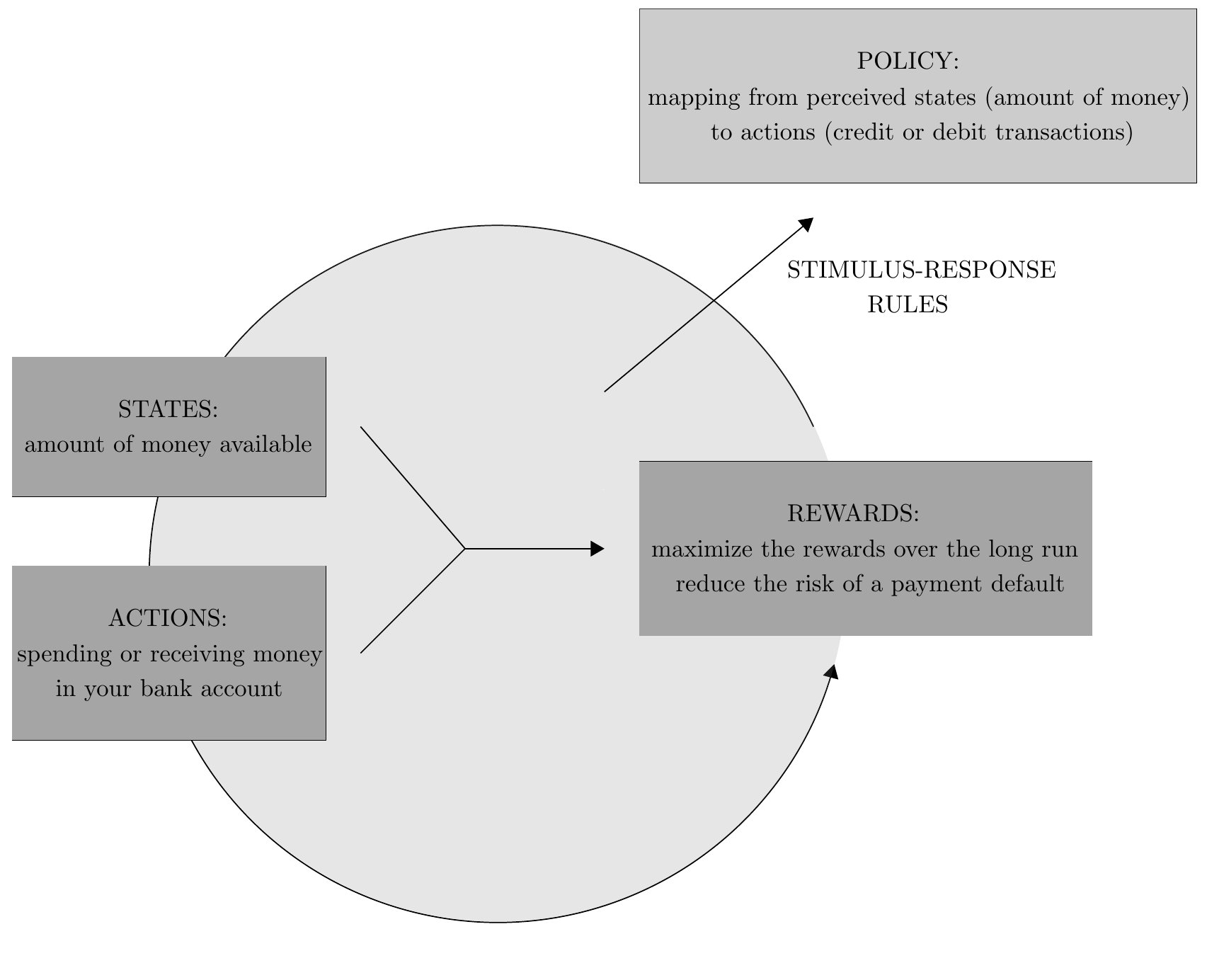}
 \caption[MQLV and reinforcement learning concept definitions]{In MQLV, reinforcement learning is applied to the retail banking. The optimal policy of money management is learned by maximizing the rewards, defined as avoiding a payment default, based on state-action sequences, respectively spending or receiving money in your bank account.}
 \label{fig::mqlv_methodo}
\end{figure}

We describe, in this section, how to derive a general recursive formulation for the optimal action. It is equivalent to an optimal hedge under a financial framework such as, for instance, portfolio or personal finance optimization. We additionally present the formulation of the action-value function, the Q-function. Both the optimal hedge and the Q-function follow the assumption of a continuous space scenario generated by the Vasicek model with Monte Carlo simulation. For the ease of understanding, we explain the meaning of the RL concepts into the retail banking world. The states are defined as the amount of money available, the actions as spending or receiving money in your bank account, the reward as the probability of avoiding a payment default, and the policy as the mapping from the perceived states (the amount of money) to actions (credit or debit transactions). Figure \ref{fig::mqlv_methodo} gathers these RL concepts. \\

By relying on the financial framework established in \cite{halperin2017qlbs}, we consider a mean reverting diffusion process, also known as the Vasicek model \cite{vasicek1977equilibrium},

\begin{equation} \label{eq::meanreverting}
dS_t = \kappa(b-S_t)dt + \sigma dB_t \quad .
\end{equation}

The term $\kappa$ is the speed reversion, $b$ the long term mean level, $\sigma$ the volatility and $B_t$ the Brownian motion. The solution of the stochastic equation is equal to

\begin{equation} \label{eq::meanrevertingsol}
S_t = S_0 e^{-\kappa t} + b(1-e^{-\kappa t}) + \sigma e^{-\kappa t} \int_0^t e^{\kappa s}dB_s \quad .
\end{equation}

Therefore, we define a new time-uniform state variable, i.e. without a drift, as 

\begin{equation} \label{eq::statevariable}
\begin{cases}
S_t = X_t + S_0e^{-\kappa t} + b(1-e^{-\kappa t}) & \\ 
\text{with } X_t = \sigma e^{-\kappa t} \int_0^t e^{\kappa s} dB_s - \left[ S_0e^{-\kappa t} + b(1-e^{-\kappa t}) \right] &
\end{cases} \quad .
\end{equation}

\begin{figure}[p]
 \centering
 \includegraphics[scale=0.525]{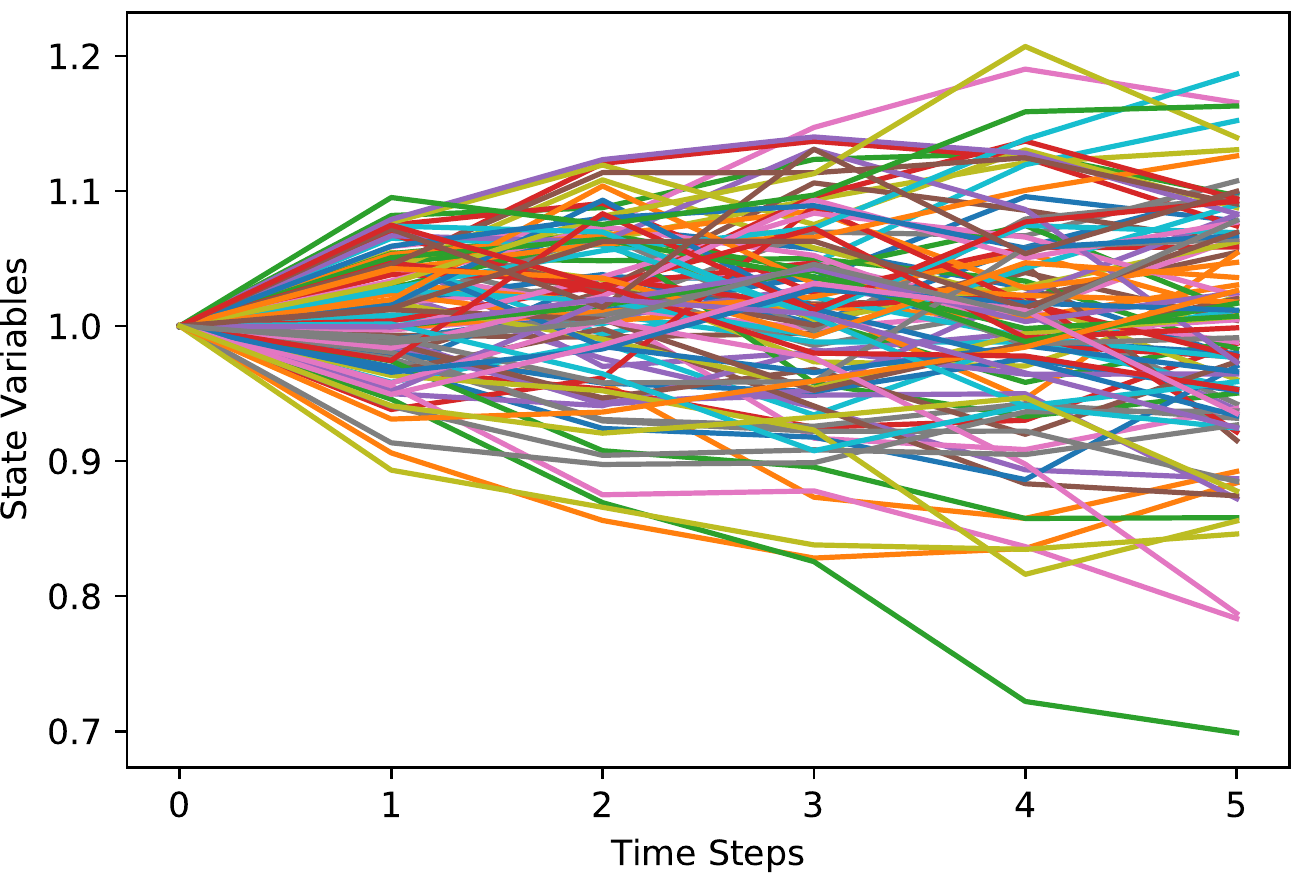}
 \hspace{0.5cm}
 \includegraphics[scale=0.525]{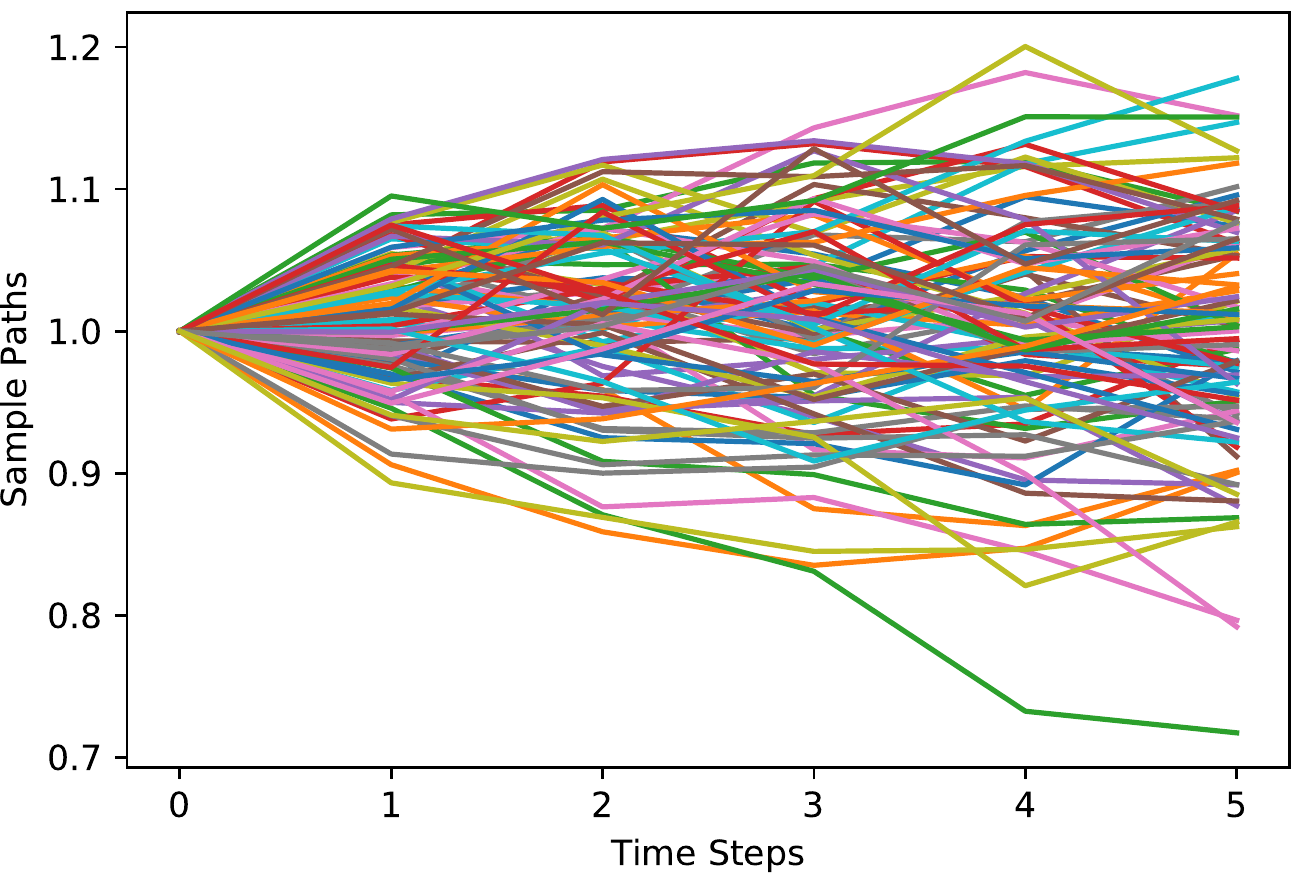}
 \caption[Time uniform state variables in the context of a null inflation]{Time uniform state variables (left) in the context of a null inflation sample paths (right). Both curves are similar since there is no time-dependency modeled here in the case of a null inflation.}
 \label{fig::timeuniform1}
\end{figure}

\begin{figure}[p]
 \centering
 \includegraphics[scale=0.525]{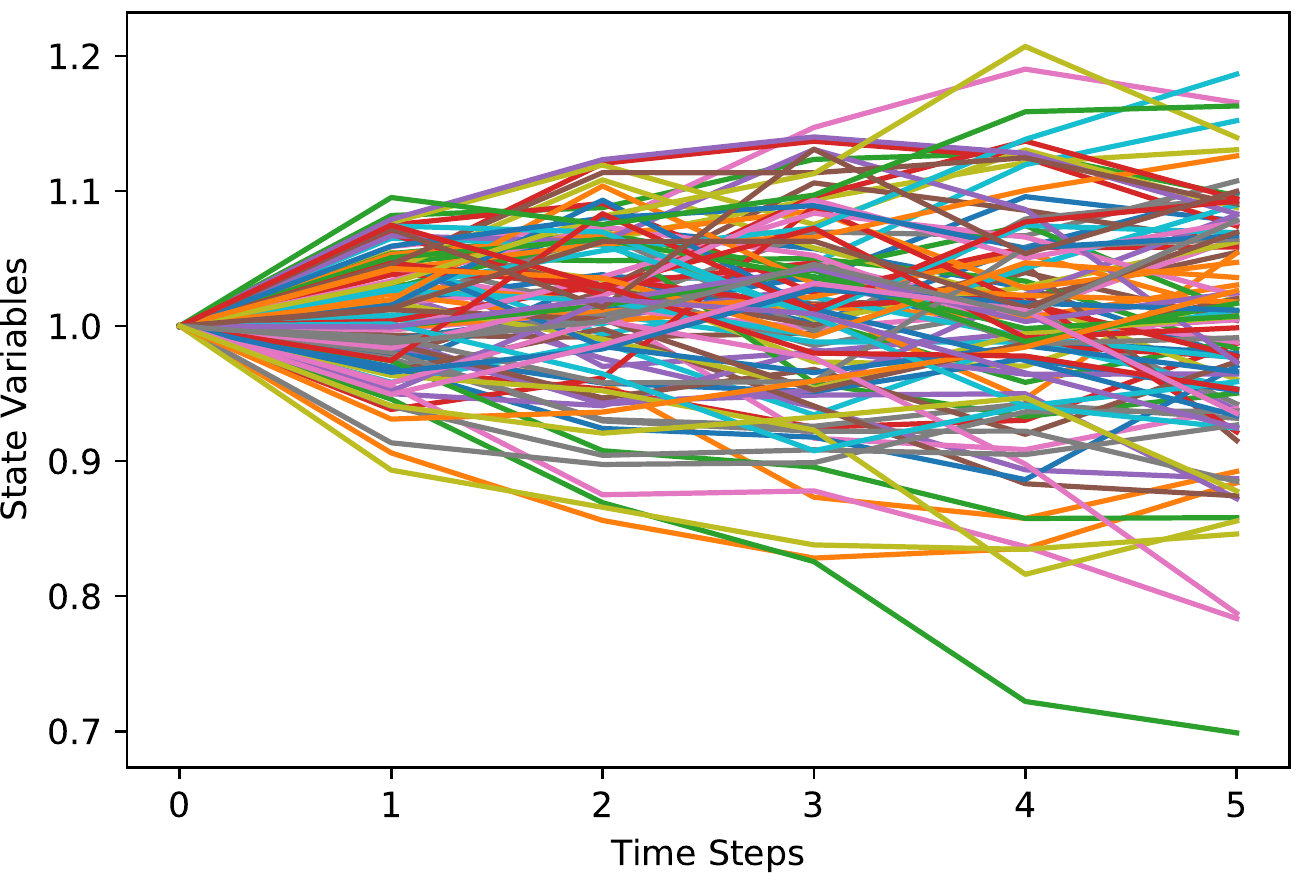}
 \hspace{0.5cm}
 \includegraphics[scale=0.525]{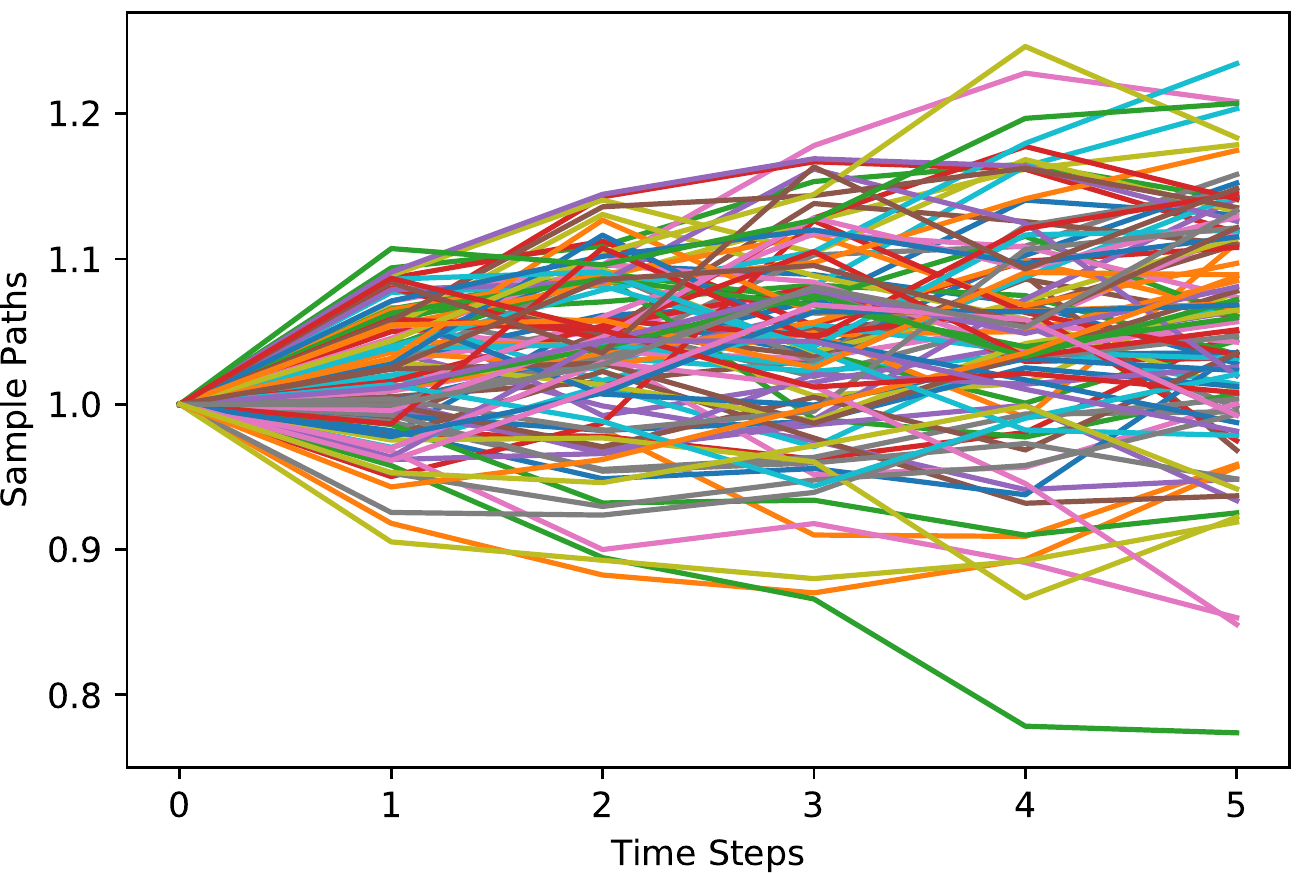}
 \caption[Time uniform state variables in the context of a strong inflation]{Time uniform state variables (left) in the context of a strong inflation (right). The time uniform state variables only capture the fluctuation of the transactions. Therefore, the long term impact of the inflation leading to the strong increase of the sample paths has no effect on the state variables.}
 \label{fig::timeuniform2}
\end{figure}

\begin{figure}[p]
 \centering
 \includegraphics[scale=0.525]{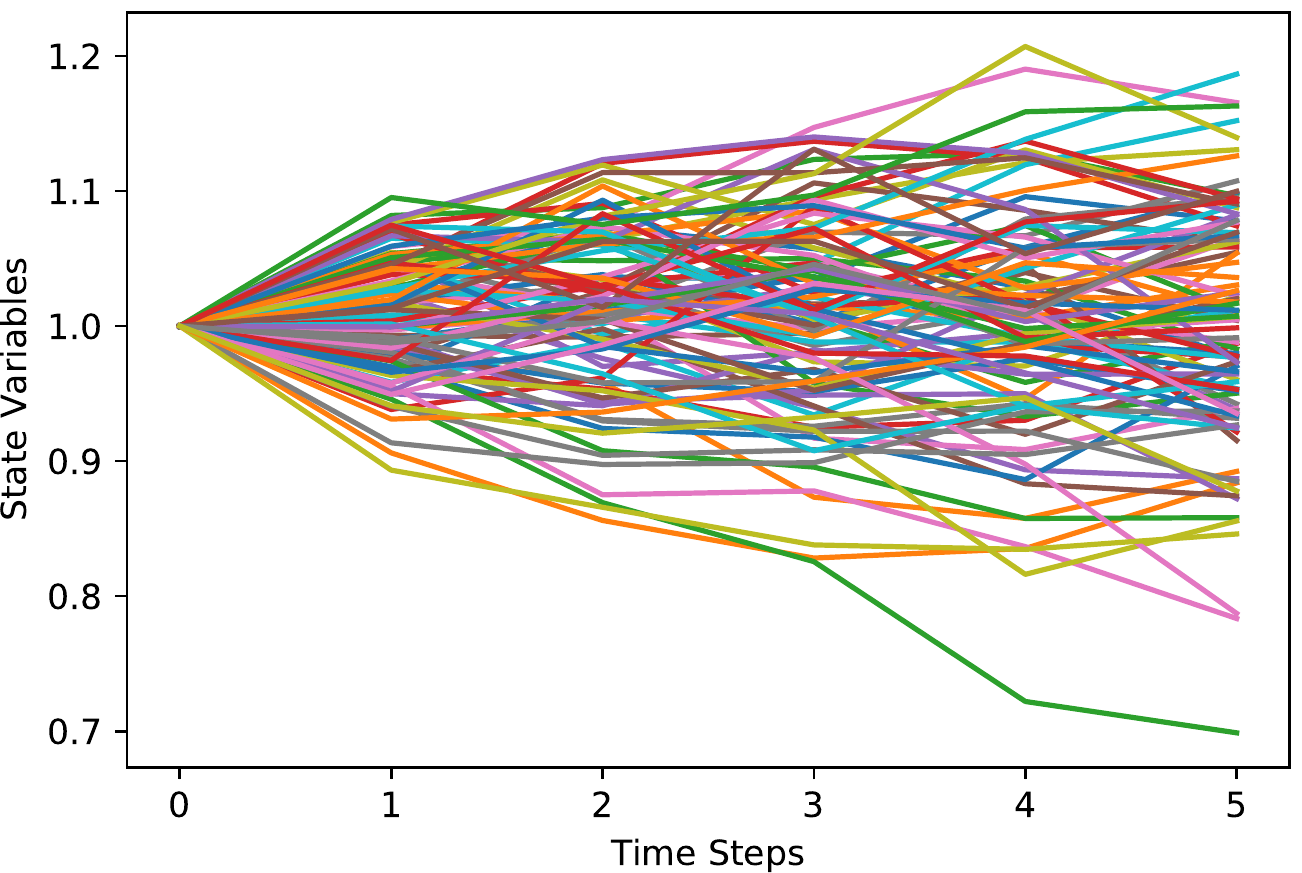}
 \hspace{0.5cm}
 \includegraphics[scale=0.525]{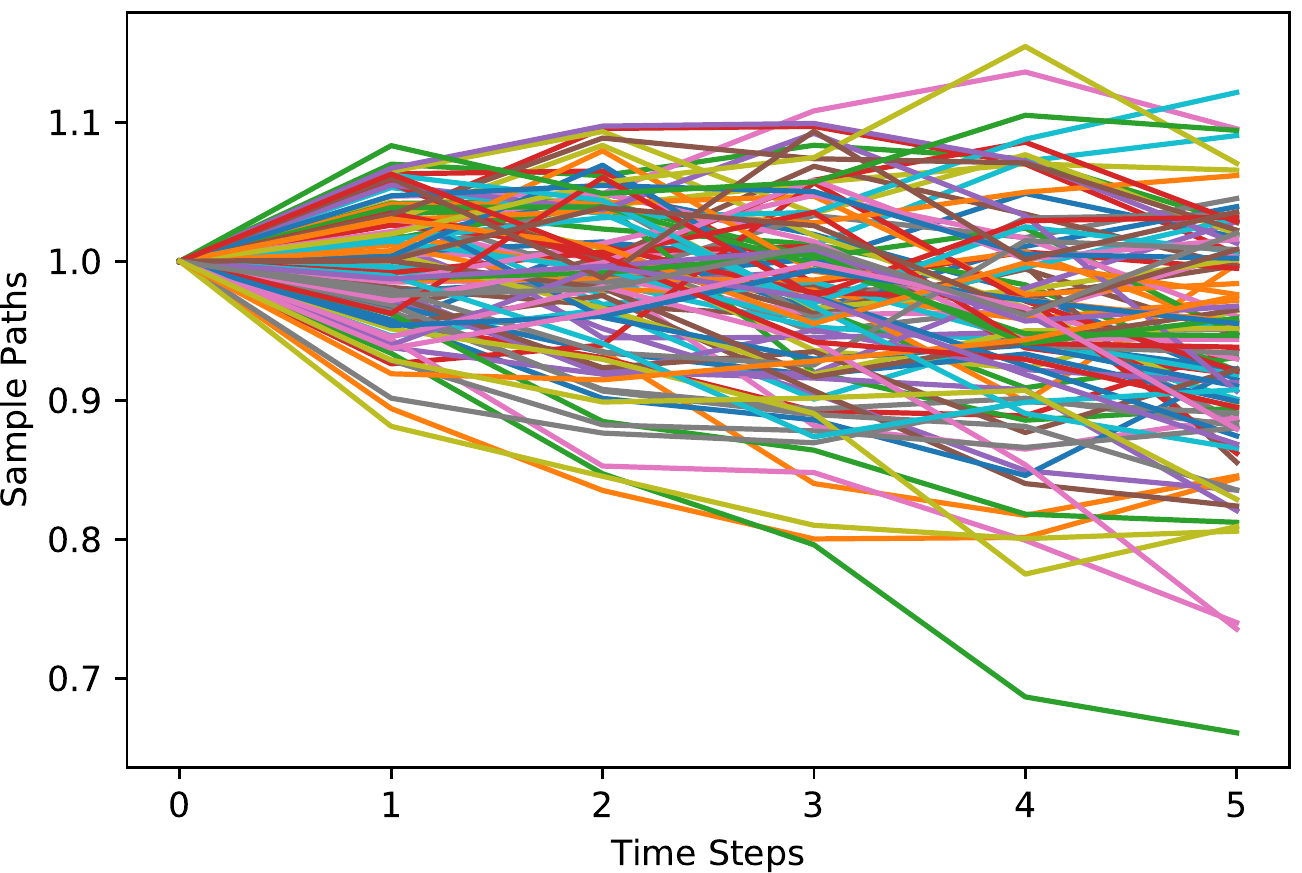}
 \caption[Time uniform state variables in the context of a strong deflation]{Time uniform state variables (left) in the context of a strong deflation (right). The deflation leads to the sample paths decrease. The time uniform state variables, however, are not impacted and only capture the fluctuation of the transactions.}
 \label{fig::timeuniform3}
\end{figure}

\begin{figure}[t!]
 \centering
 \vspace{.5cm}
 \includegraphics[height=4cm]{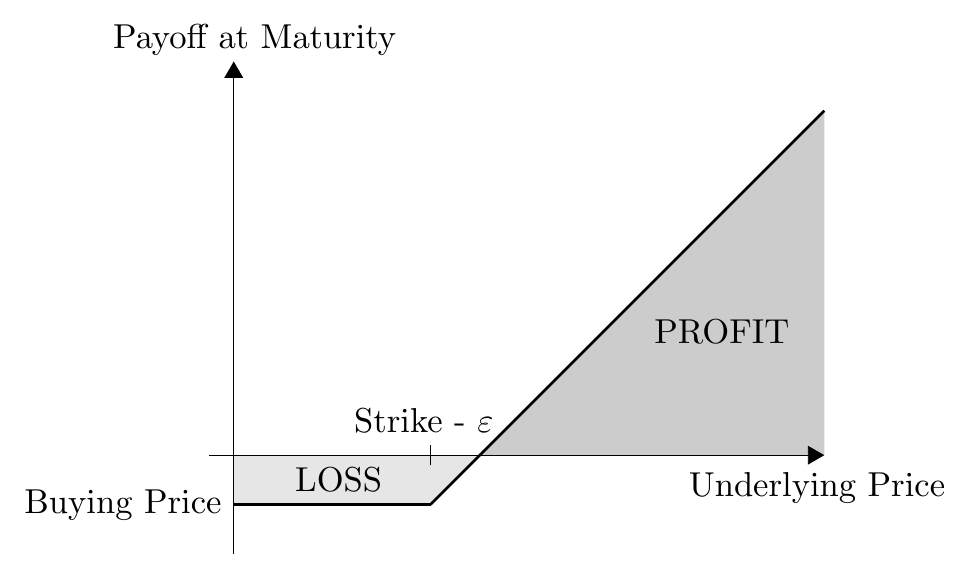}
 \includegraphics[height=4cm]{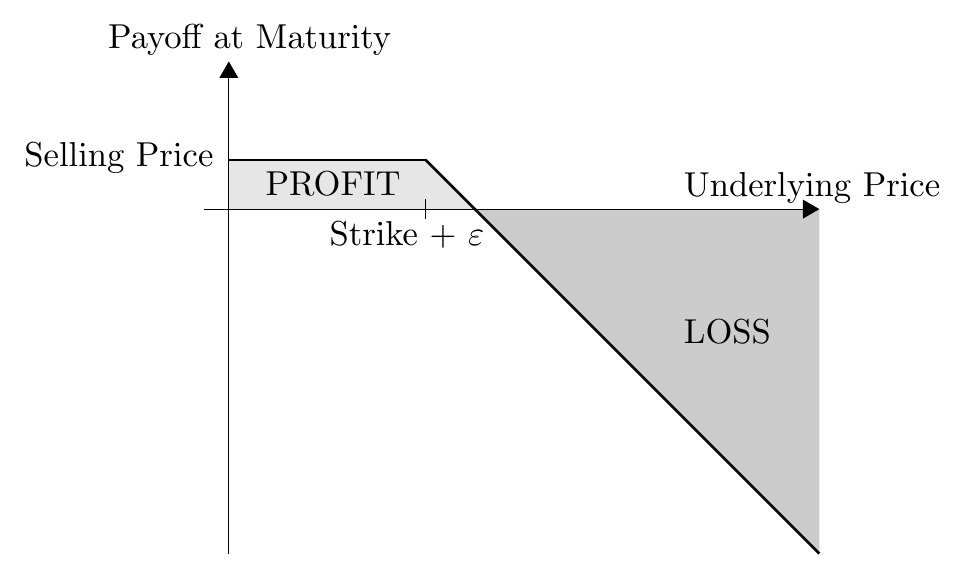}
 \includegraphics[height=5cm]{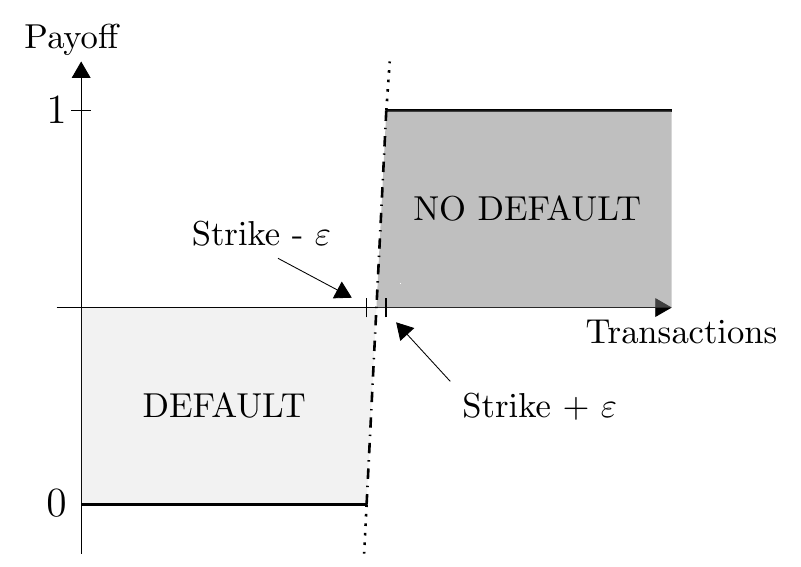}
 \caption[Replication of the digital function with vanilla options]{The BSM formula is commonly used to price vanilla options. The combination of vanilla options at different strikes allows the replication of the digital function. In our case, the BSM formula allows to replicate the thresholds for which we observe a payment default. It offers a benchmark comparison of the values found by our approach denoted by MQLV.}
 \label{fig::bsm_digital}
\end{figure}

We illustrate the idea of the time-uniform variable in Figures \ref{fig::timeuniform1}, \ref{fig::timeuniform2} and \ref{fig::timeuniform3}. In the context of finance and retail banking, a time-uniform variable is a variable that is independent of the concepts of inflation and deflation. Instead of estimating the price of a vanilla option as proposed in \cite{halperin2017qlbs}, we are interested to estimate the future probability of an event using the Q-learning algorithm and a digital function. We recall the digital function is at the center of our approach with the Vasicek model. It effectively allows the modeling of the default probabilities, a key aspect to determine the optimal policy for money management. We first define the terminal condition reflecting that with the following equation

\begin{equation} \label{eq::terminalcondition}
Q_T^*(X_T, a_T=0) = -\Pi_T - \lambda Var \left[ \Pi_T(X_T) \right] \quad ,
\end{equation} 

where $\Pi_T$ is the digital function at time $t=T$ defined such that

\begin{equation} \label{eq::digital}
\Pi_T = 1_{S_T\geq K}=
\left\{\begin{matrix}
& 1 \text{ if } S_T \geq K \\ 
& 0 \text{ otherwise}
\end{matrix}\right. \quad ,
\end{equation}

and the second term, $\lambda Var \left[ \Pi_T(X_T) \right]$, is a regularization term with $\lambda \in \mathbb{R}^+ \ll 0$. The digital function can be approximated using the BSM formula by combining vanilla options, as illustrated in Figure \ref{fig::bsm_digital}. We use a backward loop to determine the value of $\Pi_t$ for $t=T-1, ..., 0$

\begin{equation} \label{eq::backward_pi}
\Pi_t = \gamma \left( \Pi_{t+1} - a_t \Delta S_t \right) \quad \text{with} \quad \Delta S_t = S_{t+1} -\frac{S_{t}}{\gamma} = S_{t+1} - e^{r \Delta t} S_t \quad .
\end{equation}

Following the definition of the equations (\ref{eq::optimalvalue}) and (\ref{eq::backward_pi}), we express the one-step time dependent random reward with respect to the cross-sectional information $\mathcal{F}_t$ as follows

\begin{equation} \label{eq::randomreward}
\begin{split}
R_t(X_t, a_t, X_{t+1}) & =  \gamma a_t \Delta S_t(X_t, X_{t+1}) - \lambda Var \left[ \Pi_t | \mathcal{F}_t \right]\\
& \text{with } Var \left[ \Pi_t | \mathcal{F}_t \right] = \gamma^2 \mathbb{E}_t \left[ \hat{\Pi}_{t+1}^2 -2a_t \Delta \hat{S}_t \hat{\Pi}_{t+1} + a_t^2 \Delta \hat{S}_t^2 \right]
\end{split} \quad .
\end{equation}

The term $\Delta \bar{S}_t$ is defined such that $\Delta \bar{S}_t = \frac{1}{N}\Delta S$,  $\Delta \widehat{S} = \Delta S - \Delta \bar{S}_t$ and $\hat{\Pi}_{t+1}=\Pi_{t+1} - \bar{\Pi}_{t+1}$ with $\bar{\Pi}_{t+1} = \frac{1}{N} \Pi_{t+1}$. Because of the regularizer term, the expected reward $R_t$ is quadratic in $a_t$ and has a finite solution. We therefore inject the one-step time dependent random reward equation (\ref{eq::randomreward}) into the Bellman optimality equation (\ref{eq::bellmanoptimal}) to obtain the following Q-learning update, $Q^\ast$, and the optimal action, $a^\ast$, to be solved within a backward loop $\forall t = T-1, ..., 0$

\begin{equation} \label{eq::qlearningupdate}
\begin{split}
Q_t^\ast (X_t, a_t) = & \: \gamma \mathbb{E}_t \left[ Q_{t+1}^\ast (X_{t+1}, a_{t+1}^\ast ) + a_t \Delta S_t \right] - \lambda Var \left[ \Pi_t | \mathcal{F}_t \right]  \\ 
a_t^\ast (X_t) = & \: \mathbb{E}_t \left[ \Delta \hat{S}_t \hat{\Pi}_{t+1} + \frac{1}{2 \lambda \gamma} \Delta S_t \right] \left[\mathbb{E}_t \left[ \left( \Delta \hat{S}_t \right)^2 \right] \right]^{-1}
\end{split} \quad .
\end{equation}



We refer to \cite{halperin2017qlbs} for further details about the analytical solution, $a^\ast$, of the Q-learning update (\ref{eq::qlearningupdate}). Our approach uses the $N$ Monte Carlo paths simultaneously to determine the optimal action $a^*$ and the optimal action-value function $Q^*$ to learn the policy $\pi^\ast$. We thus do not need an explicit conditioning of $X_t$ at time $t$. We assume a set of $M$ basis function $\lbrace \Phi_n(x) \rbrace$, with $n = 1, \ldots, M$, for which the optimal action $a_t^*(X_t)$ and the optimal action-value function, $Q_t^*(X_t,a_t^*)$, can be expanded 

\begin{equation} \label{eq::basisfunction}
a_t^*(X_t) = \sum_n^M \phi_{nt} \Phi_n(X_t) \quad \text{and} \quad
Q_t^*(X_t, a_t^*) = \sum_n^M \omega_{nt} \Phi_n(X_t) \quad .
\end{equation}  

The coefficients $\phi$ and $\omega$ are computed recursively backward in time $\forall t = T-1, \ldots, 0$. We subsequently define the minimization problem to evaluate $\phi_{nt}$ such that

\begin{equation} \label{eq::minimization1}
G_t(\phi) = \sum_{k=1}^{N} \left[ -\sum_n^M \phi_{nt} \Phi_{n} (X_t^k) \Delta S_t^k + \gamma \lambda \left( \Pi_{t+1}^k - \sum_n^M \phi_{nt} \Phi_{n} (X_t^k) \Delta \widehat{S}_t^k\right)^2 \right] \quad .
\end{equation}

The equation (\ref{eq::minimization1}) leads to the following set of linear equations $\forall n = 1, \ldots, M$

\begin{equation} \label{eq::linearset1}
\begin{split}
\begin{dcases}
A_{nm}^{(t)} = \sum_{k=1}^N \Phi_n(X_t^k) \Phi_m(X_t^k) (\Delta \widehat{S}_{t^k})^2 & \\ 
B_n^{(t)} = \sum_{k=1}^N \Phi_n (X_t^k) \left[ \widehat{\Pi}_{t+1}^k \Delta \widehat{S}_t^k + \dfrac{1}{2\gamma \lambda} \Delta S_t^k \right] &
\end{dcases} 
\text{ with } \sum_m^M A_{nm}^{(t)} \phi_{mt} = B_n^{(t)}
\end{split} \quad .
\end{equation}

The coefficients of the optimal action $a_t^*(X_t)$ are therefore determined by 

\begin{equation} \label{eq::phi}
\phi_t^* = A_t^{-1} B_t \quad .
\end{equation}

We hereinafter use the Fitted Q Iteration (FQI) \cite{hasselt2010double,murphy2005generalization} to evaluate the coefficients $\omega$. The optimal action-value function, $Q^*(X_t, a_t)$, is represented in its matrix form according to the basis function expansion of the equation (\ref{eq::basisfunction}) such that

\begin{equation} \label{eq::expansionQ}
\begin{split}
Q_t^*(X_t,a_t) = & \left(1, a, \dfrac{1}{2} a_t^2 \right) 
\begin{pmatrix}
W_{11}(t) & W_{12}(t) & \ldots & W_{1M}(t)  \\ 
W_{21}(t) & W_{22}(t) & \ldots & W_{2M}(t)  \\ 
W_{31}(t) & W_{32}(t) & \ldots & W_{3M}(t)
\end{pmatrix}
\begin{pmatrix}
\Phi_1 (X_t) \\ \vdots \\ \Phi_M (X_t)
\end{pmatrix} \\
= & A_t^T W_t \Phi(X_t) = A_t^T U_W(t,X_t)
\end{split} \quad .
\end{equation}


Based on the least-square optimization problem, the coefficients $W_t$ are determined using backpropagation $\forall t=T-1, ..., 0$ as follows 

\begin{equation} \label{eq::leastsquareW}
\begin{split}
\mathcal{L}_t(W_t) & = \sum_{k=1}^N \left( R_t(X_t,a_t,X_{t+1}) + \gamma \max_{a_{t+1}\in \mathcal{A}} Q_{t+1}^* (X_{t+1}, a_{t+1}) - W_t \Psi_t (X_t, a_t) \right)^2 \\
& \text{with } W_t \Psi (X_t, a_t)+\epsilon \underset{\epsilon \rightarrow 0}{\longrightarrow} R_t(X_t,a_t,X_{t+1}) + \gamma \max_{a_{t+1} \in \mathcal{A}} Q_{t+1}^* (X_{t+1}, a_{t+1})
\end{split} \quad .
\end{equation}

for which we derive the following set of linear equations


\begin{equation} \label{eq::weightsW}
\begin{dcases}
M_n^{(t)} = \sum_{k=1}^{N} \Psi_n (X_t^k, a_t^k) \left[ \eta  \left( R_t(X_t,a_t,X_{t+1}) + \gamma \max_{a_{t+1} \in \mathcal{A}} Q_{t+1}^* (X_{t+1}, a_{t+1}) \right) \right]  & \\
\text{with } \eta \sim B(N, p)  &
\end{dcases} \quad .
\end{equation}

The term $B(N,p)$ represents the binomial distribution for $n$ samples with probability $p$. It plays the role of a dropout function when evaluating the matrix $M_t$ to compensate the well-known drawback of the Q-learning algorithm that is the overestimation of the Q-function values. We finally reach the definition of the optimal weights to determine the optimal action $a^\ast$ as follows

\begin{equation} \label{eq::optimalW}
W_t^* = S_t^{-1} M_t \quad .
\end{equation} 

\begin{figure}[b!]
 \centering
 \vspace{.5cm}
 \includegraphics[scale=1.025]{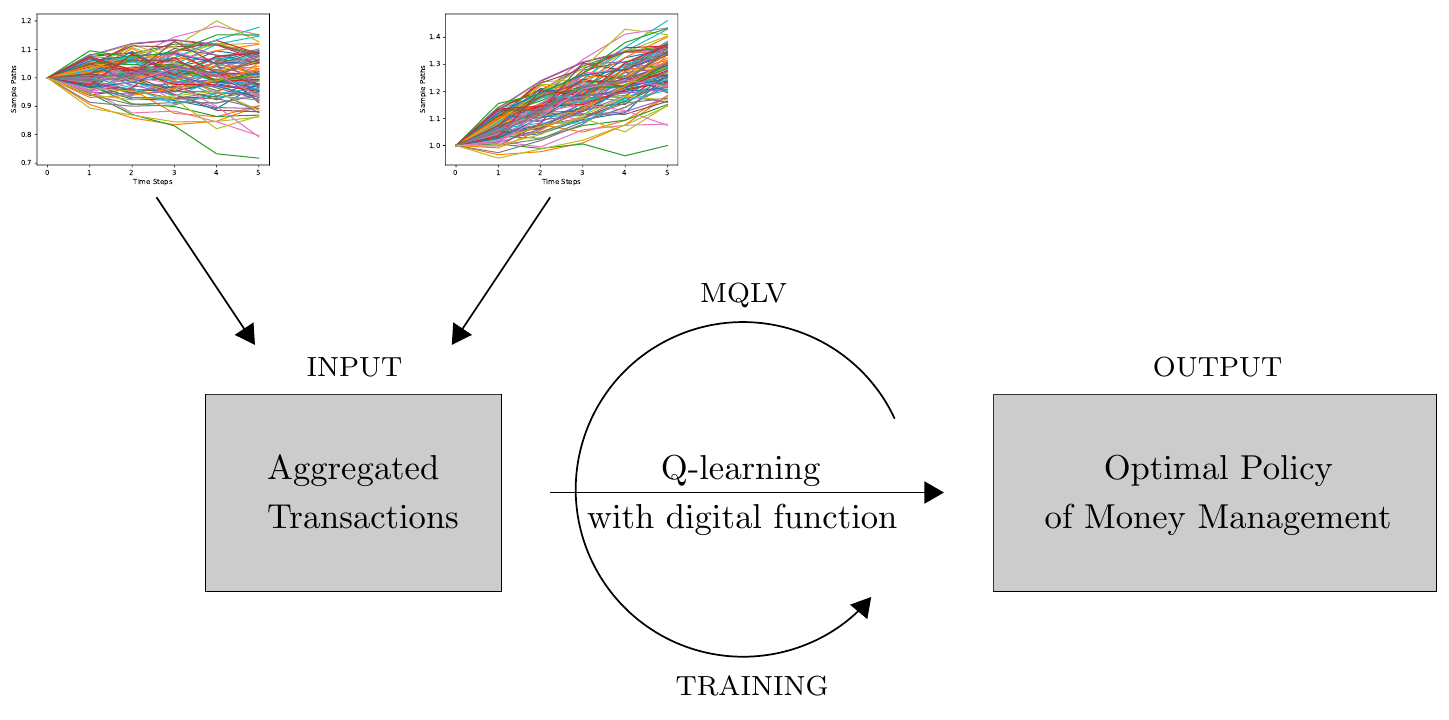}
 \caption[MQLV scheme with aggregated transactions]{MQLV takes as input the aggregated financial transactions. The training is performed using the Bellman equation updated with the digital function. At convergence, the optimal policy of money management is obtained.}
 \label{fig::mqlv_scheme}
\end{figure}

The proposed model does not require any assumption on the dynamics of the time series, neither transition probabilities nor policy or reward functions. It is an off-policy model-free approach. We highlight the overall scheme of MQLV with aggregated transactions in Figure \ref{fig::mqlv_scheme}. The computation of the optimal policy, the optimal actions and the optimal Q-function 
is summarized in Algorithm \ref{algo::mqlv}.

\SetAlFnt{\footnotesize} 
\SetAlCapFnt{\footnotesize} 
\SetAlCapNameFnt{\footnotesize} 
\SetKwFor{Case}{case}{}{}
\SetKwFunction{KwFn}{print}

\begin{algorithm}[t!] 
\setstretch{1.10} 
\DontPrintSemicolon 
\KwData{time series of maturity T, either from generated or true data} 
\KwResult{optimal Q-function $Q^\ast$, optimal action $a^\ast$, value of digital function $\Pi$}

\Begin{ 

/*\textit{\small Condition at $T$}*/ 

$a_T^*(X_T) = 0 $

$Q_T(X_T, a_T) = - \Pi_T = -1_{S_T \geq K}$ using equation (\ref{eq::digital})

$Q_T^*(X_T, a_T^*) = Q_T(X_T, a_T)$

\vspace{.2cm}
/*\textit{\small Backward Loop}*/ 

\For{$t \gets T-1$ \KwTo $0$}{

    /*\textit{\small Evaluate the coefficients $\phi$}*/ 

    compute $A_t, B_t$ using equation (\ref{eq::linearset1})

    $\phi_t^* \gets A_t^{-1} B_t$

	/*\textit{\small Evaluate the coefficients $\omega$}*/ 

    compute $S_t, M_t$ using equation (\ref{eq::weightsW})

    $W_t^* \gets S_t^{-1} M_t$

    $a_t^*(X_t) = \sum_n^M \phi_{nt}^* \Phi_n(X_t)$

    $Q^*(X_t,a_t) = A_t^T W_t^* \Phi_(X_t)$
    }

\vspace{.2cm}
/*\textit{\small Compute the digital function value to estimate the event probability at $t=0$}*/ 

\KwFn{$\Pi_0 = \text{mean}(Q_0^*)$}

} 

\KwRet{} 

\caption{Modified Q-learning for the Vasicek model with the digital value function $\Pi$ \label{algo::mqlv}} 
\end{algorithm} 

\section{Experiments} \label{sec::experiments}
We empirically evaluate the performance of MQLV. We initially highlight the similarities between historical payment transactions and Vasicek generated transactions. We then evaluate the Q-values overestimation with the closed formula of \cite{black1973pricing,merton1973theory}, hereinafter denoted by BSM's closed formula. We finally underline the MQLV's capabilities to learn the optimal policy of money management based on the estimation of future event probabilities in comparison to the BSM's closed formula. We rely on synthetic data sets because of the privacy and the confidentiality issues of the retail banking data sets.  \\

\subsection{Data Availability and Data Description}
One of our contributions is to bring a RL framework designed for retail banking. However, none of the data sets can be released publicly because of the highly sensitive information they contain. We therefore show the similarities between a small sample of anonymized transactions and Vasicek generated transactions \cite{vasicek1977equilibrium}. We then use the Vasicek mean reverting stochastic diffusion process to generate larger synthetic data sets similar to the original retail banking data sets. The mean reverting dynamic is particularly interesting since it reflects a wide range of retail banking transactions including the credit card transactions, the savings history or the clients' spendings. Three different data sets were generated to avoid any bias that could have been introduced by using only one data set. We choose to differentiate the number of Monte Carlo paths between the data sets to assess the influence of the sampling size on the results. The first, second and third data sets contain 
20,000, 30,000 and 40,000 paths. We release publicly the data sets\footnote{\label{note1}The code and the data sets are available at https://github.com/dagrate/MQLV.} to ensure the reproducibility of the experiments.  \\

\subsection{Experimental Setup and Code Availability}
In our experiments, we generate synthetic data sets using the Vasicek model with a parameter $S_0=1.0$ corresponding to the value of the time series at $t=0$, a maturity of six months $T=0.5$, a speed reversion $a=0.01$, a long term mean $b=1$ and a volatility $\sigma=0.15$. The numbers were fixed such that any limitations of the methodology would be quickly observed because the choice of the parameters of the Vasicek model does not have any influence on the results of the Q-learning approach. The number of time steps is fixed equal to 5. We additionally use different strike values for the experiments explicitly mentioned in the Results and Discussions subsection. The simulations were performed on a computer with 16GB of RAM, Intel i7 CPU and a Tesla K80 GPU accelerator. To ensure the reproducibility of the experiments, the code is available at the address\textsuperscript{\ref{note1}} aforementioned. \\

\subsection{Results and Discussions}
We first highlight the similarities between the dynamic of a small sample of anonymized transactions and Vasicek generated transactions for one client \cite{santandercreditcards} in Figure \ref{fig::real_vs_gen} because we cannot release publicly an anonymized transactions data set due to privacy, confidentiality and regulatory issues. The financial transactions in retail banking are periodic and often fluctuates around a long term mean, reflecting the frequency and the amounts of the spendings habits of the clients. The akin dynamic of the original and the generated transactions is highlighted by the small RMSE of 0.03. We also performed a least square calibration of the Vasicek parameters to assess the model's plausibility. We can observe in Table \ref{tab::vas_vs_real} that the Vasicek parameters have the same magnitude and, therefore, it supports the hypothesis that the Vasicek model could be used to generate synthetic transactions.
\\

\begin{table}[b]
\begin{minipage}{0.48\textwidth}
 \centering
 \frame{\includegraphics[scale=0.4]{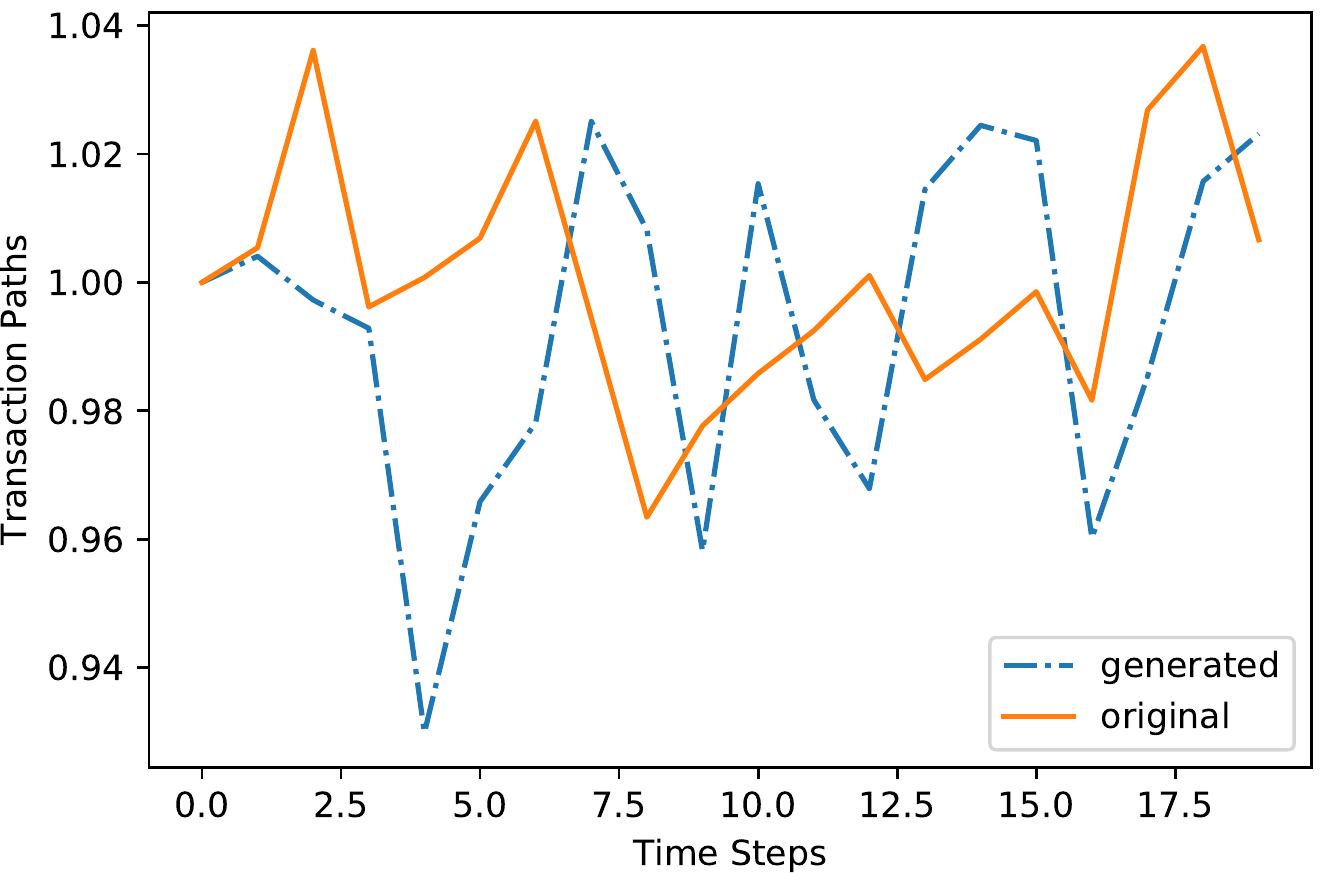}}
 \captionof{figure}[Samples of original and Vasicek generated transactions for one client]{Samples of original and Vasicek generated transactions for one client. The two samples oscillate around a long term mean of 1 and have a similar pattern, highlighted by the RMSE of 0.03 in Table \ref{tab::vas_vs_real}.}
 \label{fig::real_vs_gen}
\end{minipage} \hfill
\begin{minipage}{0.48\textwidth}
\centering 
\caption[RMSE error between the original and the generated transactions]{RMSE error between the samples of original transactions and generated Vasicek transactions of Figure \ref{fig::real_vs_gen}. We also calibrated the Vasicek parameters according to the original transactions to validate the model's plausibility. } \label{tab::vas_vs_real}
\scalebox{0.9}{ 
\begin{tabular}{cccccc} 
  \toprule 
  \; Description \hspace{.25cm} & \hspace{.25cm} Value \;\\
  \midrule
  RMSE & 0.0335 \\
  Vasicek speed reversion $a$ & 0.5444 \\
  Vasicek long term mean $b$ & 0.9001 \\
  Vasicek volatility $\sigma$ & 0.2185 \\
  \bottomrule 
\end{tabular}
} 
\end{minipage}
\end{table}

\begin{figure}[!p]
 \centering
 \frame{\includegraphics[scale=0.65]{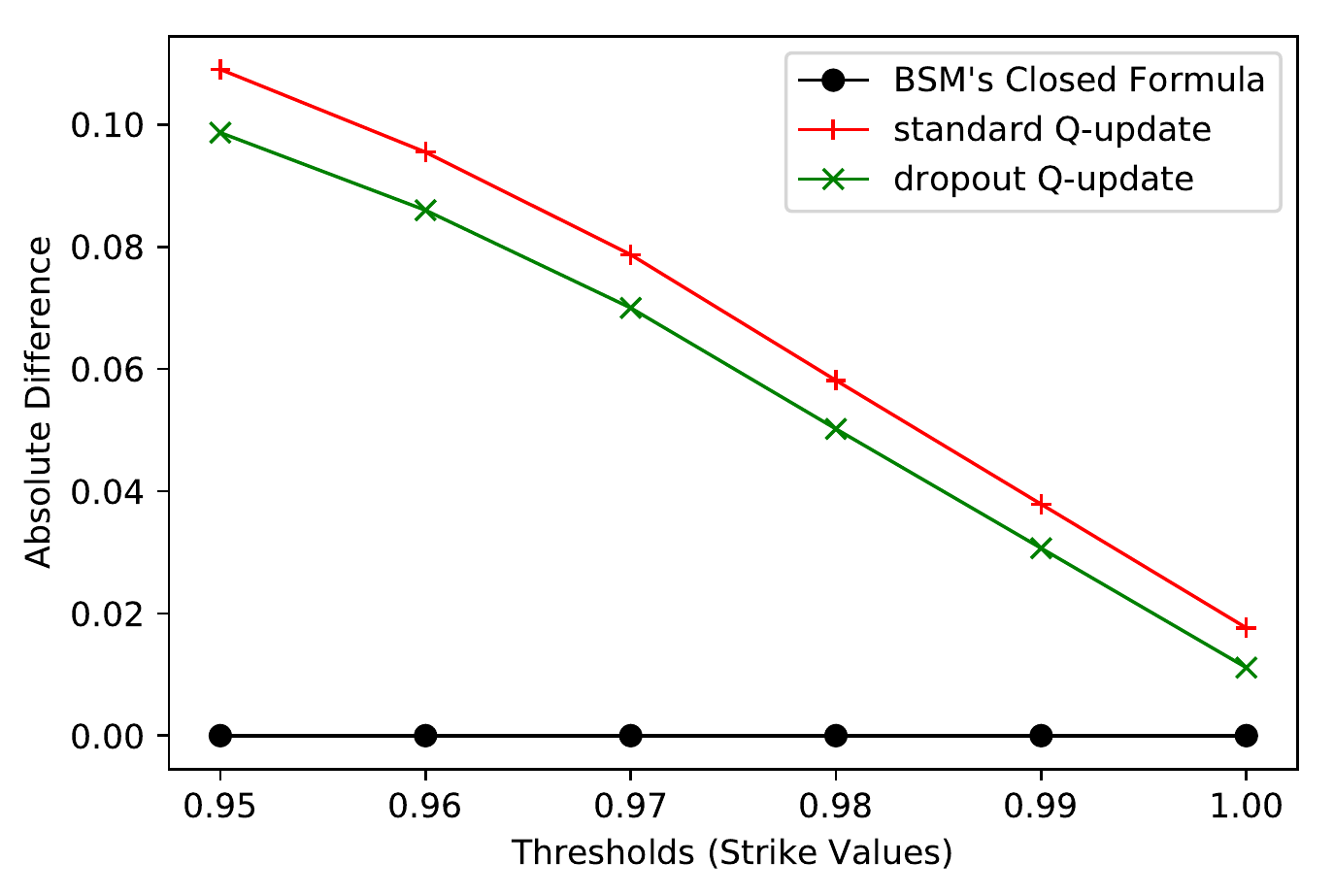}}
 \caption[Price differences between standard and dropout Q-learning updates]{Absolute differences between the standard and the dropout Q-learning updates for the price evaluation of a European vanilla call. The benchmark of the Q-learning algorithms is the BSM's closed formula and, therefore, the differences are set to 0. The dropout Q-update limits the overestimation of the Q-values leading to smaller differences with the BSM's closed formula.}
 \label{fig::dropout}
\end{figure}

We then highlight the motivations of the dropout function of MQLV, capable to limit the overfit of the Q-values. We use the standard example of vanilla option to facilitate the comparison with the BSM’s closed formula \cite{black1973pricing,merton1973theory}, used as reference values for the cross-validation of our approach. The absolute difference between the standard Q-learning and the dropout Q-learning algorithms with respect to the BSM's closed formula values are computed in Figure \ref{fig::dropout} for a European vanilla call option. The absolute differences of the BSM's closed formula are all equal to zero since they are the reference values. We can observe that the standard Q-learning update leads to higher differences for all strikes, between 1\% and 2\% depending on the threshold values. When applying the dropout function, the values obtained are closer to the target values, therefore minimizing the absolute difference. These observations are confirmed quantitatively in Table \ref{tab::dropout}. The valuation differences are reported for the BSM's closed formula, the standard and the dropout Q-learning updates for the three generated data sets. The values of the dropout Q-learning approach are always closer to the BSM's target values. We can conclude that, for our simulation, the dropout update of the equation (\ref{eq::weightsW}) reduces the discrepancy with the BSM's target values.  \\

\begin{table}[!p] 
\centering 
\caption[Valuation differences between the BSM formula and the Q-learning updates]{Valuation differences for a vanilla call option between the BSM's closed formula, the standard Q-learning update and the dropout Q-learning update. The differences are presented for the three generated data sets for different strikes. The dropout function helps to reduce the overestimation of the Q-values. The dropout Q-values are, consequently, closer to the reference values of the BSM's closed formula.} \label{tab::dropout}
\scalebox{.9}{ 
\begin{tabular}{cccccc} 
  \toprule 
  \; Data \; & \; Number \; & \; Strike \; & \; BSM's \; & \; No Dropout \; & \; Dropout \\ 
  \; Set \; & \; of Paths \; &\; Values \; & \; Values (\%) \; & \; Q-Values (\%) \; & \; Q-Values (\%) \\ 
  \midrule
  1 & 20,000 & 0.95 & \textbf{7.096} & 7.223 & \textbf{7.214} \\
  1 & 20,000 & 0.96 & \textbf{6.448} & 6.557 & \textbf{6.549} \\
  1 & 20,000 & 0.99 & \textbf{4.727} & 4.764 & \textbf{4.759} \\
  1 & 20,000 & 1.00 & \textbf{4.230} & 4.241 & \textbf{4.237} \\
  2 & 30,000 & 0.95 & \textbf{7.096} & 7.215 & \textbf{7.208} \\ 
  2 & 30,000 & 0.96 & \textbf{6.448} & 6.552 & \textbf{6.546} \\ 
  2 & 30,000 & 0.99 & \textbf{4.727} & 4.769 & \textbf{4.764} \\ 
  2 & 30,000 & 1.00 & \textbf{4.230} & 4.249 & \textbf{4.246} \\
  3 & 40,000 & 0.95 & \textbf{7.096} & 7.205 & \textbf{7.195} \\ 
  3 & 40,000 & 0.96 & \textbf{6.448} & 6.543 & \textbf{6.534} \\ 
  3 & 40,000 & 0.99 & \textbf{4.727} & 4.765 & \textbf{4.758} \\ 
  3 & 40,000 & 1.00 & \textbf{4.230} & 4.247 & \textbf{4.240} \\
  \bottomrule 
\end{tabular}
} 
\end{table}

\begin{figure}[!p]
 \centering
 \frame{\includegraphics[scale=0.7]{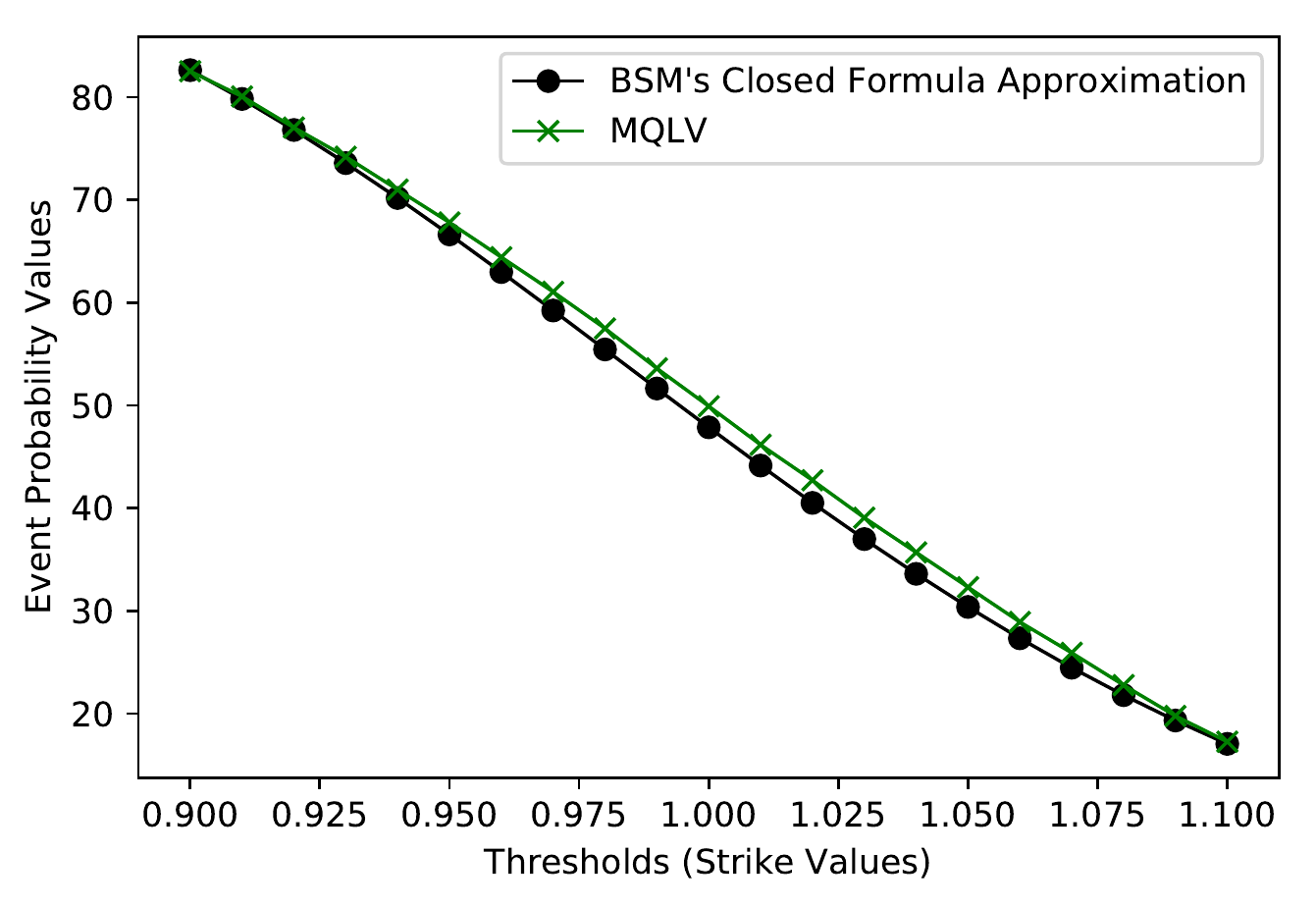}}
 \caption[Event probability values between MQLV and BSM]{Event probability values calculated by MQLV and BSM's closed formula approximation for 40,000 Monte Carlo paths with Vasicek parameters $a=0.01, b=1$ and $\sigma=0.15$. The BSM's closed formula approximation values are used as reference values. The event probabilities of MQLV are close to the BSM's values with a total RMSE of 1.502. It illustrates that MQLV is able to learn the optimal policy leading to accurate event probabilities.}
 \label{fig::digital}
\end{figure}

We present the core of our contribution in the following experiment. We aim at learning the optimal policy of money management. It is particularly interesting for bank loan applications where the differences between a client's spendings policy and the optimal policy can be compared. We show that MQLV is capable of evaluating accurately the probability of a default event using a digital function, which highlights the learning of the optimal policy of money management. Effectively, if the MQLV's learned policy is different than the optimal policy, then the probabilities of default events are not accurate. The estimation of future event probabilities for different strike values is represented in Figure \ref{fig::digital}. We rely on the BSM's closed formula for the vanilla option pricing \cite{black1973pricing,merton1973theory} to approximate the digital function values \cite{hull2003options}. We used, therefore, the BSM's values as reference values to cross-validate the MQLV's values. MQLV achieves a close representation of the event probabilities for the different strike values in Figure \ref{fig::digital}. The curves of both the MQLV and the BSM's approaches are similar with a RMSE of 1.5016. This result highlights that the learned Q-learning policy of MQLV is sufficiently close to the optimal policy to compute event probabilities almost identical to the probabilities of the BSM's formula approximation.  \\

We gathered quantitative results in Table \ref{tab::digital} for a deeper analysis of the MQLV's results. The event probability values are listed for the three data sets. We chose a set of parameters for the Vasicek model such that our configuration is free of any time-dependency. We therefore expect a probability value of 50\% at a threshold of 1 because the standard deviation of the generated data sets is only induced by the standard deviation of the normal distribution, used to simulate the Brownian motion. Surprisingly, the MQLV values at a strike of 1 are closer to 50\% than the BSM's values for all the data sets. We can conclude, subsequently, that, for our configuration, MQLV is capable to learn the optimal policy of money management which is reflected by the accurate evaluation of the event probabilities. \\

\begin{figure}[!p]
 \centering
 \frame{\includegraphics[scale=0.7]{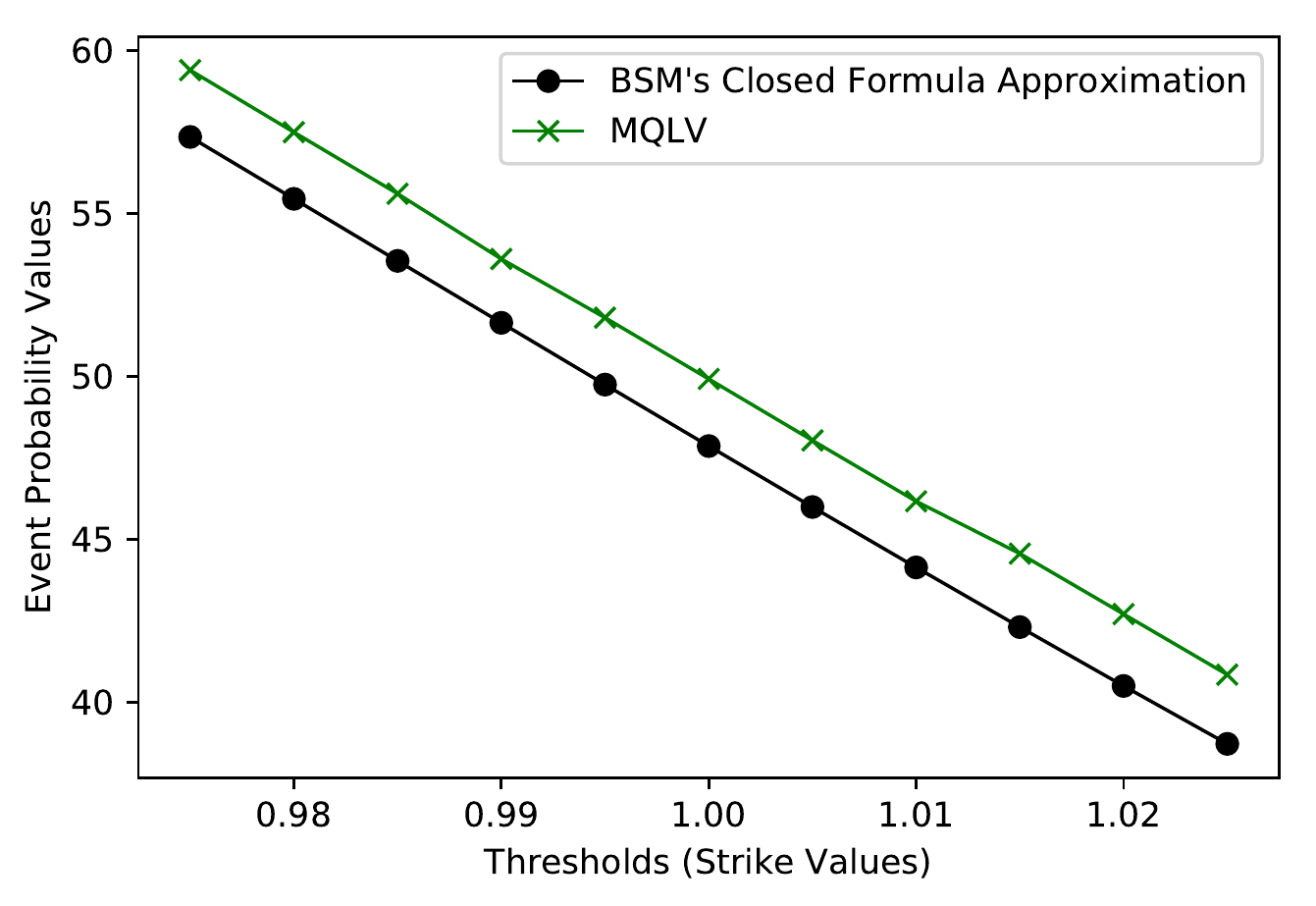}}
 \caption[Zoom in of the event probabilities between MQLV and BSM]{Zoom in of the results pictured in Figure \ref{fig::digital}. The differences between MQLV and BSM are further highlighted. 
 Both curves have the same slope of -381 between the strike values of 0.98 and 1.02. We can already highlight that the MQLV's probabilities illustrate the MQLV's ability to learn the optimal policy.}
 \label{fig::digital2}
\end{figure}

We chose to generate three new data sets with new Vasicek parameters $a$ and $\sigma$ to underline the potential of MQLV and the universality of the results. In Table \ref{tab::otherdata}, we computed the event probabilities for different strikes for the newly generated data sets. The parameter $b$ remains unchanged since we want to keep a configuration free of any time-dependency. We notice that MQLV is capable to estimate a probability of 50\% for a strike of 1 which can only be obtained if MQLV is able to learn the optimal policy. We also observe that the BSM's approximation does lead to a lower accuracy. We showed in this experiment that our model-free and off-policy RL approach, MQLV, is able to learn the optimal policy reflected by the accurate probability values independently of the data sets considered and of the Vasicek parameters.  \\

\textbf{Limitations of the BSM's closed formula used for cross validation} In our experiments, we observed, surprisingly,  that the BSM's closed formula approximation underestimates the event probability values. The volatility is the only parameter playing a significant role in the generation of the time series and, therefore, the event probability should be equal to the mean of the distribution used to generate the random numbers. The Brownian motion is simulated with a standard normal distribution with a 0.5 mean. The BSM's closed formula did not, however, lead to a probability of 0.5 but to slightly smaller values because of the limit of their theoretical framework \cite{black1973pricing,merton1973theory}. We hence observed that MQLV was more accurate than the BSM's closed formula in our configuration.

\begin{table}[!p] 
\centering 
\caption[Comparison of the event probabilities between BSM and MQLV]{Valuation differences of the digital values for event probabilities according to different strikes between the BSM's closed formula approximation and MQLV. Given our time-uniform configuration, the event probability values should be close to 50\% for a strike value of 1. The MQLV values are close to the theoretical target of 50\% at a strike of 1 highlighting the MQLV's capabilities to learn the optimal policy. The BSM's closed formula approximation slightly underestimates the probability values.} \label{tab::digital}
\scalebox{.9}{ 
\begin{tabular}{cccccc}
  \toprule 
  \; Data \; & \; Number \; & \; Strike \; & \; BSM's Approx. \; & \; MQLV \; & \; Absolute \\ 
  \; Set \; & \; of Paths \; & \; Values \; & \; Values (\%) \; & \; Values (\%) \; & \; Difference \\ 
  \midrule
  1 & 20,000 & 0.92 & 76.810 & \textbf{77.098} & 0.288 \\
  1 & 20,000 & 0.98 & 55.447 & \textbf{57.920} & 2.473 \\
  1 & 20,000 & 1.00 & 47.867 & \textbf{50.235} & 2.368 \\
  1 & 20,000 & 1.02 & 40.509 & \textbf{42.865} & 2.356 \\
  2 & 30,000 & 0.92 & 76.810 & \textbf{76.953} & 0.143 \\
  2 & 30,000 & 0.98 & 55.447 & \textbf{57.760} & 2.313 \\
  2 & 30,000 & 1.00 & 47.867 & \textbf{50.043} & 2.176 \\
  2 & 30,000 & 1.02 & 40.509 & \textbf{42.744} & 2.235 \\
  3 & 40,000 & 0.92 & 76.810 & \textbf{77.047} & 0.237 \\
  3 & 40,000 & 0.98 & 55.447 & \textbf{57.491} & 2.044 \\
  3 & 40,000 & 1.00 & 47.867 & \textbf{49.924} & 2.057 \\
  3 & 40,000 & 1.02 & 40.509 & \textbf{42.713} & 2.204 \\
  \bottomrule 
\end{tabular} 
} 
\end{table} 

\begin{table}[!p] 
\centering 
\caption[Event probabilities between BSM and MQLV for various Vasicek parameters]{Event probabilities for data sets generated with different Vasicek parameters $a$ and $\sigma$. The parameter $b$ remains unchanged to keep a time-uniform configuration to facilitate the results explainability. We can deduce that MQLV is able to learn the optimal policy because the MQLV's probabilities are close to the theoretical target of 50\% at a strike of 1. MQLV is also more accurate than BSM's formula in this configuration.} \label{tab::otherdata}
\scalebox{.9}{ 
\begin{tabular}{cccccc}
  \toprule 
  \; Parameters \; & \; Number \; & \; Strike \; & \; BSM's App. \; & \; MQLV \; & \; Absolute \\ 
  \; $a; b; \sigma$ \; & \; of Paths \; & \; Values \; & \; Values (\%) \; & \; Values (\%) \; & \; Difference \\ 
  \midrule
  0.01; 1; 0.10 & 50,000 & 0.98 & 59.856 & \textbf{61.223} & 1.366 \\
  0.01; 1; 0.10 & 50,000 & 1.00 & 48.562 & \textbf{50.001} & 1.439 \\
  0.01; 1; 0.10 & 50,000 & 1.02 & 37.596 & \textbf{39.044} & 1.447 \\
  0.01; 1; 0.30 & 50,000 & 0.98 & 49.558 & \textbf{53.647} & 4.089 \\
  0.01; 1; 0.30 & 50,000 & 1.00 & 45.767 & \textbf{49.997} & 4.230 \\
  0.01; 1; 0.30 & 50,000 & 1.02 & 42.088 & \textbf{46.194} & 4.106 \\
  0.10; 1; 0.15 & 50,000 & 0.98 & 55.447 & \textbf{57.540} & 2.093 \\
  0.10; 1; 0.15 & 50,000 & 1.00 & 47.867 & \textbf{50.015} & 2.148 \\
  0.10; 1; 0.15 & 50,000 & 1.02 & 40.509 & \textbf{42.638} & 2.129 \\
  0.30; 1; 0.15 & 50,000 & 0.98 & 55.447 & \textbf{57.586} & 2.139 \\
  0.30; 1; 0.15 & 50,000 & 1.00 & 47.867 & \textbf{50.022} & 2.155 \\
  0.30; 1; 0.15 & 50,000 & 1.02 & 40.509 & \textbf{42.542} & 2.033 \\
  \bottomrule 
\end{tabular} 
} 
\end{table}

\section{Conclusion} \label{sec::conclusion}
We introduced Modified Q-Learning for Vasicek or MQLV, a new model-free and off-policy reinforcement learning approach capable of evaluating an optimal policy of money management based on the aggregated transactions of the clients. MQLV is part of a banking strategy that looks to minimize the customer churn by including more transparency and more customization in the decision process related to bank loan applications or credit card limits. It relies on a digital function, a Heaviside step function expressed in its discrete form, to estimate the future probability of an event such as a payment default. We discuss its relation with the Bellman optimality equation and the Q-learning update. We conducted experiments on synthetic data sets because of the privacy and confidentiality issues related to the retail banking data sets. The generated data sets followed a mean reverting stochastic diffusion process, the Vasicek model, simulating retail banking data sets such as transaction payments. Our experiments showed the performance of MQLV with respect to the BSM's closed formula for vanilla options. We also highlighted that MQLV is able to determine an optimal policy, an optimal Q-function, the optimal actions and the optimal states reflected by accurate probabilities. We surprisingly observed that MQLV led to more accurate event probabilities than the popular BSM's formula in our configuration. \\

Future work will address the creation of a fully anonymized data set illustrating the retail banking daily transactions with a privacy, confidentiality and regulatory compliance. We will also evaluate the MQLV's performance for data sets that violate the Vasicek assumptions. We will furthermore assess the influence of the basis function on the results accuracy. We effectively observed that the Q-learning update could minor the real probability values for simulation involving a small temporal discretization such as $\Delta t = 200$. Preliminary results showed it is provoked by the basis function approximator error. We will address this point in future research. We will finally extend the Q-learning update to other scheme for improved accuracy and incorporate a deep learning framework in our approach. 
\chapter{Conclusion}

In this concluding chapter, we resume the accomplishments of this research project and we point out the main contributions. Throughout the different chapters, we proposed advanced analytical techniques and state of the art computer science concepts to answer the specific needs facing the retail banking industry in a context of evolving regulations and new clients' behavior. We first highlighted how to overcome the lack of financial data using generative adversarial models and persistent homology. We then addressed multidimensional financial recommendations targeting new or renewed products subscriptions for improved marketing banking strategy. We finally introduced a model-free reinforcement learning approach specifically designed for personal financial advice and custom-tailored requests, trying to limit the customer churn. 

\section{Persistent Homology for Generative Models to Generate Artificial Financial Data}
In our current modern age where digital traces and data are omnipresent, the retail banking industry frequently encounters situations where an insufficient amount of data is available to evaluate accurately the risks of the bank or to propose adequate solutions to the clients. In car loan applications for instance, some age categories of the applicants or loan amounts are historically under represented. Extending the available data is a crucial need. We therefore introduced Persistent Homology for Generative Models. In PHom-GeM, we used four different types of neural networks, including gradient penalty Wasserstein generative adversarial network, generative adversarial network, Wasserstein auto-encoders and variational auto-encoders, to generate synthetic credit card transactions based on a small sample of existing credit card transactions. PHom-GeM characterizes the generative manifolds using persistence homology to highlight manifold features of existing and artificially generated data. We chose to apply PHom-GeM on credit card transactions since the latter is particularly challenging for neural networks and persistent homology. Given the large number of features per transactions, we underline in our experiments that PHom-GeM is able to build effective representation of a high dimensional data set. We moreover introduced the bottleneck distance to compare quantitatively the differences between the original and the generated transaction samples.  \\

PHom-Gem establishes new opportunities to artificially generate and evaluate high dimensional data of retail banking. However, several questions still need to be addressed in future research. First, we used the popular Vietoris-Rips simplicial complex for the construction of the simplex tree. It is known to offer a good compromise between the computational cost and the accuracy of the results. But what is the influence of the simplicial complex? Second, we used the bottleneck distance to compare quantitatively the persistence diagram samples. Given the recent progress of the theoretical framework around optimal transport, the use of the Wasserstein distance for the quantitative comparison between the persistence diagrams of the different samples should provide interesting insights. Finally, PHom-GeM trains the neural network models and compare the original and the artificial samples in a two-steps procedure. We leave for future research how to back-propagate the persistent homology information to the objective loss function of the generative models to improve their training and the generation of adversarial samples\footnote{This chapter of the thesis was published in the 6th IEEE Swiss Conference on Data Science (SDS) and in the ATDA 2019 Workshop in conjunction with the European Conference on Machine Learning and Principles and Practice of Knowledge Discovery in Databases (ECML-PKDD).}.

\section[Accurate Tensor Resolution for Multidimensional Financial Recommendation]{Accurate Tensor Resolution for Multidimensio-nal Financial Recommendation}
The European financial regulation authorities have been increasing regulatory pressure on all the financial actors since the financial crisis of 2008. Under the revised Payment Service Directive aiming the retail banking industry, the banks are losing partly their historical privileges such as the distribution of the credit card payment solutions or the management of the personal finance. Targeting the clients' needs and proposing dynamically new financial products is part of a marketing banking strategy to retain the clients. Nonetheless, the task is complex and highly dimensional. We therefore build upon tensor decomposition, a collection of factorization techniques for multidimensional arrays, and neural networks. We first designed an optimized resolution algorithm for the resolution of complex tensor decomposition. Our resolution algorithm, VecHGrad for Vector Hessian Gradient, uses partial information of the second derivative and an adaptive strong Wolfe's line search to ensure faster convergence. We showed the superior performance of VecHGrad on several well-known research data sets, such as CIFAR-10, CIFAR-100 and MNIST, against the state of the art optimizers used in deep learning and linear algebra, including alternating least square, Adam, Adagrad and RMSProp. We then arrived at our first application in the context of financial recommendations. We predicted the next financial actions of the bank’s clients in a sparse environment. Such approach is of particular interest for the renewal of consumer loans or the renewal of credit cards subscriptions. We used a public transactions data set of the Santander bank. Our predictive method removes the sparsity of the financial transactions before predicting the future client's transactions. The CP tensor decomposition, which decomposes the initial tensor containing the financial transaction into a sum of rank-one tensors, removes the transactions sparsity. The next financial transactions are then predicted using different type of neural networks. We obtained the best predictions when using Long Short Term Memory neural network because of the recurrent nature of the financial transactions activities. In our second financial application, we addressed the authentication on mobile banking application. The traces of the mobile banking application are of particular interest since they can be used to build financial awareness profiles for every clients. We monitored and predicted imbalanced user-device authentication with the PARATUCK2 tensor decomposition and neural networks. The PARATUCK2 tensor decomposition is highly suitable for imbalanced data sets since it expresses a tensor as a multiplication of matrices and diagonal tensors. We relied on a public data set of user-device authentication proposed by the Los Alamos National Laboratory enterprise network for our quantitative experiments. Similarly to our first application, the best results were obtained with Long Short Term Memory neural network because of the recurrent and cyclic patterns of the user-device authentication.  \\

Based on the recent publications in computer science conferences, we do believe that some remaining opened questions about adaptive line search will be addressed in future research. We noticeably observed that the performance of the VecHGrad algorithm is strongly correlated to the performance of the the adaptive line search optimization. In the context of big data and very large data bases, the memory cost of the adaptive line search additionally has a crucial impact. In our client's transactions predictions case study, a different time frame discretization could furthermore be assessed. It would contribute to address a larger choice of financial product recommendations depending on the clients' mid-term and long-term interests. Concerning our second financial applications, the traces of the mobile banking application could finally be further enriched by incorporating traces such as the navigation usage or the time gap between each actions. It would further improve the design of financial awareness profile and, consequently, the potential impact of the bank's advertising campaigns\footnote{This chapter of the thesis was published in the 2017 IEEE Future Technologies Conference (FTC), the 2018 IEEE Big Data and Smart Computing (BigComp) Conference, the 6th International Workshop on Data Science and Big Data Analytics (DSBDA) in conjunction with IEEE International Conference on Data Mining (ICDM 2018) and the 32nd Canadian Conference on Artificial Intelligence. This chapter was also submitted to the thirty-fourth annual conference of the Association for the Advancement of Artificial Intelligence (AAAI).}.

\section{Reinforcement Learning for Decision Making Process in Retail Banking}
Following the new habits of younger generations and the financial regulations promoting more competition, the retail banks face a significant risk of high customer churn. Everyone nowadays can easily change from one bank to another to benefit from more attractive financial opportunities. In this chapter, we described a reinforcement learning approach at the core of the banking strategy to limit the customer churn in the retail banking market. The approach is tailored for personal policy of money management to be used for bank loan applications or credit card limits. We presented MQLV, Modified Q-Learning for the Vasicek model, a model-free and off-policy reinforcement learning approach capable of evaluating the optimal policy of money management based on the aggregated transactions of the clients. We introduced a digital function, a Heaviside step function expressed in its discrete form, to estimate the future probability of an event such as a payment default. This contributed to the evaluation of the optimal policy of money management helping to determine the limits on the credit cards or the amount of a bank loan. In our experiments, we generated artificial data sets reflecting the transaction data sets observable in retail banking because of the privacy, confidentiality and regulatory issues of the clients' historical transactions. Different data sets with different configuration were generated to ensure the legitimacy of the results. We showed the ability of MQLV to learn an optimal policy reflected by the accurate estimation of event probabilities. We compared the MQLV's probabilities with the probabilities computed with the BSM's closed formula. We successfully determined an optimal policy, an optimal Q-function, the optimal actions and the optimal states.  \\

Reinforcement learning applied to the retail banking industry for money management is a very promising concept. We established the foundations for future research in this field. In our opinion, the first research direction should address the creation of a full anonymized data set of high quality illustrating the retail banking daily transactions with a privacy, confidentiality and regulatory compliance. The second research direction should emphasize the choice of the basis function in comparison to the event probability estimation. Effectively, we observed that some fine temporal time discretization could underestimate the real probability values. Considering the recent success of deep learning with reinforcement learning applied to video games, we believe that the approach could benefit in using deep learning for function approximators. Finally, we used the popular Q-learning algorithm for our approach. Nonetheless, a large variety of other algorithms are available, such as the double Q-learning algorithm, offering wide possibilities to extend the MQLV framework\footnote{This chapter of the thesis was published in the MIDAS 2019 Workshop in conjunction with the European Conference on Machine Learning and Principles and Practice of Knowledge Discovery in Databases (ECML-PKDD).}.


\begin{spacing}{0.9}


\bibliographystyle{unsrt} 
\cleardoublepage
\bibliography{./zzz-mybibliography} 



\end{spacing}


%
%

\printthesisindex 

\end{document}